\documentclass[10pt]{article}
\usepackage[margin=1in]{geometry}
\usepackage{amsmath, amssymb, amsthm}
\usepackage{graphicx}
\usepackage{hyperref}
\usepackage{xurl}
\hypersetup{colorlinks=true, linkcolor=black, citecolor=blue!70!black, urlcolor=blue!70!black}
\usepackage{booktabs}
\usepackage{natbib}
\usepackage{xcolor}
\usepackage{caption}
\usepackage{subcaption}
\usepackage{tikz}
\usepackage{placeins}
\usepackage{float}
\usepackage{longtable}

\DeclareUnicodeCharacter{2019}{'}
\DeclareUnicodeCharacter{FF08}{(}
\DeclareUnicodeCharacter{FF09}{)}
\DeclareUnicodeCharacter{FF0C}{,}
\DeclareUnicodeCharacter{FF0E}{.}
\DeclareUnicodeCharacter{FFFD}{\texttt{<?>}}
\DeclareUnicodeCharacter{4E0E}{\texttt{[CJK]}}
\DeclareUnicodeCharacter{4E2D}{\texttt{[CJK]}}
\DeclareUnicodeCharacter{4E48}{\texttt{[CJK]}}
\DeclareUnicodeCharacter{4EC0}{\texttt{[CJK]}}
\DeclareUnicodeCharacter{4EEC}{\texttt{[CJK]}}
\DeclareUnicodeCharacter{540E}{\texttt{[CJK]}}
\DeclareUnicodeCharacter{5E08}{\texttt{[CJK]}}
\DeclareUnicodeCharacter{5F01}{\texttt{[CJK]}}
\DeclareUnicodeCharacter{601D}{\texttt{[CJK]}}
\DeclareUnicodeCharacter{610F}{\texttt{[CJK]}}
\DeclareUnicodeCharacter{6211}{\texttt{[CJK]}}
\DeclareUnicodeCharacter{6700}{\texttt{[CJK]}}
\DeclareUnicodeCharacter{6905}{\texttt{[CJK]}}
\DeclareUnicodeCharacter{6CE8}{\texttt{[CJK]}}
\DeclareUnicodeCharacter{751A}{\texttt{[CJK]}}
\DeclareUnicodeCharacter{7A7F}{\texttt{[CJK]}}
\DeclareUnicodeCharacter{8001}{\texttt{[CJK]}}
\DeclareUnicodeCharacter{81F3}{\texttt{[CJK]}}
\DeclareUnicodeCharacter{8981}{\texttt{[CJK]}}
\DeclareUnicodeCharacter{8FC7}{\texttt{[CJK]}}
\DeclareUnicodeCharacter{96EA}{\texttt{[CJK]}}
\DeclareUnicodeCharacter{98DE}{\texttt{[CJK]}}

\makeatletter
\renewcommand\paragraph{\@startsection{paragraph}{4}{\z@}%
  {2.5ex \@plus1ex \@minus.2ex}%
  {-1em}%
  {\normalfont\normalsize\itshape}}
\makeatother

\title{Can Language Models Get ``Aphasia'' From Neuronal Ablation?}
\author{Pratik Worah\thanks{pworah@uchicago.edu},~~Randall Bateman}
\date{}

\begin{document}
\maketitle

\begin{abstract}
Progressive neuronal loss is the hallmark of neurodegenerative disease, yet how loss of neurons shapes the pattern of cognitive decline remains poorly understood. Here we systematically ablate neurons (computational units) in language models spanning recurrent long short-term memory (LSTM) and Transformer architectures and apply limited fine-tuning as a simple proxy for compensatory plasticity. With whole-unit ablation, LSTMs degrade gradually with syntactic and semantic faculties declining in parallel across mild-to-moderate damage. By contrast, Transformers exhibit late threshold-like collapse under feedforward multi-layer perceptron (MLP) ablation and smoother sigmoid-like decline under attention-head ablation. Qualitatively, MLP-ablated models continue to make within-category semantic substitutions (\emph{semantic paraphasias}) alongside rising non-responses, paralleling Wernicke-type aphasia, while attention-head ablation produces a trend toward loss of semantic specificity similar to the semantic variant of primary progressive aphasia (svPPA).
These findings indicate that feedforward and attentional pathways have functionally distinct failure modes, suggesting that architecture-dependent failure profiles may offer new computational perspectives on aphasic syndromes and neurodegenerative diseases as an \emph{in silico} model.
Notably, early accuracy retention under Transformer attention-head ablation qualitatively parallels pre-symptomatic reserve before the 40--60\% neuronal loss reported by G\'omez-Isla et al.\ (1996) when clinical symptoms emerge in Alzheimer's disease, whereas the steadier LSTM decline parallels the symptomatic progression described by DeKosky and Scheff (1990) and Terry et al.\ (1991).
\end{abstract}


\section{Introduction}

Neurodegenerative diseases are characterised by the progressive loss of neurons in specific brain regions, producing architecture-dependent patterns of cognitive decline. For example, hippocampal damage in Alzheimer's disease yields memory deficits, while anterior temporal damage in semantic dementia selectively dissolves conceptual knowledge~\citep{blennow2006alzheimers, hodges2007semantic}. Identifying the nature of such damage early is important because anti-amyloid drugs may be helpful in slowing disease progression in specific cases. Until recently, however, there were few ways to model neuronal dysfunction or loss because animal models lack language and no computational systems adequately modelled human language. Whether the pattern of language decline (which faculties survive, which erode gradually, and which collapse abruptly) can be predicted from damage to a computational architecture remains an open question.

\begin{figure}[!t]
\centering
\resizebox{\textwidth}{!}{%
\begin{tikzpicture}[%
  box/.style={rounded corners=4pt, minimum width=5.2cm, minimum height=1.85cm,
              text width=4.8cm, inner sep=5pt, align=center, font=\small},
  arr/.style={->, >=stealth, line width=1.2pt, color=black!60},
  row1/.style={box, fill=black!20, draw=black!50},
  row2/.style={box, fill=black!12, draw=black!40},
  row3/.style={box, fill=black!5, draw=black!30},
]
\node[font=\bfseries\small] at (0,3.0) {Ablation Target};
\node[font=\bfseries\small] at (6.4,3.0) {Failure Mode};
\node[font=\bfseries\small] at (12.8,3.0) {Clinical Parallel};
\node[row1] (a1) at (0,1.6) {MLP / FFN\\neurons};
\node[row1] (b1) at (6.4,1.6) {Intact-rate semantic\\paraphasias, more no-responses;\\late threshold collapse.};
\node[row1] (c1) at (12.8,1.6) {Wernicke-type\\aphasia};
\draw[arr] (a1) -- (b1);
\draw[arr] (b1) -- (c1);
\node[row2] (a2) at (0,-0.75) {Attention\\heads};
\node[row2] (b2) at (6.4,-0.75) {Early plateau; trend to\\generic defaults,\\preserved syntax.};
\node[row2] (c2) at (12.8,-0.75) {svPPA / semantic\\dementia};
\draw[arr] (a2) -- (b2);
\draw[arr] (b2) -- (c2);
\node[row3] (a3) at (0,-3.10) {LSTM whole units\\(input + recurrent)};
\node[row3] (b3) at (6.4,-3.10) {Steady, gradual decline;\\syntax \& semantics\\erode in parallel.};
\node[row3] (c3) at (12.8,-3.10) {Diffuse AD-like\\decline};
\draw[arr] (a3) -- (b3);
\draw[arr] (b3) -- (c3);
\end{tikzpicture}%
}
\caption{Summary of ablation: clinical parallel mapping across the ablation experiments. Silencing Transformer feedforward (MLP/FFN) neurons, Transformer attention heads, and whole LSTM hidden units produces distinct failure modes. These patterns parallel focal aphasic syndromes versus steady diffuse decline.}
\label{fig:clinical_summary}
\end{figure}
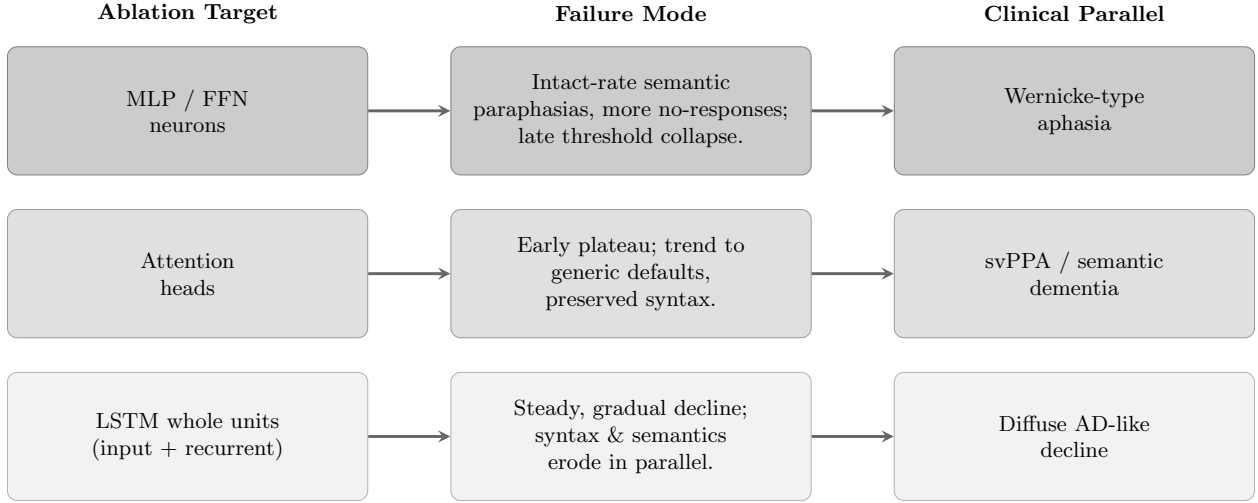

In artificial neural networks, pruning and ablation are well-studied tools for model compression~\citep{frankle2019lottery, han2015learning}, and recent interpretability work has shown that individual neurons encode semantically meaningful features~\citep{olah2020zoom, geva2021transformer}.
Moore et al.\ have demonstrated that progressive lesioning of CNNs produces visual deficits closely resembling posterior cortical atrophy~\citep{moore2023dementia, moore2023neuroplasticity, moore2025realistic}. This motivates the central question underlying our paper: can progressive neuronal ablation in language models reveal architecture-dependent failure modes analogous to clinical aphasias, and can the resulting error patterns serve as a computational model for tracking early neurodegenerative decline before overall accuracy falls?\footnote{We use clinical terms as descriptive labels for error patterns; they do not imply that language models have cognition, comprehension, or disease.}

Concurrent work has begun to apply this logic directly to language models. \citet{yang2026lesioned} perturb a multimodal model (LLaVA~1.6) by injecting noise into a chosen layer. They then search the resulting parameter space for configurations that reproduce the naming-error profiles of 278 people with post-stroke aphasia on the Philadelphia Naming Test (PNT)\footnote{The Philadelphia Naming Test (PNT)~\citep{roach1996philadelphia, mirman2010moss} is a standard 175-item clinical picture-naming assessment for aphasia that classifies patient responses into seven categories: Correct, Semantic paraphasia (meaning-related error), Formal paraphasia (sound-alike error), Mixed (both meaning- and sound-related), Unrelated real word, Neologism (non-word string), and No-response (Supplement~\ref{sec:pnt_recoding}).}, matching six of seven error categories. \citet{wang2026componentlevel} perturb functional components of Mixture-of-Experts and dense Transformers, identify components linked to Broca's and Wernicke's aphasia, and grade the resulting impairment using Western Aphasia Battery subtests. Both studies establish that perturbed language models can occupy clinically recognisable regions of behavioural space.

Our study is complementary in three respects. First, rather than asking how much a network can be compressed while maintaining performance~\citep{frankle2019lottery, han2015learning} or injecting Gaussian noise~\citep{yang2026lesioned}, we use \emph{neuronal ablation}, the targeted, permanent destruction of individual computational units by zeroing all associated incoming and outgoing weights and biases and holding them at zero during subsequent fine-tuning. Thus, we model irreversible neuronal and synaptic loss alongside local compensatory plasticity~\citep{blennow2006alzheimers, moore2023dementia, moore2023neuroplasticity}. Second, whereas Moore et al.~\citep{moore2023dementia, moore2023neuroplasticity, moore2025realistic} track how progressive damage lowers overall accuracy to model a single syndrome, we track both the full decline curve across $k \in [0, 0.9]$ and how the mistakes themselves change (which words are missed and whether grammar or word meaning fails), allowing us to distinguish multiple failure modes. Third, rather than studying a single network architecture, we silence whole units across both recurrent LSTMs and Transformers, as well as between the word-storage (MLP) and context-gathering (attention) pathways inside Transformers (see Section~\ref{sec:methods_models} for exact weight-matrix definitions). We design metrics that allow us to compare how different architectures and pathways shape language decline, and to test whether early changes in error patterns can signal mild damage before overall accuracy drops.

We evaluate this ablation protocol across four language model architectures of increasing scale and complexity (Table~\ref{tab:architectures}), from a small 10M-parameter LSTM~\citep{hochreiter1997lstm} trained from scratch to Qwen~1.5-1.8B~\citep{bai2023qwen}, a 1.8-billion-parameter pretrained Transformer~\citep{vaswani2017attention}.

\paragraph{Contributions.}
Our work makes the following contributions:
\begin{enumerate}
    \item \textbf{Architecture determines failure dynamics, not just severity.} On a shared unit-removal footing, both the from-scratch Small LSTM and the pretrained AWD-LSTM degrade gradually on next-token accuracy from $k = 0.1$ without a buffered plateau (Figures~\ref{fig:accuracy}a,b). On the Cloze battery, their syntactic and semantic scores erode in parallel without dissociating from $k = 0.1$ onward, reaching the zero floor on content words across $k = 0.4$--$0.6$ as predictions fall back to common function words (Figure~\ref{fig:cloze}, row 1). This steady co-decline parallels the broadly distributed symptomatic cognitive erosion of Alzheimer's disease\footnote{While Alzheimer's disease initially presents with a prolonged pre-symptomatic reserve phase followed by focal medial temporal deficits, its symptomatic phase progresses to broad, multi-domain cognitive decline. Our analogy to LSTM whole-unit ablation captures this second, syndromic dimension: multiple linguistic functions eroding gradually and in parallel without a selective syntactic--semantic split, in contrast to the focal pathway dissociations of Transformer sub-block ablation.}~\citep{blennow2006alzheimers}.
    Pretrained Transformers behave differently. Under MLP neuron ablation, they lose accuracy at mild damage at about the same rate as the LSTMs (no early retention advantage). They then collapse abruptly across a critical threshold ($k^* \approx 0.3$--$0.4$, or $20$--$27\%$ of non-embedding parameters removed; Figures~\ref{fig:accuracy}c,e; Supplementary Figure~\ref{sfig:accuracy_param_fraction}; Supplement~\ref{sec:supp_arch_summary}). By contrast, attention-head ablation loses less accuracy across mild-to-moderate damage ($k = 0.1$--$0.3$) before declining across $k = 0.4$--$0.7$ (Figures~\ref{fig:accuracy}d,f), and on the Cloze battery it preserves substantially higher grammatical and semantic scores than MLP or LSTM unit removal through $k = 0.5$ (Figure~\ref{fig:cloze}, rows 2--3).
    \item \textbf{Diagnostic metrics that distinguish architectures beyond raw accuracy.} Two models can have the same overall accuracy yet fail on very different words, so we pair two diagnostic tools to measure how damage reshapes predictions at matched accuracy. First, we introduce a vocabulary moment ratio ($R_4$) and Jensen--Shannon divergence benchmarked against an accuracy-matched random-failure baseline (null model; Section~\ref{sec:moments}; Figures~\ref{fig:R4}--\ref{fig:null_model_comparison}). Rather than failing at random on common function words such as \emph{``the''} or \emph{``and,''} every ablated model continues to get the most common words right and makes its new mistakes on less common content words. Moreover, for a given drop in overall accuracy, this shift toward missing rarer words is much stronger under Transformer ablation than under LSTM unit removal (Figure~\ref{fig:null_model_comparison}; Supplement~\ref{sec:supp_null_model}). Second, we use a quantified 100-prompt Cloze test that scores grammatical part-of-speech match alongside semantic word similarity, separating architectures where grammar and meaning erode together (LSTMs) from those where specific pathways diverge (Transformers; Sections~\ref{sec:cloze}--\ref{sec:clinical_evidence}; Figure~\ref{fig:cloze}). Because these non-random error shifts appear even at mild damage ($k = 0.1\text{--}0.2$) before overall accuracy collapses, applying a similar accuracy-matched comparison to patient responses may be helpful in distinguishing early neurodegeneration from benign lapses (Section~\ref{sec:discussion}).
    \item \textbf{Pathway-dependent linguistic failure profiles.} Feedforward (MLP) ablation and attention-head ablation in Transformers produce distinct degradation trajectories and directional error-profile differences that partially parallel the clinical distinction between Wernicke-type aphasia and the semantic variant of primary progressive aphasia (svPPA), respectively~\citep{wernicke1874aphasische, gornotempini2011classification} (Figure~\ref{fig:clinical_summary}; Table~\ref{tab:summary}).\footnote{Cloze tests~\citep{cloze} are fill-in-the-blank sentence-completion tasks that can separate \emph{syntactic} competence (producing grammatically well-formed sequences) from \emph{semantic} competence (selecting contextually appropriate word meanings). Clinical aphasiology has long recognised that these faculties can be independently disrupted: Wernicke's aphasia, caused by posterior temporal lesions, produces fluent but semantically empty speech~\citep{wernicke1874aphasische}, while Broca's aphasia disrupts syntactic production~\citep{broca1861remarques}.}
    MLP ablation drives a steep threshold collapse of both Cloze metrics while keeping semantic paraphasias at the intact rate and raising no-responses, whereas attention-head ablation preserves far greater Cloze competence through nominal $k = 0.5$, where half as many non-embedding weights are zeroed as in MLP removal (Figure~\ref{fig:cloze}, rows 2--3; Tables~\ref{tab:cloze_summary} and~\ref{tab:error_dist}). Attention-head ablation also shifts null-corrected $R_4$ less than MLP ablation (Figure~\ref{fig:null_model_comparison}) and shows a non-significant directional trend toward more generic substitutions (Section~\ref{sec:clinical_evidence}). In contrast to both Transformer pathways, LSTM whole-unit ablation erodes grammar and meaning together without a selective split (paralleling diffuse AD-type decline), and on the Philadelphia Naming Test (PNT) taxonomy only the larger ablated Transformers (not the LSTMs) match the overall accuracy and word-error range of post-stroke aphasia patients (Section~\ref{sec:clinical_evidence}; Figures~\ref{fig:clinical_summary} and~\ref{fig:pnt_main}; Table~\ref{tab:summary}; Supplementary Table~\ref{stab:pnt_main}).
\end{enumerate}

\noindent In short, these three contributions build on one another. How a network is wired determines whether its overall accuracy fades gradually or collapses at a threshold (Contribution~1). Our diagnostic metrics ($R_4$ with an accuracy-matched null model and the quantified Cloze test) look beyond overall accuracy to measure which words and linguistic skills fail (Contribution~2). Finally, applying those same metrics to specific pathways distinguishes word-storage (MLP) from context-gathering (attention) damage in Transformers, and contrasts both with diffuse LSTM decline, in ways that parallel clinical aphasias and dementia (Contribution~3; Figure~\ref{fig:clinical_summary} and Table~\ref{tab:summary}).

\FloatBarrier
\section{Results}

\subsection{Architectures and Ablation Protocols}

We evaluate four autoregressive language models across two architecture families (Table~\ref{tab:architectures}; Supplementary Table~\ref{stab:summary_k05}): two recurrent networks, a Small LSTM trained from scratch and a pretrained AWD-LSTM~\citep{merity2017pointer, merity2018regularizing, howard2018universal}, and two pretrained Transformers, Qwen~0.5B and Qwen~1.8B~\citep{bai2023qwen} (base training and pretraining details in Supplement~\ref{sec:supp_methods}).
To compare these architectures on equal footing, we apply a standardised unit-level ablation protocol that completely silences (fixes the output to $0$) a fraction $k \in [0, 0.9]$ of computational units per layer across all layers. In the Small LSTM and AWD-LSTM, we remove \emph{whole recurrent hidden units}, zeroing all incoming input and recurrent connections, gate biases, and outgoing connections to the next layer. In Qwen~0.5B and Qwen~1.8B, we remove either \emph{feedforward (MLP) intermediate neurons} or \emph{attention heads}. This design isolates architectural pathway properties from confounds of model size or lesion granularity, allowing us to compare degradation both as a function of $k$ and against the fraction of non-embedding parameters removed (see Methods, Section~\ref{sec:methods}, and Supplementary Figure~\ref{sfig:accuracy_param_fraction} for details).

Following ablation, all four models undergo a brief stochastic gradient descent (SGD) pass on TinyStories~\citep{eldan2023tinystories} to simulate local compensatory plasticity while gradient hooks keep all ablated weights fixed at zero.
We then evaluate each ablated model on two complementary benchmarks (see Methods, Section~\ref{sec:methods}, for details): (i)~held-out corpus next-word prediction to measure accuracy degradation and token-level error distributions (Sections~\ref{sec:accuracy}--\ref{sec:moments}), and (ii)~a shared 100-prompt Cloze sentence-completion battery to disentangle syntactic, semantic, and clinical error profiles (Sections~\ref{sec:cloze}--\ref{sec:clinical_evidence}).

\begin{table}[!htbp]
\centering
\caption{Summary of model architectures and unit-level ablation targets. Baseline Acc.\ is the top-1 next-word prediction accuracy of the undamaged ($k = 0$) model under the default protocol. Without the brief adaptation step, undamaged Qwen~0.5B and 1.8B score $51.5\%$ and $50.6\%$, while the LSTMs are unchanged at $27.5\%$ and $38.1\%$.}
\label{tab:architectures}
\begin{tabular}{@{}lllc@{}}
\toprule
Architecture & Ablated Computational Units & Params & Baseline Acc. \\
\midrule
Small LSTM   & Whole recurrent hidden units & ${\sim}$10M  & 27.5\% \\
AWD-LSTM     & Whole recurrent hidden units & ${\sim}$44M  & 38.1\% \\
Qwen 0.5B    & (1)~Feedforward (MLP) neurons; (2)~Attention heads & ${\sim}$620M & 53.6\% \\
Qwen 1.8B    & (1)~Feedforward (MLP) neurons; (2)~Attention heads & 1.8B         & 54.7\% \\
\bottomrule
\end{tabular}
\end{table}

\subsection{Accuracy Degradation Curves}
\label{sec:accuracy}

\begin{figure}[!htbp]
\centering
\includegraphics[width=0.96\textwidth,height=0.62\textheight,keepaspectratio]{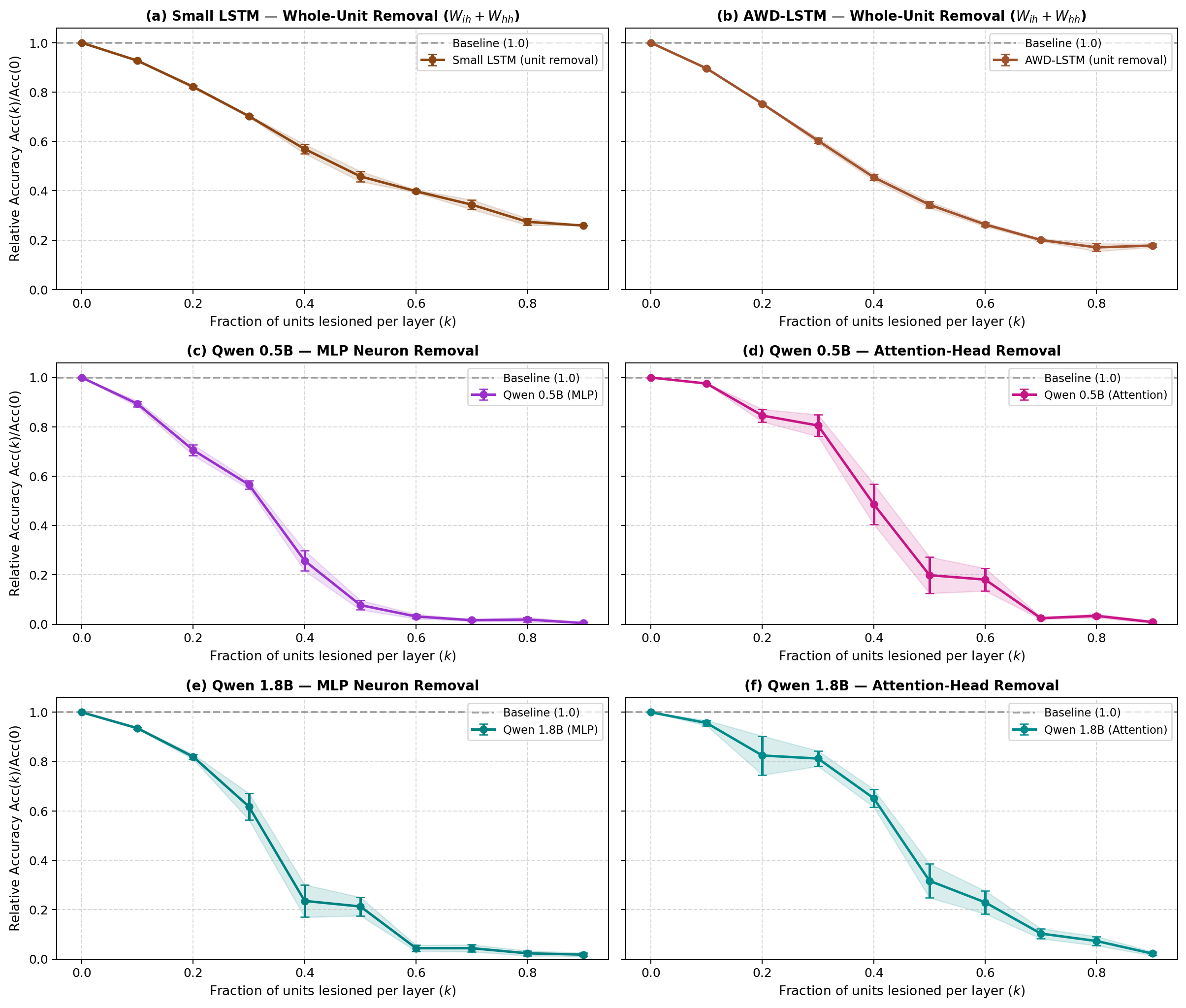}
\caption{Next-word prediction accuracy relative to the undamaged baseline ($\mathrm{Acc}(k)/\mathrm{Acc}(0)$) across unit-ablation fractions $k \in [0, 0.9]$ (means $\pm 1$ SE). The six conditions reveal three distinct degradation profiles across architectures and pathways. (a,b)~Row 1: Whole-unit removal in both the Small LSTM and pretrained AWD-LSTM causes a steady, near-monotonic decline from $k = 0.1$ onward, with neither an early plateau nor a late collapse cliff. (c,e)~Rows 2--3, left: Qwen~0.5B and 1.8B MLP neuron removal initially declines at a pace similar to the LSTMs through $k \le 0.3$. It then undergoes a steep threshold collapse across $k^* \approx 0.3$--$0.4$ to near-zero accuracy. (d,f)~Rows 2--3, right: Qwen~0.5B and 1.8B attention-head removal shows early plateau-like buffering across $k = 0.1$--$0.3$ followed by a smoother S-shaped decline. Full numerical trajectories and curve-shape tests appear in Supplement~\ref{sec:supp_arch_summary} (Supplementary Tables~\ref{stab:summary_k05}--\ref{stab:curve_shape}), connection-pruning controls appear in Supplementary Figure~\ref{sfig:accuracy_all}, and Supplementary Figure~\ref{sfig:accuracy_param_fraction} plots the same trajectories against the fraction of non-embedding parameters removed.}
\label{fig:accuracy}
\end{figure}

All models lose next-word prediction accuracy as the unit-ablation fraction $k$ increases from $0$ to $0.9$, but the shape of the decline separates into three distinct regimes across architectures and pathways (Figure~\ref{fig:accuracy}).\footnote{Full per-model accuracy numbers, curve-shape tests, and parameter-fraction scaling appear in Supplement~\ref{sec:supp_arch_summary}, Supplementary Tables~\ref{stab:summary_k05}--\ref{stab:curve_shape}, and Supplementary Figure~\ref{sfig:accuracy_param_fraction}.}
In both the Small LSTM and the pretrained AWD-LSTM, silencing whole hidden units produces a steady, near-monotonic decline in accuracy from $k = 0.1$ onward, with neither an early plateau nor a late collapse cliff, paralleling progressive, diffuse neurodegenerative decline (Figures~\ref{fig:accuracy}a,b).
Under MLP neuron removal, Qwen~0.5B and 1.8B initially lose accuracy across $k = 0.1$--$0.3$ at a pace comparable to the LSTMs (without an early plateau), but then undergo an abrupt threshold collapse across $k^* \approx 0.3$--$0.4$, falling to near-zero accuracy by $k = 0.5$--$0.6$ (Figures~\ref{fig:accuracy}c,e).
By contrast, under attention-head removal, both Qwen models show early plateau-like buffering on the nominal-$k$ axis (retaining $96$--$98\%$ of baseline accuracy at $k = 0.1$ and flattening at $81$--$85\%$ across $k = 0.2$--$0.3$), followed by a broader S-shaped (sigmoid-like) decline across $k = 0.4$--$0.7$ (Figures~\ref{fig:accuracy}d,f).

\subsection{Error Distributions Across the Vocabulary}
\label{sec:moments}

\begin{figure}[!tbp]
\centering
\includegraphics[width=0.96\textwidth,height=0.52\textheight,keepaspectratio]{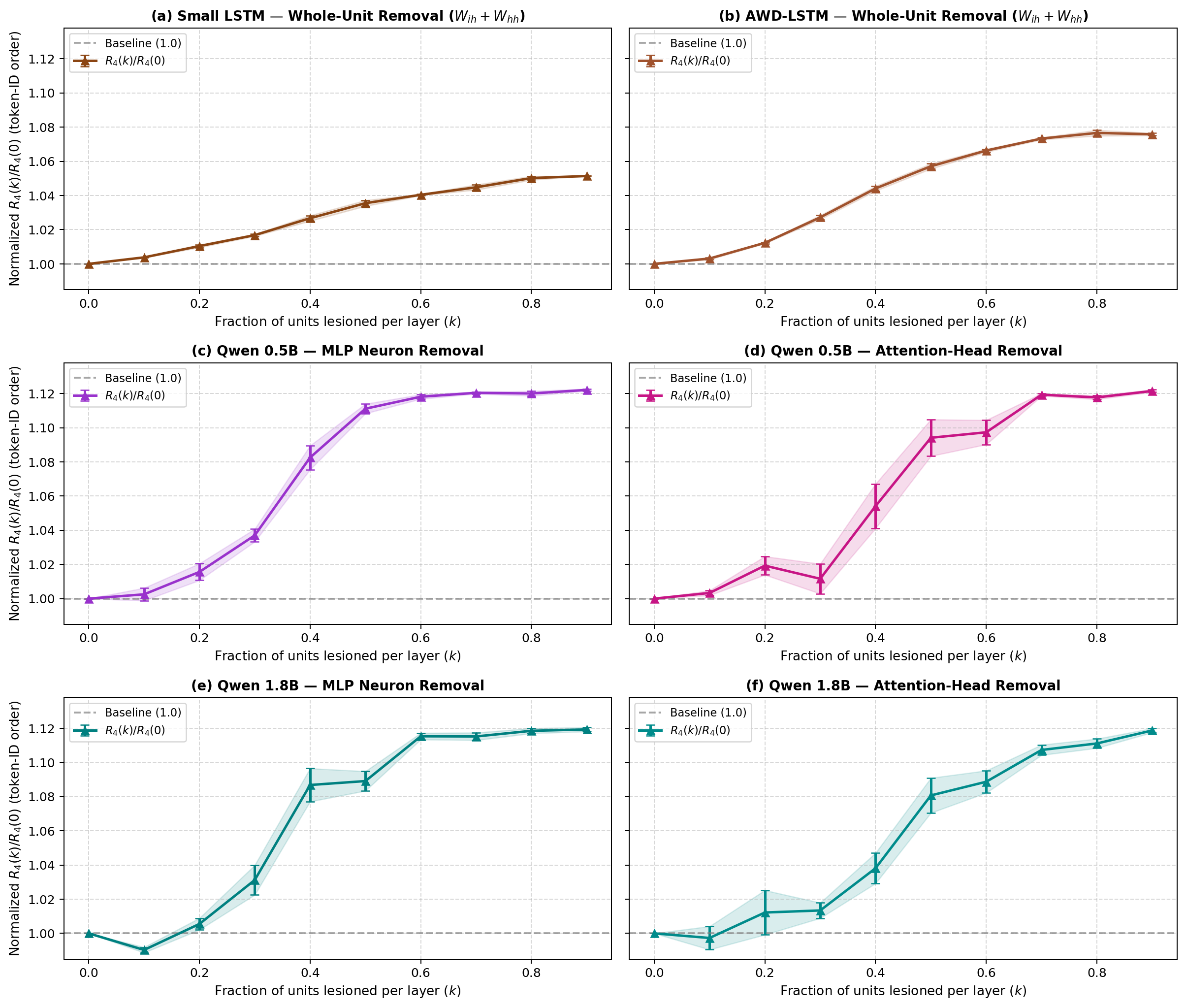}
\caption{Normalised moment ratio $R_4(k)/R_4(0)$ (Equation~\ref{eq:R4}) across unit-ablation fractions $k \in [0, 0.9]$ on a shared vertical scale (means $\pm 1$ SE; token IDs are ordered by corpus frequency in the LSTMs and BPE merge order in Qwen). Because any drop in overall accuracy adds errors on common, low-index words ($i \approx 0$), raw $R_4(k)/R_4(0)$ rises as accuracy falls, so each trajectory closely mirrors the inverse of its accuracy curve in Figure~\ref{fig:accuracy} (motivating the accuracy-matched null model in Figure~\ref{fig:null_model_comparison}). (a,b)~Row 1: Whole-unit removal in both LSTMs produces a smooth, modest rise across the sweep ($\le 5.1\%$ and $\le 7.7\%$). (c,e)~Rows 2--3, left: Qwen MLP neuron removal changes little at mild damage (dipping slightly to $0.990$ at $k = 0.1$ in Qwen~1.8B before rising to $1.031$--$1.037$ at $k = 0.3$), then jumps sharply across the $k^* \approx 0.3$--$0.4$ collapse cliff. (d,f)~Rows 2--3, right: Qwen attention-head removal stays nearly flat during the $k \in [0.1, 0.3]$ plateau ($\le 1.9\%$ change) before rising across $k = 0.4$--$0.7$ (more steeply in Qwen~0.5B than in Qwen~1.8B). Per-$k$ values and connection-pruning controls appear in Supplement~\ref{sec:supp_null_model} and Supplementary Figure~\ref{sfig:R4_all}.}
\label{fig:R4}
\end{figure}

\begin{figure}[!tbp]
\centering
\includegraphics[width=0.95\textwidth,keepaspectratio]{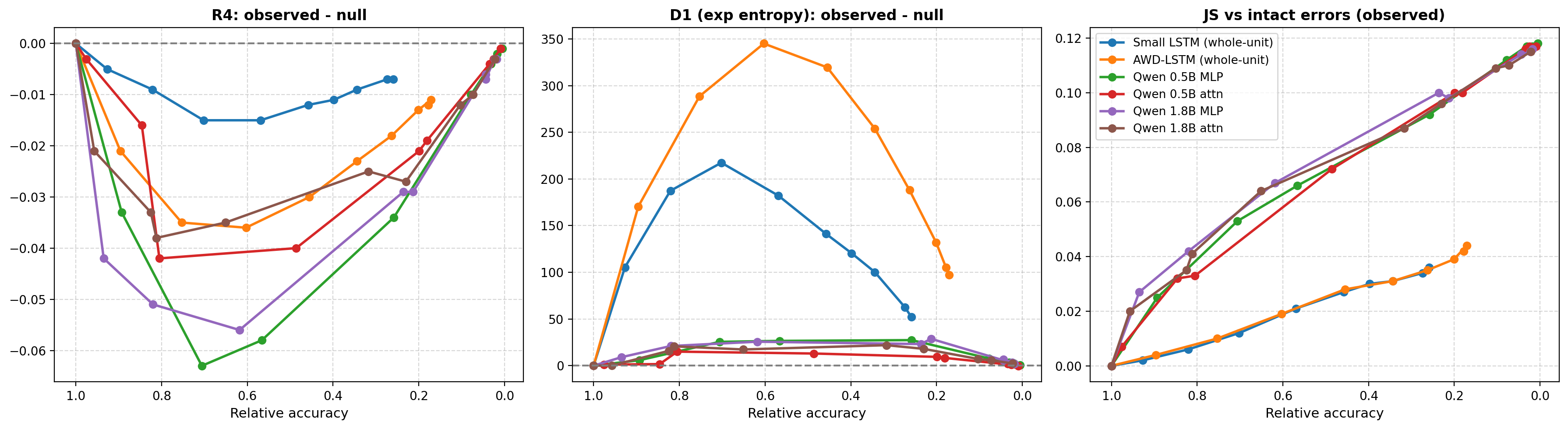}
\caption{Comparison of ablated error distributions against an accuracy-matched random-failure baseline (Equations~\ref{eq:null_model}--\ref{eq:js_div}). In all three panels, the horizontal axis is surviving relative accuracy $\mathrm{Acc}(k)/\mathrm{Acc}(0)$ (from $1.0$ at $k = 0$ on the left to $0.0$ on the right), comparing models at equal accuracy loss. The vertical axes plot: (Left)~$\Delta R_4 = R_4^{\mathrm{obs}} - R_4^{\mathrm{null}}$, the difference in moment ratio relative to random failure; (Centre)~$\Delta D_1 = D_1^{\mathrm{obs}} - D_1^{\mathrm{null}}$, the difference in effective number of distinct error words relative to random failure; and (Right)~$\mathrm{JS}(p(k)\,\|\,p(0))$, the Jensen--Shannon divergence between the damaged and undamaged ($k = 0$) error distributions. Together, the three panels establish two main findings: first, neuronal ablation fails non-randomly across all models, concentrating fewer new mistakes on ultra-common words ($\Delta R_4 < 0$) and spreading errors across more distinct words ($\Delta D_1 > 0$) than random failure; second, Transformer ablation (especially MLP removal) reshapes the error distribution much more strongly than LSTM whole-unit removal (Supplement~\ref{sec:supp_null_model}).}
\label{fig:null_model_comparison}
\end{figure}

Aggregate accuracy curves (Section~\ref{sec:accuracy}) track how many words a damaged model misses, but not which words fail, leaving open whether ablation degrades predictions uniformly across the vocabulary or selectively reshapes the error distribution.
To quantify this distinction, we introduce a vocabulary moment ratio ($R_4$) paired with an accuracy-matched random-failure null model that measures how ablation redistributes prediction errors across word-frequency ranks.
Using this framework, we show that neuronal ablation in every model redistributes errors non-randomly relative to random failure ($\Delta R_4 < 0$, $\Delta D_1 > 0$), with Transformer pathways, and particularly MLP ablation, producing larger null-corrected distributional shifts than LSTM whole-unit removal (Figures~\ref{fig:R4}--\ref{fig:null_model_comparison}).

Let $\mathbf{e}(k) = (e_0, e_1, \ldots, e_{C-1})$ record how many times the model fails to predict each word $i$ in a vocabulary of $C$ tokens at ablation fraction $k$, evaluated on the held-out WikiText-2 test set for the LSTMs and held-out TinyStories passages for the Qwen models (Methods, Section~\ref{sec:methods_metrics}). Here, $i$ denotes the vocabulary index, ordered by word frequency (most common $= 0$) for the LSTMs and by tokenizer merge order for Qwen, where common word pieces also have low indices.
We normalise $\mathbf{e}$ to a probability distribution $p_i = e_i / \sum_j e_j$ and define the $m$-th moment as:
\begin{equation}
    \mu_m(k) = \sum_{i=0}^{C-1} i^m \cdot p_i(k)\,.
\end{equation}
The key diagnostic is the \emph{moment ratio}:
\begin{equation}
    R_4(k) = \frac{\mu_4(k)^{1/4}}{\mu_2(k)^{1/2}}\,.
    \label{eq:R4}
\end{equation}
A single moment such as $\mu_1$ or $\mu_2^{1/2}$ only measures the average rank scale of errors, whereas the ratio $R_4 = \mu_4^{1/4}/\mu_2^{1/2}$ cancels out vocabulary scale to measure distribution shape (analogous to kurtosis, since $R_2 = \mu_2^{1/2}/\mu_2^{1/2} \equiv 1$ is trivial). Specifically, the denominator $\mu_2^{1/2}$ tracks the bulk error rank, while the $i^4$ weighting makes the numerator $\mu_4^{1/4}$ sensitive to the high-index rare-word tail.
When errors are spread uniformly across a band of words $[0, K]$, the bulk and tail scale together ($R_4 \approx 1.158$ for any $K$; $R_4 = 1$ on a single word). By contrast, when most errors pile up near $i \approx 0$ (common words) alongside a rare-word tail at large $i$, the mass near $i \approx 0$ pulls down $\mu_2^{1/2}$ while $\mu_4^{1/4}$ stays pinned by the tail, making $R_4$ large ($1.845$--$1.881$ for a Zipfian distribution~\citep{zipf1949human} $p_i \propto 1/(i+1)$).

\paragraph{Raw $R_4$ trajectories.}
Across progressive unit ablation, the raw ratio $R_4(k)/R_4(0)$ rises gradually and modestly in both LSTMs (by at most $5.1\%$ and $7.7\%$), indicating that error counts scale up nearly proportionally across the vocabulary. In the Transformers under MLP ablation, $R_4(k)/R_4(0)$ rises gradually at mild damage, jumps by about $5$ further percentage points across the threshold cliff, and reaches a maximum rise of $11.9$--$12.2\%$ at high $k$ (Figure~\ref{fig:R4}; Supplement~\ref{sec:supp_null_model}).
Because an undamaged model mostly gets common, low-index words right and misses harder words, any drop in overall accuracy adds new errors near $i \approx 0$, automatically pulling down $\mu_2^{1/2}$ relative to $\mu_4^{1/4}$ and pushing raw $R_4$ upward even under random failure.

\paragraph{Accuracy-matched random-failure baseline (null model) and distributional shift.}
{\emergencystretch=1.5em
To separate genuine pathway-specific changes from the automatic effect of making more errors, we compare each damaged model against a simulated \emph{random-failure baseline} (null model) matched to the exact same error count.
Let $\mathbf{e}(0)$ and $\mathbf{e}(k)$ denote the per-token error count vectors of the undamaged ($k = 0$) and damaged ($k > 0$) models across the test set.
Their sums, $M(0) = \sum_i e_i(0)$ and $M(k) = \sum_i e_i(k)$ (with $M(k) \geq M(0)$), give the total number of prediction errors made by the undamaged and damaged models, so $M(k) - M(0)$ is the number of additional errors caused by ablation.
Let $\mathbf{c}(0) = (c_0(0), \ldots, c_{C-1}(0))$ be the per-token count vector of test-set tokens that the undamaged ($k = 0$) model predicted \emph{correctly}. Under the random-failure null hypothesis, the $M(k) - M(0)$ additional mistakes caused by ablation occur by randomly failing on $M(k) - M(0)$ of the test-token occurrences that the undamaged model got right. Specifically, in each simulation we pick $M(k) - M(0)$ individual word occurrences uniformly at random \emph{without replacement} from the pool of test words the undamaged model got right, turn those into errors, and tally how many times each vocabulary word $i$ was picked. Adding that vector of newly missed word counts, denoted $\mathrm{SampleWithoutReplacement}\!\bigl(M(k) - M(0),\; \mathbf{c}(0)\bigr)$, to $\mathbf{e}(0)$ gives the simulated error counts:\par}
\begin{equation}
    \mathbf{e}^{\mathrm{null}}(k) = \mathbf{e}(0) + \mathrm{SampleWithoutReplacement}\!\bigl(M(k) - M(0),\; \mathbf{c}(0)\bigr)\,.
    \label{eq:null_model}
\end{equation}
Alongside $R_4$, which depends on vocabulary rank order $i$, we also track how many distinct words the errors span regardless of rank order using Hill's diversity index~\citep{hill1973diversity, jost2006entropy}, $D_1(p) = \exp(H(p))$, where $H(p) = -\sum_{i:\,p_i>0} p_i \ln p_i$ is the Shannon entropy (in nats).\footnote{Because $H(p)$ uses natural logarithms, $D_1 = \exp(H(p))$ equals $K$ when errors are distributed uniformly across $K$ distinct tokens.}
From each simulated null error vector $\mathbf{e}^{\mathrm{null}}(k)$, we compute both $R_4^{\mathrm{null}}(k)$ (Equation~\ref{eq:R4}) and the effective error-vocabulary size $D_1^{\mathrm{null}}(k) = \exp\!\bigl(H(p^{\mathrm{null}}(k))\bigr)$ (with $p^{\mathrm{null}}_i(k) = e^{\mathrm{null}}_i(k)/M(k)$), and denote the corresponding observed quantities for $\mathbf{e}(k)$ by $R_4^{\mathrm{obs}}(k)$ and $D_1^{\mathrm{obs}}(k)$.
Repeating Equation~\ref{eq:null_model} for $B = 100$ independent random draws per trial and averaging $R_4^{\mathrm{null}}(k)$ and $D_1^{\mathrm{null}}(k)$ across those $B$ draws yields the null expectations $\mathbb{E}[R_4^{\mathrm{null}}(k)]$ and $\mathbb{E}[D_1^{\mathrm{null}}(k)]$.
Subtracting these expectations from the observed values defines the \emph{null-corrected $R_4$ difference} $\Delta R_4(k)$ and the \emph{null-corrected error-vocabulary difference} $\Delta D_1(k)$:
\begin{equation}
    \Delta R_4(k) = R_4^{\mathrm{obs}}(k) - \mathbb{E}\bigl[R_4^{\mathrm{null}}(k)\bigr]\,, \qquad
    \Delta D_1(k) = D_1^{\mathrm{obs}}(k) - \mathbb{E}\bigl[D_1^{\mathrm{null}}(k)\bigr]\,.
    \label{eq:delta_R4}
\end{equation}
To measure how much the damaged error distribution $p_i(k) = e_i(k)/M(k)$ differs overall from the undamaged error distribution $p_i(0) = e_i(0)/M(0)$ on a 0-to-1 scale, we compute the base-2 Jensen--Shannon divergence:
\begin{equation}
    \mathrm{JS}\bigl(p(k)\,\|\,p(0)\bigr) = \frac{1}{2}\,\mathrm{KL}\bigl(p(k)\,\|\,m\bigr) + \frac{1}{2}\,\mathrm{KL}\bigl(p(0)\,\|\,m\bigr)\,, \qquad m = \frac{1}{2}\bigl(p(k) + p(0)\bigr)\,,
    \label{eq:js_div}
\end{equation}
where $\mathrm{KL}(a\,\|\,b) = \sum_{i:\,a_i>0} a_i \log_2(a_i/b_i)$ is the base-2 Kullback--Leibler divergence between two probability distributions $a$ and $b$.

Across all six unit-level conditions, neuronal ablation fails non-randomly compared with the random-failure baseline ($\Delta R_4 < 0$, Figure~\ref{fig:null_model_comparison} left; $\Delta D_1 > 0$, centre; full tables and Fourier spectral analysis appear in Supplements~\ref{sec:supp_null_model}--\ref{sec:supp_spectral_details}). The reason is straightforward: in natural text, a handful of common words (such as \emph{the} and \emph{and}, near $i \approx 0$) make up a large share of all tokens, so randomly failing on previously correct predictions puts most new mistakes onto those very common words and drives $R_4^{\mathrm{null}}$ sharply upward. An ablated network, by contrast, still gets the most common function words right and instead makes its new mistakes on less common content words (ranks $i$ in the hundreds to thousands). As a result, its errors sit at higher word ranks ($\Delta R_4 < 0$) and are spread across more distinct words ($\Delta D_1 > 0$) than under random failure. In the LSTMs and Qwen MLP, $\Delta R_4$ is already negative at $10\%$ damage ($k = 0.1$), showing that non-random error shifts appear even at mild damage before overall accuracy collapses.

Figure~\ref{fig:null_model_comparison} also reveals differences across pathways and architectures. Within the Transformers, MLP ablation shifts $\Delta R_4 / R_4(0)$ more negatively than attention-head ablation when trials are pooled across $k \in [0.1, 0.4]$ (permutation $p < 10^{-4}$ in Qwen~0.5B and $p = 0.0238$ in Qwen~1.8B), though at a single matched accuracy level ($\mathrm{RelAcc} = 0.80$) their confidence intervals overlap (Supplement~\ref{sec:supp_null_model}). This pattern is consistent with the hypothesis that feedforward layers store word-specific associations~\citep{geva2021transformer} whereas attention heads route broader context. Across architectures at equal surviving accuracy ($\mathrm{RelAcc} = 0.80$), the overall shift from the undamaged error distribution ($\mathrm{JS}(p(k)\,\|\,p(0))$, Figure~\ref{fig:null_model_comparison} right) is larger in the four Qwen curves than in the two LSTM whole-unit curves, and this difference remains after subtracting the random-failure baseline ($\Delta\mathrm{JS} = \mathrm{JS}^{\mathrm{obs}} - \mathrm{JS}^{\mathrm{null}}$ of $+0.010$ to $+0.020$ in Qwen vs.\ $+0.002$ to $+0.003$ in the LSTMs; Supplement~\ref{sec:supp_null_model}). Together, these results show that pairing the moment ratio and distributional metrics ($\Delta R_4$, $\Delta D_1$, and $\Delta\mathrm{JS}$) with an accuracy-matched random-failure null model separates non-random error concentration from simple accuracy loss, distinguishing targeted Transformer pathway damage from more uniform LSTM decline when accuracy scores alone cannot.

\subsection{Quantified Cloze Test: Dissociating Syntax from Semantics}
\label{sec:cloze}

To test whether grammatical structure and word meaning degrade together or separate by pathway, as in clinical aphasia~\citep{goodglass1993understanding}, we evaluate all models on a 100-prompt fill-in-the-blank sentence-completion (Cloze) battery (where each model predicts the missing final word of a sentence), stratified across nouns, verbs, adjectives, and adverbs. We measure \emph{syntactic accuracy} by part-of-speech (POS) match rate and \emph{semantic relevance} by GloVe~\citep{pennington2014glove} word-embedding cosine similarity to the target word. Because cosine similarity is set to zero whenever the predicted word is not tagged as a content word (Methods, Section~\ref{sec:methods_metrics}), the two metrics are not fully independent.

\begin{figure}[!htbp]
\centering
\includegraphics[width=\textwidth,height=0.50\textheight,keepaspectratio]{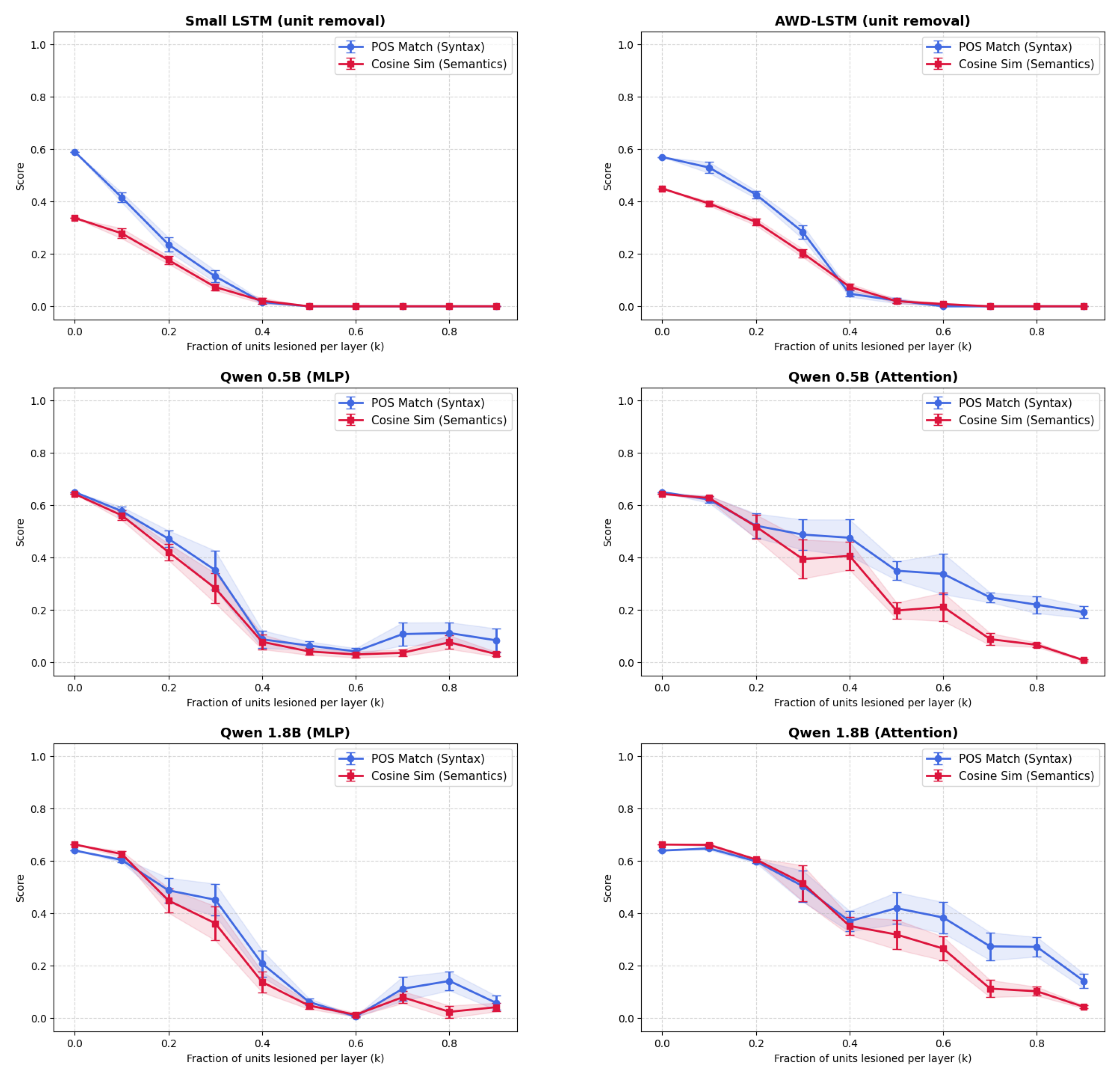}
\caption{Syntactic and semantic performance on the 100-prompt fill-in-the-blank (Cloze) test under progressive unit-level ablation. Each panel plots grammatical part-of-speech match rate (blue circles, syntax) and GloVe cosine similarity to the target word (red squares, semantics) on the vertical axis against the fraction of units removed $k \in [0.0, 0.9]$ on the horizontal axis (means $\pm 1$ SE across trials; $k = 0.0$ is the undamaged model). Row 1: Under whole-unit removal in both the Small LSTM (left) and pretrained AWD-LSTM (right), syntax and semantics erode in parallel from $k = 0.1$ onward, reaching near zero at $k = 0.4$ and $0.0\%$ POS by $k = 0.5$--$0.6$ as outputs default to common function words. Rows 2--3: In both Qwen~0.5B and Qwen~1.8B, MLP neuron ablation (left) declines moderately through $k = 0.3$ before a steep threshold collapse across $k^* \approx 0.3$--$0.4$ (near zero by $k = 0.5$--$0.6$), whereas attention-head ablation (right) declines much more gradually on both metrics through $k = 0.5$--$0.6$.}
\label{fig:cloze}
\end{figure}

Across architectures, the unit-level Cloze trajectories separate pathway-selective Transformer breakdown from non-dissociated LSTM decline (Figure~\ref{fig:cloze} and Table~\ref{tab:cloze_summary}, with full per-$k$ curves in Supplement~\ref{sec:supp_cloze_summary} and Supplementary Table~\ref{stab:cloze_values}).
In both Qwen Transformers (Figure~\ref{fig:cloze}, rows 2--3), MLP neuron ablation retains moderate performance at mild damage before collapsing both syntax and semantics across $k^* \approx 0.3$--$0.4$ to near zero by $k = 0.5$. Attention-head ablation declines much more gradually on the nominal-$k$ axis, retaining roughly $4.7$- to $7$-fold higher grammatical and semantic competence at nominal $k = 0.5$ (Supplement~\ref{sec:supp_cloze_summary}).
In both LSTMs under whole-unit removal (Figure~\ref{fig:cloze}, row 1), syntax and semantics always erode in parallel without dissociating. Both Cloze metrics (POS match and cosine similarity) decline proportionally across $k = 0.1$--$0.3$ and reach the zero floor on content words across $k = 0.4$--$0.6$, as cumulative unit loss across the serial 3-layer stack reduces predictions to common function words.

\begin{table}[!htbp]
\footnotesize
\centering
\caption{Cloze syntactic (POS match) and semantic (GloVe cosine) benchmarks and representative errors across the six unit-level conditions, grouped by clinical parallel. Transformer MLP and attention pathways diverge sharply at severe damage ($k = 0.5$), whereas both LSTMs erode syntax and semantics in lockstep until reaching the content-word floor across $k = 0.4$--$0.6$ (means across trials; full trajectories appear in Supplement~\ref{sec:supp_cloze_summary} and Supplementary Table~\ref{stab:cloze_values}).}
\label{tab:cloze_summary}
\resizebox{\textwidth}{!}{%
\begin{tabular}{@{}llccccl@{}}
\toprule
\textbf{Architecture} & \textbf{Target} & \textbf{$k$} & \textbf{POS Match} & \textbf{Cos Sim} & \textbf{Trajectory} & \textbf{Qualitative Error Profile and Examples ($k=0.2$)} \\
\midrule
\multicolumn{7}{l}{\textit{Baselines}} \\
Small LSTM & -- & 0.0 & 59.0\% & 0.337 & -- & grammatical continuations, no exact matches (4\% correct, all via top-5, e.g.\ \textit{neck}$\to$\textit{head}) \\
AWD-LSTM & -- & 0.0 & 57.0\% & 0.450 & -- & moderate accuracy (25\% correct; \textit{time} \checkmark, \textit{sky} \checkmark, \textit{song} \checkmark, \textit{ocean} \checkmark) \\
Qwen 0.5B & -- & 0.0 & 65.0\% & 0.643 & -- & strong (64\% correct; \textit{time} \checkmark, \textit{dress} \checkmark, \textit{hug} \checkmark) \\
Qwen 1.8B & -- & 0.0 & 64.0\% & 0.663 & -- & strong (66\% correct; \textit{time} \checkmark, \textit{sky} \checkmark, \textit{hug} \checkmark) \\
\midrule
\multicolumn{7}{l}{\textit{Wernicke-type parallel: related-word substitutions at the intact rate, more no-responses \& late threshold collapse}} \\
Qwen 1.8B & MLP & 0.2 & 48.8\% & 0.449 & threshold & \textit{ocean}$\to$\textit{bay}, \textit{sea}$\to$\textit{river} (same-category); \textit{cold}$\to$\textit{shallow} (attribute substitution) \\
Qwen 0.5B & MLP & 0.2 & 47.2\% & 0.421 & threshold & \textit{cheer}$\to$\textit{sing}, \textit{storm}$\to$\textit{dawn} (contextually plausible substitutions); \textit{stunning}$\to$\textit{fine} (generic evaluative default) \\
\midrule
\multicolumn{7}{l}{\textit{svPPA-type parallel: buffered Cloze trajectory \& non-significant trend toward generic defaulting}} \\
Qwen 1.8B & Attn & 0.2 & 59.9\% & 0.606 & sigmoid & \textit{gentle}$\to$\textit{small} (generic); \textit{keyboard}$\to$\textit{desktop}, \textit{pond}$\to$\textit{tank} (within-domain) \\
Qwen 0.5B & Attn & 0.2 & 52.2\% & 0.519 & sigmoid & \textit{ball}$\to$\textit{following}, \textit{clear}$\to$\textit{good}, \textit{heart}$\to$\textit{name} (concrete$\to$abstract or generic trend) \\
\midrule
\multicolumn{7}{l}{\textit{AD-type diffuse stage: steady parallel decline across mild-to-moderate whole-unit ablation}} \\
AWD-LSTM & Whole units & 0.2 & 42.7\% & 0.322 & gradual & parallel syntax/semantics decline (\textit{field}$\to$\textit{country}, \textit{storm}$\to$\textit{event}, \textit{ball}$\to$\textit{number}) \\
AWD-LSTM & Whole units & 0.3 & 28.4\% & 0.203 & gradual & continued parallel erosion without syntactic--semantic dissociation \\
Small LSTM & Whole units & 0.2 & 23.6\% & 0.177 & gradual & parallel decline toward high-frequency defaults (\textit{clouds}$\to$\textit{first}; one chance hit, \textit{field}, missed by the intact model) \\
Small LSTM & Whole units & 0.3 & 11.4\% & 0.073 & gradual & continued parallel decline of POS and cosine prior to serial input exhaustion \\
\midrule
\multicolumn{7}{l}{\textit{Late-stage / post-threshold comparison at nominal $k = 0.5$ ($70$--$73\%$ non-emb.\ params removed for LSTM units, $33$--$34\%$ for Qwen MLP, $16$--$17\%$ for Qwen Attn)}} \\
AWD-LSTM & Whole units & 0.5 & 2.2\% & 0.020 & floor & content-word floor at severe $k$ (POS 0.0\%, cos 0.008 at $k = 0.6$; 0.000 for $k \geq 0.7$) \\
Small LSTM & Whole units & 0.5 & 0.0\% & 0.000 & floor & content-word floor at severe $k$ (POS 0.0\%, cos 0.000 for $k \geq 0.5$) \\
Qwen 1.8B & MLP & 0.5 & 6.2\% & 0.048 & post-thresh.\ & post-threshold language collapse after MLP cliff \\
Qwen 0.5B & MLP & 0.5 & 6.4\% & 0.042 & post-thresh.\ & post-threshold language collapse after MLP cliff \\
Qwen 1.8B & Attn & 0.5 & 42.0\% & 0.319 & sigmoid & buffered mid-range survival (${\sim}7\times$ higher POS \& cosine than MLP at nominal $k{=}0.5$) \\
Qwen 0.5B & Attn & 0.5 & 35.0\% & 0.198 & sigmoid & buffered syntactic survival (${>}5\times$ higher POS \& ${\sim}4.7\times$ higher cosine than MLP at nominal $k{=}0.5$) \\
\bottomrule
\end{tabular}%
}
\end{table}

\subsection{Clinical Parallel Assignments: Converging Evidence}
\label{sec:clinical_evidence}

We test three hypotheses linking network architecture to clinical language disorders: (1)~in Transformers, damaging the feedforward (MLP) layers disrupts word retrieval while leaving related-word mistakes intact, resembling \emph{Wernicke-type aphasia}; (2)~damaging the attention heads preserves sentence structure across moderate damage while shifting wrong words toward more general terms, resembling \emph{semantic dementia (svPPA)}; and (3)~in serial LSTMs, which lack separate pathways or skip connections, removing whole units erodes grammar and word meaning together, resembling \emph{diffuse Alzheimer's-type (AD) decline}.

\begin{figure}[!tbp]
\centering
\includegraphics[width=0.98\textwidth,keepaspectratio]{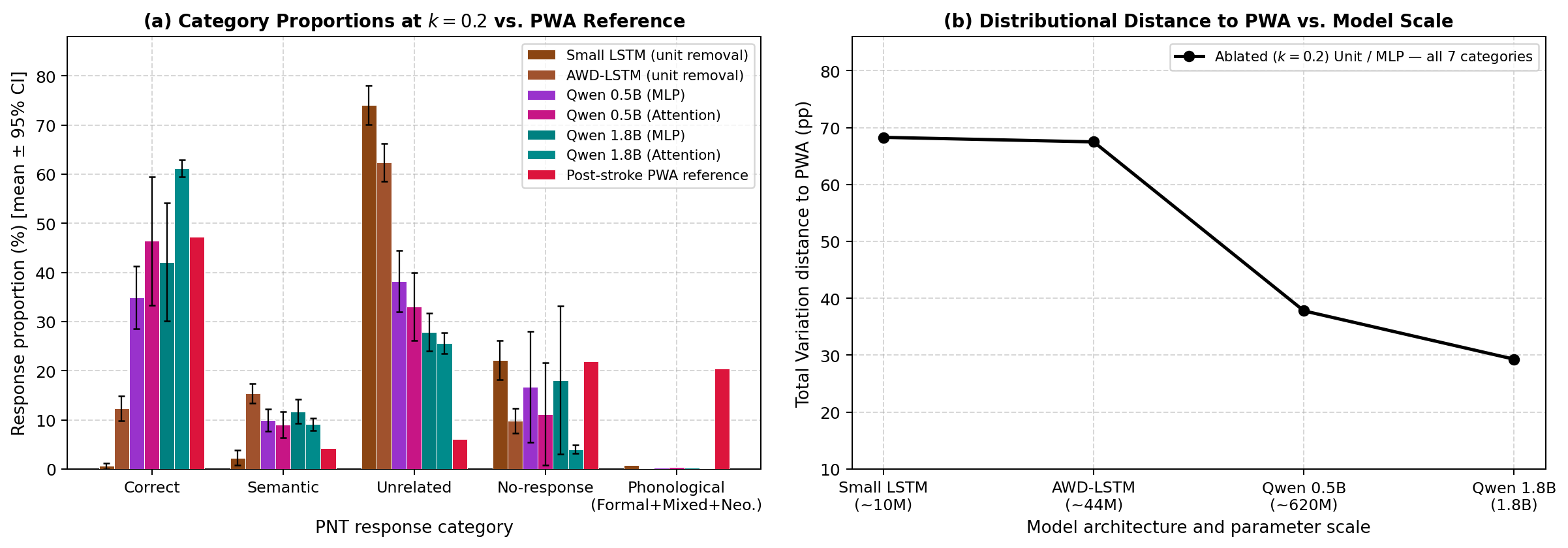}
\caption{Comparison of undamaged ($k = 0$) and unit-ablated ($k = 0.2$) model predictions with 278 post-stroke aphasia patients (PWA)~\citep{yang2026lesioned} on the Philadelphia Naming Test (PNT) taxonomy~\citep{roach1996philadelphia, mirman2010moss} (full seven-category numerical values appear in Supplementary Table~\ref{stab:pnt_main}, Supplement~\ref{sec:pnt_recoding}). (a)~Category proportions at $k = 0.2$ (mean $\pm$ 95\% CI across 10 independent ablation trials, $100$ prompts per trial) ordered by model scale against the post-stroke PWA reference, with the three sound-based categories (Formal, Mixed, and Neologism) grouped as Phonological. Qwen~1.8B MLP ablation preserves its undamaged rate of semantic paraphasias while No-response failures rise into the patient range, whereas LSTMs remain dominated by unrelated defaults. (b)~Overall difference (Total Variation distance across all seven PNT categories, in percentage points) between the ablated ($k = 0.2$) unit/MLP model predictions and the post-stroke patient percentages as model size increases.}
\label{fig:pnt_main}
\end{figure}

The four paragraphs below test these hypotheses by combining the damage curves from Sections~\ref{sec:accuracy}--\ref{sec:cloze} with the specific word mistakes each model makes at $20\%$ ablation (Tables~\ref{tab:summary}--\ref{tab:error_dist}, building on the Cloze benchmarks in Table~\ref{tab:cloze_summary}). The first two paragraphs examine the two Transformer pathways in turn, comparing MLP damage against Wernicke-type aphasia (Table~\ref{tab:summary}, rows 1--2; Table~\ref{tab:error_dist}) and attention-head damage against svPPA (rows 3--4). The third paragraph contrasts both Transformer pathways with whole-unit removal in the two LSTMs, which parallels diffuse AD-type decline (rows 5--6). The fourth paragraph compares the mistake patterns of all undamaged and damaged models directly against clinical naming data from stroke patients with aphasia (Figure~\ref{fig:pnt_main}; Supplementary Table~\ref{stab:pnt_main}).

\begin{table}[!htbp]
\footnotesize
\centering
\caption{Clinical parallel synthesis across the six unit-level ablation conditions. Each condition is classified along two converging axes: quantitative degradation dynamics ($k \in [0, 0.9]$) and qualitative Cloze error profiles at $k = 0.2$ (detailed error-type breakdowns appear in Table~\ref{tab:error_dist}, Figure~\ref{fig:pnt_main}, and Supplements~\ref{sec:supp_clinical_stats} and~\ref{sec:pnt_recoding}).}
\label{tab:summary}
\resizebox{\textwidth}{!}{%
\begin{tabular}{@{}llp{5.0cm}p{5.2cm}p{2.9cm}@{}}
\toprule
Architecture & Target & Quantitative Signature & Qualitative \& Lexical Evidence & Clinical Parallel \\
\midrule
Qwen 1.8B & MLP & Initial accuracy decline comparable to LSTMs ($93.5\%$ and $82.0\%$ of baseline at $k{=}0.1$, $0.2$), followed by late threshold collapse ($k^* {\approx} 0.3$--$0.4$); largest negative shift in null-corrected $R_4$ & Preserves the intact PNT Semantic rate ($11.7\%$ vs.\ $11.0\%$ intact) while No-response rises ($18.1\%$ vs.\ $1.0\%$ intact); produces within-category substitutions (``ocean''$\to$``bay'', ``sea''$\to$``river'') and attribute substitutions (``cold''$\to$``shallow'') with unchanged word concreteness & Wernicke-type \\
Qwen 0.5B & MLP & Same pattern: initial decline comparable to LSTMs ($89.3\%$ and $70.6\%$ at $k{=}0.1$, $0.2$) and late threshold collapse ($k^* {\approx} 0.3$--$0.4$); strong negative shift in null-corrected $R_4$ & Preserves the intact PNT Semantic rate ($9.9\%$ vs.\ $9.0\%$ intact) while No-response rises ($16.7\%$ vs.\ $1.0\%$ intact); produces contextually plausible substitutions (``cheer''$\to$``sing'', ``storm''$\to$``dawn'') and generic evaluative defaults (``stunning''$\to$``fine'') & Wernicke-type \\
\midrule
Qwen 1.8B & Attn & Early retention at $k{=}0.1$ followed by a smoother sigmoid-like decline; preserves substantial Cloze syntax and semantics through $k{=}0.5$; milder null-corrected $R_4$ shift than MLP & Slightly fewer semantic paraphasias than intact ($9.1\%$ vs.\ $11.0\%$) and a borderline shift toward more abstract words ($\Delta\text{Conc.}{=}{+}0.24$ vs.\ ${+}0.13$ intact): generic substitutions (``gentle''$\to$``small'') alongside within-domain ones (``keyboard''$\to$``desktop'') & svPPA \\
Qwen 0.5B & Attn & Early retention at $k{=}0.1$ followed by a smoother sigmoid-like decline; Cloze syntax and semantics buffered through $k{=}0.4$--$0.5$; milder null-corrected $R_4$ shift than MLP & Shifts substitutions toward more abstract words than the intact model ($\Delta\text{Conc.}{=}{+}0.43$ vs.\ ${+}0.15$ intact), with generic defaults such as ``ball''$\to$``following'' and ``clear''$\to$``good'' & svPPA \\
\midrule
AWD-LSTM & Whole units & Steady, gradual accuracy decline across $k \in [0, 0.9]$ without a late cliff ($75.3\%$ of baseline at $k{=}0.2$, $34.4\%$ at $k{=}0.5$); modest gradual $R_4$ rise ($\le 7.7\%$; $\mathrm{JS}{=}0.008$ at $\mathrm{RelAcc}{=}0.80$) & Syntax (POS) and semantics (cosine) erode in lockstep across $k{=}0.1$--$0.3$ without dissociating ($46.6\%$ content sub.\ at $k{=}0.2$), reaching the content-word Cloze floor across $k{=}0.4$--$0.6$ ($0.0\%$ POS by $k{=}0.6$) & Diffuse AD-type \\
Small LSTM & Whole units & Steady, gradual accuracy decline across $k \in [0, 0.9]$ without a late cliff ($82.2\%$ of baseline at $k{=}0.2$, $45.8\%$ at $k{=}0.5$); modest gradual $R_4$ rise ($\le 5.1\%$; $\mathrm{JS}{=}0.007$ at $\mathrm{RelAcc}{=}0.80$) & Syntax and semantics erode in parallel across $k{=}0.1$--$0.3$ without dissociating, progressing rapidly to non-content defaults ($84.3\%$ non-content at $k{=}0.2$) and reaching the content-word Cloze floor across $k{=}0.4$--$0.5$ ($0.0\%$ POS by $k{=}0.5$) & Diffuse AD-type \\
\bottomrule
\end{tabular}%
}
\end{table}

\paragraph{MLP ablation $\leftrightarrow$ Wernicke-type aphasia.}
Under MLP neuron ablation, both Qwen Transformers lose exact word-prediction accuracy and eventually collapse across $k^* \approx 0.3\text{--}0.4$, yet at moderate damage ($k = 0.2$) their wrong content-word guesses remain similar in specificity and concreteness to those of the undamaged model and preserve the undamaged rate of meaning-related word substitutions (semantic paraphasias: $11.7\%$ vs.\ $11.0\%$ undamaged in Qwen~1.8B).\footnote{Word specificity is measured using the human concreteness ratings of~\citet{brysbaert2014concreteness} ($1 = \text{abstract}$ to $5 = \text{concrete}$), where MLP-ablated substitutions show no detectable drop in concreteness relative to the undamaged model. Meaning-related errors are scored under the Philadelphia Naming Test (PNT) taxonomy: MLP ablation preserves the undamaged semantic paraphasia rate ($11.7\%$ vs.\ $11.0\%$ in Qwen~1.8B), which exceeds attention-head ablation ($9.1\%$) because attention ablation slightly reduces semantic paraphasias. Examples cited in the text are absent from the undamaged model's top-5 predictions; attribute changes such as \emph{``shallow''} for \emph{``cold''} are scored as unrelated rather than same-category semantic paraphasias under PNT rules (Supplements~\ref{sec:supp_clinical_stats}--\ref{sec:pnt_recoding}).}
For example, the MLP-damaged model replaces target words with other specific words from the same category (such as \emph{``bay''} for \emph{``ocean''} and \emph{``river''} for \emph{``sea''}) or with related attributes (such as \emph{``shallow''} for \emph{``cold''}).
This error pattern resembles \emph{Wernicke-type (receptive) aphasia} after posterior temporal damage~\citep{wernicke1874aphasische, goodglass1993understanding}: in both cases, retrieving the exact target word fails, yet substitutions remain concrete, meaning-related words without a selective loss of grammar over meaning.
The parallel is partial because clinical Wernicke's aphasia is fluent and elevates semantic paraphasias, whereas MLP ablation keeps semantic paraphasias at the undamaged rate while increasing blank and function-word outputs and lowering grammatical match.\footnote{In Table~\ref{tab:cloze_summary}, MLP ablation at $k = 0.2$ lowers Cloze grammatical POS match ($64.0\% \to 48.8\%$ in Qwen~1.8B; $65.0\% \to 47.2\%$ in Qwen~0.5B) and semantic cosine similarity ($0.663 \to 0.449$; $0.643 \to 0.421$) in parallel before collapsing at $k = 0.5$ ($6.2\%\text{--}6.4\%$ POS), while non-content outputs rise from $7.0\%\text{--}9.0\%$ undamaged to $30.7\%\text{--}34.8\%$ at $k = 0.2$ (Table~\ref{tab:error_dist}).}
Overall, this pattern is consistent with Transformer MLP layers storing specific word meanings (Table~\ref{tab:summary}, rows 1--2; Tables~\ref{tab:cloze_summary} and~\ref{tab:error_dist}; Figures~\ref{fig:accuracy}c,e, \ref{fig:null_model_comparison}--\ref{fig:pnt_main}; Supplement~\ref{sec:supp_clinical_stats} and Supplementary Tables~\ref{stab:error_comparison} and~\ref{stab:pnt_main}).

\begin{table}[!htbp]
\footnotesize
\centering
\caption{Cloze error-type distribution and Brysbaert concreteness drop ($\Delta\text{Conc.}$, target minus prediction~\citep{brysbaert2014concreteness}; positive values indicate more abstract substitutions) for undamaged ($k = 0$) and unit-ablated ($k = 0.2$) models. MLP ablation lowers accuracy and raises non-content outputs while keeping substitution concreteness near the undamaged baseline, whereas attention ablation shifts substitutions toward more abstract words. For the ablated models, values show the mean and 95\% confidence interval across 10 independent ablation trials (coding and exact-match details appear in Supplements~\ref{sec:supp_clinical_stats}--\ref{sec:concreteness_method}).}
\label{tab:error_dist}
\resizebox{\textwidth}{!}{%
\begin{tabular}{@{}llcccc@{}}
\toprule
Architecture & Condition & Correct & Content Substitution & Non-content & Mean~$\Delta$Conc. \\
\midrule
\multicolumn{6}{@{}l}{\textit{Undamaged baselines ($k = 0$, $n = 100$ prompts)}} \\
Qwen 1.8B & Intact ($k{=}0$) & 66.0\% & 27.0\% & 7.0\% & $+0.13$ \\
Qwen 0.5B & Intact ($k{=}0$) & 64.0\% & 27.0\% & 9.0\% & $+0.15$ \\
AWD-LSTM & Intact ($k{=}0$) & 25.0\% & 53.0\% & 22.0\% & $+0.26$ \\
Small LSTM & Intact ($k{=}0$) & 4.0\% & 45.0\% & 51.0\% & $+0.53$ \\
\midrule
\multicolumn{6}{@{}l}{\textit{Unit-ablated conditions at $k = 0.2$ (mean [95\% CI] across 10 independent ablation trials, $100$ prompts per trial)}} \\
Qwen 1.8B & MLP ($k{=}0.2$) & 42.1\% [30.1, 54.1] & 27.2\% [21.7, 32.7] & 30.7\% [14.2, 47.2] & $+0.19$ [$+0.07$, $+0.30$] \\
Qwen 0.5B & MLP ($k{=}0.2$) & 34.9\% [28.6, 41.2] & 30.3\% [24.8, 35.8] & 34.8\% [23.5, 46.1] & $+0.29$ [$+0.13$, $+0.45$] \\
\midrule
Qwen 1.8B & Attn ($k{=}0.2$) & 61.2\% [59.4, 63.0] & 27.4\% [26.1, 28.7] & 11.4\% [9.8, 13.0] & $+0.24$ [$+0.13$, $+0.34$] \\
Qwen 0.5B & Attn ($k{=}0.2$) & 46.4\% [33.3, 59.5] & 29.1\% [23.7, 34.5] & 24.5\% [7.0, 42.0] & $+0.43$ [$+0.33$, $+0.53$] \\
\midrule
AWD-LSTM & Whole units ($k{=}0.2$) & 12.3\% [9.8, 14.9] & 46.6\% [41.5, 51.7] & 41.1\% [36.3, 46.0] & $+0.34$ [$+0.21$, $+0.46$] \\
Small LSTM & Whole units ($k{=}0.2$) & 0.6\% [0.0, 1.2] & 15.1\% [11.3, 18.9] & 84.3\% [80.6, 88.0] & $+0.52$ [$+0.20$, $+0.85$] \\
\bottomrule
\end{tabular}%
}
\end{table}

\paragraph{Attention-head ablation $\leftrightarrow$ svPPA.}
Under attention-head ablation, both Qwen Transformers lose performance much more gradually than under MLP ablation, retaining substantial grammatical and meaning-related Cloze competence even when $40\%\text{--}50\%$ of attention heads are removed ($k = 0.4\text{--}0.5$).\footnote{In Table~\ref{tab:cloze_summary}, attention-head ablation preserves high Cloze grammatical POS match and semantic cosine similarity at $k = 0.2$ ($59.9\%$ and $0.606$ in Qwen~1.8B; $52.2\%$ and $0.519$ in Qwen~0.5B) and retains $42.0\%$ POS / $0.319$ cosine (Qwen~1.8B) and $35.0\%$ POS / $0.198$ cosine (Qwen~0.5B) at $k = 0.5$, noting that $k = 0.4\text{--}0.5$ removes $12\%\text{--}17\%$ of non-embedding weights because attention blocks contain half as many parameters as MLP blocks (Supplement~\ref{sec:supp_cloze_summary}).}
Yet when the attention-damaged model makes content-word mistakes at moderate damage ($k = 0.2$), its guesses shift toward more general, abstract words than those of the undamaged model.\footnote{On the Brysbaert concreteness scale (Table~\ref{tab:error_dist}), the shift toward more abstract substitutions ($\Delta\text{Conc.}$) at $k = 0.2$ is statistically significant relative to the undamaged model in Qwen~0.5B ($+0.43$ vs.\ $+0.15$ undamaged) and borderline in Qwen~1.8B ($+0.24$ vs.\ $+0.13$ undamaged; Supplement~\ref{sec:supp_clinical_stats}).}
For example, the attention-damaged model replaces specific words with broader or more generic terms (such as \emph{``small''} for \emph{``gentle''} in Qwen~1.8B and \emph{``good''} for \emph{``clear''} in Qwen~0.5B), alongside some same-domain substitutions (such as \emph{``desktop''} for \emph{``keyboard''} in Qwen~1.8B).
This combination of preserved sentence grammar and a drift from specific toward more general words resembles the \emph{semantic variant of primary progressive aphasia (svPPA, or semantic dementia)}, in which anterior temporal atrophy erodes specific word knowledge into generic or superordinate terms while sparing syntax~\citep{warrington1975selective, hodges2007semantic, gornotempini2011classification, lambonralph2017semantic}.
The parallel is partial because clinical svPPA involves a progressive, category-by-category loss of conceptual knowledge, and while attention and MLP ablation separate sharply in their overall Cloze curves across $k = 0.3\text{--}0.5$, their difference in individual error types at $k = 0.2$ is a non-significant directional trend (see also Section~\ref{sec:discussion}, Limitations; Table~\ref{tab:summary}, rows 3--4; Tables~\ref{tab:cloze_summary} and~\ref{tab:error_dist}; Figure~\ref{fig:cloze}, rows 2--3; Supplement~\ref{sec:supp_clinical_stats}).

\paragraph{LSTM whole-unit ablation $\leftrightarrow$ Diffuse AD-type decline.}
Silencing whole hidden units in either the Small LSTM or the pretrained AWD-LSTM produces a smooth, steady decline in next-word accuracy across $k \in [0, 0.9]$ (Figures~\ref{fig:accuracy}a,b).
Unlike the Transformers, the damaged LSTMs keep their mistakes spread across common and rare words in roughly the same proportions as the undamaged model (Figures~\ref{fig:R4}a,b and~\ref{fig:null_model_comparison}), and their grammatical and meaning-related Cloze scores decline together at every damage level without separating (Figure~\ref{fig:cloze}, row 1).\footnote{In Table~\ref{tab:cloze_summary} and Supplementary Table~\ref{stab:cloze_values}, whole-unit ablation lowers Cloze grammatical POS match and semantic cosine similarity in parallel across mild-to-moderate damage. In the AWD-LSTM, POS match and cosine similarity drop from $57.0\%$ and $0.450$ undamaged to $42.7\%$ and $0.322$ at $k = 0.2$, $28.4\%$ and $0.203$ at $k = 0.3$, and $2.2\%$ and $0.020$ at $k = 0.5$. In the Small LSTM, they drop from $59.0\%$ and $0.337$ undamaged to $23.6\%$ and $0.177$ at $k = 0.2$, $11.4\%$ and $0.073$ at $k = 0.3$, and $0.0\%$ and $0.000$ at $k = 0.5$.}
Across mild-to-moderate damage ($k \le 0.2\text{--}0.3$), this steady, across-the-board erosion of grammar and word meaning resembles the broad cognitive-linguistic decline of \emph{Alzheimer's-type dementia (AD)}~\citep{blennow2006alzheimers}, reflecting an architecture in which grammar and meaning share a single recurrent pathway.
At heavier damage ($k \ge 0.4$), because the 3-layer LSTM has no skip connections to bypass damaged layers, cumulative unit loss across the three layers reduces predictions to common function words (which still score hits on general text in Figures~\ref{fig:accuracy}a,b, but bring content-word Cloze scores to the zero floor by $k = 0.4\text{--}0.6$; Table~\ref{tab:summary}, rows 5--6; Tables~\ref{tab:cloze_summary} and~\ref{tab:error_dist}).

\paragraph{Comparison of intact and ablated models with post-stroke patient error distributions.}
Finally, we compare the mistake patterns of all four undamaged ($k = 0$) and moderately damaged ($k = 0.2$) models on the 100-prompt Cloze test against clinical naming data from 278 post-stroke aphasia patients~\citep{yang2026lesioned} using the Philadelphia Naming Test (PNT) error categories~\citep{roach1996philadelphia, mirman2010moss} (Figure~\ref{fig:pnt_main}; Supplementary Table~\ref{stab:pnt_main}, Supplement~\ref{sec:pnt_recoding}).
This comparison yields two main findings.
First, a damaged model only resembles clinical aphasia if the undamaged model can already perform the task: at $20\%$ ablation, Qwen~1.8B matches the patient range in overall accuracy and blank responses while preserving its meaning-related mistakes, whereas the smaller LSTMs start well below the patient average and collapse into unrelated common words (Figure~\ref{fig:pnt_main}a).\footnote{In Figure~\ref{fig:pnt_main}a, $20\%$ unit ablation shifts Qwen~1.8B into an accuracy range ($42.1\%$ MLP, $61.2\%$ attention) bracketing the $47.3\%$ patient mean when Correct includes top-5 predictions (first-response exact match is lower; Supplementary Table~\ref{stab:pnt_main}), while MLP No-response outputs (punctuation, digits, and, in bilingual Qwen models, occasional Chinese or byte-fallback tokens) rise to $18.1\%$ versus $21.9\%$ in patients.}
Second, across all seven PNT categories, the overall gap between the ablated model and the post-stroke patient profile drops sharply as model scale increases from the LSTMs to Qwen~1.8B (Figure~\ref{fig:pnt_main}b).

\section{Discussion}
\label{sec:discussion}

\paragraph{Why Transformer pathways split: word storage versus context routing.}
One hypothesis for the divergence between MLP and attention-head damage comes from their proposed roles inside a Transformer block.
Prior interpretability studies suggest that feedforward MLP layers act as distributed memory stores for specific word associations~\citep{geva2021transformer}, whereas attention heads gather context from earlier words in the sentence~\citep{elhage2021mathematical, olsson2022context} along a shared skip-connection (residual) stream.
Under this hypothesis, damaging MLP neurons while leaving attention intact would allow sentence context to steer the model toward the right topic, while impairing its ability to retrieve the exact target word from memory.
This could explain why the MLP-damaged model continues to make same-category word substitutions at the undamaged rate before collapsing abruptly at $k^* \approx 0.3\text{--}0.4$, partially paralleling Wernicke-type word-selection errors (though without preserved fluency, as blank and function-word outputs also rise)~\citep{wernicke1874aphasische, goodglass1993understanding}.
Conversely, damaging attention heads while leaving the MLP store intact would allow skip connections to preserve basic grammar and vocabulary access through moderate damage, while weakened sentence context could cause the model to fall back onto safer, more general words, resembling the loss of word specificity in svPPA~\citep{warrington1975selective, hodges2007semantic, gornotempini2011classification, lambonralph2017semantic}.

\paragraph{Reserve, plasticity, and detecting mild damage before accuracy drops.}
These pathway-specific failure curves persist across all four post-damage fine-tuning regimes, indicating that architectural wiring, rather than short-term weight recovery, governs the shape of decline (Section~\ref{sec:methods_plasticity}; Supplement~\ref{sec:supp_plasticity}).
In clinical neurodegeneration such as Alzheimer's disease (AD), overall test scores often stay near normal during a pre-symptomatic reserve phase despite substantial early entorhinal pathology and neuron loss~\citep{braak1991staging, GomezIsla4491}, before entering a symptomatic phase in which cortical synapse loss tracks steady cognitive decline~\citep{dekosky1990, terry1991}.
Our experiments show an analogous split between buffered early accuracy (under Transformer attention-head removal on the unit-fraction axis) and steady accuracy loss (under LSTM unit removal).
Crucially, however, even at mild damage ($k = 0.1\text{--}0.2$), ablated models' new mistakes already shift toward less common content words rather than ultra-common function words relative to an accuracy-matched random-failure baseline ($\Delta R_4 < 0$, $\Delta D_1 > 0$; significant from $k = 0.1$ in both LSTMs and Qwen MLP and from $k = 0.2$ in Qwen~1.8B attention; Figure~\ref{fig:null_model_comparison}).
Because modern disease-modifying therapies for AD~\citep{vandyck2023lecanemab, sims2023donanemab} require intervention before tau-driven cortical atrophy becomes widespread~\citep{lajoie2020tau}, comparing patients' word-frequency error patterns on naming or Cloze tests against an accuracy-matched random baseline could help distinguish early structural decline from benign attentional lapses before overall test scores fall.

\paragraph{Why LSTMs decline steadily without a pathway split.}
In contrast to the Transformer, both the Small LSTM and the much larger AWD-LSTM pass information through a single chain of three recurrent layers without separate MLP and attention sub-blocks or skip connections.
Making the LSTM larger raises its undamaged accuracy ($27.5\%$ to $38.1\%$), but does not change its failure pattern: next-word accuracy declines smoothly across the entire ablation range (Figures~\ref{fig:accuracy}a,b), and on the Cloze test grammar and word meaning erode steadily together from $k = 0.1$ until reaching the zero floor on content words by $k = 0.4\text{--}0.6$, where the models fall back to common function words (Figure~\ref{fig:cloze}, row 1).
Our supplementary experiments that cut specific types of LSTM connections one at a time (Section~\ref{sec:methods_models}; Supplements~\ref{sec:supp_arch_summary}--\ref{sec:supp_clinical_stats}) explain both aspects of this profile: cutting only the memory links from earlier words causes a slow, steady co-decline of grammar and meaning across the full range, whereas cutting the links that feed the current word up the three-layer chain causes a rapid drop onto function-word defaults once enough serial capacity is lost.

\paragraph{Limitations.}
Three main limitations qualify our conclusions.
First, uniform binary neuron zeroing across all layers captures only one idealised mode of neural damage.
Real brain disease and stroke can damage circuits in many other ways, including gradual synapse weakening and noisy signalling before cell death~\citep{blennow2006alzheimers, yang2026lesioned}, damage concentrated in specific layers or regions rather than spread evenly across layers~\citep{braak1991staging, moore2025realistic, wang2026componentlevel}, broken connections between regions, and brain recovery mechanisms that differ from standard model fine-tuning~\citep{moore2023dementia, moore2023neuroplasticity}.
Accordingly, our clinical parallels describe output-level behavioural analogies rather than one-to-one mappings onto the neurocognitive components of controlled semantic cognition~\citep{jefferies2006semantic, lambonralph2017semantic, robson2012wernicke}.
Second, our empirical scope is bounded by model scale, training corpus, and prompt sample size. The LSTMs (${\sim}10$--$44$M parameters, trained on WikiText) and Transformers ($0.6$--$1.8$B parameters, evaluated on TinyStories) differ in size, corpus, and tokeniser (Supplement~\ref{sec:supp_arch_summary}). In addition, while our 100-prompt Cloze battery clearly separates the overall degradation curves of the MLP and attention pathways, larger prompt batteries will be needed to test fine-grained differences between pathways in individual error subtypes (Supplement~\ref{sec:supp_clinical_stats}).
Third, single-word Cloze completions over a fixed text vocabulary isolate word-choice errors (Semantic, Unrelated, and No-response) but cannot reproduce spoken-sound slips or made-up non-words (Formal, Mixed, and Neologism ${\leq}1\%$ vs.\ ${\approx}20.5\%$ in patients~\citep{roach1996philadelphia, mirman2010moss, yang2026lesioned}) because the models lack a speech-production stage; they also do not assess sentence-level fluency or pragmatics. Moreover, automated scoring depends on operational rules (top-5 accuracy matching, POS-gated GloVe similarity, and WordNet/CMUdict thresholds; Supplements~\ref{sec:concreteness_method}--\ref{sec:pnt_recoding}) on open-ended prompts that admit multiple valid continuations. Extending these analyses to larger models and connected patient speech from AphasiaBank~\citep{macwhinney2011aphasiabank} is an important direction for future work.

\section{Methods}
\label{sec:methods}

Full architectural specifications, weight-matrix indexing equations, hyperparameter schedules, and scoring algorithms are provided in Extended Materials and Methods (Supplement~\ref{sec:supp_methods}; schematic overview in Supplementary Figure~\ref{fig:methods_schematics}).

\paragraph{Model architectures and lesion definitions.}
\label{sec:methods_models}
We evaluate four autoregressive language models spanning two architecture families: a 3-layer Small LSTM (${\sim}10$M parameters) trained from scratch, a pretrained 3-layer AWD-LSTM (${\sim}44$M parameters)~\citep{merity2018regularizing, howard2018universal}, and two 24-layer pretrained Transformers, Qwen~1.5-0.5B (${\sim}620$M parameters) and Qwen~1.5-1.8B ($1.8$B parameters)~\citep{bai2023qwen}. Full layer dimensions, non-embedding parameter counts, and pretraining details are listed in Table~\ref{tab:architectures} and Supplement~\ref{sec:supp_methods}.
In our six primary unit-level conditions, ablating a fraction $k \in \{0.0, 0.1, \ldots, 0.9\}$ per layer completely silences $\lfloor k \times n \rfloor$ randomly selected units in every layer (Supplementary Figure~\ref{fig:methods_schematics}).
First, \emph{LSTM whole-unit removal} zeroes all incoming and outgoing weights and gate biases of the selected hidden units across both the input-to-hidden and recurrent hidden-to-hidden pathways (removing $70.4\%$ and $72.7\%$ of non-embedding parameters at $k = 0.5$ in the Small LSTM and AWD-LSTM).
Second, \emph{Transformer MLP neuron removal} zeroes the input and output projections of the selected intermediate feedforward neurons ($33.7\%$ and $33.4\%$ of non-embedding parameters at $k = 0.5$).
Third, \emph{Transformer attention-head removal} zeroes the query, key, value, and output projections (and biases) of the selected attention heads ($16.3\%$ and $16.6\%$ of non-embedding parameters at $k = 0.5$, in discrete steps of $1/16$ corresponding to the 16 heads per layer).
To separate the two incoming LSTM pathways, we also run two connection-level controls across both LSTMs: \emph{recurrent-connection pruning}, which zeroes only the hidden-to-hidden weights and gate biases of selected units ($37.4\%$ and $41.7\%$ of non-embedding parameters at $k = 0.5$), and \emph{input-connection pruning}, which zeroes only their input-to-hidden weights and gate biases ($25.0\%$ and $22.2\%$ at $k = 0.5$).

\paragraph{Ablation, replication, and fine-tuning protocol.}
\label{sec:methods_plasticity}
We run $5$ independent random-seed trials per $k \in \{0.0, \allowbreak 0.1, \allowbreak \ldots, \allowbreak 0.9\}$ for all unit- and connection-level conditions, increasing replication to $15$ trials per $k$ across $k \in \{0.1, \allowbreak 0.2, \allowbreak 0.3, \allowbreak 0.4\}$ ($n = 60$ trials per pathway) for the Qwen null-model tests (Section~\ref{sec:moments}) and $10$ trials at $k = 0.2$ ($N = 1{,}000$ prompt evaluations per condition) for the Cloze, concreteness, and PNT analyses (Section~\ref{sec:clinical_evidence}).
In each trial, ablated weights and biases are zeroed and kept frozen at zero during a brief stochastic gradient descent (SGD) pass ($\mathrm{lr} = 10^{-3}$, gradient clipping $0.25$) that models limited local recovery ($40$ and $50$ sequence steps for the Small LSTM and AWD-LSTM; $62$ weight updates across $500$ TinyStories passages for Qwen).
Sensitivity controls comparing zero fine-tuning, brief local SGD, and 500-step AdamW retraining are reported in Supplement~\ref{sec:supp_plasticity}.

\paragraph{Evaluation corpora and metrics.}
\label{sec:methods_metrics}
Next-token accuracy and token-level error distributions $\mathbf{e}(k)$ are evaluated on held-out WikiText-2~\citep{merity2017pointer} for the LSTMs ($241{,}200$ lexical word tokens, of which ${\sim}225{,}700$ in-vocabulary tokens are scored after excluding \texttt{xxunk}). For Qwen, they are evaluated on $500$ held-out TinyStories passages~\citep{eldan2023tinystories} truncated to $32$ tokens (up to $31$ next-token predictions per passage), and all curves are reported as relative accuracy $\mathrm{RelAcc}(k) = \mathrm{Acc}(k)/\mathrm{Acc}(0)$. To test whether degradation curves bend non-linearly, we compare 3-parameter logistic and 2-parameter linear fits using $\Delta\mathrm{AIC}$ against both $k$ and the fraction of non-embedding parameters removed (Supplementary Table~\ref{stab:curve_shape}).
From $\mathbf{e}(k)$, we compute the fourth-to-second moment ratio $R_4(k)$ over native vocabulary indices $i$ (ordered by word frequency for the LSTMs and BPE merge order for Qwen; Equation~\ref{eq:R4}), alongside the order-invariant effective error vocabulary $D_1(k) = \exp(H(p(k)))$ and base-2 Jensen--Shannon divergence $\mathrm{JS}(p(k)\,\|\,p(0))$ (Equation~\ref{eq:js_div}). Each metric is compared against $B = 100$ accuracy-matched random-failure draws sampled without replacement from the undamaged model's correct predictions $\mathbf{c}(0)$ (Equations~\ref{eq:null_model}--\ref{eq:delta_R4}; auxiliary Fourier spectral slopes $\beta$ and gaps $G$, computed after explicitly reordering the top $1{,}000$ tokens by corpus frequency across all models, appear in Supplement~\ref{sec:supp_spectral_details}).
Syntactic and semantic degradation are evaluated on a shared 100-prompt Cloze battery (48 nouns, 25 verbs, 22 adjectives, 5 adverbs) using part-of-speech (POS) match on the top-1 predicted word (tagged in isolation via NLTK~\citep{bird2009nltk}) and 50-dimensional GloVe~\citep{pennington2014glove} cosine similarity (set to zero for non-content predictions). Finally, we compute word-concreteness shifts $\Delta\text{Conc.}$ from the Brysbaert norms~\citep{brysbaert2014concreteness} and classify responses into the seven-category PNT error taxonomy~\citep{mirman2010moss, fellbaum1998wordnet}, evaluating pathway differences via two-sided permutation tests ($20{,}000$ permutations with Holm correction; Supplements~\ref{sec:supp_methods} and~\ref{sec:concreteness_method}--\ref{sec:pnt_recoding}).

\paragraph{AI usage.}
AI was used for code generation and polishing. The initial code for one of the models was written by P.W. and then AI was used to replicate it for other models and write code for generating the figures. AI was also used to draft and polish the document initially, including extracting figures and tabular data from the notebook output. P.W. wrote the final draft of the paper and also verified the code. Finally, we note that the paper does not provide medical advice.


\setlength{\bibsep}{2.2pt plus 0.3ex}
\bibliographystyle{plainnat}
\bibliography{references}

@article{broca1861remarques,
  title={Remarques sur le si\`{e}ge de la facult\'{e} du langage articul\'{e}},
  author={Broca, Paul},
  journal={Bulletins de la Soci\'{e}t\'{e} Anatomique de Paris},
  volume={36},
  pages={330--357},
  year={1861}
}

@article{frankle2019lottery,
  title={The Lottery Ticket Hypothesis: Finding Sparse, Trainable Neural Networks},
  author={Frankle, Jonathan and Carbin, Michael},
  journal={Proceedings of the International Conference on Learning Representations (ICLR)},
  year={2019}
}

@inproceedings{han2015learning,
  title={Learning Both Weights and Connections for Efficient Neural Networks},
  author={Han, Song and Pool, Jeff and Tran, John and Dally, William J.},
  booktitle={Advances in Neural Information Processing Systems (NeurIPS)},
  pages={1135--1143},
  year={2015}
}

@book{wernicke1874aphasische,
  title={Der aphasische Symptomencomplex: Eine psychologische Studie auf anatomischer Basis},
  author={Wernicke, Carl},
  publisher={Cohn und Weigert},
  address={Breslau},
  year={1874}
}

@article{hodges2007semantic,
  title={Semantic dementia: A unique clinicopathological syndrome},
  author={Hodges, John R. and Patterson, Karalyn},
  journal={The Lancet Neurology},
  volume={6},
  number={11},
  pages={1004--1014},
  year={2007},
  publisher={Elsevier}
}

@article{blennow2006alzheimers,
  title={Alzheimer's disease},
  author={Blennow, Kaj and de Leon, Mony J. and Zetterberg, Henrik},
  journal={The Lancet},
  volume={368},
  number={9533},
  pages={387--403},
  year={2006},
  publisher={Elsevier}
}

@article{vaswani2017attention,
  title={Attention Is All You Need},
  author={Vaswani, Ashish and Shazeer, Noam and Parmar, Niki and Uszkoreit, Jakob and Jones, Llion and Gomez, Aidan N. and Kaiser, {\L}ukasz and Polosukhin, Illia},
  journal={Advances in Neural Information Processing Systems (NeurIPS)},
  volume={30},
  year={2017}
}

@article{hochreiter1997lstm,
  title={Long Short-Term Memory},
  author={Hochreiter, Sepp and Schmidhuber, J{\"u}rgen},
  journal={Neural Computation},
  volume={9},
  number={8},
  pages={1735--1780},
  year={1997},
  publisher={MIT Press}
}

@article{eldan2023tinystories,
  title={{TinyStories}: How Small Can Language Models Be and Still Speak Coherent {E}nglish?},
  author={Eldan, Ronen and Li, Yuanzhi},
  journal={arXiv preprint arXiv:2305.07759},
  year={2023}
}

@article{bai2023qwen,
  title={Qwen Technical Report},
  author={Bai, Jinze and Bai, Shuai and Chu, Yunfei and Cui, Zeyu and Dang, Kai and Deng, Xiaodong and Fan, Yang and Ge, Wenbin and Han, Yu and Huang, Fei and others},
  journal={arXiv preprint arXiv:2309.16609},
  year={2023}
}

@article{pennington2014glove,
  title={{GloVe}: Global Vectors for Word Representation},
  author={Pennington, Jeffrey and Socher, Richard and Manning, Christopher D.},
  journal={Proceedings of the Conference on Empirical Methods in Natural Language Processing (EMNLP)},
  pages={1532--1543},
  year={2014}
}

@article{olah2020zoom,
  title={Zoom In: An Introduction to Circuits},
  author={Olah, Chris and Cammarata, Nick and Schubert, Ludwig and Goh, Gabriel and Petrov, Michael and Carter, Shan},
  journal={Distill},
  volume={5},
  number={3},
  pages={e00024.001},
  year={2020}
}

@inproceedings{geva2021transformer,
  title={Transformer Feed-Forward Layers Are Key-Value Memories},
  author={Geva, Mor and Schuster, Roei and Berant, Jonathan and Levy, Omer},
  booktitle={Proceedings of the 2021 Conference on Empirical Methods in Natural Language Processing},
  pages={5484--5495},
  year={2021}
}

@article{hill1973diversity,
  title={Diversity and Evenness: A Unifying Notation and Its Consequences},
  author={Hill, Mark O.},
  journal={Ecology},
  volume={54},
  number={2},
  pages={427--432},
  year={1973},
  publisher={Wiley}
}

@article{jost2006entropy,
  title={Entropy and Diversity},
  author={Jost, Lou},
  journal={Oikos},
  volume={113},
  number={2},
  pages={363--375},
  year={2006},
  publisher={Wiley}
}

@book{zipf1949human,
  title={Human Behavior and the Principle of Least Effort},
  author={Zipf, George Kingsley},
  year={1949},
  publisher={Addison-Wesley}
}

@article{moore2023dementia,
  title={Dementia in Convolutional Neural Networks: Using Deep Learning Models to Simulate Neurodegeneration of the Visual System},
  author={Moore, Jasmine A. and Tuladhar, Anup and Ismail, Zahinoor and Mouches, Pauline and Wilms, Matthias and Forkert, Nils D.},
  journal={Neuroinformatics},
  volume={21},
  pages={45--55},
  year={2023},
  publisher={Springer}
}

@article{moore2023neuroplasticity,
  title={Simulation of Neuroplasticity in a {CNN}-Based In-Silico Model of Neurodegeneration of the Visual System},
  author={Moore, Jasmine A. and Wilms, Matthias and Gutierrez, Alejandro and Ismail, Zahinoor and Fakhar, Kayson and Hadaeghi, Fatemeh and Hilgetag, Claus C. and Forkert, Nils D.},
  journal={Frontiers in Computational Neuroscience},
  volume={17},
  pages={1274824},
  year={2023},
  publisher={Frontiers}
}

@article{moore2025realistic,
  title={Towards Realistic Simulation of Disease Progression in the Visual Cortex with {CNN}s},
  author={Moore, Jasmine A. and Kang, Chris and Vigneshwaran, Vibujithan and Stanley, Emma A. M. and Memon, Ashar and Wilms, Matthias and Forkert, Nils D.},
  journal={Scientific Reports},
  volume={15},
  pages={6099},
  year={2025},
  publisher={Nature Publishing Group}
}

@book{goodglass1993understanding,
  title={Understanding Aphasia},
  author={Goodglass, Harold},
  publisher={Academic Press},
  year={1993},
  address={San Diego}
}

@article{warrington1975selective,
  title={The selective impairment of semantic memory},
  author={Warrington, Elizabeth K.},
  journal={Quarterly Journal of Experimental Psychology},
  volume={27},
  number={4},
  pages={635--657},
  year={1975}
}

@article{gornotempini2011classification,
  title={Classification of primary progressive aphasia and its variants},
  author={Gorno-Tempini, Maria Luisa and Hillis, Argye E. and Weintraub, Sandra and Kertesz, Andrew and Mendez, Mario and Cappa, Stefano F. and Ogar, Jennifer M. and Rohrer, Jonathan D. and Black, Sandra and Boeve, Bradley F. and Manes, Facundo and Dronkers, Nina F. and Vandenberghe, Rik and Rascovsky, Katya and Patterson, Karalyn and Miller, Bruce L. and Knopman, David S. and Hodges, John R. and Mesulam, M.-Marsel and Grossman, Murray},
  journal={Neurology},
  volume={76},
  number={11},
  pages={1006--1014},
  year={2011}
}

@inproceedings{merity2017pointer,
  title={Pointer Sentinel Mixture Models},
  author={Merity, Stephen and Xiong, Caiming and Bradbury, James and Socher, Richard},
  booktitle={International Conference on Learning Representations ({ICLR})},
  year={2017}
}

@inproceedings{merity2018regularizing,
  title={Regularizing and Optimizing {LSTM} Language Models},
  author={Merity, Stephen and Keskar, Nitish Shirish and Socher, Richard},
  booktitle={International Conference on Learning Representations ({ICLR})},
  year={2018}
}

@article{elhage2021mathematical,
  title={A Mathematical Framework for Transformer Circuits},
  author={Elhage, Nelson and Nanda, Neel and Olsson, Catherine and Henighan, Tom and Joseph, Nicholas and Mann, Ben and Askell, Amanda and Bai, Yuntao and Chen, Anna and Conerly, Tom and others},
  journal={Transformer Circuits Thread},
  year={2021}
}

@article{olsson2022context,
  title={In-context Learning and Induction Heads},
  author={Olsson, Catherine and Elhage, Nelson and Nanda, Neel and Joseph, Nicholas and DasSarma, Nova and Henighan, Tom and Mann, Ben and Askell, Amanda and Bai, Yuntao and Chen, Anna and others},
  journal={Transformer Circuits Thread},
  year={2022}
}

@inproceedings{howard2018universal,
  title={Universal Language Model Fine-tuning for Text Classification},
  author={Howard, Jeremy and Ruder, Sebastian},
  booktitle={Proceedings of the 56th Annual Meeting of the Association for Computational Linguistics (Volume 1: Long Papers)},
  pages={328--339},
  year={2018}
}

@article{brysbaert2014concreteness,
  title={Concreteness ratings for 40 thousand generally known {English} word lemmas},
  author={Brysbaert, Marc and Warriner, Amy Beth and Kuperman, Victor},
  journal={Behavior Research Methods},
  volume={46},
  number={3},
  pages={904--911},
  year={2014},
  publisher={Springer},
  doi={10.3758/s13428-013-0403-5}
}

@article{cloze,
  author={Taylor, Wilson L.},
  title={{Cloze procedure: A new tool for measuring readability}},
  journal={Journalism Quarterly},
  volume={30},
  number={4},
  pages={415--433},
  year={1953}
}

@inproceedings{kingma2015adam,
  author    = {Kingma, Diederik P. and Ba, Jimmy},
  title     = {{Adam: A Method for Stochastic Optimization}},
  booktitle = {International Conference on Learning Representations (ICLR)},
  year      = {2015},
  url       = {https://arxiv.org/abs/1412.6980}
}

@article{terry1991,
author = {Terry, Robert D. and Masliah, Eliezer and Salmon, David P. and Butters, Nelson and DeTeresa, Richard and Hill, Robert and Hansen, Lawrence A. and Katzman, Robert},
title = {Physical basis of cognitive alterations in {Alzheimer}'s disease: Synapse loss is the major correlate of cognitive impairment},
journal = {Annals of Neurology},
volume = {30},
number = {4},
pages = {572--580},
year = {1991}
}

@article{dekosky1990,
author = {DeKosky, Steven T. and Scheff, Stephen W.},
title = {Synapse loss in frontal cortex biopsies in {Alzheimer}'s disease: Correlation with cognitive severity},
journal = {Annals of Neurology},
volume = {27},
number = {5},
pages = {457--464},
year = {1990}
}

@article{GomezIsla4491,
author = {G{\'o}mez-Isla, Teresa and Price, Joseph L. and McKeel, Jr., Daniel W. and Morris, John C. and Growdon, John H. and Hyman, Bradley T.},
title = {Profound loss of layer {II} entorhinal cortex neurons occurs in very mild {Alzheimer's} disease},
volume = {16},
number = {14},
pages = {4491--4500},
year = {1996},
publisher = {Society for Neuroscience},
journal = {Journal of Neuroscience}
}

@article{lambonralph2017semantic,
  title={The Neural and Computational Bases of Semantic Cognition},
  author={Lambon Ralph, Matthew A. and Jefferies, Elizabeth and Patterson, Karalyn and Rogers, Timothy T.},
  journal={Nature Reviews Neuroscience},
  volume={18},
  number={1},
  pages={42--55},
  year={2017},
  publisher={Nature Publishing Group}
}

@article{macwhinney2011aphasiabank,
  title={{AphasiaBank}: Methods for Studying Discourse},
  author={MacWhinney, Brian and Fromm, Davida and Forbes, Margaret and Holland, Audrey},
  journal={Aphasiology},
  volume={25},
  number={11},
  pages={1286--1307},
  year={2011},
  publisher={Taylor \& Francis}
}

@article{yang2026lesioned,
  title={Lesioned Multimodal Language Models Reproduce Aphasic Picture-Naming Patterns},
  author={Yang, Yong and Guan, Xiang and Arheix-Parras, Sophie and Ahmadi, Saeed and Newman-Norlund, Roger and Bonilha, Leonardo and Rorden, Christopher and Fridriksson, Julius and Desai, Rutvik H. and Nelakuditi, Srihari},
  journal={arXiv preprint arXiv:2607.11621},
  year={2026}
}

@article{wang2026componentlevel,
  title={Component-Level Lesioning of Language Models Reveals Clinically Aligned Aphasia Phenotypes},
  author={Wang, Yifan and Zheng, Jichen and Sun, Jingyuan and Zhang, Yunhao and Ye, Chunyu and Li, Jixing and Zong, Chengqing and Wang, Shaonan},
  journal={arXiv preprint arXiv:2601.19723},
  year={2026}
}

@article{mirman2010moss,
  title={A large, searchable, web-based database of aphasic performance on picture naming and other tests of cognitive function},
  author={Mirman, Daniel and Strauss, Ted J. and Brecher, Adelyn and Walker, Grant M. and Sobel, Paula and Dell, Gary S. and Schwartz, Myrna F.},
  journal={Cognitive Neuropsychology},
  volume={27},
  number={6},
  pages={495--504},
  year={2010}
}

@book{fellbaum1998wordnet,
  title={{WordNet}: An Electronic Lexical Database},
  author={Fellbaum, Christiane},
  publisher={MIT Press},
  year={1998}
}

@article{vandyck2023lecanemab,
  title={Lecanemab in Early {Alzheimer}'s Disease},
  author={van Dyck, Christopher H. and Swanson, Chad J. and Aisen, Paul and Bateman, Randall J. and Chen, Christopher and Gee, Michelle and Kanekiyo, Michio and Li, David and Reyderman, Larisa and Cohen, Sharon and Froelich, Lutz and Katayama, Sadao and Sabbagh, Marwan and Vellas, Bruno and Watson, David and Dhadda, Shobha and Irizarry, Michael C. and Kramer, Lynn D. and Iwatsubo, Takeshi},
  journal={New England Journal of Medicine},
  volume={388},
  number={1},
  pages={9--21},
  year={2023},
  publisher={Massachusetts Medical Society}
}

@article{sims2023donanemab,
  title={Donanemab in Early Symptomatic {Alzheimer} Disease: The {TRAILBLAZER-ALZ}~2 Randomized Clinical Trial},
  author={Sims, John R. and Zimmer, Jennifer A. and Evans, Cynthia D. and Lu, Ming and Ardayfio, Paul and Sparks, JonDavid and Wessels, Alette M. and Shcherbinin, Sergey and Wang, Hong and Monkul Nery, Emel Serap and Collins, Emily C. and Solomon, Paul and Salloway, Stephen and Apostolova, Liana G. and Hansson, Oskar and Ritchie, Craig and Brooks, Dawn A. and Mintun, Mark and Skovronsky, Daniel M. and others},
  journal={JAMA},
  volume={330},
  number={6},
  pages={512--527},
  year={2023},
  publisher={American Medical Association}
}

@article{braak1991staging,
  title={Neuropathological stageing of {Alzheimer}-related changes},
  author={Braak, Heiko and Braak, Eva},
  journal={Acta Neuropathologica},
  volume={82},
  number={4},
  pages={239--259},
  year={1991},
  publisher={Springer}
}

@article{lajoie2020tau,
  title={Prospective longitudinal atrophy in {Alzheimer}'s disease correlates with the intensity and topography of baseline tau-{PET}},
  author={La Joie, Renaud and Visani, Adrienne V. and Baker, Suzanne L. and Brown, Jesse A. and Bourakova, Viktoriya and Cha, Jungho and Chaudhary, Kiran and Edwards, Lauren and Iaccarino, Leonardo and Janabi, Mustafa and Lesman-Segev, Orit H. and Miller, Zachary A. and Perry, David C. and O'Neil, James P. and Pham, Julie and Rojas, Julio C. and Rosen, Howard J. and Seeley, William W. and Tsai, Richard M. and Miller, Bruce L. and Jagust, William J. and Rabinovici, Gil D.},
  journal={Science Translational Medicine},
  volume={12},
  number={524},
  pages={eaau5732},
  year={2020},
  publisher={American Association for the Advancement of Science}
}

@book{bird2009nltk,
  title={Natural Language Processing with {Python}: Analyzing Text with the Natural Language Toolkit},
  author={Bird, Steven and Klein, Ewan and Loper, Edward},
  year={2009},
  publisher={O'Reilly Media},
  address={Sebastopol, CA},
  isbn={978-0-596-51649-9}
}

@article{jefferies2006semantic,
  title={Semantic impairment in stroke aphasia versus semantic dementia: a case-series comparison},
  author={Jefferies, Elizabeth and Lambon Ralph, Matthew A.},
  journal={Brain},
  volume={129},
  number={8},
  pages={2132--2147},
  year={2006}
}

@article{robson2012wernicke,
  title={Wernicke's aphasia reflects a combination of acoustic-phonological and semantic control deficits: a case-series comparison of {W}ernicke's aphasia, semantic dementia and semantic aphasia},
  author={Robson, Holly and Sage, Karen and Lambon Ralph, Matthew A.},
  journal={Neuropsychologia},
  volume={50},
  number={2},
  pages={266--275},
  year={2012}
}

@article{roach1996philadelphia,
  title={The {P}hiladelphia {N}aming {T}est: Scoring and rationale},
  author={Roach, April and Schwartz, Myrna F. and Martin, Nadine and Grewal, Rita S. and Brecher, Adelyn},
  journal={Clinical Aphasiology},
  volume={24},
  pages={121--133},
  year={1996}
}

\clearpage
\section*{Supplementary Material}
\setcounter{subsection}{0}
\renewcommand{\thesubsection}{S\arabic{subsection}}
\renewcommand{\theHsubsection}{Supp.S\arabic{subsection}}

\paragraph{Organization of the Supplementary Material.}
This supplement is organized into three parts so that all methodological details and quantitative analyses appear before the archival prompt-level lookup tables.
First, Section~\ref{sec:supp_methods} (\emph{Extended Materials and Methods}) provides the complete architectural specifications, weight-matrix ablation definitions (Supplementary Figure~\ref{fig:methods_schematics}), post-lesion fine-tuning protocols, and metric implementations.
Second, Sections~\ref{sec:supp_arch_summary}--\ref{sec:supp_plasticity} present the supporting quantitative analyses and sensitivity controls: parameter-fraction scaling, curve-shape model selection, and LSTM connection-pruning controls (Section~\ref{sec:supp_arch_summary}); $R_4$ scale-invariance derivation and accuracy-matched null-model breakdowns (Section~\ref{sec:supp_null_model}); frequency-rank Fourier spectral analysis and complete per-$k$ metric tables (Section~\ref{sec:supp_spectral_details}); quantitative Cloze trajectories and qualitative cross-model comparison (Section~\ref{sec:supp_cloze_summary}); statistical formalisation of the clinical parallel assignments (Section~\ref{sec:supp_clinical_stats}); Brysbaert concreteness scoring methodology (Section~\ref{sec:concreteness_method}); automated re-coding into the seven-category Philadelphia Naming Test (PNT) taxonomy and patient cohort comparisons (Section~\ref{sec:pnt_recoding}); and four-arm compensatory plasticity sensitivity controls across zero, local SGD, and 500-step AdamW fine-tuning (Section~\ref{sec:supp_plasticity}).
Third, Sections~\ref{sec:supp_qwen18_cloze}--\ref{sec:supp_small_lstm_cloze} at the end of the supplement archive the complete 100-prompt first-trial Cloze predictions across all undamaged and ablated conditions for Qwen~1.8B, Qwen~0.5B, AWD-LSTM, and Small LSTM for completeness and reproducibility.

\subsection{Extended Materials and Methods}\label{sec:supp_methods}

\subsubsection*{Model Architectures and Lesion Definitions}

\begin{figure}[!htb]
\centering
\includegraphics[width=\textwidth,keepaspectratio]{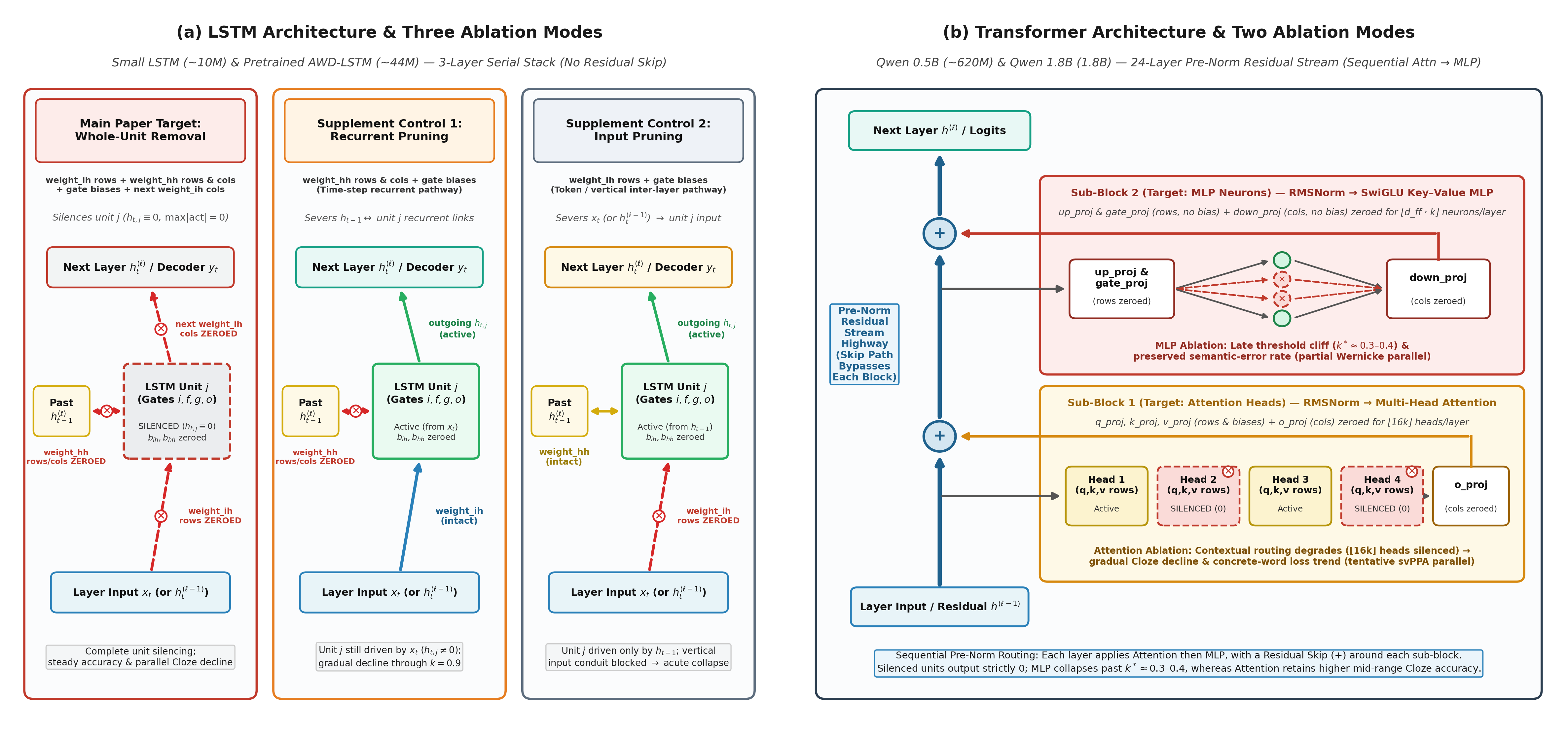}
\caption{Schematic overview of model architectures and ablation modes. (a)~In the 3-layer Small LSTM and pretrained AWD-LSTM (serial layer stack without residual skip connections), our primary target is \emph{whole-unit removal} (left card), which zeroes \texttt{weight\_ih} rows, \texttt{weight\_hh} incoming rows and outgoing columns, gate biases \texttt{bias\_ih} and \texttt{bias\_hh}, and outgoing columns in the next layer's \texttt{weight\_ih} to silence unit $j$ completely ($h_{t,j} \equiv 0$). To separate the two incoming pathways (Section~\ref{sec:discussion}; Supplements~\ref{sec:supp_arch_summary}--\ref{sec:supp_clinical_stats}), we also evaluate two connection-level controls: \emph{recurrent-connection pruning} (centre card), which zeroes \texttt{weight\_hh} rows/columns and gate biases while unit $j$ still receives current input $x_t$, and \emph{input-connection pruning} (right card), which zeroes \texttt{weight\_ih} rows and gate biases while unit $j$ still receives recurrent state $h_{t-1}$. (b)~In the 24-layer Qwen~0.5B and Qwen~1.8B Transformers, a residual stream ($+$) bypasses each layer's attention and MLP sub-blocks. We compare (1)~attention-head removal ($Q,K,V$ projection rows/biases and $O$ projection columns zeroed) and (2)~MLP neuron removal (\texttt{up\_proj}/\texttt{gate\_proj} rows and \texttt{down\_proj} columns zeroed).}
\label{fig:methods_schematics}
\end{figure}

Figure~\ref{fig:methods_schematics} illustrates the computational pathways and ablation modes evaluated across the recurrent (LSTM) and Transformer model families. In each LSTM layer, every hidden unit receives two incoming linear projections that sum across its four gates (input, forget, cell, and output): an input-to-hidden weight matrix (\texttt{weight\_ih}, with gate bias \texttt{bias\_ih}) that projects the current-step input $x_t$ from the embedding or preceding layer, and a recurrent hidden-to-hidden weight matrix (\texttt{weight\_hh}, with gate bias \texttt{bias\_hh}) that projects the previous hidden state $h_{t-1}$ (Figure~\ref{fig:methods_schematics}a).

\paragraph{Small LSTM (${\sim}$10M parameters; 580,320 non-embedding parameters).}
This model is a 3-layer LSTM with 155-dimensional embeddings and hidden states, a tied decoder sharing encoder embedding weights, and dropout 0.2.
It is trained from scratch on the WikiText-103 training split~\citep{merity2017pointer} for 10 epochs with the Adam optimizer~\citep{kingma2015adam}. Text is tokenized with lowercased word and punctuation splitting mapped to the shared $60{,}000$-token AWD-LSTM vocabulary, using \texttt{xxunk} for out-of-vocabulary words without \texttt{xxbos}/\texttt{xxmaj}/\texttt{xxup} prefix tokens.
Across all 3 layers simultaneously, we evaluate complete whole-unit removal alongside two incomplete (connection-level) controls discussed in Section~\ref{sec:discussion} and reported in the Supplement, defined in terms of the input-to-hidden matrix \texttt{weight\_ih} and recurrent hidden-to-hidden matrix \texttt{weight\_hh} (Figure~\ref{fig:methods_schematics}a):
\begin{enumerate}
    \item \emph{Whole-unit removal} (\texttt{weight\_ih} + \texttt{weight\_hh}): For each selected unit index $j$, we zero rows $[g \cdot H + j, :]$ of both \texttt{weight\_ih} and \texttt{weight\_hh} across all four gates ($g \in \{0,1,2,3\}$, where $g$ indexes the four stacked LSTM gates: input, forget, cell, and output, each of height $H$). We simultaneously zero columns $[:, j]$ of \texttt{weight\_hh}, columns $[:, j]$ of the next layer's \texttt{weight\_ih} (for layers $\ell < L-1$), and the corresponding entries of \texttt{bias\_ih} and \texttt{bias\_hh}. This completely silences unit $j$ ($h_{t,j} \equiv 0$, verified by forward hooks showing $\max|\text{activity}| = 0.0$) and removes $408{,}408$ parameters, or 70.4\% of non-embedding parameters, at $k = 0.5$.
    \item \emph{Recurrent-connection pruning} (incomplete ablation control 1; \texttt{weight\_hh}): We zero rows $[g \cdot H + j, :]$ (incoming recurrent) and columns $[:, j]$ (outgoing recurrent) of \texttt{weight\_hh}, together with the corresponding gate entries of \texttt{bias\_ih} and \texttt{bias\_hh}, across all four gates. This leaves \texttt{weight\_ih} and next-layer projections intact so unit $j$ can still be driven by $x_t$ (removing 37.4\% of non-embedding parameters at $k = 0.5$).
    \item \emph{Input-connection pruning} (incomplete ablation control 2; \texttt{weight\_ih}): We zero rows $[g \cdot H + j, :]$ of \texttt{weight\_ih} and the corresponding gate entries of \texttt{bias\_ih} and \texttt{bias\_hh} across all four gates, leaving \texttt{weight\_hh} intact (removing 25.0\% of non-embedding parameters at $k = 0.5$).
\end{enumerate}
Gradient hooks maintain all ablated entries at zero during fine-tuning.

\paragraph{AWD-LSTM (${\sim}$44M parameters; 20.3M non-embedding parameters).}
This model is a 3-layer Average Stochastic Gradient Descent Weight-Dropped LSTM (AWD-LSTM)~\citep{merity2018regularizing} with 400-dimensional embeddings, 1,152 hidden units in the first two layers, 400 in the third, and a tied decoder.
It uses WeightDrop~\citep{merity2018regularizing} with $w_{\mathrm{drop}} = 0.5$ during training. At ablation time, we disable WeightDrop ($w_{\mathrm{drop}} \to 0$) so that dropout-trained resilience does not mask ablation effects.
Pretrained weights are loaded from fastai~\citep{howard2018universal} trained on WikiText-103, sharing the exact $60{,}000$-token vocabulary of the Small LSTM.
We evaluate the same three lesion types as for the Small LSTM (Figure~\ref{fig:methods_schematics}a):
\begin{enumerate}
    \item \emph{Whole-unit removal}: Zeroes \texttt{weight\_ih}, \texttt{weight\_hh}, gate biases, and next-layer \texttt{weight\_ih} columns, completely silencing ablated units ($\max|\text{activity}| = 0.0$) and removing $14{,}738{,}752$ parameters (72.7\% of non-embedding parameters) at $k = 0.5$.
    \item \emph{Recurrent-connection pruning} (incomplete control 1): Zeroes the recurrent weight matrix \texttt{weight\_hh} and gate biases (41.7\% of non-embedding parameters at $k = 0.5$).
    \item \emph{Input-connection pruning} (incomplete control 2): Zeroes \texttt{weight\_ih} plus gate biases (22.2\% of non-embedding parameters at $k = 0.5$).
\end{enumerate}

\paragraph{Qwen 0.5B (${\sim}$620M parameters; 308.4M non-embedding parameters).}
Qwen~1.5-0.5B~\citep{bai2023qwen} is loaded from its off-the-shelf pretrained checkpoint (\texttt{Qwen/Qwen1.5-0.5B}) and is a dense Transformer with SwiGLU MLP activations, RoPE positional embeddings, and multi-head attention (24 layers; hidden size 1{,}024; MLP intermediate size 2{,}816; 16 query heads and 16 key--value heads).
We evaluate two ablation targets across all 24 layers. First, MLP neuron removal zeroes rows of the bias-free \texttt{up\_proj} and \texttt{gate\_proj} matrices and columns of the bias-free \texttt{down\_proj} matrix ($\max|\text{activity}|=0.0$; removing 33.7\% of non-embedding parameters at $k = 0.5$). Second, attention-head removal zeroes $Q$, $K$, $V$ projection rows (and their additive biases) and the bias-free $O$ projection columns for selected heads ($\max|\text{activity}|=0.0$; removing 16.3\% of non-embedding parameters at $k = 0.5$).
The two targets differ in granularity: each layer contains $2{,}816$ MLP neurons ($5{,}504$ in Qwen~1.8B) but only $16$ attention heads. Because attention ablation removes $\lfloor 16\,k \rfloor$ whole heads per layer, the removed head fraction increases in steps of $1/16$ ($6.25\%$) and is slightly below $k$ when $16\,k$ is not an integer (for example, $k = 0.3$ removes $4$ of $16$ heads, or $25\%$ of heads and $8.2\%$ of non-embedding parameters). To account for both discrete head counts and pathway parameter size, Supplementary Figure~\ref{sfig:accuracy_param_fraction} and Supplementary Table~\ref{stab:curve_shape} plot and fit degradation curves against both $k$ and the exact fraction of non-embedding parameters removed.

\paragraph{Qwen 1.8B (1.8B parameters; 1,214.5M non-embedding parameters).}
Qwen~1.5-1.8B~\citep{bai2023qwen} is likewise loaded from its off-the-shelf pretrained checkpoint (\texttt{Qwen/Qwen1.5-1.8B}); it shares the architecture family and depth of the 0.5B model (24 layers; 16 query and 16 key--value heads) but is wider, with hidden size 2{,}048 and MLP intermediate size 5{,}504.
It follows the same ablation protocol as Qwen~0.5B, removing 33.4\% of non-embedding parameters at $k = 0.5$ for MLP ablation and 16.6\% for attention-head ablation.

\subsubsection*{Ablation, Fine-Tuning, and Plasticity Sensitivity Protocol}

For each model--target combination and ablation fraction $k \in \{0.0, 0.1, 0.2, \ldots, 0.9\}$, we run 5 independent trials across the full sweep to support all main and supplementary degradation curves (Supplementary Tables~\ref{stab:summary_k05}--\ref{stab:qwen18_attn_full}). We increase replication to 15 independent trials per $k$ across the shared transition region $k \in \{0.1, 0.2, 0.3, 0.4\}$ ($n = 60$ trials per pathway) for the Qwen null-model pathway tests in Section~\ref{sec:moments} and Supplement~\ref{sec:supp_null_model}. For the Cloze, Brysbaert word-concreteness~\citep{brysbaert2014concreteness}, and Philadelphia Naming Test (PNT)~\citep{mirman2010moss} analyses (Section~\ref{sec:clinical_evidence}; Supplements~\ref{sec:concreteness_method}--\ref{sec:pnt_recoding}), we evaluate 10 independent trials at $k = 0.2$, alongside the deterministic greedy baseline at $k = 0.0$ and 5 trials at $k \in \{0.1, 0.3, \ldots, 0.9\}$. Each trial proceeds in five steps:

\begin{enumerate}
    \item \textit{Deep copy} the pretrained model and seed all random number generators deterministically per $(k, \text{trial})$.
    \item \textit{Ablate}: randomly select $\lfloor k \times n \rfloor$ target units per layer independently across layers and zero all associated weights and gate/head biases across all layers (systemic ablation).
    \item \textit{Freeze ablated weights at zero} during fine-tuning (via gradient hooks and post-update masking) so that silenced units or pruned connections cannot recover.
    \item \textit{Fine-tune} briefly at every $k \in \{0.0, 0.1, \ldots, 0.9\}$ (so undamaged $k = 0$ and damaged $k > 0$ models undergo the exact same adaptation step) as a simple proxy for local compensatory plasticity (primary \texttt{weak} SGD arm):
    \begin{itemize}
        \item LSTM-based models use SGD ($\mathrm{lr} = 10^{-3}$, gradient clipping 0.25, $\mathrm{bptt} = 70$) on TinyStories for 40 sequence steps (Small LSTM, plain lowercased tokens) or 50 sequence steps (AWD-LSTM, with fastai \texttt{xxbos}/\texttt{xxmaj}/\texttt{xxup} markers).
        \item Qwen models use SGD ($\mathrm{lr} = 10^{-3}$) across the first 500 TinyStories passages (passages 1--500, max sequence length 64), clipping gradients to 0.25 after each passage's backward pass and accumulating over 8 passages per update (62 weight updates in total).
    \end{itemize}
    Because 40--50 steps of clipped SGD at $\mathrm{lr} = 10^{-3}$ on out-of-domain TinyStories move the LSTM weights by only $\|\Delta\theta\|_2 \le 0.012$ (relative change ${\sim}10^{-5}$), this step leaves LSTM WikiText-2 accuracy essentially unchanged. Thus, the \texttt{weak} and \texttt{none} arms coincide for the LSTMs, whereas 500 steps of AdamW on the same TinyStories passages in the \texttt{strong} arm do shift WikiText-2 accuracy (Supplement~\ref{sec:supp_plasticity}).
    \item \textit{Evaluate} on the held-out test set.
\end{enumerate}

\paragraph{Fine-tuning sensitivity control.}
Because biological recovery after acute damage operates through local weight adjustments rather than global network retraining, we use brief fixed-step SGD without momentum (\texttt{weak} arm) as our primary protocol.
To test whether the contrast between Transformer and LSTM curve shapes depends on this choice, we compare zero fine-tuning (\texttt{none}), brief local SGD (\texttt{weak}), and 500-step AdamW retraining (\texttt{strong}, plus in-domain WikiText-103 \texttt{strong\_wt103} for both LSTMs) across $k \in \{0.0, 0.1, \ldots, 0.9\}$. This comparison covers both recurrent architectures across all three ablation modes and the pretrained Transformers (Qwen~0.5B across both pathways, plus Qwen~1.8B MLP across \texttt{none}, \texttt{weak}, and $k \le 0.2$ of \texttt{strong}).
With zero fine-tuning, Qwen MLP ablation still exhibits a sharp late threshold collapse without an early retention advantage over the LSTMs. Conversely, the Small LSTM and AWD-LSTM whole-unit and recurrent pathways decline smoothly without a cliff even under 500 steps of in-domain AdamW retraining (Supplement~\ref{sec:supp_plasticity}; Supplementary Figures~\ref{sfig:plasticity_small_lstm}--\ref{sfig:plasticity_qwen}).

For accuracy evaluation, LSTM models are tested on WikiText-2~\citep{merity2017pointer} ($241{,}200$ lowercased lexical tokens for the Small LSTM; $271{,}480$ tokens for the AWD-LSTM, comprising $241{,}202$ lexical tokens and $30{,}278$ unscored \texttt{xxbos}/\texttt{xxmaj}/\texttt{xxup} prefix markers). Excluding out-of-vocabulary \texttt{xxunk} targets yields $225{,}702$ scored in-vocabulary tokens for the Small LSTM and $225{,}704$ for the AWD-LSTM. Qwen models are tested on the first 32 tokens of each held-out TinyStories passage (passages 501--1,000, yielding up to 31 next-token predictions per passage, or $15{,}268$ scored targets in total). The passages used to rank tokens for the auxiliary Qwen spectral analysis (passages 1--2,000; Supplement~\ref{sec:supp_spectral_details}) overlap these evaluation passages.
To ensure cross-architecture comparability, we report \emph{relative accuracy} ($\mathrm{Acc}(k)/\mathrm{Acc}(0)$), evaluate error distributions against an accuracy-matched random-failure baseline, and base all clinical parallel assignments (Section~\ref{sec:clinical_evidence}) on the shared 100-prompt Cloze battery.

\subsubsection*{Evaluation Metrics}

\paragraph{Accuracy and Curve-Shape Model Selection.}
Next-word prediction accuracy (top-1) is evaluated over all valid in-vocabulary test tokens and reported as relative accuracy $\mathrm{RelAcc}(k) = \mathrm{Acc}(k) / \mathrm{Acc}(0)$, with bootstrap standard errors across trials. For both LSTMs, out-of-vocabulary (\texttt{xxunk}) targets are excluded from scoring. For the AWD-LSTM, fastai formatting tokens (\texttt{xxbos}, \texttt{xxmaj}, \texttt{xxup}) are also excluded from target scoring; however, because they remain in the input sequence and indicate whether the following word is capitalized, they contribute to the AWD-LSTM's higher baseline accuracy relative to the Small LSTM.
To test whether each degradation curve follows a straight line or bends non-linearly, we fit both a 2-parameter line $y(x) = a + b x$ and a 3-parameter logistic S-curve $y(x) = a / (1 + \exp(b(x - x_0)))$ to the trial-level relative accuracy points. We perform these fits against both $x = k$ and $x = \text{fraction of non-embedding parameters removed}$, comparing them via $\Delta\mathrm{AIC} = \mathrm{AIC}_{\mathrm{logistic}} - \mathrm{AIC}_{\mathrm{linear}}$ (Supplementary Table~\ref{stab:curve_shape}). A $\Delta\mathrm{AIC} < -2$ indicates non-linear curvature, while the presence of an early buffered plateau versus an immediate drop is assessed from surviving relative accuracy across $k \in [0.1, 0.3]$. By this criterion, only attention-head ablation shows an early plateau. Because Qwen MLP relative accuracy at $k = 0.1\text{--}0.2$ is not systematically higher than that of the LSTMs ($71\%\text{--}94\%$ of baseline across $k = 0.1\text{--}0.2$ in Qwen MLP vs.\ $75\%\text{--}93\%$ in the LSTMs; Supplementary Table~\ref{stab:curve_shape}), we describe the MLP curves as a late threshold collapse.

\paragraph{Moment Ratio $R_4$ and Accuracy-Matched Null Model.}
$R_4(k)$ is computed from the per-token error count vector $\mathbf{e}(k)$ ordered by native vocabulary index $i$ (word-frequency rank for the LSTMs and BPE merge order for Qwen, where common subwords have low $i$; Equation~\ref{eq:R4}), whereas $D_1(k)$ and $\mathrm{JS}(p(k)\,\|\,p(0))$ do not depend on token ordering and the auxiliary spectral metrics $\beta, G$ explicitly reorder the top $1{,}000$ tokens by corpus frequency across all models (Supplement~\ref{sec:supp_spectral_details}).
For the accuracy-matched random-failure baseline (Equations~\ref{eq:null_model}--\ref{eq:js_div}), at each $(k, \text{trial})$ we draw $B = 100$ independent samples without replacement of size $\Delta M = M(k) - M(0)$ from the undamaged model's correct-token count vector $\mathbf{c}(0)$ and add each draw to $\mathbf{e}(0)$. From these draws we compute the mean null expectations $\mathbb{E}[R_4^{\mathrm{null}}(k)]$ and $\mathbb{E}[D_1^{\mathrm{null}}(k)]$, along with the base-2 Jensen--Shannon divergence $\mathrm{JS}(p(k) \,\|\, p(0))$.

\paragraph{Quantified Cloze Test.}
We use 100 sentence prompts with predictable final words, stratified by POS tag (48 nouns, 25 verbs, 22 adjectives, 5 adverbs).
Each prompt is fed to the model and the top-1 predicted word is extracted. For the LSTMs, we skip fastai control tokens (\texttt{xxunk}, \texttt{xxmaj}, \texttt{xxup}, \texttt{xxbos}, \texttt{xxeos}, \texttt{xxfld}, \texttt{xxpad}, \texttt{xxrep}, \texttt{xxwrep}) when selecting the top lexical prediction; for Qwen, we strip leading and trailing whitespace from the decoded subword token.
Part-of-speech (POS) tags are assigned using NLTK's~\citep{bird2009nltk} standard POS tagger (Universal tagset), applied to the predicted word on its own rather than in sentence context. Prompts whose predicted token remains a non-word symbol are scored as POS mismatches and assigned zero semantic similarity. Semantic similarity is also set to zero whenever the predicted word's tag is not a content-word tag, even if the word has a GloVe vector, which couples the semantic score to the syntactic score.
For content-tagged predictions, semantic similarity is the cosine similarity between 50-dimensional GloVe~\citep{pennington2014glove} vectors (\texttt{glove.6B.50d}) of the predicted and target words; out-of-vocabulary predictions receive similarity 0.0.
Across $k \in \{0.0, 0.2\}$, we evaluate 10 independent trials ($N = 1{,}000$ prompt evaluations per condition). We use these to compute 95\% confidence intervals and two-sided permutation tests ($20{,}000$ permutations, with Holm correction across the planned contrasts) for the Brysbaert concreteness shift $\Delta\text{Conc.}$ (Supplement~\ref{sec:concreteness_method}) and the Philadelphia Naming Test (PNT) error proportions (Supplement~\ref{sec:pnt_recoding}).

\subsection{Architecture Summary, Parameter-Fraction Scaling, Detailed Accuracy Analysis, and LSTM Connection-Pruning Controls}\label{sec:supp_arch_summary}

\begin{figure}[!htbp]
\centering
\includegraphics[width=0.78\textwidth,height=0.58\textheight,keepaspectratio]{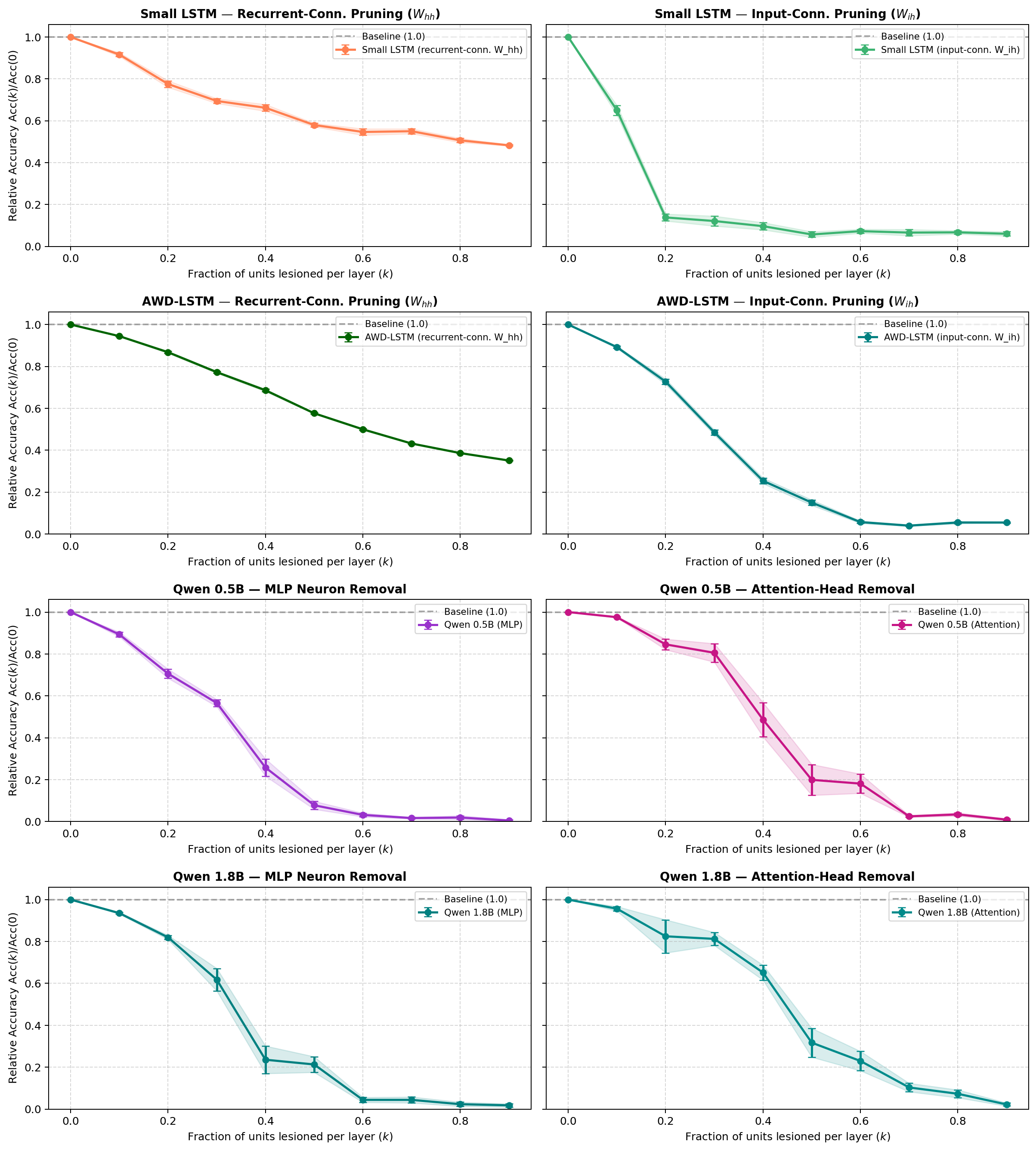}
\caption{Relative accuracy ($\mathrm{Acc}(k)/\mathrm{Acc}(0)$, normalised to baseline) degradation curves for the supplementary LSTM connection-pruning controls (top two rows: Small LSTM and AWD-LSTM recurrent-connection pruning \texttt{weight\_hh} in the left column and input-connection pruning \texttt{weight\_ih} in the right column) alongside the Qwen~0.5B and Qwen~1.8B MLP and Attention pathways (bottom two rows; primary LSTM whole-unit curves are shown in Figure~\ref{fig:accuracy}a,b). Isolating \texttt{weight\_hh} vs.\ \texttt{weight\_ih} shows that recurrent-connection pruning degrades gradually through $k = 0.9$, whereas input-connection pruning collapses steeply by $k = 0.2$--$0.4$. All curves show means $\pm 1$ SE across trials.}
\label{sfig:accuracy_all}
\end{figure}

\begin{figure}[!htbp]
\centering
\includegraphics[width=0.82\textwidth,height=0.62\textheight,keepaspectratio]{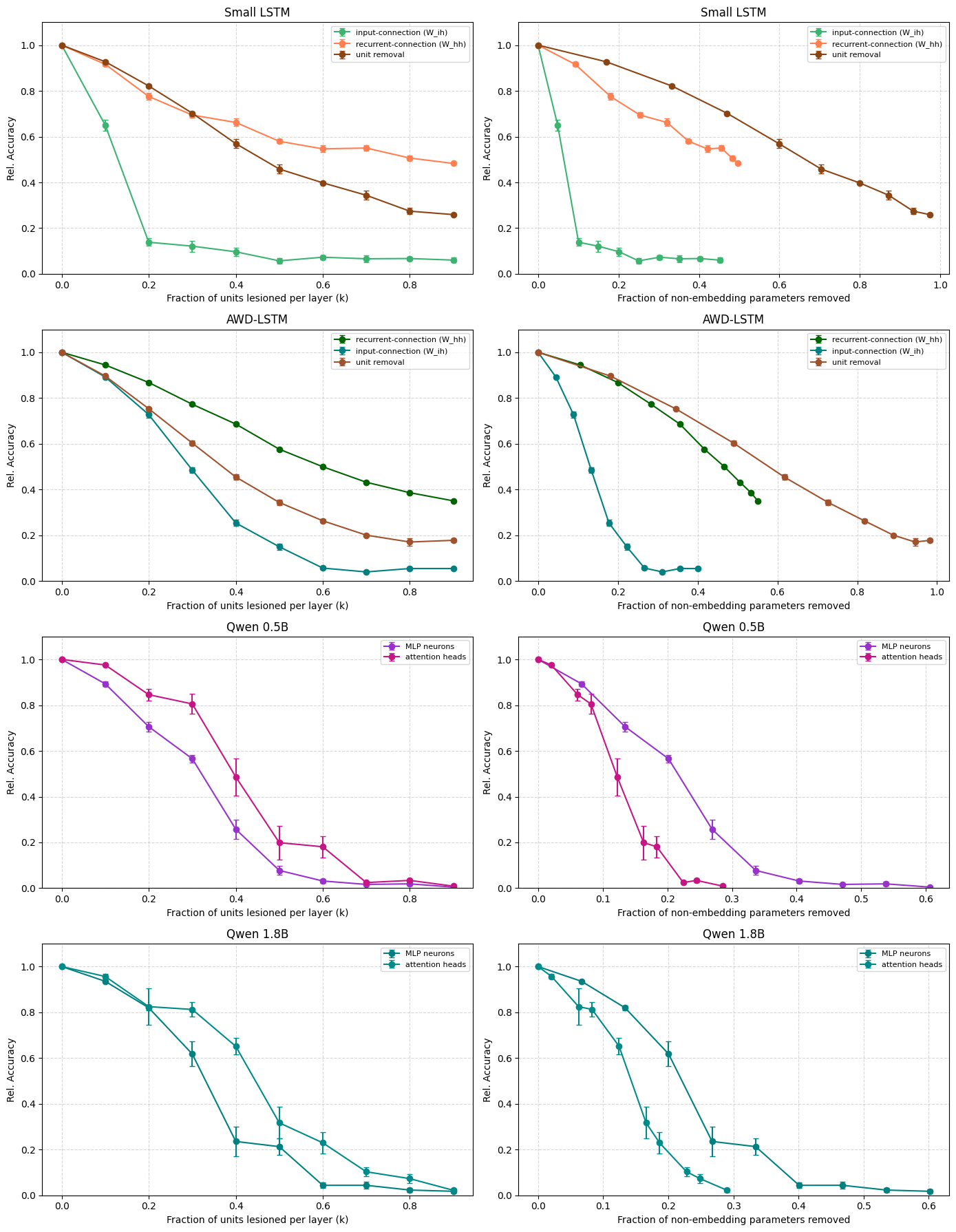}
\caption{Relative accuracy plotted against nominal unit/connection fraction $k$ (left column) and against the exact fraction of non-embedding parameters removed (right column) across all four architectures, comparing whole-unit removal (\texttt{weight\_ih} + \texttt{weight\_hh}, solid brown curves) against recurrent-connection pruning (\texttt{weight\_hh}, coral in row~1 and dark green in row~2) and input-connection pruning (\texttt{weight\_ih}, medium sea green in row~1 and teal in row~2) in both LSTMs, and MLP vs.\ attention-head removal in both Qwen models (rows~3--4). Plotted against parameter fraction, attention-head ablation degrades faster per parameter removed than MLP ablation in both Qwen models, and both Qwen pathways degrade faster per parameter removed than LSTM whole-unit removal and recurrent-connection pruning, which decline monotonically without a cliff.}
\label{sfig:accuracy_param_fraction}
\end{figure}

Table~\ref{stab:summary_k05} provides the full summary of baseline and ablated performance across all four architectures (including both the main-paper whole-unit removal, \texttt{weight\_ih} + \texttt{weight\_hh} + gate biases, and the supplementary pathway-specific connection-pruning controls, \texttt{weight\_hh} recurrent-only and \texttt{weight\_ih} input-only, for the LSTMs) at the benchmark level $k = 0.5$. Table~\ref{stab:curve_shape} reports the curve-shape comparison between a 3-parameter S-curve fit and a 2-parameter straight-line fit against both nominal $k$ and the fraction of non-embedding parameters removed.
Figures~\ref{sfig:accuracy_all} and~\ref{sfig:accuracy_param_fraction} display the complete next-token accuracy degradation curves across $k \in [0, 0.9]$ for the pathway-specific connection controls and parameter-fraction scaling.

\paragraph{Detailed per-architecture accuracy trajectories and LSTM pathway-connection controls (supporting Section~\ref{sec:accuracy}).}
Across the six main-paper unit-level conditions and the four supplementary LSTM connection-pruning controls, next-word prediction accuracy evolves as follows:
\begin{itemize}
    \item \textit{Small LSTM} (baseline 27.5\%): Under whole-unit removal (\texttt{weight\_ih} + \texttt{weight\_hh} + gate biases, main paper Figure~\ref{fig:accuracy}a), next-token accuracy degrades smoothly and monotonically across the entire range without an initial plateau: 25.5\% at $k = 0.1$ ($\mathrm{RelAcc} = 0.927$), 22.6\% at $k = 0.2$ ($0.822$), 19.3\% at $k = 0.3$ ($0.702$), 15.7\% at $k = 0.4$ ($0.570$), 12.6\% at $k = 0.5$ ($0.458$), 10.9\% at $k = 0.6$ ($0.398$), 9.5\% at $k = 0.7$ ($0.344$), 7.5\% at $k = 0.8$ ($0.274$), and 7.12\% at $k = 0.9$ ($0.259$, where all five trials settle on always guessing the single most common word in the text, \emph{``the''}, which makes up 7.12\% of test tokens; Table~\ref{stab:lstm_unit_full}). Dissecting the two incoming pathways separately in our supplementary controls (Figure~\ref{sfig:accuracy_all}, row 1) shows why: hidden-state (\texttt{weight\_hh}) pruning alone tracks whole-unit removal closely through $k = 0.3$ and degrades more gradually thereafter because unit $j$ remains driven by the current input word $x_t$ via \texttt{weight\_ih} (25.2\% at $k = 0.1$, 21.3\% at $k = 0.2$, 15.9\% at $k = 0.5$, and 13.3\% at $k = 0.9$, preserving 48.3\% of baseline accuracy at extreme ablation). When plotted against the fraction of non-embedding parameters removed, this trajectory is nearly a straight line with slope $-1.02$ and the smallest departure from linearity of any condition ($\Delta\mathrm{AIC} = -7.75$; Figure~\ref{sfig:accuracy_param_fraction} and Table~\ref{stab:curve_shape}). Conversely, input-projection (\texttt{weight\_ih}) pruning alone produces a steep initial collapse: accuracy drops to 17.9\% at $k = 0.1$ and 3.8\% at $k = 0.2$, then remains at a near-zero floor between 1.6--3.3\% for $k \geq 0.3$ (e.g., 1.6\% at $k = 0.5$, 1.8\% at $k = 0.8$, and 1.7\% at $k = 0.9$).
    \item \textit{AWD-LSTM} (baseline 38.1\%): Under whole-unit removal (\texttt{weight\_ih} + \texttt{weight\_hh} + gate biases, main paper Figure~\ref{fig:accuracy}b), next-token accuracy likewise degrades smoothly and near-monotonically (apart from a floor at $k \ge 0.8$) without a buffered plateau: 34.1\% at $k = 0.1$ ($\mathrm{RelAcc} = 0.896$), 28.7\% at $k = 0.2$ ($0.753$), 23.0\% at $k = 0.3$ ($0.603$), 17.3\% at $k = 0.4$ ($0.455$), 13.1\% at $k = 0.5$ ($0.344$), 10.0\% at $k = 0.6$ ($0.264$), 7.6\% at $k = 0.7$ ($0.201$), 6.5\% at $k = 0.8$ ($0.171$), and 6.8\% at $k = 0.9$ ($0.178$; Table~\ref{stab:awd_unit_full}). In the supplementary pathway-connection controls (Figure~\ref{sfig:accuracy_all}, row 2), hidden-state (\texttt{weight\_hh}) pruning degrades gradually from 36.0\% at $k = 0.1$ and 33.0\% at $k = 0.2$ to 21.9\% at $k = 0.5$ (57.6\% of baseline) and 13.4\% at $k = 0.9$ (35.1\% of baseline, closely matching the Small LSTM recurrent pathway's 58.0\% relative accuracy at $k = 0.5$; Table~\ref{stab:awd_hidden_full}), whereas input-projection (\texttt{weight\_ih}) pruning drops steeply from 27.7\% at $k = 0.2$ and 18.5\% at $k = 0.3$ to 9.7\% at $k = 0.4$, 5.7\% at $k = 0.5$, and 1.5--2.2\% for $k \geq 0.6$ (Table~\ref{stab:awd_input_full}).
    \item \textit{Qwen 0.5B} (baseline 53.6\%): MLP neuron ablation exhibits a late threshold collapse: accuracy is 89.3\% of baseline at $k = 0.1$ (47.9\%) and 70.6\% at $k = 0.2$ (37.9\%), then collapses sharply between $k = 0.3$ (30.4\%) and $k = 0.4$ (13.8\%), reaching 4.2\% at $k = 0.5$ and near zero (${\approx}1\%$) for $k \geq 0.7$. Attention-head ablation shows an early plateau: accuracy remains at 52.3\% at $k = 0.1$ (97.6\% of baseline), 45.4\% at $k = 0.2$ (84.6\%), and 43.2\% at $k = 0.3$ (80.5\%), before dropping to 26.1\% at $k = 0.4$ and 10.7\% at $k = 0.5$.
    \item \textit{Qwen 1.8B} (baseline 54.7\%): MLP neuron ablation produces a late threshold collapse similar to the 0.5B model: accuracy remains at 51.1\% at $k = 0.1$ (93.5\% of baseline) and 44.8\% at $k = 0.2$ (82.0\% of baseline), before dropping to 33.8\% at $k = 0.3$, collapsing to 12.9\% at $k = 0.4$ and 11.6\% at $k = 0.5$, and falling to 1.0--2.4\% for $k \geq 0.6$. Attention-head ablation follows a smoother sigmoid-like decline: 52.3\% at $k = 0.1$, 45.1\% at $k = 0.2$ (82.5\% of baseline), 44.4\% at $k = 0.3$ (81.2\%), 35.6\% at $k = 0.4$ (65.1\%), 17.3\% at $k = 0.5$ (31.7\%), and 1.2\% at $k = 0.9$. When plotted against the fraction of non-embedding parameters removed rather than nominal $k$, the Qwen MLP and attention curves do not align: attention ablation degrades faster per parameter removed (Qwen~0.5B: $0.486$ of baseline at $12.25\%$ removed and $0.199$ at $16.3\%$ under attention, vs.\ $0.706$ at $13.5\%$ and $0.566$ at $20.2\%$ under MLP; Qwen~1.8B: $0.651$ at $12.4\%$ under attention vs.\ $0.820$ at $13.4\%$ under MLP), and both Qwen pathways lose more accuracy per parameter removed than LSTM whole-unit removal (Small LSTM $0.822$ at $33.25\%$ removed; Supplementary Figure~\ref{sfig:accuracy_param_fraction}).
\end{itemize}

\paragraph{Mild-damage and cross-threshold comparison (supporting the main-text summary).}
At $k = 0.1$ and $0.2$, Qwen MLP relative accuracy ($0.893$ and $0.706$ in Qwen~0.5B; $0.935$ and $0.820$ in Qwen~1.8B) is no higher than that of LSTM whole-unit removal ($0.927$ and $0.822$ in the Small LSTM; $0.896$ and $0.753$ in the AWD-LSTM). Attention-head ablation retains $0.976$ (Qwen~0.5B) and $0.957$ (Qwen~1.8B) of baseline at $k = 0.1$, where only one head per layer is removed, $0.846$ and $0.825$ at $k = 0.2$ (vs.\ $0.822$ and $0.753$ in the Small LSTM and AWD-LSTM), and $0.805$ and $0.812$ at $k = 0.3$ (vs.\ $0.702$ and $0.603$). Across $k = 0.3 \to 0.4$, MLP ablation lowers relative accuracy by $0.31$ in Qwen~0.5B and $0.38$ in Qwen~1.8B, versus $0.13$ and $0.15$ for whole-unit removal in the Small LSTM and AWD-LSTM; MLP relative accuracy is $0.258$ and $0.236$ at $k = 0.4$ and near zero by $k = 0.5$--$0.6$. Under attention ablation, Qwen~0.5B falls steeply once $k > 0.3$ ($0.805 \to 0.486 \to 0.199$ at $k = 0.3$, $0.4$, $0.5$), whereas Qwen~1.8B falls more gently ($0.812 \to 0.651$ from $k = 0.3$ to $0.4$). Per parameter removed, the LSTMs are more robust: the Small LSTM retains $0.822$ of baseline with $33.25\%$ of non-embedding parameters removed, whereas Qwen MLP ablation retains only $0.078$--$0.213$ at about $33.4$--$33.7\%$ removed.

\begin{table}[!htbp]
\centering
\caption{Full ablation summary at $k = 0.5$ across the six primary unit-level conditions and the four supplementary LSTM connection-pruning controls.}
\label{stab:summary_k05}
\resizebox{\textwidth}{!}{%
\begin{tabular}{@{}llcccc@{}}
\toprule
Model & Ablation & Acc@$k{=}0$ & Acc@$k{=}0.5$ & RelAcc & Clinical Parallel \\
\midrule
Small LSTM & Unit (w\_ih + w\_hh) & 0.2747 & 0.1260 & 0.458 & Diffuse AD-type ($\to$ global collapse) \\
AWD-LSTM & Unit (w\_ih + w\_hh) & 0.3806 & 0.1307 & 0.344 & Diffuse AD-type ($\to$ global collapse) \\
Qwen 0.5B & MLP neurons & 0.5363 & 0.0417 & 0.078 & Wernicke-type \\
Qwen 0.5B & Attention heads & 0.5363 & 0.1067 & 0.199 & svPPA \\
Qwen 1.8B & MLP neurons & 0.5465 & 0.1163 & 0.213 & Wernicke-type \\
Qwen 1.8B & Attention heads & 0.5465 & 0.1732 & 0.317 & svPPA \\
\midrule
Small LSTM & Suppl.\ control (w\_hh recurrent) & 0.2747 & 0.1593 & 0.580 & Diffuse AD-type (sustained) \\
AWD-LSTM & Suppl.\ control (w\_hh recurrent) & 0.3806 & 0.2193 & 0.576 & Diffuse AD-type (sustained) \\
Small LSTM & Suppl.\ control (w\_ih input) & 0.2747 & 0.0157 & 0.057 & Input bottleneck collapse \\
AWD-LSTM & Suppl.\ control (w\_ih input) & 0.3806 & 0.0570 & 0.150 & Input bottleneck collapse \\
\bottomrule
\end{tabular}%
}
\end{table}

\begin{table}[!htbp]
\small
\centering
\caption{Curve-shape comparison across trial-level relative accuracy points. Each row compares a 3-parameter S-curve (logistic) fit $y(x) = a/(1+\exp(b(x-x_0)))$ against a 2-parameter straight-line fit $y(x) = a + bx$ as a function of $k$ or fraction of non-embedding parameters removed. Because the 3-parameter curve bends toward zero at high $k$, any trajectory that curves (whether through an initial flat plateau followed by a cliff, a steep early drop, or a smooth bend) achieves $\Delta\mathrm{AIC} = \mathrm{AIC}_{\mathrm{logistic}} - \mathrm{AIC}_{\mathrm{linear}} < -2$ (labelled \emph{non-linear} in the table). The rightmost column notes the qualitative shape seen in Figures~\ref{fig:accuracy} and~\ref{sfig:accuracy_param_fraction}.}
\label{stab:curve_shape}
\resizebox{\textwidth}{!}{%
\begin{tabular}{@{}llcccc@{}}
\toprule
Model & Lesion & Axis $x$ & Linear Slope & $\Delta\mathrm{AIC}$ & Fit Verdict (Qualitative Shape) \\
\midrule
Small LSTM & Unit removal (w\_ih + w\_hh) & $k$ & $-0.888$ & $-39.04$ & non-linear (gradual bend) \\
Small LSTM & Unit removal (w\_ih + w\_hh) & frac.\ params & $-0.812$ & $-23.71$ & non-linear (gradual bend) \\
Small LSTM & Recurrent-connection (w\_hh) & $k$ & $-0.556$ & $-26.23$ & non-linear (gradual bend) \\
Small LSTM & Recurrent-connection (w\_hh) & frac.\ params & $-1.019$ & $-7.75$ & non-linear (nearly straight line) \\
Small LSTM & Input-connection (w\_ih) & $k$ & $-0.793$ & $-104.98$ & non-linear (steep early drop) \\
Small LSTM & Input-connection (w\_ih) & frac.\ params & $-1.577$ & $-106.07$ & non-linear (steep early drop) \\
\midrule
AWD-LSTM & Unit removal (w\_ih + w\_hh) & $k$ & $-0.992$ & $-79.52$ & non-linear (gradual bend) \\
AWD-LSTM & Unit removal (w\_ih + w\_hh) & frac.\ params & $-0.914$ & $-38.00$ & non-linear (gradual bend) \\
AWD-LSTM & Recurrent-connection (w\_hh) & $k$ & $-0.779$ & $-38.33$ & non-linear (gradual bend) \\
AWD-LSTM & Recurrent-connection (w\_hh) & frac.\ params & $-1.231$ & $-144.68$ & non-linear (gradual bend) \\
AWD-LSTM & Input-connection (w\_ih) & $k$ & $-1.163$ & $-144.12$ & non-linear (steep early drop) \\
AWD-LSTM & Input-connection (w\_ih) & frac.\ params & $-2.617$ & $-143.97$ & non-linear (steep early drop) \\
\midrule
Qwen 0.5B & MLP neurons & $k$ & $-1.231$ & $-100.79$ & non-linear (late threshold collapse) \\
Qwen 0.5B & MLP neurons & frac.\ params & $-1.828$ & $-100.90$ & non-linear (late threshold collapse) \\
Qwen 0.5B & Attention heads & $k$ & $-1.320$ & $-31.38$ & non-linear (broad plateau + S-curve) \\
Qwen 0.5B & Attention heads & frac.\ params & $-4.130$ & $-30.69$ & non-linear (compressed plateau / S-curve) \\
\midrule
Qwen 1.8B & MLP neurons & $k$ & $-1.264$ & $-60.02$ & non-linear (late threshold collapse) \\
Qwen 1.8B & MLP neurons & frac.\ params & $-1.891$ & $-59.99$ & non-linear (late threshold collapse) \\
Qwen 1.8B & Attention heads & $k$ & $-1.253$ & $-18.71$ & non-linear (broad plateau + S-curve) \\
Qwen 1.8B & Attention heads & frac.\ params & $-3.853$ & $-17.07$ & non-linear (compressed plateau / S-curve) \\
\bottomrule
\end{tabular}%
}
\end{table}

\subsection{Detailed \texorpdfstring{$R_4$}{R4} Trajectories, Scale-Invariance Derivation, and Accuracy-Matched Null Model Analysis}\label{sec:supp_null_model}

\begin{figure}[!htbp]
\centering
\includegraphics[width=0.78\textwidth,height=0.58\textheight,keepaspectratio]{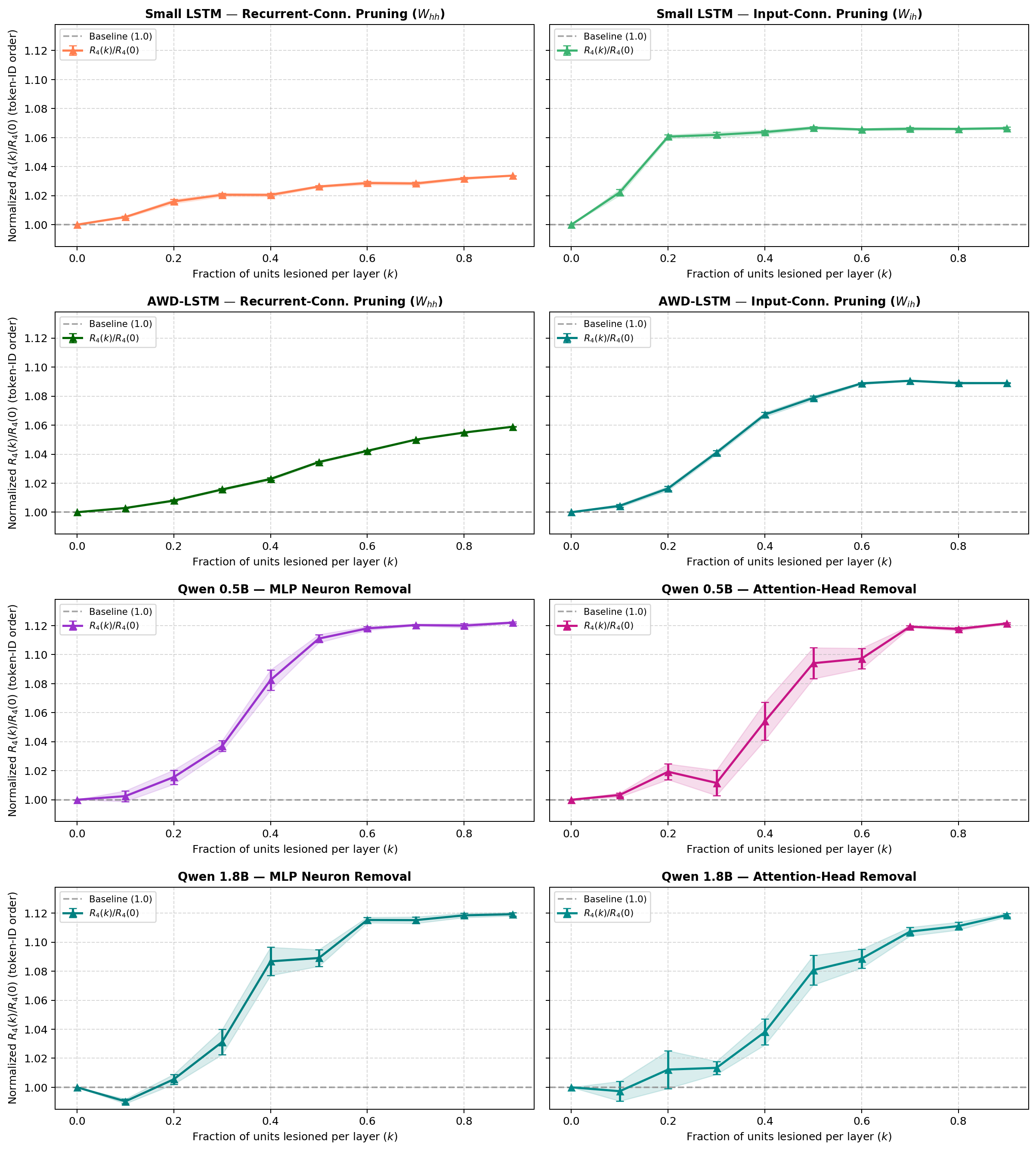}
\caption{Moment ratio $R_4(k)/R_4(0)$ for the supplementary LSTM connection-pruning pathways (top two rows: Small LSTM and AWD-LSTM recurrent-connection pruning \texttt{weight\_hh} in the left column and input-connection pruning \texttt{weight\_ih} in the right column) alongside Qwen~0.5B and Qwen~1.8B MLP and Attention pathways (bottom two rows; shared vertical scale; means $\pm 1$ SE; primary LSTM whole-unit $R_4$ curves are shown in Figure~\ref{fig:R4}a,b).}
\label{sfig:R4_all}
\end{figure}

\paragraph{Raw $R_4$ trajectories and scale invariance (supporting Section~\ref{sec:moments}).}
Plotted against $k$, the raw ratio $R_4(k)/R_4(0)$ rises above $1.0$ in a pattern that mirrors each architecture's accuracy drop (Figure~\ref{fig:R4}; Supplementary Figure~\ref{sfig:R4_all}).
Under whole-unit removal, the Small LSTM's $R_4$ rises smoothly by at most 5.14\% across the entire ablation range ($2.2575 \to 2.3735$ at $k = 0.9$; Table~\ref{stab:lstm_unit_full}), and the AWD-LSTM's by at most 7.66\% ($2.2084 \to 2.3775$ at $k = 0.8$, $2.3758$ at $k = 0.9$; Table~\ref{stab:awd_unit_full}).
In the supplementary LSTM connection-pruning controls (Supplementary Figure~\ref{sfig:R4_all}, top two rows), recurrent-connection (\texttt{weight\_hh}) pruning rises by at most 3.38\% in the Small LSTM ($2.2575 \to 2.3339$; Table~\ref{stab:lstm_hidden_full}) and 5.89\% in the AWD-LSTM ($2.2084 \to 2.3385$; Table~\ref{stab:awd_hidden_full}), whereas input-connection (\texttt{weight\_ih}) pruning reaches 6.68\% (Small LSTM; Table~\ref{stab:lstm_input_full}) and 9.06\% (AWD-LSTM; Table~\ref{stab:awd_input_full}).
The Qwen models show the largest raw shifts (up to 12.21\% for Qwen~0.5B MLP, 12.15\% for Qwen~0.5B Attention, 11.93\% for Qwen~1.8B MLP, and 11.87\% for Qwen~1.8B Attention; Tables~\ref{stab:qwen05_mlp_full}--\ref{stab:qwen18_attn_full}), rising gradually at mild damage (MLP: $+3.7\%$ and $+3.1\%$ by $k = 0.3$ in Qwen~0.5B and 1.8B) and by about $5$ further points across the MLP threshold cliff ($k = 0.3 \to 0.4$: $1.0370 \to 1.0825$ in Qwen~0.5B and $1.0312 \to 1.0868$ in Qwen~1.8B, i.e.\ $+4.6$ and $+5.6$ points).
Because $R_4$ is computed from the normalised distribution $p_i$, it is strictly invariant if every word's error count scales up by the same factor ($\mathbf{e}_{\mathrm{abl}} = \alpha \cdot \mathbf{e}_{\mathrm{base}}$, $\alpha > 0$):
\begin{equation}
    R_4(\mathrm{abl}) = \frac{\left(\sum_i i^4 \frac{\alpha e_i}{\alpha \sum_j e_j}\right)^{1/4}}{\left(\sum_i i^2 \frac{\alpha e_i}{\alpha \sum_j e_j}\right)^{1/2}} = R_4(\mathrm{base})\,.
\end{equation}
Raw $R_4$ rises as accuracy falls because an undamaged model mostly gets common, low-index words right and misses harder words; as damage increases and the model misses many common words, the average error index in the denominator ($\mu_2^{1/2}$) drops faster than the extreme tail in the numerator ($\mu_4^{1/4}$), pushing $R_4$ upward.

\paragraph{Accuracy-matched null model breakdown.}
Subtracting the accuracy-matched random-failure baseline (Equations~\ref{eq:null_model}--\ref{eq:js_div}; Figure~\ref{fig:null_model_comparison}) yields three detailed quantitative findings across the six primary unit-level conditions and four supplementary connection-pruning controls:
\begin{enumerate}
    \item \textit{Negative null-corrected $\Delta R_4 = R_4^{\mathrm{obs}} - R_4^{\mathrm{null}}$ across all conditions}: Whereas raw $R_4(k)/R_4(0)$ increases as accuracy drops, the null-corrected difference $\Delta R_4$ is consistently \emph{negative} across all ten conditions. At matched relative accuracy of $0.80$, $\Delta R_4$ is $-0.010$ ($95\%$ CI $[-0.015, -0.005]$) for Small LSTM whole-unit removal ($-0.004$ to $-0.007$ for \texttt{weight\_ih}/\texttt{weight\_hh}), $-0.030$ ($[-0.033, -0.028]$) for AWD-LSTM whole-unit removal ($-0.021$ to $-0.026$ for \texttt{weight\_hh}/\texttt{weight\_ih}), and $-0.037$ to $-0.052$ for the Qwen pathways. Because the undamaged model's correct predictions $\mathbf{c}(0)$ are heavily concentrated on the most common, lowest-index tokens ($i \approx 0$), random failure dumps new errors onto $i \approx 0$ and drives $R_4^{\mathrm{null}}$ strongly upward. Relative to this null, neuronal ablation is less concentrated at $i \approx 0$ ($\Delta R_4 < 0$) and instead spreads error mass across a wider effective number of distinct tokens ($\Delta D_1 = D_1^{\mathrm{obs}} - D_1^{\mathrm{null}} > 0$ across all conditions). At $\mathrm{RelAcc} = 0.80$, $\Delta D_1$ is $+192.8$ ($[+180.9, +204.6]$) for Small LSTM unit removal and $+249.8$ ($[+222.4, +277.2]$) for AWD-LSTM unit removal, versus $+15.0$ to $+21.5$ tokens in Qwen. This raw token-count difference reflects the larger WikiText-2 evaluation set ($D_1^{\mathrm{null}} \approx 1{,}354$--$1{,}359$ vs.\ $278$--$288$ on TinyStories), corresponding to relative increases $\Delta D_1 / D_1^{\mathrm{null}}$ of $+14.2\%$ to $+18.4\%$ in the LSTMs versus $+5.2\%$ to $+7.7\%$ in Qwen. As $\mathrm{RelAcc} \to 0$, both $\mathbf{e}^{\mathrm{obs}}(k)$ and $\mathbf{e}^{\mathrm{null}}(k)$ approach the full test-corpus unigram vector $\mathbf{e}(0)+\mathbf{c}(0)$, pinning $\Delta R_4$, $\Delta D_1$, and $\Delta\mathrm{JS}$ to zero at both endpoints. Negative $\Delta R_4$ already appears at mild damage ($k = 0.1$) in the LSTMs (e.g., $-0.021$ [$-0.024$, $-0.018$] for AWD-LSTM whole-unit removal), whereas it is not significant in either Qwen attention condition at $k = 0.1$ ($-0.003$ [$-0.009$, $+0.003$] in Qwen~0.5B; $-0.021$ [$-0.062$, $+0.020$] in Qwen~1.8B).
    \item \textit{Separation of MLP and attention pathways under null correction}: Comparing the uncorrected ratio $R_4(k)/R_4(0)$ across the common relative-accuracy range ($[0.207, 0.935]$ for Qwen~0.5B, $56/60$ MLP and $45/60$ attention trials, $n = 101$; $[0.301, 0.948]$ for Qwen~1.8B, $49/60$ MLP and $48/60$ attention trials, $n = 97$) yields small, non-significant differences between attention and MLP ablation ($+0.0040$, permutation $p = 0.5009$ in Qwen~0.5B; $+0.0020$, $p = 0.7130$ in Qwen~1.8B). When the random-failure baseline is subtracted, however, the normalised difference $(R_4^{\mathrm{obs}} - R_4^{\mathrm{null}})/R_4(0)$ pooled across that shared transition range separates the two pathways significantly in both models: in Qwen~0.5B, attention ablation shifts by $-0.0109$ vs.\ $-0.0242$ for MLP (difference $+0.0132$, two-sided permutation $p < 10^{-4}$); in Qwen~1.8B, attention shifts by $-0.0209$ vs.\ $-0.0251$ for MLP (difference $+0.0042$, permutation $p = 0.0238$). Note that this test pools trial points across the full transition range (where MLP trials have lower mean relative accuracy than attention trials); when interpolated at a single fixed slice ($\mathrm{RelAcc} = 0.80$), the point-level 95\% confidence intervals of $\Delta R_4$ for the two pathways overlap ($-0.048$, 95\% CI $[-0.066, -0.030]$ for MLP vs.\ $-0.042$, $[-0.064, -0.020]$ for Attention in Qwen~0.5B; $-0.052$, $[-0.078, -0.025]$ for MLP vs.\ $-0.037$, $[-0.053, -0.022]$ for Attention in Qwen~1.8B). Because the pooled trials differ in mean relative accuracy ($0.618$ for MLP vs.\ $0.689$ for attention in Qwen~0.5B; $0.734$ vs.\ $0.755$ in Qwen~1.8B) and no multiplicity correction is applied, we treat the pooled test as indicative rather than as a relative-accuracy-matched comparison.
    \item \textit{Architectural separation in Jensen--Shannon divergence}: The Jensen--Shannon divergence $\mathrm{JS}(p(k) \,\|\, p(0))$ from the intact error distribution grows to $0.034$--$0.045$ at relative accuracy $0.80$ ($0.070$--$0.077$ at relative accuracy $0.50$) in the Qwen models, larger than in LSTM whole-unit removal ($\mathrm{JS} = 0.007$ in Small LSTM unit and $0.008$ in AWD-LSTM unit at relative accuracy $0.80$; $0.025$ in both at relative accuracy $0.50$) and LSTM recurrent-connection pruning ($0.006$--$0.007$ at relative accuracy $0.80$; $0.015$--$0.019$ at relative accuracy $0.50$). After subtracting the accuracy-matched null ($\mathrm{JS}^{\mathrm{null}} = 0.024$--$0.025$ for Qwen vs.\ $0.004$--$0.006$ for the LSTMs at relative accuracy $0.80$), the excess remains larger in Qwen ($\Delta\mathrm{JS} = +0.015$ [$+0.006$, $+0.024$] and $+0.010$ [$+0.001$, $+0.019$] for Qwen~0.5B MLP and attention; $+0.020$ [$+0.010$, $+0.029$] and $+0.018$ [$+0.011$, $+0.025$] for Qwen~1.8B MLP and attention) than in LSTM whole-unit removal ($+0.003$ [$+0.002$, $+0.003$] Small LSTM; $+0.002$ [$+0.001$, $+0.003$] AWD-LSTM) or recurrent-connection pruning ($+0.002$ [$+0.000$, $+0.004$] Small LSTM; $+0.001$ [$+0.000$, $+0.001$] AWD-LSTM), so Transformer unit ablation reshapes the error distribution more than LSTM ablation at equal accuracy loss, although the absolute excess is small.
\end{enumerate}

\subsection{Spectral Analysis of Error Distributions and Complete Per-\texorpdfstring{$k$}{k} Metrics Tables Across All Ten Conditions}\label{sec:supp_spectral_details}\label{sec:supp_awd_metrics}\label{sec:spectral}

\paragraph{Definition and methodology.}
To measure whether prediction errors vary smoothly or jaggedly across word-frequency ranks, we reorder the vocabulary by corpus frequency across all four architectures (most frequent word $= 0$ across the top $N = 1{,}000$ lexical tokens, ranked on the WikiText-103 training split for the Small LSTM, the WikiText-2 test split for the AWD-LSTM, and the first $2{,}000$ TinyStories passages (passages 1--2,000) for the Qwen models, excluding fastai control markers for the LSTMs) and compute the discrete Fourier power spectrum $P(f) = |\hat{e}(f)|^2$ of the reordered error counts:
\begin{equation}
    \hat{e}(f) = \sum_{i=0}^{N-1} e_i \, e^{-2\pi i f \cdot i / N}\,, \qquad
    G = \log_2 \frac{\sum_{f=1}^{\lfloor N/4 \rfloor} P(f)}{\sum_{f=\lfloor N/4 \rfloor+1}^{\lfloor N/2 \rfloor} P(f)}\,.
\end{equation}
We summarise the spectrum by its log--log slope $\beta$ ($\log P(f) \approx \alpha + \beta \log f$, estimated by ordinary least squares regression of $\log P(f)$ on $\log f$ for $f \in [1, \lfloor N/2 \rfloor]$) and the low-to-high frequency spectral gap $G$: a more negative slope $\beta$ and a larger gap $G$ indicate that error counts vary smoothly across neighbouring word-frequency ranks rather than jumping sharply from word to word. Because the AWD-LSTM rank source is the WikiText-2 test split and the Qwen rank source (passages 1--2,000) includes the Qwen evaluation passages (passages 501--1,000), both overlap their evaluation sets; however, because this ranking only orders tokens for the auxiliary Fourier analysis, the overlap affects only the spectral metrics and none of the accuracy, $R_4$, null-model, Cloze, or error-type results.

\begin{figure}[!htbp]
\centering
\includegraphics[width=0.92\textwidth,height=0.58\textheight,keepaspectratio]{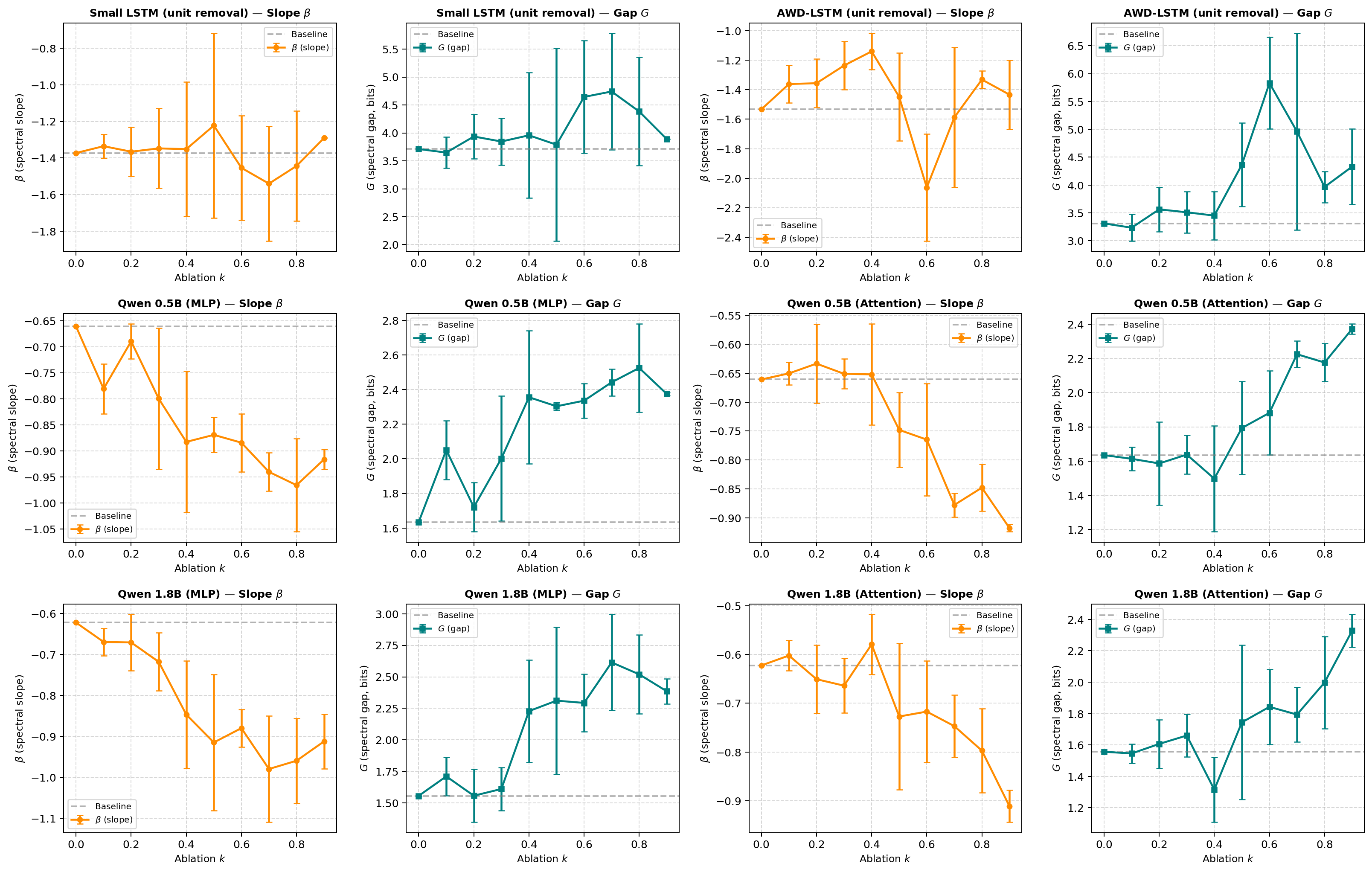}
\caption{Fourier spectral analysis of the frequency-ranked error distribution across unit-ablation fractions $k \in [0, 0.9]$ for the six primary unit-level conditions (means $\pm 1$ SD): within each panel, the left graph shows spectral slope $\beta$ (orange) and the right graph shows the low-to-high-frequency spectral gap $G$ (teal). Row 1: Small LSTM whole-unit removal (left) and AWD-LSTM whole-unit removal (right). Rows 2--3: Qwen~0.5B and Qwen~1.8B MLP neuron ablation (left) and attention-head ablation (right), where Qwen~1.8B and Qwen~0.5B attention remain buffered through $k \le 0.3$--$0.4$ before shifting at higher $k$, while Qwen~0.5B MLP and AWD-LSTM whole-unit removal show non-monotonic fluctuations across $k$ (per-$k$ values in Tables~\ref{stab:awd_unit_full}--\ref{stab:qwen18_attn_full}).}
\label{fig:spectral}
\end{figure}

\begin{figure}[!htbp]
\centering
\includegraphics[width=0.88\textwidth,height=0.56\textheight,keepaspectratio]{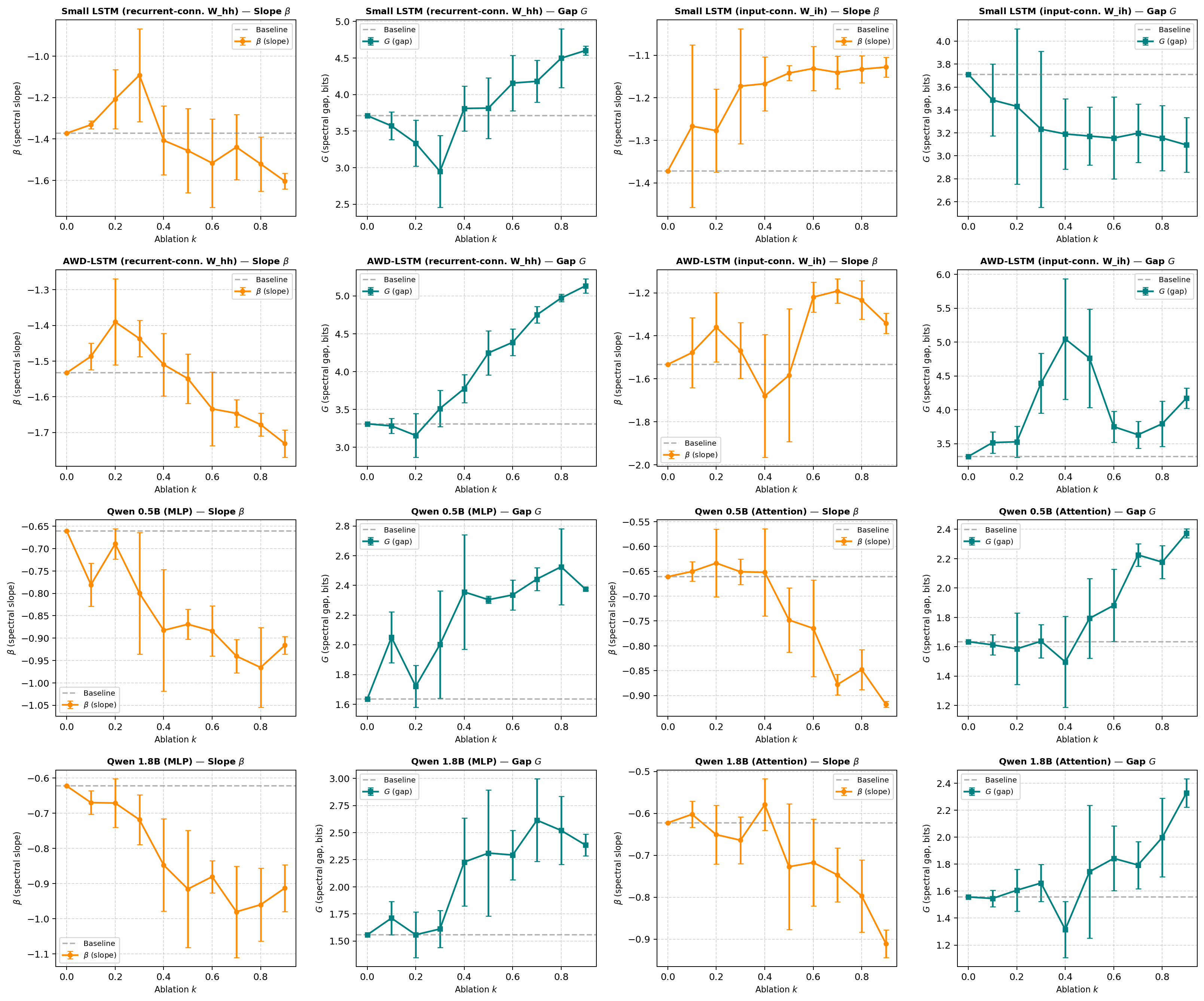}
\caption{Spectral slope $\beta$ (orange, left graph of each pair) and spectral gap $G$ (teal, right graph of each pair) across $k \in [0, 0.9]$ for the supplementary LSTM connection-pruning controls (top two rows: Small LSTM and AWD-LSTM recurrent-connection pruning \texttt{weight\_hh} in the left column and input-connection pruning \texttt{weight\_ih} in the right column) alongside Qwen~0.5B and Qwen~1.8B MLP and Attention pathways (bottom two rows; means $\pm 1$ SD).}
\label{sfig:spectral_all}
\end{figure}

\paragraph{Detailed spectral trajectories across architectures.}
The Fourier spectral metrics $\beta$ and $G$ track how smoothly prediction errors vary across word-frequency ranks as ablation increases (Figures~\ref{fig:spectral} and~\ref{sfig:spectral_all}; Tables~\ref{stab:awd_unit_full}--\ref{stab:qwen18_attn_full}). They also show why accuracy-matched null correction (Section~\ref{sec:moments}) is needed when comparing pathways.

Both LSTMs start with steep baseline spectra ($\beta = -1.37$, $G = 3.71$ for the Small LSTM; $\beta = -1.53$, $G = 3.31$ for the AWD-LSTM). Under whole-unit removal (Figure~\ref{fig:spectral}, top row; Tables~\ref{stab:awd_unit_full}--\ref{stab:lstm_unit_full}), the Small LSTM spectrum changes little across mild-to-moderate damage ($\beta$ between $-1.37$ and $-1.22$, $G$ between $3.65$ and $3.96$ for $k \in [0, 0.5]$), steepens to $\beta = -1.54$ and $G = 4.74$ at $k = 0.7$, and settles at $\beta = -1.29$ and $G = 3.89$ at $k = 0.9$. In the AWD-LSTM, whole-unit removal produces a non-monotonic curve: $\beta$ first flattens from $-1.53$ ($G = 3.31$) at $k = 0$ to $-1.14$ ($G = 3.45$) at $k = 0.4$, then steepens sharply at $k = 0.6$ ($\beta = -2.06$, $G = 5.83$) as Cloze output collapses, and finally moves back toward $\beta = -1.43$ and $G = 4.33$ at $k = 0.9$.

In the supplementary LSTM connection-pruning controls (Figure~\ref{sfig:spectral_all}, top two rows; Tables~\ref{stab:awd_hidden_full}--\ref{stab:lstm_input_full}), recurrent-connection (\texttt{weight\_hh}) pruning in the AWD-LSTM flattens the spectrum slightly at mild damage ($\beta = -1.39$, $G = 3.16$ at $k = 0.2$) before steepening steadily to $\beta = -1.73$ and $G = 5.13$ at $k = 0.9$. Small LSTM \texttt{weight\_hh} pruning follows a similar path, moving from $\beta = -1.09$ and $G = 2.95$ at $k = 0.3$ to $\beta = -1.60$ and $G = 4.60$ at $k = 0.9$ (Table~\ref{stab:lstm_hidden_full}). By contrast, input-connection (\texttt{weight\_ih}) pruning flattens the Small LSTM spectrum to $\beta = -1.13$ and $G = 3.10$ at $k = 0.9$ (Table~\ref{stab:lstm_input_full}), whereas in the AWD-LSTM it causes an early steepening ($\beta = -1.68$, $G = 5.05$ at $k = 0.4$) followed by a partial reversal at $k = 0.9$ ($\beta = -1.34$, $G = 4.17$; Table~\ref{stab:awd_input_full}).

The Qwen models start with flatter baseline spectra ($\beta = -0.66$, $G = 1.63$ for Qwen~0.5B; $\beta = -0.62$, $G = 1.56$ for Qwen~1.8B; Tables~\ref{stab:qwen05_mlp_full}--\ref{stab:qwen18_attn_full}). In Qwen~1.8B (Figure~\ref{fig:spectral}, row 3), MLP ablation stays near baseline through $k = 0.3$ ($\beta = -0.72$, $G = 1.61$) before steepening at $k = 0.4$ ($\beta = -0.85$, $G = 2.23$) and $k = 0.7$ ($\beta = -0.98$, $G = 2.61$). Attention-head ablation in Qwen~1.8B stays near baseline through $k = 0.4$ ($\beta = -0.58$, $G = 1.32$) and steepens to $\beta = -0.91$ and $G = 2.33$ by $k = 0.9$. In Qwen~0.5B (Figure~\ref{fig:spectral}, row 2), attention-head ablation similarly stays near baseline through $k = 0.4$ ($\beta = -0.65$, $G = 1.50$) and steepens for $k \ge 0.5$ ($\beta = -0.75$ to $-0.92$). Under Qwen~0.5B MLP ablation, the spectrum fluctuates across mild damage ($\beta = -0.78$, $G = 2.05$ at $k = 0.1$; $\beta = -0.69$, $G = 1.72$ at $k = 0.2$; $\beta = -0.80$, $G = 2.00$ at $k = 0.3$) and then settles at $\beta = -0.87$ to $-0.97$ ($G = 2.30$--$2.53$) for $k \ge 0.4$.

Because raw $\beta$ and $G$ do not control for total error count and show non-monotonic fluctuations and wider trial-to-trial variation than $R_4$, we base our main cross-pathway comparisons on accuracy-matched, null-corrected $\Delta R_4$ and Jensen--Shannon divergence (Contribution~2; Section~\ref{sec:moments}).

\begin{table}[!htbp]
\centering
\caption{AWD-LSTM Whole-Unit Removal (\texttt{weight\_ih} + \texttt{weight\_hh} + biases): full metrics.}
\label{stab:awd_unit_full}
\resizebox{\textwidth}{!}{%
\begin{tabular}{@{}cccccccccc@{}}
\toprule
$k$ & Accuracy & $\pm$SE & RelAcc & $R_4$ & $\pm R_4$ SE & $R_4/R_4(0)$ & $\beta$ & $\pm\beta$ SD & $G$ \\
\midrule
0.00 & 0.3806 & 0.0000 & 1.0000 & 2.2084 & 0.0000 & 1.0000 & $-$1.5328 & 0.0002 & 3.3106 \\
0.10 & 0.3410 & 0.0007 & 0.8960 & 2.2153 & 0.0014 & 1.0031 & $-$1.3621 & 0.1268 & 3.2343 \\
0.20 & 0.2866 & 0.0009 & 0.7531 & 2.2355 & 0.0009 & 1.0123 & $-$1.3563 & 0.1657 & 3.5633 \\
0.30 & 0.2295 & 0.0041 & 0.6030 & 2.2684 & 0.0028 & 1.0272 & $-$1.2366 & 0.1620 & 3.5111 \\
0.40 & 0.1731 & 0.0046 & 0.4548 & 2.3059 & 0.0031 & 1.0441 & $-$1.1410 & 0.1227 & 3.4511 \\
0.50 & 0.1307 & 0.0049 & 0.3435 & 2.3345 & 0.0033 & 1.0571 & $-$1.4492 & 0.2976 & 4.3659 \\
0.60 & 0.1003 & 0.0031 & 0.2635 & 2.3548 & 0.0020 & 1.0663 & $-$2.0622 & 0.3626 & 5.8311 \\
0.70 & 0.0763 & 0.0021 & 0.2006 & 2.3703 & 0.0013 & 1.0733 & $-$1.5878 & 0.4728 & 4.9594 \\
0.80 & 0.0649 & 0.0061 & 0.1707 & 2.3775 & 0.0039 & 1.0766 & $-$1.3316 & 0.0604 & 3.9658 \\
0.90 & 0.0678 & 0.0026 & 0.1780 & 2.3758 & 0.0016 & 1.0758 & $-$1.4336 & 0.2332 & 4.3306 \\
\bottomrule
\end{tabular}%
}
\end{table}

\begin{table}[!htbp]
\centering
\caption{Small LSTM Whole-Unit Removal (\texttt{weight\_ih} + \texttt{weight\_hh} + biases): full metrics. At $k = 0.90$, all five trials settle on predicting the single most common word in the test text (\emph{``the''}, $7.12\%$ of tokens), giving zero variation across trials ($\pm 0.0000$).}
\label{stab:lstm_unit_full}
\resizebox{\textwidth}{!}{%
\begin{tabular}{@{}cccccccccc@{}}
\toprule
$k$ & Accuracy & $\pm$SE & RelAcc & $R_4$ & $\pm R_4$ SE & $R_4/R_4(0)$ & $\beta$ & $\pm\beta$ SD & $G$ \\
\midrule
0.00 & 0.2747 & 0.0000 & 1.0000 & 2.2575 & 0.0000 & 1.0000 & $-$1.3724 & 0.0001 & 3.7113 \\
0.10 & 0.2548 & 0.0008 & 0.9274 & 2.2661 & 0.0005 & 1.0038 & $-$1.3352 & 0.0652 & 3.6480 \\
0.20 & 0.2257 & 0.0015 & 0.8215 & 2.2810 & 0.0017 & 1.0104 & $-$1.3652 & 0.1346 & 3.9358 \\
0.30 & 0.1930 & 0.0011 & 0.7024 & 2.2955 & 0.0014 & 1.0168 & $-$1.3473 & 0.2175 & 3.8458 \\
0.40 & 0.1565 & 0.0052 & 0.5695 & 2.3177 & 0.0035 & 1.0267 & $-$1.3513 & 0.3677 & 3.9571 \\
0.50 & 0.1260 & 0.0057 & 0.4585 & 2.3377 & 0.0038 & 1.0355 & $-$1.2225 & 0.5036 & 3.7887 \\
0.60 & 0.1094 & 0.0015 & 0.3981 & 2.3488 & 0.0010 & 1.0404 & $-$1.4535 & 0.2858 & 4.6439 \\
0.70 & 0.0945 & 0.0053 & 0.3438 & 2.3585 & 0.0035 & 1.0448 & $-$1.5397 & 0.3136 & 4.7413 \\
0.80 & 0.0754 & 0.0037 & 0.2743 & 2.3709 & 0.0024 & 1.0502 & $-$1.4432 & 0.3002 & 4.3827 \\
0.90 & 0.0712 & 0.0000 & 0.2593 & 2.3735 & 0.0000 & 1.0514 & $-$1.2903 & 0.0000 & 3.8883 \\
\bottomrule
\end{tabular}%
}
\end{table}

\begin{table}[!htbp]
\centering
\caption{AWD-LSTM Hidden (\texttt{weight\_hh}) connection pruning (supplementary control 1): full metrics.}
\label{stab:awd_hidden_full}
\resizebox{\textwidth}{!}{%
\begin{tabular}{@{}cccccccccc@{}}
\toprule
$k$ & Accuracy & $\pm$SE & RelAcc & $R_4$ & $\pm R_4$ SE & $R_4/R_4(0)$ & $\beta$ & $\pm\beta$ SD & $G$ \\
\midrule
0.00 & 0.3806 & 0.0000 & 1.0000 & 2.2084 & 0.0000 & 1.0000 & $-$1.5327 & 0.0002 & 3.3103 \\
0.10 & 0.3595 & 0.0005 & 0.9445 & 2.2148 & 0.0005 & 1.0029 & $-$1.4866 & 0.0375 & 3.2828 \\
0.20 & 0.3303 & 0.0018 & 0.8678 & 2.2260 & 0.0014 & 1.0080 & $-$1.3900 & 0.1209 & 3.1562 \\
0.30 & 0.2941 & 0.0018 & 0.7727 & 2.2431 & 0.0014 & 1.0157 & $-$1.4368 & 0.0510 & 3.5124 \\
0.40 & 0.2611 & 0.0022 & 0.6861 & 2.2591 & 0.0019 & 1.0229 & $-$1.5096 & 0.0874 & 3.7748 \\
0.50 & 0.2193 & 0.0010 & 0.5763 & 2.2848 & 0.0012 & 1.0346 & $-$1.5491 & 0.0695 & 4.2473 \\
0.60 & 0.1902 & 0.0014 & 0.4997 & 2.3019 & 0.0011 & 1.0423 & $-$1.6339 & 0.1035 & 4.3868 \\
0.70 & 0.1644 & 0.0009 & 0.4319 & 2.3189 & 0.0007 & 1.0500 & $-$1.6463 & 0.0383 & 4.7502 \\
0.80 & 0.1470 & 0.0010 & 0.3862 & 2.3297 & 0.0007 & 1.0549 & $-$1.6780 & 0.0321 & 4.9724 \\
0.90 & 0.1335 & 0.0007 & 0.3507 & 2.3385 & 0.0005 & 1.0589 & $-$1.7311 & 0.0386 & 5.1322 \\
\bottomrule
\end{tabular}%
}
\end{table}

\begin{table}[!htbp]
\centering
\caption{AWD-LSTM Input (\texttt{weight\_ih}) connection pruning (supplementary control 2): full metrics.}
\label{stab:awd_input_full}
\resizebox{\textwidth}{!}{%
\begin{tabular}{@{}cccccccccc@{}}
\toprule
$k$ & Accuracy & $\pm$SE & RelAcc & $R_4$ & $\pm R_4$ SE & $R_4/R_4(0)$ & $\beta$ & $\pm\beta$ SD & $G$ \\
\midrule
0.00 & 0.3806 & 0.0000 & 1.0000 & 2.2085 & 0.0000 & 1.0000 & $-$1.5329 & 0.0001 & 3.3108 \\
0.10 & 0.3395 & 0.0020 & 0.8921 & 2.2182 & 0.0023 & 1.0044 & $-$1.4784 & 0.1623 & 3.5164 \\
0.20 & 0.2768 & 0.0047 & 0.7274 & 2.2448 & 0.0030 & 1.0164 & $-$1.3600 & 0.1614 & 3.5278 \\
0.30 & 0.1847 & 0.0045 & 0.4852 & 2.2993 & 0.0031 & 1.0411 & $-$1.4685 & 0.1298 & 4.3924 \\
0.40 & 0.0966 & 0.0051 & 0.2538 & 2.3572 & 0.0033 & 1.0673 & $-$1.6801 & 0.2852 & 5.0464 \\
0.50 & 0.0570 & 0.0049 & 0.1498 & 2.3825 & 0.0031 & 1.0788 & $-$1.5833 & 0.3095 & 4.7614 \\
0.60 & 0.0216 & 0.0023 & 0.0567 & 2.4046 & 0.0014 & 1.0888 & $-$1.2200 & 0.0701 & 3.7510 \\
0.70 & 0.0152 & 0.0013 & 0.0399 & 2.4086 & 0.0008 & 1.0906 & $-$1.1914 & 0.0569 & 3.6317 \\
0.80 & 0.0209 & 0.0021 & 0.0548 & 2.4051 & 0.0013 & 1.0890 & $-$1.2331 & 0.0894 & 3.7939 \\
0.90 & 0.0209 & 0.0011 & 0.0548 & 2.4051 & 0.0007 & 1.0890 & $-$1.3414 & 0.0471 & 4.1736 \\
\bottomrule
\end{tabular}%
}
\end{table}

\begin{table}[!htbp]
\centering
\caption{Small LSTM Recurrent-Connection Pruning (\texttt{weight\_hh}, supplementary control 1): full metrics.}
\label{stab:lstm_hidden_full}
\resizebox{\textwidth}{!}{%
\begin{tabular}{@{}cccccccccc@{}}
\toprule
$k$ & Accuracy & $\pm$SE & RelAcc & $R_4$ & $\pm R_4$ SE & $R_4/R_4(0)$ & $\beta$ & $\pm\beta$ SD & $G$ \\
\midrule
0.00 & 0.2747 & 0.0000 & 1.0000 & 2.2575 & 0.0000 & 1.0000 & $-$1.3724 & 0.0001 & 3.7112 \\
0.10 & 0.2517 & 0.0025 & 0.9161 & 2.2695 & 0.0015 & 1.0053 & $-$1.3314 & 0.0194 & 3.5717 \\
0.20 & 0.2131 & 0.0043 & 0.7755 & 2.2937 & 0.0033 & 1.0160 & $-$1.2073 & 0.1430 & 3.3351 \\
0.30 & 0.1908 & 0.0032 & 0.6945 & 2.3040 & 0.0027 & 1.0206 & $-$1.0916 & 0.2242 & 2.9486 \\
0.40 & 0.1819 & 0.0045 & 0.6621 & 2.3037 & 0.0026 & 1.0205 & $-$1.4068 & 0.1665 & 3.8097 \\
0.50 & 0.1593 & 0.0023 & 0.5798 & 2.3168 & 0.0015 & 1.0263 & $-$1.4565 & 0.2034 & 3.8150 \\
0.60 & 0.1502 & 0.0042 & 0.5466 & 2.3224 & 0.0028 & 1.0287 & $-$1.5167 & 0.2135 & 4.1573 \\
0.70 & 0.1512 & 0.0032 & 0.5502 & 2.3217 & 0.0023 & 1.0284 & $-$1.4395 & 0.1577 & 4.1807 \\
0.80 & 0.1391 & 0.0027 & 0.5065 & 2.3295 & 0.0018 & 1.0319 & $-$1.5210 & 0.1310 & 4.4977 \\
0.90 & 0.1327 & 0.0008 & 0.4829 & 2.3339 & 0.0005 & 1.0338 & $-$1.6041 & 0.0381 & 4.6018 \\
\bottomrule
\end{tabular}%
}
\end{table}

\begin{table}[!htbp]
\centering
\caption{Small LSTM Input-Connection Pruning (\texttt{weight\_ih}, supplementary control 2): full metrics.}
\label{stab:lstm_input_full}
\resizebox{\textwidth}{!}{%
\begin{tabular}{@{}cccccccccc@{}}
\toprule
$k$ & Accuracy & $\pm$SE & RelAcc & $R_4$ & $\pm R_4$ SE & $R_4/R_4(0)$ & $\beta$ & $\pm\beta$ SD & $G$ \\
\midrule
0.00 & 0.2747 & 0.0000 & 1.0000 & 2.2575 & 0.0000 & 1.0000 & $-$1.3724 & 0.0001 & 3.7115 \\
0.10 & 0.1786 & 0.0065 & 0.6500 & 2.3078 & 0.0046 & 1.0223 & $-$1.2667 & 0.1908 & 3.4870 \\
0.20 & 0.0380 & 0.0046 & 0.1385 & 2.3945 & 0.0028 & 1.0607 & $-$1.2772 & 0.0980 & 3.4318 \\
0.30 & 0.0333 & 0.0065 & 0.1214 & 2.3973 & 0.0040 & 1.0619 & $-$1.1726 & 0.1351 & 3.2326 \\
0.40 & 0.0266 & 0.0047 & 0.0968 & 2.4016 & 0.0029 & 1.0638 & $-$1.1669 & 0.0633 & 3.1909 \\
0.50 & 0.0157 & 0.0036 & 0.0572 & 2.4082 & 0.0022 & 1.0668 & $-$1.1419 & 0.0177 & 3.1727 \\
0.60 & 0.0201 & 0.0026 & 0.0730 & 2.4056 & 0.0016 & 1.0656 & $-$1.1309 & 0.0519 & 3.1554 \\
0.70 & 0.0182 & 0.0041 & 0.0662 & 2.4067 & 0.0025 & 1.0661 & $-$1.1405 & 0.0385 & 3.1979 \\
0.80 & 0.0185 & 0.0022 & 0.0673 & 2.4065 & 0.0014 & 1.0660 & $-$1.1326 & 0.0316 & 3.1552 \\
0.90 & 0.0166 & 0.0029 & 0.0604 & 2.4077 & 0.0018 & 1.0665 & $-$1.1279 & 0.0230 & 3.0952 \\
\bottomrule
\end{tabular}%
}
\end{table}

\begin{table}[!htbp]
\centering
\caption{Qwen~0.5B MLP Neuron Removal: full metrics.}
\label{stab:qwen05_mlp_full}
\resizebox{\textwidth}{!}{%
\begin{tabular}{@{}cccccccccc@{}}
\toprule
$k$ & Accuracy & $\pm$SE & RelAcc & $R_4$ & $\pm R_4$ SE & $R_4/R_4(0)$ & $\beta$ & $\pm\beta$ SD & $G$ \\
\midrule
0.00 & 0.5363 & 0.0000 & 1.0000 & 1.8861 & 0.0000 & 1.0000 & $-$0.6608 & 0.0000 & 1.6345 \\
0.10 & 0.4790 & 0.0060 & 0.8932 & 1.8908 & 0.0069 & 1.0025 & $-$0.7806 & 0.0481 & 2.0512 \\
0.20 & 0.3786 & 0.0117 & 0.7059 & 1.9155 & 0.0093 & 1.0156 & $-$0.6893 & 0.0336 & 1.7207 \\
0.30 & 0.3035 & 0.0089 & 0.5658 & 1.9558 & 0.0070 & 1.0370 & $-$0.7995 & 0.1359 & 2.0021 \\
0.40 & 0.1382 & 0.0221 & 0.2576 & 2.0416 & 0.0131 & 1.0825 & $-$0.8825 & 0.1356 & 2.3562 \\
0.50 & 0.0417 & 0.0103 & 0.0777 & 2.0957 & 0.0053 & 1.1111 & $-$0.8690 & 0.0335 & 2.3038 \\
0.60 & 0.0169 & 0.0047 & 0.0315 & 2.1087 & 0.0025 & 1.1181 & $-$0.8840 & 0.0558 & 2.3363 \\
0.70 & 0.0089 & 0.0022 & 0.0165 & 2.1131 & 0.0012 & 1.1204 & $-$0.9401 & 0.0371 & 2.4425 \\
0.80 & 0.0102 & 0.0049 & 0.0190 & 2.1123 & 0.0026 & 1.1200 & $-$0.9656 & 0.0892 & 2.5255 \\
0.90 & 0.0026 & 0.0022 & 0.0048 & 2.1164 & 0.0012 & 1.1221 & $-$0.9160 & 0.0194 & 2.3750 \\
\bottomrule
\end{tabular}%
}
\end{table}

\begin{table}[!htbp]
\centering
\caption{Qwen~0.5B Attention-Head Removal: full metrics.}
\label{stab:qwen05_attn_full}
\resizebox{\textwidth}{!}{%
\begin{tabular}{@{}cccccccccc@{}}
\toprule
$k$ & Accuracy & $\pm$SE & RelAcc & $R_4$ & $\pm R_4$ SE & $R_4/R_4(0)$ & $\beta$ & $\pm\beta$ SD & $G$ \\
\midrule
0.00 & 0.5363 & 0.0000 & 1.0000 & 1.8861 & 0.0000 & 1.0000 & $-$0.6608 & 0.0000 & 1.6345 \\
0.10 & 0.5232 & 0.0011 & 0.9756 & 1.8923 & 0.0027 & 1.0033 & $-$0.6503 & 0.0196 & 1.6131 \\
0.20 & 0.4536 & 0.0135 & 0.8458 & 1.9225 & 0.0102 & 1.0193 & $-$0.6334 & 0.0682 & 1.5857 \\
0.30 & 0.4320 & 0.0236 & 0.8055 & 1.9079 & 0.0165 & 1.0116 & $-$0.6510 & 0.0258 & 1.6377 \\
0.40 & 0.2608 & 0.0438 & 0.4863 & 1.9880 & 0.0245 & 1.0541 & $-$0.6520 & 0.0875 & 1.4967 \\
0.50 & 0.1067 & 0.0395 & 0.1990 & 2.0635 & 0.0203 & 1.0941 & $-$0.7480 & 0.0646 & 1.7933 \\
0.60 & 0.0971 & 0.0246 & 0.1810 & 2.0695 & 0.0134 & 1.0973 & $-$0.7648 & 0.0972 & 1.8819 \\
0.70 & 0.0132 & 0.0023 & 0.0247 & 2.1111 & 0.0012 & 1.1193 & $-$0.8778 & 0.0207 & 2.2251 \\
0.80 & 0.0182 & 0.0040 & 0.0340 & 2.1081 & 0.0022 & 1.1177 & $-$0.8479 & 0.0403 & 2.1763 \\
0.90 & 0.0048 & 0.0025 & 0.0090 & 2.1152 & 0.0013 & 1.1215 & $-$0.9177 & 0.0062 & 2.3729 \\
\bottomrule
\end{tabular}%
}
\end{table}

\begin{table}[!htbp]
\centering
\caption{Qwen~1.8B MLP Neuron Removal: full metrics.}
\label{stab:qwen18_mlp_full}
\resizebox{\textwidth}{!}{%
\begin{tabular}{@{}cccccccccc@{}}
\toprule
$k$ & Accuracy & $\pm$SE & RelAcc & $R_4$ & $\pm R_4$ SE & $R_4/R_4(0)$ & $\beta$ & $\pm\beta$ SD & $G$ \\
\midrule
0.00 & 0.5465 & 0.0000 & 1.0000 & 1.8874 & 0.0000 & 1.0000 & $-$0.6225 & 0.0000 & 1.5568 \\
0.10 & 0.5112 & 0.0022 & 0.9354 & 1.8691 & 0.0031 & 0.9903 & $-$0.6697 & 0.0334 & 1.7113 \\
0.20 & 0.4479 & 0.0054 & 0.8197 & 1.8978 & 0.0065 & 1.0055 & $-$0.6710 & 0.0687 & 1.5578 \\
0.30 & 0.3376 & 0.0294 & 0.6178 & 1.9462 & 0.0164 & 1.0312 & $-$0.7183 & 0.0708 & 1.6114 \\
0.40 & 0.1287 & 0.0360 & 0.2355 & 2.0513 & 0.0184 & 1.0868 & $-$0.8472 & 0.1312 & 2.2284 \\
0.50 & 0.1163 & 0.0205 & 0.2129 & 2.0556 & 0.0108 & 1.0891 & $-$0.9153 & 0.1664 & 2.3113 \\
0.60 & 0.0239 & 0.0066 & 0.0438 & 2.1049 & 0.0036 & 1.1153 & $-$0.8803 & 0.0460 & 2.2934 \\
0.70 & 0.0240 & 0.0078 & 0.0438 & 2.1049 & 0.0042 & 1.1152 & $-$0.9801 & 0.1298 & 2.6144 \\
0.80 & 0.0126 & 0.0054 & 0.0231 & 2.1110 & 0.0029 & 1.1185 & $-$0.9599 & 0.1035 & 2.5199 \\
0.90 & 0.0097 & 0.0041 & 0.0177 & 2.1126 & 0.0022 & 1.1193 & $-$0.9126 & 0.0665 & 2.3857 \\
\bottomrule
\end{tabular}%
}
\end{table}

\begin{table}[!htbp]
\centering
\caption{Qwen~1.8B Attention-Head Removal: full metrics.}
\label{stab:qwen18_attn_full}
\resizebox{\textwidth}{!}{%
\begin{tabular}{@{}cccccccccc@{}}
\toprule
$k$ & Accuracy & $\pm$SE & RelAcc & $R_4$ & $\pm R_4$ SE & $R_4/R_4(0)$ & $\beta$ & $\pm\beta$ SD & $G$ \\
\midrule
0.00 & 0.5465 & 0.0000 & 1.0000 & 1.8874 & 0.0000 & 1.0000 & $-$0.6225 & 0.0000 & 1.5568 \\
0.10 & 0.5227 & 0.0059 & 0.9566 & 1.8823 & 0.0128 & 0.9973 & $-$0.6022 & 0.0310 & 1.5457 \\
0.20 & 0.4507 & 0.0432 & 0.8248 & 1.9104 & 0.0244 & 1.0122 & $-$0.6507 & 0.0699 & 1.6066 \\
0.30 & 0.4440 & 0.0171 & 0.8125 & 1.9127 & 0.0085 & 1.0134 & $-$0.6640 & 0.0558 & 1.6601 \\
0.40 & 0.3559 & 0.0196 & 0.6514 & 1.9593 & 0.0167 & 1.0381 & $-$0.5792 & 0.0615 & 1.3152 \\
0.50 & 0.1732 & 0.0376 & 0.3169 & 2.0397 & 0.0193 & 1.0807 & $-$0.7272 & 0.1497 & 1.7441 \\
0.60 & 0.1254 & 0.0254 & 0.2295 & 2.0547 & 0.0122 & 1.0887 & $-$0.7171 & 0.1038 & 1.8428 \\
0.70 & 0.0563 & 0.0111 & 0.1030 & 2.0899 & 0.0056 & 1.1073 & $-$0.7469 & 0.0641 & 1.7936 \\
0.80 & 0.0398 & 0.0102 & 0.0729 & 2.0970 & 0.0052 & 1.1111 & $-$0.7971 & 0.0863 & 1.9979 \\
0.90 & 0.0120 & 0.0042 & 0.0220 & 2.1115 & 0.0022 & 1.1187 & $-$0.9110 & 0.0327 & 2.3280 \\
\bottomrule
\end{tabular}%
}
\end{table}

\subsection{Detailed Quantitative Cloze Analysis, LSTM Pathway Asymmetry, and Representative Examples}\label{sec:supp_cloze_summary}\label{sec:supp_qual_cloze}

\paragraph{Detailed per-model Cloze trajectories (supporting Section~\ref{sec:cloze}).}
For Qwen~0.5B, baseline Cloze performance is strong: POS match = 65.0\%, cosine similarity = 0.643 (Table~\ref{stab:cloze_values}).
Under MLP ablation, both metrics decline through $k = 0.2$ (POS 47.2\%, cosine 0.421) and $k = 0.3$ (POS 35.2\%, cosine 0.283), then collapse sharply at $k = 0.4$ (POS 8.8\%, cosine 0.078) and $k = 0.5$ (POS 6.4\%, cosine 0.042).
Under attention-head ablation, Qwen~0.5B retains substantially higher competence across the mid-range on the nominal-$k$ axis: at $k = 0.4$, POS is 47.6\% and cosine is 0.407 (more than fivefold higher than MLP ablation at the same nominal $k$), and at $k = 0.5$, POS is 35.0\% with a larger syntactic--semantic gap (cosine 0.198).
Qwen~1.8B shows the same contrast: from a baseline of POS = 64.0\% and cosine = 0.663, MLP ablation drops to POS = 48.8\% and cosine = 0.449 at $k = 0.2$, 45.2\% and 0.362 at $k = 0.3$, 20.8\% and 0.138 at $k = 0.4$, and collapses to POS = 6.2\% and cosine = 0.048 at $k = 0.5$ (Figure~\ref{fig:cloze}, rows 2 and 3).
In contrast, Qwen~1.8B attention-head ablation retains POS = 59.9\% and cosine = 0.606 at $k = 0.2$, POS = 37.0\% and cosine = 0.352 at $k = 0.4$ (about $1.8\times$ and $2.6\times$ the MLP values of 20.8\% and 0.138), and still preserves POS = 42.0\% and cosine = 0.319 at nominal $k = 0.5$, outperforming MLP ablation by nearly sevenfold on both metrics at nominal $k = 0.5$. Because nominal $k = 0.5$ removes $16.3\%$ (0.5B) and $16.6\%$ (1.8B) of non-embedding parameters under attention-head ablation versus $33.7\%$ and $33.4\%$ under MLP ablation (and $70.4\%$ and $72.7\%$ under LSTM whole-unit removal), at a matched parameter-removal fraction of ${\sim}13.4$--$20.2\%$ Qwen MLP ($k = 0.2$--$0.3$: POS $35.2\%$--$48.8\%$, cosine $0.283$--$0.449$) and AWD-LSTM whole-unit removal ($k = 0.1$, $18.1\%$ removed: POS $53.0\%$, cosine $0.393$) match or exceed attention at $k = 0.5$. Similarly, at ${\sim}34\%$ of non-embedding parameters removed, AWD-LSTM whole-unit removal ($k = 0.2$, $34.5\%$ removed: POS $42.7\%$, cosine $0.322$) exceeds Qwen MLP at $k = 0.5$ (POS $6.2$--$6.4\%$, cosine $0.042$--$0.048$).

\begin{figure}[!htbp]
\centering
\includegraphics[width=0.92\textwidth,height=0.58\textheight,keepaspectratio]{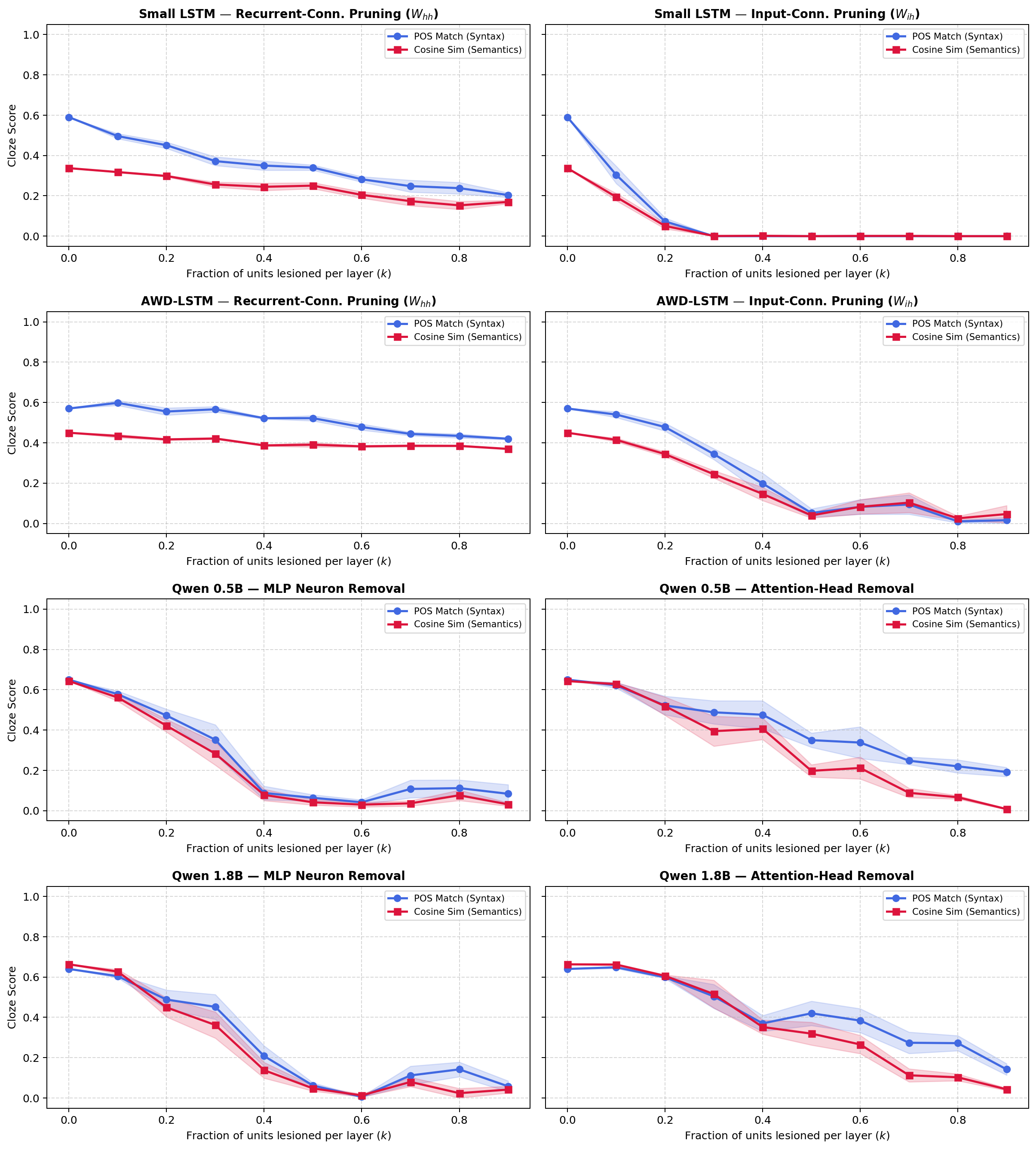}
\caption{Syntactic (POS match, blue) and semantic (GloVe cosine similarity, red) Cloze performance for the supplementary LSTM connection-pruning controls (rows 1--2: Small LSTM and AWD-LSTM recurrent-connection pruning \texttt{weight\_hh} in the left column and input-connection pruning \texttt{weight\_ih} in the right column) alongside Qwen~0.5B and Qwen~1.8B MLP and Attention pathways (rows 3--4; primary unit-level Cloze trajectories are shown in Figure~\ref{fig:cloze}). Isolating \texttt{weight\_hh} (rows 1--2, left) confirms that pruning recurrent memory connections alone produces a gradual, non-dissociated decline all the way to $k = 0.9$, whereas pruning \texttt{weight\_ih} input connections (rows 1--2, right) drives the rapid bilateral Cloze collapse seen at high $k$ under whole-unit removal.}
\label{sfig:cloze_all}
\end{figure}

In the intact AWD-LSTM (baseline POS = 57.0\%, cosine = 0.450; Figure~\ref{fig:cloze}, row 1 right), whole-unit removal (\texttt{weight\_ih} + \texttt{weight\_hh} + gate biases) degrades both POS match and cosine similarity in parallel without dissociation across mild-to-moderate damage: POS is 53.0\% (cosine 0.393) at $k = 0.1$, 42.7\% (cosine 0.322) at $k = 0.2$, and 28.4\% (cosine 0.203) at $k = 0.3$, before collapsing to 4.8\% (cosine 0.074) at $k = 0.4$, 2.2\% (cosine 0.020) at $k = 0.5$, 0.0\% (cosine 0.008) at $k = 0.6$, and 0.0\% (cosine 0.000) for $k \geq 0.7$ (Table~\ref{stab:cloze_values}).
Our supplementary connection-pruning controls (Supplementary Figure~\ref{sfig:cloze_all}, row 2) isolate the two mechanisms behind this trajectory: under pure recurrent-connection (\texttt{weight\_hh}) pruning, POS and cosine degrade gradually and in parallel across the entire sweep (POS 55.5\%, cosine 0.417 at $k = 0.2$; 52.2\%, 0.390 at $k = 0.5$; and 42.0\%, 0.370 at $k = 0.9$, preserving roughly three-quarters of baseline competence even at 90\% recurrent pruning), whereas under pure input-connection (\texttt{weight\_ih}) pruning, POS and cosine collapse steeply together (47.8\% and 0.344 at $k = 0.2$; 34.4\% and 0.244 at $k = 0.3$; 19.8\% and 0.147 at $k = 0.4$; 5.2\% and 0.040 at $k = 0.5$; and 1.6\% and 0.047 at $k = 0.9$).

The Small LSTM exhibits the exact same pattern from a baseline of POS = 59.0\% and cosine = 0.337 (Figure~\ref{fig:cloze}, row 1 left, and Supplementary Figure~\ref{sfig:cloze_all}, row 1).
Under whole-unit removal, POS and cosine decline in parallel across $k = 0.1$ (41.6\%, 0.279), $k = 0.2$ (23.6\%, 0.177), and $k = 0.3$ (11.4\%, 0.073), before collapsing to 1.6\% (0.021) at $k = 0.4$ and 0.0\% (0.000) for all $k \geq 0.5$.
Under recurrent-connection (\texttt{weight\_hh}) pruning alone, both POS and cosine decline gradually and in parallel across the full sweep (45.1\% and 0.298 at $k = 0.2$; 34.0\% and 0.251 at $k = 0.5$; and 20.4\% and 0.170 at $k = 0.9$), whereas under input-connection (\texttt{weight\_ih}) pruning alone, both metrics collapse rapidly (30.4\% and 0.194 at $k = 0.1$; 7.3\% and 0.051 at $k = 0.2$; and 0.0\% and $\leq 0.002$ for all $k \geq 0.3$).

\paragraph{Architectural basis for pathway asymmetry and whole-unit removal in LSTMs.}
In an LSTM, each hidden unit receives input from two additive sources at every time step: the input projection (\texttt{weight\_ih}), which transforms the current token embedding $x_t$ (or previous layer's hidden state $h_t^{(\ell-1)}$), and the recurrent projection (\texttt{weight\_hh}), which propagates the previous time step's hidden state $h_{t-1}^{(\ell)}$.
When \texttt{weight\_hh} connections are pruned for a subset of units (Supplement control 1), those units still receive intact current-token projections via \texttt{weight\_ih}: the feedforward pathway from the current token embedding through the LSTM gates to the tied output decoder remains unbroken, allowing the network to act as a local short-context predictor whose performance degrades gradually in proportion to the lost recurrent memory.
Conversely, when \texttt{weight\_ih} is pruned (Supplement control 2), or when whole units are completely silenced in the main paper by zeroing both \texttt{weight\_ih} and \texttt{weight\_hh} across all layers (including outgoing columns feeding the next layer), the conduit through which current token information traverses the serial 3-layer stack is progressively narrowed and, at $k \ge 0.4$ (reaching $0.0\%$ POS by $k = 0.5$--$0.6$), starved of prompt information.
In the Transformer, Pre-Norm residual stream connections ($+$) bypass each layer's MLP and attention sub-blocks across all 24 layers, which may contribute to the early retention under attention-head ablation and to the abrupt threshold collapse under MLP ablation across $k^* \approx 0.3$--$0.4$ (after an early decline comparable to the LSTMs), whereas in the 3-layer LSTM, no residual stream bypasses the serial layer stack.

\begin{table}[!htbp]
\centering
\caption{Quantified Cloze test: POS match rate and GloVe cosine similarity at selected ablation fractions across all ten conditions (six primary unit-level conditions plus four supplementary LSTM connection-pruning controls). Values are means across trials (deterministic baseline at $k = 0.0$, $n = 10$ trials at $k = 0.2$, and $n = 5$ trials at $k \in \{0.5, 0.9\}$).}
\label{stab:cloze_values}
\resizebox{\textwidth}{!}{%
\begin{tabular}{@{}llcccccccc@{}}
\toprule
& & \multicolumn{4}{c}{POS Match (Syntax)} & \multicolumn{4}{c}{Cosine Similarity (Semantics)} \\
\cmidrule(lr){3-6} \cmidrule(lr){7-10}
Model & Target & $k{=}0$ & $k{=}0.2$ & $k{=}0.5$ & $k{=}0.9$ & $k{=}0$ & $k{=}0.2$ & $k{=}0.5$ & $k{=}0.9$ \\
\midrule
Small LSTM & unit & 0.590 & 0.236 & 0.000 & 0.000 & 0.337 & 0.177 & 0.000 & 0.000 \\
AWD-LSTM & unit & 0.570 & 0.427 & 0.022 & 0.000 & 0.450 & 0.322 & 0.020 & 0.000 \\
Qwen 0.5B & MLP & 0.650 & 0.472 & 0.064 & 0.084 & 0.643 & 0.421 & 0.042 & 0.031 \\
Qwen 0.5B & Attn & 0.650 & 0.522 & 0.350 & 0.192 & 0.643 & 0.519 & 0.198 & 0.008 \\
Qwen 1.8B & MLP & 0.640 & 0.488 & 0.062 & 0.058 & 0.663 & 0.449 & 0.048 & 0.042 \\
Qwen 1.8B & Attn & 0.640 & 0.599 & 0.420 & 0.142 & 0.663 & 0.606 & 0.319 & 0.043 \\
\midrule
Small LSTM & w\_hh (Suppl.) & 0.590 & 0.451 & 0.340 & 0.204 & 0.337 & 0.298 & 0.251 & 0.170 \\
AWD-LSTM & w\_hh (Suppl.) & 0.570 & 0.555 & 0.522 & 0.420 & 0.450 & 0.417 & 0.390 & 0.370 \\
Small LSTM & w\_ih (Suppl.) & 0.590 & 0.073 & 0.000 & 0.000 & 0.337 & 0.051 & 0.000 & 0.000 \\
AWD-LSTM & w\_ih (Suppl.) & 0.570 & 0.478 & 0.052 & 0.016 & 0.450 & 0.344 & 0.040 & 0.047 \\
\bottomrule
\end{tabular}%
}
\end{table}

\begin{table}[!htbp]
\footnotesize
\centering
\caption{Representative qualitative Cloze predictions across architectures at $k = 0.2$ (from the first ablation trial), showing the undamaged ($k = 0$) Qwen~1.8B prediction (\texttt{Qwen1.8 Intact}) alongside Qwen~1.8B MLP and Attention, primary LSTM whole-unit removal (\texttt{AWD Unit}, \texttt{LSTM Unit}), and supplementary LSTM recurrent-connection pruning (\texttt{AWD Hid.}, \texttt{LSTM Hid.}). Errors quoted as examples in the main text were checked against the top-1 and top-5 predictions of the corresponding undamaged model for the same prompt and do not occur there; some ablated errors in this table (for example \textit{flower}$\to$\textit{girl} and \textit{mouse}$\to$\textit{number}) are also produced by the undamaged model and should not be read as ablation-induced. {\color{green!50!black}$\checkmark$} = exact match; {\color{orange}$\sim$} = in top-5; {\color{red}$\times$} = miss.}
\label{stab:error_comparison}
\begin{tabular}{@{}p{2.3cm}p{0.75cm}p{0.95cm}p{0.95cm}p{0.95cm}p{0.95cm}p{0.95cm}p{0.95cm}p{0.95cm}@{}}
\toprule
\scriptsize Prompt & \scriptsize Exp. & \scriptsize Qwen1.8 Intact & \scriptsize Qwen1.8 MLP & \scriptsize Qwen1.8 Attn & \scriptsize AWD Unit & \scriptsize LSTM Unit & \scriptsize AWD Hid. & \scriptsize LSTM Hid. \\
\midrule
once upon a & time & time {\color{green!50!black}$\checkmark$} & day {\color{orange}$\sim$} & time {\color{green!50!black}$\checkmark$} & a {\color{red}$\times$} & " {\color{red}$\times$} & time {\color{green!50!black}$\checkmark$} & new {\color{red}$\times$} \\
the sun was shining bright in the & sky & sky {\color{green!50!black}$\checkmark$} & sky {\color{green!50!black}$\checkmark$} & sky {\color{green!50!black}$\checkmark$} & summer {\color{red}$\times$} & first {\color{red}$\times$} & sky {\color{green!50!black}$\checkmark$} & same {\color{red}$\times$} \\
she picked up the & ball & phone {\color{orange}$\sim$} & ball {\color{green!50!black}$\checkmark$} & 1 {\color{red}$\times$} & number {\color{red}$\times$} & first {\color{red}$\times$} & first {\color{red}$\times$} & end {\color{red}$\times$} \\
the bird flew up into the & sky & sky {\color{green!50!black}$\checkmark$} & trees {\color{orange}$\sim$} & tree {\color{red}$\times$} & air {\color{red}$\times$} & first {\color{red}$\times$} & air {\color{orange}$\sim$} & first {\color{red}$\times$} \\
she sang a beautiful & song & song {\color{green!50!black}$\checkmark$} & song {\color{green!50!black}$\checkmark$} & song {\color{green!50!black}$\checkmark$} & , {\color{orange}$\sim$} & , {\color{red}$\times$} & song {\color{green!50!black}$\checkmark$} & - {\color{red}$\times$} \\
he wanted to go & home & to {\color{orange}$\sim$} & to {\color{red}$\times$} & to {\color{red}$\times$} & on {\color{red}$\times$} & to {\color{red}$\times$} & to {\color{red}$\times$} & to {\color{red}$\times$} \\
he was very happy and started to & laugh & sing {\color{orange}$\sim$} & play {\color{red}$\times$} & cry {\color{red}$\times$} & make {\color{red}$\times$} & be {\color{red}$\times$} & make {\color{red}$\times$} & be {\color{red}$\times$} \\
the dog saw a cat and started to & bark & bark {\color{green!50!black}$\checkmark$} & chase {\color{orange}$\sim$} & bark {\color{green!50!black}$\checkmark$} & make {\color{red}$\times$} & be {\color{red}$\times$} & show {\color{red}$\times$} & be {\color{red}$\times$} \\
he ran as fast as he & could & could {\color{green!50!black}$\checkmark$} & could {\color{green!50!black}$\checkmark$} & could {\color{green!50!black}$\checkmark$} & was {\color{red}$\times$} & was {\color{orange}$\sim$} & could {\color{green!50!black}$\checkmark$} & was {\color{red}$\times$} \\
the baby started to & cry & cry {\color{green!50!black}$\checkmark$} & cry {\color{green!50!black}$\checkmark$} & cry {\color{green!50!black}$\checkmark$} & fall {\color{red}$\times$} & be {\color{red}$\times$} & appear {\color{red}$\times$} & be {\color{red}$\times$} \\
the surprisingly & gentle & good {\color{red}$\times$} & good {\color{red}$\times$} & small {\color{red}$\times$} & first {\color{red}$\times$} & - {\color{red}$\times$} & successful {\color{red}$\times$} & the {\color{red}$\times$} \\
the boat sailed across the & ocean & river {\color{red}$\times$} & bay {\color{red}$\times$} & river {\color{red}$\times$} & western {\color{red}$\times$} & city {\color{red}$\times$} & sea {\color{orange}$\sim$} & country {\color{red}$\times$} \\
she poured the milk into the & glass & bowl {\color{red}$\times$} & bottle {\color{red}$\times$} & blender {\color{red}$\times$} & water {\color{red}$\times$} & first {\color{red}$\times$} & water {\color{red}$\times$} & house {\color{red}$\times$} \\
the chef chopped the & vegetables & onion {\color{orange}$\sim$} & meat {\color{red}$\times$} & vegetables {\color{green!50!black}$\checkmark$} & way {\color{red}$\times$} & first {\color{red}$\times$} & ice {\color{red}$\times$} & house {\color{red}$\times$} \\
he found a beautiful & flower & girl {\color{red}$\times$} & girl {\color{red}$\times$} & dress {\color{orange}$\sim$} & , {\color{red}$\times$} & , {\color{red}$\times$} & woman {\color{red}$\times$} & - {\color{red}$\times$} \\
the really big & bear & question {\color{red}$\times$} & ones {\color{red}$\times$} & deal {\color{red}$\times$} & , {\color{red}$\times$} & , {\color{red}$\times$} & - {\color{red}$\times$} & - {\color{red}$\times$} \\
a very small & mouse & number {\color{red}$\times$} & , {\color{red}$\times$} & number {\color{red}$\times$} & , {\color{red}$\times$} & , {\color{red}$\times$} & small {\color{red}$\times$} & - {\color{red}$\times$} \\
the horse galloped across the & field & field {\color{green!50!black}$\checkmark$} & ground {\color{orange}$\sim$} & field {\color{green!50!black}$\checkmark$} & country {\color{red}$\times$} & field {\color{green!50!black}$\checkmark$} & street {\color{red}$\times$} & city {\color{red}$\times$} \\
she stirred the soup with a & spoon & spoon {\color{green!50!black}$\checkmark$} & spoon {\color{green!50!black}$\checkmark$} & wooden {\color{orange}$\sim$} & " {\color{red}$\times$} & " {\color{red}$\times$} & black {\color{red}$\times$} & " {\color{red}$\times$} \\
the plane disappeared behind the & clouds & mountain {\color{orange}$\sim$} & clouds {\color{green!50!black}$\checkmark$} & mountain {\color{orange}$\sim$} & body {\color{red}$\times$} & first {\color{red}$\times$} & scenes {\color{red}$\times$} & the {\color{red}$\times$} \\
dark clouds gathered before the & storm & sun {\color{orange}$\sim$} & stars {\color{red}$\times$} & sun {\color{orange}$\sim$} & event {\color{red}$\times$} & first {\color{red}$\times$} & night {\color{orange}$\sim$} & first {\color{red}$\times$} \\
the river flowed into the & sea & sea {\color{green!50!black}$\checkmark$} & river {\color{red}$\times$} & city {\color{red}$\times$} & river {\color{orange}$\sim$} & area {\color{red}$\times$} & river {\color{orange}$\sim$} & area {\color{red}$\times$} \\
the doctor examined the patient's & heart & chest {\color{orange}$\sim$} & body {\color{red}$\times$} & tongue {\color{orange}$\sim$} & body {\color{red}$\times$} & " {\color{red}$\times$} & body {\color{red}$\times$} & first {\color{red}$\times$} \\
\bottomrule
\end{tabular}
\end{table}

\subsection{Detailed Statistical Formalisation of Clinical Parallel Assignments}\label{sec:supp_clinical_stats}

This section provides the complete numerical and statistical breakdowns supporting the clinical parallel subsections in Section~\ref{sec:clinical_evidence}, along with Supplementary Table~\ref{stab:lstm_conn_clinical} reporting the error-distribution, Brysbaert concreteness, and PNT taxonomy profiles for the four supplementary LSTM connection-pruning controls (\texttt{weight\_hh} and \texttt{weight\_ih}):

\begin{table}[!htbp]
\footnotesize
\centering
\caption{Error-type distribution, Brysbaert concreteness shift $\Delta\text{Conc.}$, and seven-category Philadelphia Naming Test (PNT) taxonomy at $k = 0.2$ for the four supplementary LSTM connection-pruning controls (\texttt{weight\_hh} recurrent pruning and \texttt{weight\_ih} input pruning; compare with the primary unit-level results in Table~\ref{tab:error_dist}, Figure~\ref{fig:pnt_main}, and Supplementary Table~\ref{stab:pnt_main}). Values show the mean and 95\% confidence interval (or mean percentage for the PNT categories) across 10 independent ablation trials ($100$ prompts per trial).}
\label{stab:lstm_conn_clinical}
\resizebox{\textwidth}{!}{%
\begin{tabular}{@{}llcccc|ccccccc@{}}
\toprule
& & \multicolumn{4}{c|}{Three-Way Cloze Error \& Concreteness Profile} & \multicolumn{7}{c}{Seven-Category PNT Taxonomy (\%)} \\
Architecture & Pathway ($k{=}0.2$) & Correct & Content Sub. & Non-content & Mean $\Delta$Conc. & Corr. & Sem. & Form. & Mix. & Unrel. & Neol. & No-resp. \\
\midrule
AWD-LSTM & w\_hh (recurrent) & 20.2\% [18.7, 21.7] & 53.1\% [51.0, 55.2] & 26.7\% [24.7, 28.7] & $+0.34$ [$+0.27$, $+0.41$] & 20.2\% & 21.6\% & 0.1\% & 0.0\% & 54.9\% & 0.0\% & 3.2\% \\
Small LSTM & w\_hh (recurrent) & 1.8\% [1.1, 2.5] & 37.2\% [32.5, 41.9] & 61.0\% [55.9, 66.1] & $+0.67$ [$+0.54$, $+0.80$] & 1.8\% & 7.8\% & 0.9\% & 0.0\% & 81.5\% & 0.0\% & 8.0\% \\
AWD-LSTM & w\_ih (input) & 10.7\% [8.8, 12.6] & 53.9\% [47.4, 60.4] & 35.4\% [28.5, 42.3] & $+0.30$ [$+0.18$, $+0.42$] & 10.7\% & 18.4\% & 0.2\% & 0.0\% & 64.7\% & 0.0\% & 6.0\% \\
Small LSTM & w\_ih (input) & 0.7\% [0.4, 1.1] & 23.0\% [11.6, 34.4] & 76.3\% [65.0, 87.6] & $+0.85$ [$+0.32$, $+1.39$] & 0.7\% & 1.6\% & 0.1\% & 0.0\% & 42.1\% & 0.0\% & 55.5\% \\
\bottomrule
\end{tabular}%
}
\end{table}

\begin{itemize}
    \item \textit{Comparison of intact and ablated models with post-stroke patient error distributions}: In Figure~\ref{fig:pnt_main} and Supplementary Table~\ref{stab:pnt_main}, intact ($k = 0$) Qwen~1.8B and Qwen~0.5B achieve $66.0\%$ and $64.0\%$ Correct, with $1.0\%$ No-response and $11.0\%$ and $9.0\%$ Semantic substitutions. When Correct is restricted to an exact top-1 match rather than top-5, first-trial rates are $42\%$, $38\%$, $7\%$, and $0\%$ for the four intact baselines, and $22\%$, $17\%$, $37\%$, $16\%$, $1\%$, and $1\%$ for the six unit-ablated conditions (listed in the row order of Table~\ref{tab:error_dist}). At $k = 0.2$ across the 10 independent ablation trials ($N = 1{,}000$ predictions per condition), Qwen~1.8B Correct rates bracket the $47.3\%$ PWA patient mean ($42.1\%$ under MLP and $61.2\%$ under attention; $45\%$ and $61\%$ in the first trial). Under Qwen~1.8B MLP ablation, No-response errors jump from $1.0\%$ to $18.1\%$ (95\% CI [$3.0$, $33.2$]; $16.7\%$ in Qwen~0.5B MLP), a wide interval that overlaps the $21.9\%$ patient rate. Unrelated word errors rise from $22.0\%$ to $27.9\%$, while PNT Semantic paraphasias remain at $11.7\%$ under MLP versus $9.1\%$ under attention ($18\%$ vs.\ $11\%$ in the first trial; patient reference $4.2\%$). By contrast, intact AWD-LSTM and Small LSTM start at $25.0\%$ and $4.0\%$ Correct at $k = 0$ and drop under whole-unit removal at $k = 0.2$ to $12.3\%$ and $0.6\%$ Correct across the 10 trials ($9\%$ and $2\%$ in the first trial), with $62.4\%$ and $74.1\%$ Unrelated defaults and $9.8\%$ and $22.2\%$ No-response. Across the supplementary connection-pruning controls (Table~\ref{stab:lstm_conn_clinical}), Correct rates range from $0.7\%$ to $20.2\%$ and No-response rates reach up to $55.5\%$.
    \item \textit{Null-corrected $R_4$ and Jensen--Shannon divergence}: At matched relative accuracy ($0.80$), the Jensen--Shannon divergence from the intact error distribution is larger in the Qwen unit-ablation pathways ($\mathrm{JS} = 0.034$--$0.045$; null-corrected $\Delta\mathrm{JS} = +0.010$ to $+0.020$) than in LSTM whole-unit removal ($\mathrm{JS} = 0.007$--$0.008$; $\Delta\mathrm{JS} = +0.002$ to $+0.003$, Figure~\ref{fig:null_model_comparison}) or LSTM recurrent-connection pruning ($\mathrm{JS} = 0.006$--$0.007$, Supplement~\ref{sec:supp_null_model}). When trials are pooled across the shared transition range (without matching for relative accuracy or correcting for multiple comparisons), the normalised null-corrected shift $(R_4^{\mathrm{obs}} - R_4^{\mathrm{null}})/R_4(0)$ is more negative under MLP ablation than under attention-head ablation in both Qwen~0.5B ($-0.0242$ vs.\ $-0.0109$, two-sided permutation $p < 10^{-4}$) and Qwen~1.8B ($-0.0251$ vs.\ $-0.0209$, $p = 0.0238$). At a single interpolated accuracy slice ($0.80$), however, their 95\% confidence intervals overlap (Supplement~\ref{sec:supp_null_model}).
    \item \textit{MLP ablation $\leftrightarrow$ Wernicke-type aphasia}: Across the 10 ablation trials at $k = 0.2$ (Table~\ref{tab:error_dist}), Qwen~1.8B MLP retains $42.1\%$ [95\% CI: $30.1\%, 54.1\%$] correct predictions and $27.2\%$ [$21.7\%, 32.7\%$] content-word substitutions (in the first trial: $45\%$ correct, consisting of $22\%$ exact match plus $23\%$ top-5, and $38\%$ content substitutions). Its mean Brysbaert concreteness drop is $\Delta\text{Conc.} = +0.19$ [$+0.07, +0.30$], which does not differ significantly from the intact baseline ($+0.13$; difference $+0.06$, 95\% CI $[-0.06, +0.17]$; Supplement~\ref{sec:concreteness_method}). For Qwen~0.5B MLP, the corresponding difference from intact is borderline ($+0.29$ [$+0.13$, $+0.45$] vs.\ $+0.15$ intact; difference $+0.14$, 95\% CI $[-0.02, +0.30]$). On the PNT taxonomy (Figure~\ref{fig:pnt_main}; Supplementary Table~\ref{stab:pnt_main}), Qwen~1.8B MLP averages $11.7\%$ [$9.2\%, 14.2\%$] Semantic errors across the 10 trials, maintaining its intact $k = 0$ rate of $11.0\%$ (difference $+0.70$ [$-1.78$, $+3.18$]). Over the same trials, Non-content responses rise by $23.7$ [$7.2$, $40.2$] percentage points above intact and PNT No-responses rise by $17.1$ [$2.0$, $32.2$] points (from $1.0\%$ to $18.1\%$; in Qwen~0.5B MLP, Non-content and No-response rise by $25.8$ and $15.7$ points). Compared with attention ablation ($9.1\%$ [$7.8\%, 10.4\%$] Semantic errors), Qwen~1.8B MLP shows a directional difference (two-sided permutation $p = 0.0613$, Holm-adjusted $p = 0.1227$). In the first trial alone, Qwen~1.8B MLP produces 18 PNT Semantic paraphasias out of 38 content substitutions at $\theta_s = 0.60$ (rising to 27 of 38 at $\theta_s = 0.50$) and exceeds attention-head ablation at every similarity threshold $\theta_s \in [0.50, 0.80]$ in both Qwen~1.8B and Qwen~0.5B (Table~\ref{stab:pnt_sensitivity}). On the quantitative Cloze sweep (Table~\ref{tab:cloze_summary}), Qwen~1.8B MLP retains moderate Cloze performance at $k = 0.2$ (POS $48.8\%$, cosine $0.449$) and $k = 0.3$ ($45.2\%$, $0.362$) before collapsing at $k = 0.5$ to POS $6.2\%$ and cosine $0.048$ (Qwen~0.5B MLP at $k = 0.5$: POS $6.4\%$, cosine $0.042$), accompanied by the largest negative shift in null-corrected $\Delta R_4 / R_4(0)$ ($-0.0251$ in Qwen~1.8B and $-0.0242$ in Qwen~0.5B).
    \item \textit{Attention-head ablation $\leftrightarrow$ svPPA}: At $k = 0.2$, Qwen~1.8B attention ablation retains $61.2\%$ [$59.4\%, 63.0\%$] correct predictions across the 10 trials ($61\%$ in the first trial) and preserves Cloze POS at $59.9\%$ and cosine at $0.606$. (In Qwen~0.5B attention, the first trial is an atypically degraded outlier with $24\%$ Correct and $54\%$ Non-content, compared with the 10-trial means of $46.4\%$ and $24.5\%$.) By nominal $k = 0.5$, attention ablation removes $16.3$--$16.6\%$ of non-embedding weights (vs.\ $33.4$--$33.7\%$ under MLP ablation) and retains POS $= 42.0\%$ and cosine $= 0.319$ in Qwen~1.8B, compared with $6.2\%$ and $0.048$ for MLP. Similarly, in Qwen~0.5B at nominal $k = 0.5$, attention retains POS $= 35.0\%$ and cosine $= 0.198$ versus $6.4\%$ and $0.042$ for MLP; when matched at ${\sim}13$--$20\%$ parameter removal (MLP $k = 0.2$--$0.3$), MLP retains POS $35.2\%$--$48.8\%$ and cosine $0.283$--$0.449$ (Table~\ref{tab:cloze_summary}). Across the 10 trials at $k = 0.2$ (Table~\ref{tab:error_dist}), the lemmatised Brysbaert concreteness drop under attention ablation rises above the intact baseline in Qwen~0.5B ($\Delta\text{Conc.} = +0.43$ [$0.33, 0.53$] vs.\ $+0.15$ intact; difference $+0.28$, 95\% CI $[+0.18, +0.38]$; first trial $+0.60$) and is borderline in Qwen~1.8B ($\Delta\text{Conc.} = +0.24$ [$0.13, 0.34$] vs.\ $+0.13$ intact; difference $+0.10$, 95\% CI $[+0.00, +0.21]$; first trial $+0.21$). Because these CIs treat the deterministic intact baseline as fixed, and the Qwen~0.5B MLP difference from intact ($+0.14$ [$-0.02$, $+0.30$]) is of comparable size, concreteness alone does not cleanly separate the two pathways. LSTM recurrent-connection pruning also raises the concreteness drop significantly above its intact baseline ($+0.14$ [$+0.02$, $+0.27$] in the Small LSTM and $+0.09$ [$+0.02$, $+0.16$] in the AWD-LSTM), showing that this shift is not unique to attention ablation; conversely, AWD-LSTM \texttt{weight\_hh} pruning is the only condition with a significant rise in PNT Semantic errors ($+2.6$ [$+0.6$, $+4.6$] points vs.\ intact). When we compare the two Qwen pathways directly across the 10 trials at $k = 0.2$, point estimates shift in the clinically predicted direction in both models: PNT Semantic paraphasia rates are higher under MLP ($11.7\%$ vs.\ $9.1\%$ in Qwen~1.8B, permutation $p = 0.0613$; $9.9\%$ vs.\ $9.0\%$ in Qwen~0.5B, $p = 0.6058$) and concreteness drops are larger under attention ($+0.43$ vs.\ $+0.29$ in Qwen~0.5B, $p = 0.1114$; $+0.24$ vs.\ $+0.19$ in Qwen~1.8B, $p = 0.5019$). None of these direct error-type contrasts reaches statistical significance (Holm-adjusted $p = 0.1227$ and $0.5019$ for the two Qwen~1.8B contrasts). These directional trends sit alongside the clear pathway separation in Cloze trajectories on the nominal-$k$ axis ($k = 0.3$--$0.6$, Figure~\ref{fig:cloze}) and in the pooled null-corrected $R_4$ test ($p < 10^{-4}$ in Qwen~0.5B; $p = 0.0238$ in Qwen~1.8B; not matched for relative accuracy).
    \item \textit{LSTM whole-unit removal and pathway-connection controls $\leftrightarrow$ Diffuse AD-type decline and late-stage collapse}: Under whole-unit removal (Figures~\ref{fig:accuracy}a,b, \ref{fig:R4}a,b, \ref{fig:null_model_comparison}, and~\ref{fig:cloze}, row 1), next-token accuracy declines steadily across $k \in [0, 0.9]$, retaining $82.2\%$ (Small LSTM) and $75.3\%$ (AWD-LSTM) of baseline at $k = 0.2$, and $45.8\%$ and $34.4\%$ at $k = 0.5$. Across the full sweep, raw $R_4$ stays within $5.1\%$ (Small LSTM) and $7.7\%$ (AWD-LSTM) of baseline, and $\mathrm{JS}$ at $\mathrm{RelAcc} = 0.80$ is only $0.007$--$0.008$. At $k = 0.2$ across the 10 trials (Table~\ref{tab:error_dist}), AWD-LSTM whole-unit removal produces $12.3\%$ Correct predictions, $46.6\%$ Content substitutions, and $41.1\%$ Non-content responses ($\Delta\text{Conc.} = +0.34$), while Small LSTM whole-unit removal drops to $0.6\%$ Correct, $15.1\%$ Content substitutions, and $84.3\%$ Non-content ($\Delta\text{Conc.} = +0.52$). The supplementary connection-pruning controls (Table~\ref{stab:lstm_conn_clinical}) separate the two mechanisms behind this behaviour. Pruning recurrent connections (\texttt{weight\_hh}) alone preserves $48.3\%$ (Small LSTM) and $35.1\%$ (AWD-LSTM) of baseline accuracy even at $k = 0.9$, keeps raw $R_4$ within $3.4\%$ and $5.9\%$ of baseline ($\mathrm{JS} = 0.006$--$0.007$ at $\mathrm{RelAcc} = 0.80$), and reduces Cloze POS and cosine gradually through $k = 0.9$ (Small LSTM POS $59.0\% \to 34.0\% \to 20.4\%$; AWD-LSTM POS $57.0\% \to 52.2\% \to 42.0\%$; Table~\ref{stab:cloze_values}). By contrast, pruning input connections (\texttt{weight\_ih}) alone collapses Cloze POS to $7.3\%$ at $k = 0.2$ ($0.0\%$ for $k \ge 0.3$) in the Small LSTM and to $5.2\%$ at $k = 0.5$ in the AWD-LSTM.
\end{itemize}

\subsection{Concreteness Classification Methodology}
\label{sec:concreteness_method}

To quantify whether content substitutions under ablation preserve the concreteness level of the target word (as in within-category paraphasia) or shift toward abstract vocabulary (as in generic defaulting), we use the concreteness norms of Brysbaert et al.~\citep{brysbaert2014concreteness}.
This dataset provides human-rated concreteness scores on a 1--5 scale (1 = highly abstract, 5 = highly concrete) for 39,954 English word lemmas, collected from over 4,000 participants via Amazon Mechanical Turk.

For each content word substitution (i.e., predictions classified as neither correct, non-content, nor function words), we compute the \emph{concreteness drop}: $\Delta\text{Conc.} = \text{Conc}(\text{target}) - \text{Conc}(\text{prediction})$.
Words are looked up first in surface form and then via WordNet lemmatisation (across noun, verb, adjective, and adverb parts of speech) so that regular inflected forms (such as \emph{vegetables} $\to$ \emph{vegetable}, 4.89, or \emph{walked} $\to$ \emph{walk}, 4.07) are scored rather than discarded.
A positive value indicates that the prediction is more abstract than the target (loss of specificity); a value near zero indicates that the prediction preserves the target's concreteness level (consistent with within-category substitution); a negative value indicates a shift toward more concrete vocabulary.

Content substitutions are further sub-classified as follows:
\begin{itemize}
\item \textit{Within-category}: Both target and prediction have concreteness $\geq 3.5$ (both concrete) and GloVe 50d cosine similarity $\geq 0.30$ (semantically related). Examples from $k = 0.2$ (Qwen~1.8B MLP, first ablation trial): ``ocean'' $\to$ ``bay'' and ``sea'' $\to$ ``river'' (intact $k = 0$ calibration examples: ``ocean'' $\to$ ``river'', Conc: 4.86 $\to$ 4.89, cos = 0.66; ``glass'' $\to$ ``bowl'', Conc: 4.82 $\to$ 4.87).
\item \textit{Generic defaulting}: Target concreteness $\geq 3.5$ but prediction concreteness $< 3.5$ (concrete $\to$ abstract). Examples from $k = 0.2$ (first ablation trial): ``ball'' $\to$ ``following'' (Conc: 5.00 $\to$ 1.80) and ``heart'' $\to$ ``name'' (Qwen~0.5B attention; calibration example in Section~\ref{sec:pnt_recoding}: ``ball'' $\to$ ``opportunity'', Conc: 5.00 $\to$ 1.88).
\end{itemize}

The mean $\Delta$Conc.\ values reported in Table~\ref{tab:error_dist} and Supplementary Table~\ref{stab:lstm_conn_clinical} are computed across all 10 independent ablation trials ($N = 1{,}000$ responses per condition). In the first ablation trial alone ($n = 100$ responses per condition), the numbers of scored content substitutions and their mean $\Delta$Conc.\ values are as follows: Qwen~1.8B MLP ($n = 38$, $+0.21$), Qwen~0.5B MLP ($n = 29$, $+0.40$), Qwen~1.8B Attn ($n = 29$, $+0.21$), Qwen~0.5B Attn ($n = 18$, $+0.60$), AWD-LSTM unit removal ($n = 49$, $+0.50$), Small LSTM unit removal ($n = 6$, $+0.84$), Small LSTM \texttt{weight\_hh} ($n = 31$, $+0.55$), AWD-LSTM \texttt{weight\_hh} ($n = 50$, $+0.38$), AWD-LSTM \texttt{weight\_ih} ($n = 55$, $+0.43$), and Small LSTM \texttt{weight\_ih} ($n = 0$, because 8 of its 9 substitutions in that trial are hyphens and 1 is \texttt{'s}). Because the three-way classifier allows hyphens and apostrophes inside Content substitutions, single non-alphabetic tokens such as \texttt{-} are counted as Content in Table~\ref{tab:error_dist} and Supplementary Table~\ref{stab:lstm_conn_clinical}. In the first trial, such tokens account for 4 of 10 content substitutions in Small LSTM unit removal, 3 of 22 in Qwen~0.5B Attn, 2 of 52 in AWD-LSTM \texttt{weight\_hh}, 5 of 36 in Small LSTM \texttt{weight\_hh}, 10 of 65 in AWD-LSTM \texttt{weight\_ih}, and all 9 in Small LSTM \texttt{weight\_ih}. By contrast, the PNT coding (Section~\ref{sec:pnt_recoding}) classifies these non-alphabetic tokens as No-response. Finally, expanding the closed-class list to include \emph{much}, \emph{two}, \emph{ones}, and \emph{toward} changes the first-trial values in five conditions (Correct/Content Sub./Non-content, $\Delta$Conc.): Qwen~1.8B MLP ($45/37/18\%$, $+0.19$), Qwen~0.5B Attn ($24/21/55\%$, $+0.45$), AWD-LSTM unit ($9/48/43\%$, $+0.52$), Small LSTM \texttt{weight\_hh} ($1/32/67\%$, $+0.39$), and AWD-LSTM \texttt{weight\_hh} ($22/51/27\%$, $+0.36$).

\subsection{Re-coding into the Philadelphia Naming Test Taxonomy}
\label{sec:pnt_recoding}

\paragraph{Motivation and data.}
The three-way classification of Table~\ref{tab:error_dist} (Correct / Content substitution / Non-content) was designed for this study and does not correspond to any coding scheme in the clinical literature, which prevents direct numerical comparison with published patient data.
We therefore re-coded the same predictions into the seven-category taxonomy conventionally applied to the Philadelphia Naming Test~\citep{roach1996philadelphia, mirman2010moss}, as visualised in Figure~\ref{fig:pnt_main} in Section~\ref{sec:clinical_evidence} and tabulated in Supplementary Table~\ref{stab:pnt_main} below (and Supplementary Table~\ref{stab:lstm_conn_clinical} for the LSTM connection-pruning controls).
We evaluate both the full 10-trial evaluation at $k = 0.2$ ($1{,}000$ predictions per condition across 10 independent ablation trials, reported in Table~\ref{tab:error_dist}, Figure~\ref{fig:pnt_main}, and Supplementary Tables~\ref{stab:lstm_conn_clinical} and~\ref{stab:pnt_main}) and the individual predictions from the first ablation trial listed in Supplements~\ref{sec:supp_qwen18_cloze}--\ref{sec:supp_small_lstm_cloze} (100 prompts for each of the six primary unit-level ablation conditions and four LSTM connection-pruning controls at $k = 0.2$, together with the four undamaged baselines: 14 tables $\times$ 100 prompts $= 1{,}400$ predictions).
Each record consists of the prompt, the expected target, the model's top-1 prediction, and the top-5 candidate list.

\paragraph{Automated decision procedure.}
In clinical neuropsychology, a patient's spoken responses on the Philadelphia Naming Test are categorized by expert examiners based on whether an error shares the target's semantic category, sound structure, both, or neither~\citep{roach1996philadelphia, mirman2010moss}.
To compare computational model predictions directly against clinical patient distributions without subjective scoring bias or manual labeling inconsistency, we operationalize the clinical diagnostic criteria into a fully automatic, deterministic decision procedure.
The procedure evaluates each (target, prediction) pair along the two primary linguistic dimensions defining aphasic speech errors: semantic category relatedness ($s$) and phonological similarity ($p$).

Semantic relatedness, denoted $s \in [0, 1]$, is the maximum Wu--Palmer taxonomy similarity in WordNet~\citep{fellbaum1998wordnet} over synset pairs restricted to the target's annotated part of speech, measuring conceptual proximity in the category hierarchy. Because WordNet organises nouns and verbs into hypernym trees but represents adjectives and adverbs as satellite clusters without hypernym paths, Wu--Palmer similarity returns $s = 0$ for all non-synonymous adjective and adverb pairs ($s = 1$ only when both words share an exact synset). Consequently, across the 27 adjective and adverb prompts (27\% of the battery), non-synonymous substitutions cannot be coded as Semantic or Mixed and fall into Unrelated (or Formal). Closed-class predictions are not eligible to count as semantic paraphasias, since WordNet assigns them senses that yield spuriously high similarity.
We use WordNet category similarity rather than word-embedding cosine because clinical semantic paraphasia means substituting a word from the same category (such as ``river'' or ``bay'' for ``ocean''), whereas embedding cosine also rewards words that merely appear in similar contexts (such as ``boat'' for ``ocean'').
Phonological similarity, denoted $p \in [0, 1]$, measures normalized phonemic overlap: $p = 1 - d/\max(|a|,|b|)$, where $a$ and $b$ are phoneme sequences from the CMU Pronouncing Dictionary (with lexical stress removed), $d$ is the Levenshtein edit distance between them, and $|a|, |b|$ are sequence lengths.
A pair is treated as phonologically related only if $p \geq \theta_p$ \emph{and} the two words share their initial phoneme, which prevents unrelated short words from registering as formal errors.
For example, substituting ``bowl'' (/B OW L/) for target ``ball'' (/B AO L/) alters only one vowel out of three phonemes ($d = 1$, $\max(|a|,|b|) = 3$), giving $p = 1 - 1/3 = 0.67$. Because $p \geq \theta_p = 0.50$ and both share the initial phoneme /B/, this substitution satisfies the phonological criterion.

Using these metrics, the classifier assigns each prediction to one of the seven mutually exclusive PNT categories in hierarchical diagnostic order:

\begin{enumerate}
\item \textit{Correct}: the prediction is an exact match or the target appears in the top-5 list. This is the same criterion used for the Correct column of Table~\ref{tab:error_dist}.
\item \textit{No-response}: the prediction is a tokeniser control token (\texttt{xxunk}, \texttt{xxmaj}, \texttt{xxbos}, \texttt{xxeos}, \texttt{xxpad}, \texttt{unk}, \texttt{eos}, \texttt{pad}), an underscore placeholder, punctuation, an empty string, or a non-Latin token. Specifically, any token failing the lowercase ASCII alphabetic check \texttt{[a-z]+} (such as digits, byte-fallback tokens \texttt{<?>}, or Chinese \texttt{[CJK]} tokens in the bilingual Qwen tokenizers; Supplements~\ref{sec:supp_qwen18_cloze}--\ref{sec:supp_qwen05_cloze}) is assigned to this category.
\item \textit{Neologism}: the prediction is alphabetic but appears in neither WordNet nor the CMU dictionary, and is therefore not an English word.
\item \textit{Mixed}: $s \geq \theta_s$ and the pair is phonologically related.
\item \textit{Semantic}: $s \geq \theta_s$ and the pair is not phonologically related.
\item \textit{Formal}: $s < \theta_s$ and the pair is phonologically related.
\item \textit{Unrelated}: a real word satisfying none of the above.
\end{enumerate}

The parameters $\theta_s$ and $\theta_p$ serve as the decision thresholds (cutoffs) for semantic and phonological relatedness, set to defaults $\theta_s = 0.60$ and $\theta_p = 0.50$.
As a check on the implementation, the resulting Correct percentages in Supplementary Tables~\ref{stab:pnt_main} and~\ref{stab:lstm_conn_clinical} reproduce the Correct column of Table~\ref{tab:error_dist} exactly for all conditions.%
Calibration on canonical examples illustrates how $\theta_s$ discriminates within-category paraphasias from generic defaults: textbook semantic paraphasias yield high similarity exceeding $\theta_s$ ($s = 0.92$ for \emph{moth}~$\to$~\emph{butterfly}; $s = 0.73$ for \emph{ocean}~$\to$~\emph{river}, classified as Semantic), whereas generic or abstract defaults fall well below $\theta_s$ ($s = 0.43$ for \emph{ball}~$\to$~\emph{opportunity}, classified as Unrelated).

\begin{table}[!htbp]
\footnotesize
\centering
\caption{Comparison of undamaged ($k = 0$) and unit-ablated ($k = 0.2$) model predictions with 278 post-stroke aphasia patients (PWA)~\citep{yang2026lesioned} on the seven-category Philadelphia Naming Test (PNT) taxonomy~\citep{roach1996philadelphia, mirman2010moss} (visualised in Figure~\ref{fig:pnt_main} in Section~\ref{sec:clinical_evidence}). At $k = 0.2$, Qwen~1.8B MLP ablation preserves its undamaged rate of semantic paraphasias while non-word and punctuation failures (No-response) rise into the patient range, whereas LSTMs collapse onto unrelated defaults. For the ablated models, values are mean percentages across 10 independent ablation trials ($100$ prompts per trial; 95\% confidence intervals appear in Supplement~\ref{sec:supp_clinical_stats}).}
\label{stab:pnt_main}
\resizebox{\textwidth}{!}{%
\begin{tabular}{@{}llccccccc@{}}
\toprule
Architecture & Condition & Correct & Semantic & Formal & Mixed & Unrelated & Neologism & No-response \\
\midrule
\multicolumn{9}{@{}l}{\textit{Undamaged baselines ($k = 0$, $n = 100$ prompts)}} \\
Qwen 1.8B & Intact ($k{=}0$) & 66.0\% & 11.0\% & 0.0\% & 0.0\% & 22.0\% & 0.0\% & 1.0\% \\
Qwen 0.5B & Intact ($k{=}0$) & 64.0\% & 9.0\% & 0.0\% & 0.0\% & 26.0\% & 0.0\% & 1.0\% \\
AWD-LSTM & Intact ($k{=}0$) & 25.0\% & 19.0\% & 0.0\% & 0.0\% & 53.0\% & 0.0\% & 3.0\% \\
Small LSTM & Intact ($k{=}0$) & 4.0\% & 11.0\% & 1.0\% & 0.0\% & 80.0\% & 0.0\% & 4.0\% \\
\midrule
\multicolumn{9}{@{}l}{\textit{Unit-ablated conditions at $k = 0.2$ (mean across 10 independent ablation trials, $100$ prompts per trial)}} \\
Qwen 1.8B & MLP ($k{=}0.2$) & 42.1\% & 11.7\% & 0.1\% & 0.0\% & 27.9\% & 0.1\% & 18.1\% \\
Qwen 1.8B & Attn ($k{=}0.2$) & 61.2\% & 9.1\% & 0.1\% & 0.0\% & 25.6\% & 0.0\% & 4.0\% \\
Qwen 0.5B & MLP ($k{=}0.2$) & 34.9\% & 9.9\% & 0.2\% & 0.0\% & 38.2\% & 0.1\% & 16.7\% \\
Qwen 0.5B & Attn ($k{=}0.2$) & 46.4\% & 9.0\% & 0.1\% & 0.1\% & 33.0\% & 0.2\% & 11.2\% \\
AWD-LSTM & Whole units ($k{=}0.2$) & 12.3\% & 15.4\% & 0.1\% & 0.0\% & 62.4\% & 0.0\% & 9.8\% \\
Small LSTM & Whole units ($k{=}0.2$) & 0.6\% & 2.3\% & 0.8\% & 0.0\% & 74.1\% & 0.0\% & 22.2\% \\
\midrule
\multicolumn{2}{@{}l}{Post-stroke PWA reference ($N = 278$)~\citep{yang2026lesioned}} & 47.3\% & 4.2\% & 11.6\% & 0.9\% & 6.1\% & 8.0\% & 21.9\% \\
\bottomrule
\end{tabular}%
}
\end{table}

\paragraph{Threshold sensitivity.}
Because the cutoff $\theta_s$ decides whether a real-word mistake counts as Semantic or Unrelated, we re-counted the first-trial Semantic errors across cutoffs from $\theta_s = 0.50$ to $0.80$ (Table~\ref{stab:pnt_sensitivity}).
In that trial, MLP ablation produces more Semantic errors than attention ablation at every cutoff tested in both Qwen~1.8B and Qwen~0.5B.
Across all 10 ablation trials at the default cutoff ($\theta_s = 0.60$), MLP ablation also averages slightly more Semantic errors than attention ablation in both Qwen~1.8B ($11.7\%$ vs.\ $9.1\%$, $+2.6$ percentage points; two-sided permutation $p = 0.0613$) and Qwen~0.5B ($9.9\%$ vs.\ $9.0\%$, $+0.9$ points; $p = 0.6058$). Because this difference does not reach the $\alpha = 0.05$ significance level after multiple-testing correction (Holm-adjusted $p = 0.1227$), we treat the difference in error types between the two pathways as a directional trend rather than a statistically significant split.

\begin{table}[!tbp]
\small
\centering
\caption{Number of predictions classified as Semantic (out of 100 in the first ablation trial) as a function of the semantic threshold $\theta_s$, with $\theta_p$ fixed at 0.50.}
\label{stab:pnt_sensitivity}
\begin{tabular}{@{}llccccccc@{}}
\toprule
Architecture & Pathway & 0.50 & 0.55 & 0.60 & 0.65 & 0.70 & 0.75 & 0.80 \\
\midrule
Qwen 1.8B & MLP & 27 & 18 & 18 & 14 & 10 & 7 & 6 \\
Qwen 1.8B & Attn & 21 & 12 & 11 & 9 & 7 & 4 & 3 \\
Qwen 0.5B & MLP & 17 & 11 & 10 & 8 & 6 & 5 & 3 \\
Qwen 0.5B & Attn & 13 & 7 & 5 & 4 & 2 & 2 & 1 \\
\bottomrule
\end{tabular}
\end{table}

\paragraph{Data and resource availability.}
The single-trial model predictions re-coded here are those listed in Supplements~\ref{sec:supp_qwen18_cloze}--\ref{sec:supp_small_lstm_cloze} of this manuscript, and the 10-trial statistics are aggregated over the 10 independent ablation trials at $k = 0.2$.
Semantic relatedness uses Princeton WordNet~3.0~\citep{fellbaum1998wordnet}, and phonological similarity uses version~0.7a of the Carnegie Mellon University Pronouncing Dictionary.
Both were obtained through the NLTK corpus distribution~\citep{bird2009nltk} (\url{https://www.nltk.org}), at
\url{https://raw.githubusercontent.com/nltk/nltk_data/gh-pages/packages/corpora/wordnet.zip}
and
\url{https://raw.githubusercontent.com/nltk/nltk_data/gh-pages/packages/corpora/cmudict.zip};
the upstream CMUdict repository is \url{https://github.com/cmusphinx/cmudict}.
The patient reference distribution in the final row of Supplementary Table~\ref{stab:pnt_main} is taken from the $N = 278$ C-STAR post-stroke aphasia cohort reported in Table~1 of~\citet{yang2026lesioned} (\url{https://arxiv.org/abs/2607.11621}; mean item counts $82.7$ Correct, $7.3$ Semantic, $20.3$ Formal, $1.6$ Mixed, $10.7$ Unrelated, $14.0$ Neologism, and $38.4$ No-response divided by the $175$ PNT items), scored under the standard Philadelphia Naming Test taxonomy~\citep{roach1996philadelphia, mirman2010moss}.

\paragraph{Limitations of this comparison.}
Four caveats apply when comparing our Cloze predictions with clinical PNT naming data.
First, clinical PNT responses come from picture naming, where a stimulus picture specifies a single target word. Our Cloze prompts provide a sentence context that allows several plausible continuations. As a result, predictions that fit the sentence context without matching the target or its WordNet category are coded as Unrelated. This property of the elicitation task (rather than of the models themselves) explains why Unrelated responses are much more frequent in our models ($25.6$--$81.5\%$ across the 10 ablation trials; $23$--$84\%$ in the first trial) than in patients ($6.1\%$).
Second, the patient reference pools data across severity levels and aphasia subtypes, so it shows whether ablated models fall within the overall aphasic range rather than whether a specific ablation matches a specific syndrome.
Third, the two studies use different scoring instruments: our deterministic WordNet/CMUdict rules versus a classifier trained on speech-language pathologist annotations, so part of the difference reflects measurement rather than model behaviour.
Fourth, WordNet Wu--Palmer similarity is relatively high for pairs of abstract nouns because abstract concepts sit near the top of the WordNet hierarchy, whereas for adjectives and adverbs ($27\%$ of the prompts) WordNet lacks hypernym trees and returns $s = 0$ for all non-synonymous pairs. Consequently, abstract noun pairs slightly inflate the Semantic count near the $\theta_s = 0.60$ cutoff. In the first ablation trial, 34 of 639 classified real-word responses across the $1{,}000$ ablated predictions fall within $0.05$ of $\theta_s = 0.60$ (see Supplementary Table~\ref{stab:pnt_sensitivity} for the full $\theta_s \in [0.50, 0.80]$ sweep).

\subsection{Compensatory Plasticity Sensitivity: Zero Fine-Tuning, Local SGD, and AdamW}\label{sec:supp_plasticity}

To determine how much of the post-lesion behaviour reflects pre-existing structural reserve versus gradient-based weight adaptation, we compare up to four fine-tuning regimes across the full damage sweep $k \in \{0.0, 0.1, \ldots, 0.9\}$ across both recurrent architectures (Small LSTM and AWD-LSTM) and the Transformer models (Qwen~0.5B across both pathways and Qwen~1.8B MLP):
\begin{enumerate}
    \item \texttt{none}: zero fine-tuning steps ($\|\Delta\theta\|_2 = 0$), measuring pure pre-existing architectural redundancy immediately after zeroing the ablated weights.
    \item \texttt{weak} (default main-paper protocol): brief fixed-step SGD without momentum ($\mathrm{lr} = 10^{-3}$, gradient clipping $0.25$; 40 BPTT steps for Small LSTM and 50 BPTT steps for AWD-LSTM, and 62 weight updates for Qwen accumulating gradients over 8 stories across 500 TinyStories passages with per-passage gradient clipping to $0.25$), producing small, strictly local parameter adjustments ($\|\Delta\theta\|_2 \approx 0.0016$--$0.0094$ in Small LSTM, $0.0018$--$0.0119$ in AWD-LSTM, $0.003$--$0.010$ in Qwen~0.5B, and $0.004$--$0.009$ in Qwen~1.8B MLP).
    \item \texttt{strong}: 500 gradient steps of AdamW ($\mathrm{lr} = 10^{-4}$, gradient clipping $1.0$) on TinyStories ($\|\Delta\theta\|_2 \approx 46$--$80$ in Small LSTM [$3.0$--$77$ under \texttt{weight\_ih}], $73$--$176$ in AWD-LSTM, $87$--$116$ in Qwen~0.5B, and $72$--$78$ in Qwen~1.8B MLP for $k \leq 0.2$), representing an upper bound of global weight re-optimisation across all surviving parameters.
    \item \texttt{strong\_wt103} (Small LSTM and AWD-LSTM): 500 gradient steps of AdamW ($\mathrm{lr} = 10^{-4}$, gradient clipping $1.0$) on in-domain WikiText-103 ($\|\Delta\theta\|_2 \approx 34$--$67$ in Small LSTM [$3.4$--$70$ under \texttt{weight\_ih}] and $68$--$168$ in AWD-LSTM), testing whether prolonged in-domain retraining induces a buffered plateau in recurrent architectures.
\end{enumerate}

\begin{figure}[!tbp]
\centering
\includegraphics[width=0.92\textwidth,keepaspectratio]{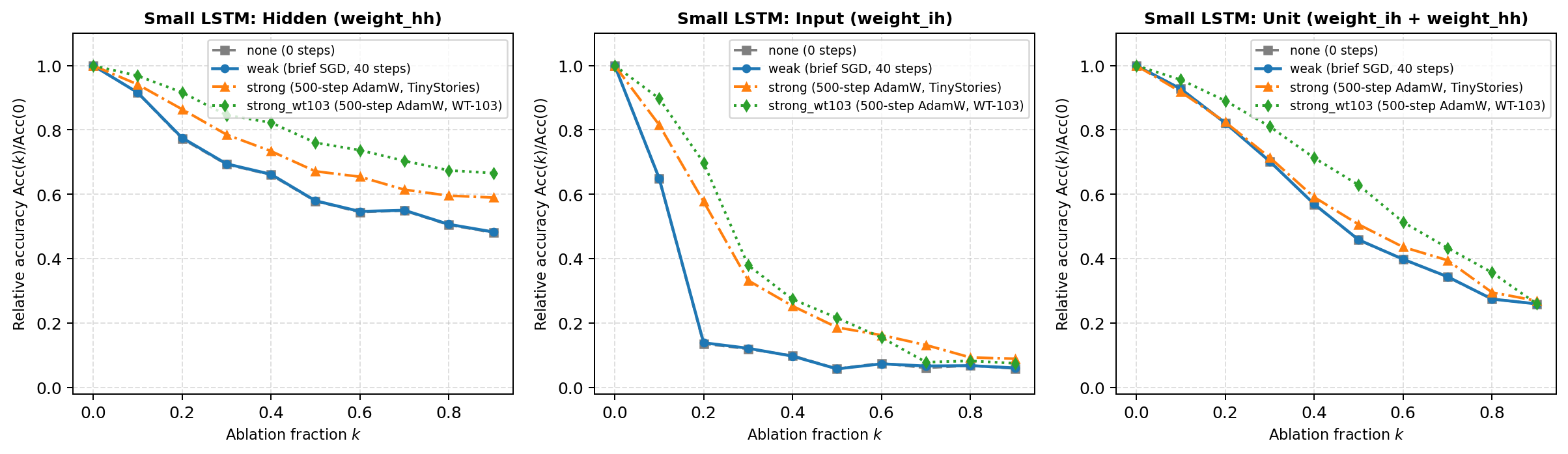}
\caption{Fine-tuning sensitivity across all three ablation modes of the Small LSTM ($k \in [0, 0.9]$ across all four arms): (a)~recurrent-connection pruning (\texttt{weight\_hh}), (b)~input-connection pruning (\texttt{weight\_ih}), and (c)~whole-unit removal (\texttt{weight\_ih} + \texttt{weight\_hh}, primary unit-level target). Curves compare zero fine-tuning (\texttt{none}), brief local SGD (\texttt{weak}, default protocol), 500-step AdamW on TinyStories (\texttt{strong}), and 500-step AdamW on in-domain WikiText-103 (\texttt{strong\_wt103}). As in the AWD-LSTM, \texttt{none} and \texttt{weak} are nearly indistinguishable, and whole-unit and recurrent ablation decline smoothly without a buffered plateau across all four arms.}
\label{sfig:plasticity_small_lstm}
\end{figure}

\begin{figure}[!tbp]
\centering
\includegraphics[width=0.92\textwidth,keepaspectratio]{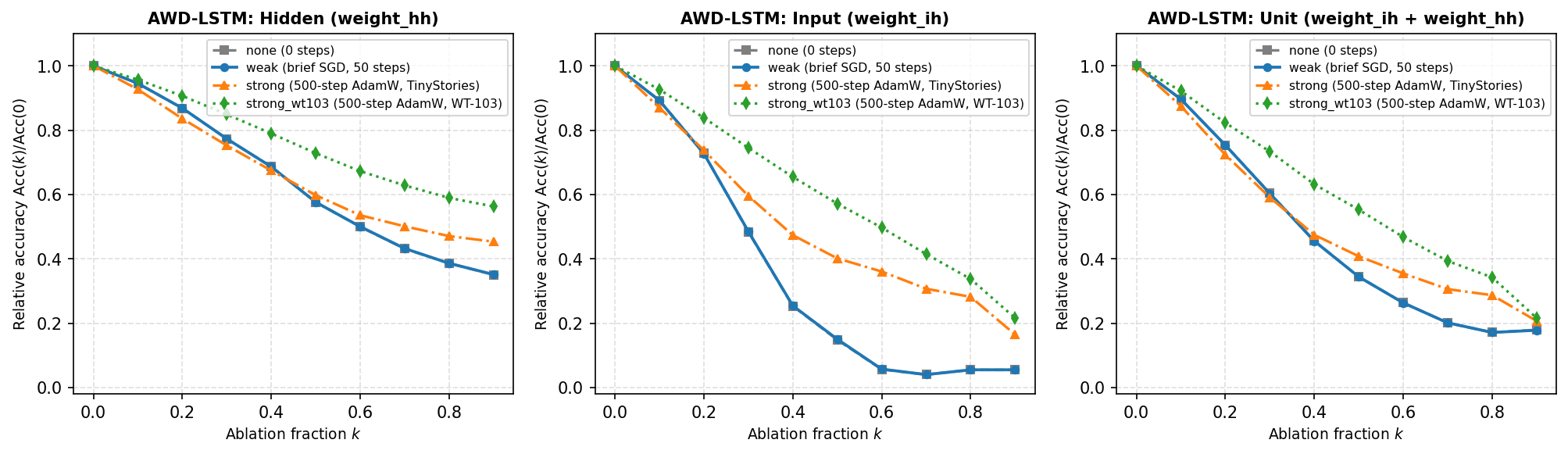}
\caption{Fine-tuning sensitivity across all three ablation modes of the pretrained AWD-LSTM: (a)~recurrent-connection pruning (\texttt{weight\_hh}), (b)~input-connection pruning (\texttt{weight\_ih}), and (c)~whole-unit removal (\texttt{weight\_ih} + \texttt{weight\_hh}, primary unit-level target). Curves compare zero fine-tuning (\texttt{none}), brief local SGD (\texttt{weak}, default protocol), 500-step AdamW on TinyStories (\texttt{strong}), and 500-step AdamW on in-domain WikiText-103 (\texttt{strong\_wt103}). Across all arms, whole-unit and recurrent ablation decline smoothly and monotonically without a buffered plateau.}
\label{sfig:plasticity_lstm}
\end{figure}

\begin{figure}[!tbp]
\centering
\includegraphics[width=0.84\textwidth,keepaspectratio]{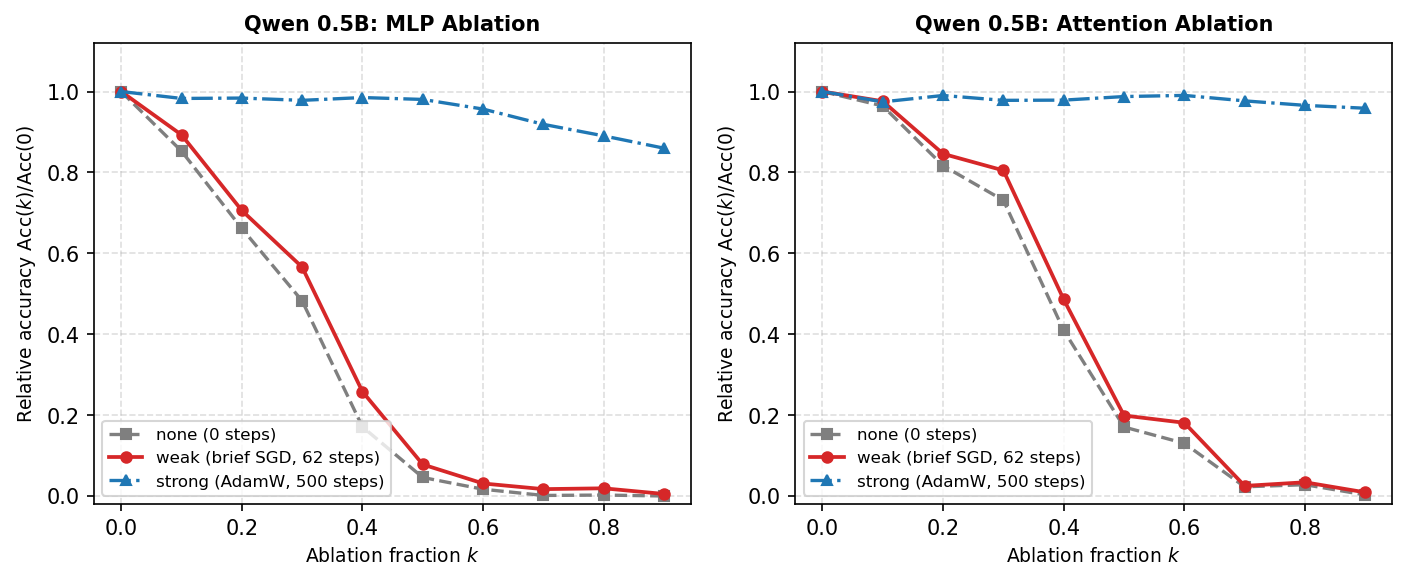}
\caption{Fine-tuning sensitivity across both pathways of the pretrained Qwen~0.5B Transformer ($k \in [0, 0.9]$ across all three arms): (a)~MLP neuron removal and (b)~attention-head removal. Under zero fine-tuning (\texttt{none}) and brief local SGD (\texttt{weak}, 62 updates), MLP ablation declines at mild damage and then collapses sharply at $k \geq 0.4$, while attention ablation shows an early plateau before declining, whereas 500-step global AdamW (\texttt{strong}) re-optimises surviving residual pathways across layers to restore near-baseline accuracy.}
\label{sfig:plasticity_qwen}
\end{figure}

Across all models, this comparison establishes three main findings (Supplementary Figures~\ref{sfig:plasticity_small_lstm}--\ref{sfig:plasticity_qwen}). First, zero fine-tuning (\texttt{none}) and brief local SGD (\texttt{weak}) produce similar curve shapes: even at $\|\Delta\theta\|_2 = 0$, Qwen shows a late threshold collapse under MLP ablation and an early plateau under attention ablation (scoring $2\text{--}6$ percentage points lower in raw accuracy than \texttt{weak}, whereas \texttt{none} and \texttt{weak} coincide in the LSTMs). Second, in the LSTMs, even 500 steps of in-domain AdamW retraining (\texttt{strong\_wt103}) lift overall accuracy at high $k$ without inducing a buffered plateau. Third, in the Transformers, 500 steps of global AdamW (\texttt{strong}) act as wholesale network retraining across surviving layers (recovering $86\%\text{--}99\%$ of baseline accuracy even at $k = 0.9$), which is why we use brief local SGD (\texttt{weak}) as our primary protocol.

Supplementary Figures~\ref{sfig:plasticity_small_lstm} and~\ref{sfig:plasticity_lstm} show all three ablation modes (\texttt{weight\_hh} recurrent connections, \texttt{weight\_ih} input connections, and \texttt{weight\_ih} + \texttt{weight\_hh} whole-unit removal) for the Small LSTM and the pretrained AWD-LSTM, respectively. In both recurrent models and across all three ablation modes, the \texttt{none} and \texttt{weak} curves are virtually identical. In the Small LSTM (baseline $27.5\%$), whole-unit accuracy is $22.6\%$ vs.\ $22.6\%$ at $k = 0.2$ and $12.6\%$ vs.\ $12.6\%$ at $k = 0.5$ (and under \texttt{weight\_hh} pruning, $21.3\%$ vs.\ $21.3\%$ at $k = 0.2$ and $15.9\%$ vs.\ $15.9\%$ at $k = 0.5$). In the AWD-LSTM (baseline $38.1\%$), whole-unit accuracy is $28.7\%$ vs.\ $28.7\%$ at $k = 0.2$ and $13.1\%$ vs.\ $13.1\%$ at $k = 0.5$ (and under \texttt{weight\_hh} pruning, $33.0\%$ vs.\ $33.0\%$ at $k = 0.2$ and $21.9\%$ vs.\ $21.9\%$ at $k = 0.5$). Prolonged 500-step AdamW retraining on in-domain WikiText-103 (\texttt{strong\_wt103}) lifts accuracy at higher damage levels in both LSTMs. In the Small LSTM ($k = 0$ baseline $27.6\%$), whole-unit accuracy at $k = 0.5$ rises from $12.6\%$ under \texttt{none}/\texttt{weak} to $17.4\%$ ($62.9\%$ of baseline), and \texttt{weight\_hh} accuracy rises from $15.9\%$ to $21.0\%$ ($76.0\%$). In the AWD-LSTM, the $k = 0$ baseline falls slightly from $38.05\%$ to $37.02\%$, while at $k = 0.5$ whole-unit accuracy rises from $13.1\%$ to $20.5\%$ and \texttt{weight\_hh} accuracy rises from $21.9\%$ to $26.9\%$. Even so, the whole-unit and recurrent trajectories in both LSTMs remain smooth and gradual without any buffered plateau. Because 40--50 steps of clipped SGD at $\mathrm{lr} = 10^{-3}$ on out-of-domain TinyStories move the LSTM weights by only $\|\Delta\theta\|_2 \le 0.0119$ (relative change ${\sim}10^{-5}$), the LSTM \texttt{weak} arm leaves WikiText-2 accuracy unchanged and is effectively equivalent to \texttt{none}. By contrast, 500 steps of AdamW on the same TinyStories tensor in the \texttt{strong} arm do shift WikiText-2 accuracy (for example, lifting Small LSTM \texttt{weight\_ih} accuracy at $k = 0.2$ from $3.7\%$ to $15.4\%$). Thus, the LSTM and Qwen \texttt{weak} arms are not symmetric.

Supplementary Figure~\ref{sfig:plasticity_qwen} shows the two Qwen~0.5B pathways (MLP neuron removal and attention-head removal). With zero fine-tuning (\texttt{none}), the late threshold collapse is already present, although MLP retention at mild damage is lower than for Small LSTM whole-unit removal under the same arm ($92.8\%$ and $82.2\%$ at $k = 0.1$ and $0.2$). Specifically, Qwen~0.5B retains $85.3\%$ (MLP, $44.0\%$ raw) and $96.4\%$ (Attention, $49.7\%$ raw) of its unadapted $k = 0$ baseline ($51.5\%$) at $k = 0.1$, and $66.1\%$ (MLP, $34.1\%$ raw) and $81.7\%$ (Attention, $42.1\%$ raw) at $k = 0.2$, before dropping steeply to $17.1\%$ ($8.8\%$ raw) at $k = 0.4$ under MLP ablation and to the same level at $k = 0.5$ under attention ablation. Brief local SGD (\texttt{weak}) shifts accuracy modestly upward (for example, MLP at $k = 0.2$: $34.1\% \to 37.9\%$; at $k = 0.3$: $24.9\% \to 30.4\%$) while preserving the sharp cliff at $k \geq 0.4$. The larger Qwen~1.8B MLP pathway exhibits the same pattern. With zero fine-tuning (\texttt{none}, $k = 0$ baseline $50.6\%$), accuracy remains at $48.5\%$ ($96.0\%$ of baseline) at $k = 0.1$, $41.5\%$ ($82.1\%$) at $k = 0.2$, and $27.4\%$ ($54.2\%$) at $k = 0.3$, then drops sharply to $8.9\%$ ($17.6\%$) at $k = 0.4$ and $7.5\%$ ($14.8\%$) at $k = 0.5$. This closely matches the \texttt{weak} trajectory ($51.1\%$ [$93.5\%$] at $k = 0.1$, $44.8\%$ [$82.0\%$] at $k = 0.2$, and $12.9\%$ [$23.5\%$] at $k = 0.4$). By contrast, 500 steps of AdamW (\texttt{strong}, $\|\Delta\theta\|_2 \approx 87$--$116$ in Qwen~0.5B and $72$--$78$ in Qwen~1.8B MLP for $k \leq 0.2$) globally re-optimise the surviving residual stream and un-lesioned layers to bypass the ablated sub-blocks almost completely. This restores Qwen~0.5B accuracy to $98.0\%$ of baseline ($52.2\%$ raw) at $k = 0.5$ and $86.0\%$ ($45.8\%$ raw) at $k = 0.9$ under MLP ablation, and to $95.9\%$--$99.0\%$ ($51.1\%$--$52.7\%$ raw) across all $k \in [0.1, 0.9]$ under attention ablation. In Qwen~1.8B MLP, it restores accuracy to $51.2\%$ raw ($103.1\%$ of the $49.7\%$ \texttt{strong} $k = 0$ baseline) at $k = 0.1$ and $50.7\%$ raw ($102.1\%$) at $k = 0.2$. For Qwen~1.8B, the 500-step \texttt{strong} MLP sweep beyond $k = 0.2$ and the \texttt{none}/\texttt{strong} attention sweeps were omitted due to compute constraints, while the complete 10-point $k \in [0, 0.9]$ sweep across all three arms in Qwen~0.5B and across \texttt{none}/\texttt{weak} in Qwen~1.8B MLP is reported above.

\FloatBarrier
\subsection{Qwen 1.8B: Full Qualitative Cloze Data}\label{sec:supp_qwen18_cloze}

The 14 tables in Supplements~\ref{sec:supp_qwen18_cloze}--\ref{sec:supp_small_lstm_cloze} list the complete 100-prompt predictions from the first ablation trial (\texttt{trial 0}) of each condition. In each table, the fourth column indicates whether the target word (\texttt{Exp.}) matches the top-1 prediction (\texttt{Pred.}; {\color{green!50!black}$\checkmark$}), appears among the model's top-5 predictions ({\color{orange}$\sim$}), or is absent from the top-5 ({\color{red}$\times$}).

\begin{longtable}{@{}p{3.0cm}p{0.9cm}p{1.1cm}cp{3.5cm}@{}}
\caption{Qwen 1.8B: Baseline ($k=0$), all 100 Cloze prompts.} \\
\toprule
Prompt & Exp. & Pred. & & Top-5 \\
\midrule
\endfirsthead
\toprule
Prompt & Exp. & Pred. & & Top-5 \\
\midrule
\endhead
\scriptsize once upon a & \scriptsize time & \scriptsize time & {\color{green!50!black}$\checkmark$} & \scriptsize [time, midnight, lonely, summer, winter] \\
\scriptsize the sun was shining bright in the & \scriptsize sky & \scriptsize sky & {\color{green!50!black}$\checkmark$} & \scriptsize [sky, clear, blue, garden, morning] \\
\scriptsize she picked up the & \scriptsize ball & \scriptsize phone & {\color{orange}$\sim$} & \scriptsize [phone, book, pen, ball, bag] \\
\scriptsize he put the food in his & \scriptsize mouth & \scriptsize mouth & {\color{green!50!black}$\checkmark$} & \scriptsize [mouth, bag, pocket, hand, basket] \\
\scriptsize the little girl was wearing a pretty & \scriptsize dress & \scriptsize dress & {\color{green!50!black}$\checkmark$} & \scriptsize [dress, skirt, white, pink, red] \\
\scriptsize the bird flew up into the & \scriptsize sky & \scriptsize sky & {\color{green!50!black}$\checkmark$} & \scriptsize [sky, air, tree, trees, blue] \\
\scriptsize she gave her friend a big & \scriptsize hug & \scriptsize hug & {\color{green!50!black}$\checkmark$} & \scriptsize [hug, surprise, smile, present, kiss] \\
\scriptsize the cat jumped on the & \scriptsize table & \scriptsize table & {\color{green!50!black}$\checkmark$} & \scriptsize [table, bed, chair, mat, mouse] \\
\scriptsize he found a beautiful & \scriptsize flower & \scriptsize girl & {\color{red}$\times$} & \scriptsize [girl, painting, house, vase, dress] \\
\scriptsize they played together in the & \scriptsize park & \scriptsize park & {\color{green!50!black}$\checkmark$} & \scriptsize [park, same, , school, past] \\
\scriptsize she sang a beautiful & \scriptsize song & \scriptsize song & {\color{green!50!black}$\checkmark$} & \scriptsize [song, , piece, \_\_\_\_\_\_, and] \\
\scriptsize he wanted to go & \scriptsize home & \scriptsize to & {\color{orange}$\sim$} & \scriptsize [to, out, home, on, there] \\
\scriptsize the really big & \scriptsize bear & \scriptsize question & {\color{red}$\times$} & \scriptsize [question, news, one, deal, thing] \\
\scriptsize a very small & \scriptsize mouse & \scriptsize number & {\color{red}$\times$} & \scriptsize [number, and, ,, but, example] \\
\scriptsize the extremely bright & \scriptsize light & \scriptsize light & {\color{green!50!black}$\checkmark$} & \scriptsize [light, and, star, object, white] \\
\scriptsize he was very happy and started to & \scriptsize laugh & \scriptsize sing & {\color{orange}$\sim$} & \scriptsize [sing, dance, play, cry, laugh] \\
\scriptsize the dog saw a cat and started to & \scriptsize bark & \scriptsize bark & {\color{green!50!black}$\checkmark$} & \scriptsize [bark, run, chase, cry, play] \\
\scriptsize they went outside to & \scriptsize play & \scriptsize play & {\color{green!50!black}$\checkmark$} & \scriptsize [play, see, have, enjoy, look] \\
\scriptsize he ran as fast as he & \scriptsize could & \scriptsize could & {\color{green!50!black}$\checkmark$} & \scriptsize [could, can, \_\_\_\_\_\_, c, cou] \\
\scriptsize the baby started to & \scriptsize cry & \scriptsize cry & {\color{green!50!black}$\checkmark$} & \scriptsize [cry, walk, laugh, sleep, speak] \\
\scriptsize he carefully & \scriptsize placed & \scriptsize , & {\color{red}$\times$} & \scriptsize [,, ，, listened, and, to] \\
\scriptsize they quickly & \scriptsize ran & \scriptsize became & {\color{red}$\times$} & \scriptsize [became, got, went, moved, made] \\
\scriptsize she slowly & \scriptsize walked & \scriptsize walked & {\color{green!50!black}$\checkmark$} & \scriptsize [walked, \_\_\_\_\_\_, moved, came, got] \\
\scriptsize he gently & \scriptsize touched & \scriptsize touched & {\color{green!50!black}$\checkmark$} & \scriptsize [touched, shook, took, pulled, placed] \\
\scriptsize he was so tired that he fell & \scriptsize asleep & \scriptsize asleep & {\color{green!50!black}$\checkmark$} & \scriptsize [asleep, a, \_\_\_\_\_\_, , into] \\
\scriptsize she looked at the stars and felt & \scriptsize happy & \scriptsize the & {\color{red}$\times$} & \scriptsize [the, a, that, very, like] \\
\scriptsize she was really & \scriptsize excited & \scriptsize a & {\color{red}$\times$} & \scriptsize [a, good, tired, upset, angry] \\
\scriptsize the water was very & \scriptsize cold & \scriptsize cold & {\color{green!50!black}$\checkmark$} & \scriptsize [cold, clear, dirty, hot, clean] \\
\scriptsize he was feeling very & \scriptsize sad & \scriptsize tired & {\color{orange}$\sim$} & \scriptsize [tired, sad, ill, hungry, sick] \\
\scriptsize it was an incredibly & \scriptsize beautiful & \scriptsize long & {\color{red}$\times$} & \scriptsize [long, hot, busy, difficult, exciting] \\
\scriptsize she had a particularly & \scriptsize special & \scriptsize bad & {\color{red}$\times$} & \scriptsize [bad, difficult, good, hard, strong] \\
\scriptsize the surprisingly & \scriptsize gentle & \scriptsize good & {\color{red}$\times$} & \scriptsize [good, large, short, simple, long] \\
\scriptsize an unusually & \scriptsize quiet & \scriptsize large & {\color{red}$\times$} & \scriptsize [large, strong, high, long, good] \\
\scriptsize the remarkably & \scriptsize clever & \scriptsize good & {\color{red}$\times$} & \scriptsize [good, large, simple, high, low] \\
\scriptsize a wonderfully & \scriptsize warm & \scriptsize simple & {\color{orange}$\sim$} & \scriptsize [simple, unique, creative, warm, illustrated] \\
\scriptsize she opened the door and walked & \scriptsize inside & \scriptsize into & {\color{red}$\times$} & \scriptsize [into, in, out, \_\_\_\_\_\_, to] \\
\scriptsize the story ended & \scriptsize happily & \scriptsize with & {\color{red}$\times$} & \scriptsize [with, in, up, ,, ，] \\
\scriptsize the children were playing & \scriptsize outside & \scriptsize happily & {\color{orange}$\sim$} & \scriptsize [happily, in, on, outside, games] \\
\scriptsize the boat sailed across the & \scriptsize ocean & \scriptsize river & {\color{red}$\times$} & \scriptsize [river, sea, lake, atl, channel] \\
\scriptsize she hung the picture on the & \scriptsize wall & \scriptsize wall & {\color{green!50!black}$\checkmark$} & \scriptsize [wall, \_\_\_\_, , back, \_\_\_\_\_\_] \\
\scriptsize the keys were left on the & \scriptsize counter & \scriptsize table & {\color{red}$\times$} & \scriptsize [table, desk, floor, ground, chair] \\
\scriptsize he parked the car in the & \scriptsize garage & \scriptsize garage & {\color{green!50!black}$\checkmark$} & \scriptsize [garage, driveway, parking, car, street] \\
\scriptsize the plane disappeared behind the & \scriptsize clouds & \scriptsize mountain & {\color{orange}$\sim$} & \scriptsize [mountain, clouds, mountains, hill, trees] \\
\scriptsize they arrived early in the & \scriptsize morning & \scriptsize morning & {\color{green!50!black}$\checkmark$} & \scriptsize [morning, afternoon, city, evening, day] \\
\scriptsize the concert lasted until & \scriptsize midnight & \scriptsize  & {\color{orange}$\sim$} & \scriptsize [, midnight, ten, the, eleven] \\
\scriptsize we will meet again next & \scriptsize week & \scriptsize week & {\color{green!50!black}$\checkmark$} & \scriptsize [week, month, year, time, friday] \\
\scriptsize she poured the milk into the & \scriptsize glass & \scriptsize bowl & {\color{red}$\times$} & \scriptsize [bowl, blender, cup, pot, bottle] \\
\scriptsize he spread butter on the & \scriptsize bread & \scriptsize bread & {\color{green!50!black}$\checkmark$} & \scriptsize [bread, toast, cake, sandwich, slices] \\
\scriptsize the chef chopped the & \scriptsize vegetables & \scriptsize onion & {\color{orange}$\sim$} & \scriptsize [onion, meat, vegetables, chicken, onions] \\
\scriptsize she stirred the soup with a & \scriptsize spoon & \scriptsize spoon & {\color{green!50!black}$\checkmark$} & \scriptsize [spoon, wooden, lad, fork, long] \\
\scriptsize the horse galloped across the & \scriptsize field & \scriptsize field & {\color{green!50!black}$\checkmark$} & \scriptsize [field, grass, open, pr, yard] \\
\scriptsize the fish swam in the & \scriptsize pond & \scriptsize water & {\color{orange}$\sim$} & \scriptsize [water, river, pond, lake, sea] \\
\scriptsize the eagle soared above the & \scriptsize mountains & \scriptsize city & {\color{orange}$\sim$} & \scriptsize [city, mountains, valley, mountain, fields] \\
\scriptsize after the long hike he was extremely & \scriptsize tired & \scriptsize tired & {\color{green!50!black}$\checkmark$} & \scriptsize [tired, exhausted, hungry, \_\_\_\_\_\_, t] \\
\scriptsize the news made everyone very & \scriptsize happy & \scriptsize excited & {\color{orange}$\sim$} & \scriptsize [excited, happy, \_\_\_\_\_\_, angry, \_\_\_\_] \\
\scriptsize the dark alley made her feel & \scriptsize afraid & \scriptsize like & {\color{red}$\times$} & \scriptsize [like, \_\_\_\_\_\_, as, very, uneasy] \\
\scriptsize the puzzle was incredibly & \scriptsize difficult & \scriptsize difficult & {\color{green!50!black}$\checkmark$} & \scriptsize [difficult, hard, easy, challenging, complex] \\
\scriptsize the audience was completely & \scriptsize silent & \scriptsize \_\_\_\_\_\_ & {\color{orange}$\sim$} & \scriptsize [\_\_\_\_\_\_, capt, \_\_\_\_, silent, absorbed] \\
\scriptsize the teacher asked the students to & \scriptsize listen & \scriptsize write & {\color{red}$\times$} & \scriptsize [write, answer, stop, do, \_\_\_\_\_\_] \\
\scriptsize she decided to finally & \scriptsize leave & \scriptsize get & {\color{red}$\times$} & \scriptsize [get, go, take, give, make] \\
\scriptsize the alarm clock started to & \scriptsize ring & \scriptsize ring & {\color{green!50!black}$\checkmark$} & \scriptsize [ring, go, r, \_\_\_\_\_\_, \_\_\_\_] \\
\scriptsize the snow began to & \scriptsize melt & \scriptsize fall & {\color{orange}$\sim$} & \scriptsize [fall, melt, , \_\_\_\_\_\_, fly] \\
\scriptsize he promised he would never & \scriptsize forget & \scriptsize forget & {\color{green!50!black}$\checkmark$} & \scriptsize [forget, do, let, leave, be] \\
\scriptsize the children love to & \scriptsize swim & \scriptsize play & {\color{red}$\times$} & \scriptsize [play, go, watch, see, have] \\
\scriptsize she wanted to learn how to & \scriptsize dance & \scriptsize play & {\color{red}$\times$} & \scriptsize [play, make, drive, use, swim] \\
\scriptsize the mechanic tried to & \scriptsize repair & \scriptsize fix & {\color{orange}$\sim$} & \scriptsize [fix, repair, get, sell, make] \\
\scriptsize dark clouds gathered before the & \scriptsize storm & \scriptsize sun & {\color{orange}$\sim$} & \scriptsize [sun, moon, storm, dawn, rain] \\
\scriptsize the flowers bloomed in the & \scriptsize spring & \scriptsize garden & {\color{orange}$\sim$} & \scriptsize [garden, spring, park, sun, morning] \\
\scriptsize the leaves fall from the trees in & \scriptsize autumn & \scriptsize autumn & {\color{green!50!black}$\checkmark$} & \scriptsize [autumn, the, september, fall, a] \\
\scriptsize the river flowed into the & \scriptsize sea & \scriptsize sea & {\color{green!50!black}$\checkmark$} & \scriptsize [sea, ocean, lake, city, bay] \\
\scriptsize the student wrote an & \scriptsize essay & \scriptsize essay & {\color{green!50!black}$\checkmark$} & \scriptsize [essay, article, excellent, application, assignment] \\
\scriptsize she typed the email on her & \scriptsize keyboard & \scriptsize computer & {\color{red}$\times$} & \scriptsize [computer, phone, mobile, laptop, smart] \\
\scriptsize the professor gave a & \scriptsize lecture & \scriptsize lecture & {\color{green!50!black}$\checkmark$} & \scriptsize [lecture, talk, detailed, speech, very] \\
\scriptsize he submitted his final & \scriptsize report & \scriptsize paper & {\color{red}$\times$} & \scriptsize [paper, essay, papers, thesis, exam] \\
\scriptsize the crowd started to & \scriptsize cheer & \scriptsize gather & {\color{orange}$\sim$} & \scriptsize [gather, get, cheer, \_\_\_\_\_\_, move] \\
\scriptsize he bent down to & \scriptsize pick & \scriptsize pick & {\color{green!50!black}$\checkmark$} & \scriptsize [pick, look, see, examine, take] \\
\scriptsize she leaned forward to & \scriptsize whisper & \scriptsize see & {\color{red}$\times$} & \scriptsize [see, look, take, kiss, give] \\
\scriptsize the wind continued to & \scriptsize blow & \scriptsize blow & {\color{green!50!black}$\checkmark$} & \scriptsize [blow, how, \_\_\_\_\_\_, be, pick] \\
\scriptsize they gathered around to & \scriptsize celebrate & \scriptsize hear & {\color{red}$\times$} & \scriptsize [hear, listen, watch, see, talk] \\
\scriptsize he struggled to & \scriptsize breathe & \scriptsize keep & {\color{red}$\times$} & \scriptsize [keep, get, find, make, \_\_\_\_\_\_] \\
\scriptsize the sunset was absolutely & \scriptsize stunning & \scriptsize beautiful & {\color{orange}$\sim$} & \scriptsize [beautiful, gorgeous, stunning, breathtaking, amazing] \\
\scriptsize the old house looked rather & \scriptsize creepy & \scriptsize small & {\color{red}$\times$} & \scriptsize [small, empty, the, strange, sh] \\
\scriptsize his explanation was perfectly & \scriptsize clear & \scriptsize clear & {\color{green!50!black}$\checkmark$} & \scriptsize [clear, reasonable, \_\_\_\_\_\_, correct, in] \\
\scriptsize the cake tasted absolutely & \scriptsize delicious & \scriptsize delicious & {\color{green!50!black}$\checkmark$} & \scriptsize [delicious, wonderful, fantastic, amazing, terrible] \\
\scriptsize the mountain trail was extremely & \scriptsize steep & \scriptsize difficult & {\color{orange}$\sim$} & \scriptsize [difficult, steep, challenging, beautiful, hard] \\
\scriptsize the instructions were surprisingly & \scriptsize simple & \scriptsize easy & {\color{orange}$\sim$} & \scriptsize [easy, clear, simple, straightforward, straight] \\
\scriptsize the scientist published an important & \scriptsize discovery & \scriptsize paper & {\color{red}$\times$} & \scriptsize [paper, scientific, new, article, book] \\
\scriptsize the city was known for its rich & \scriptsize history & \scriptsize history & {\color{green!50!black}$\checkmark$} & \scriptsize [history, cultural, culture, and, \_\_\_\_\_\_] \\
\scriptsize she always valued honesty and & \scriptsize kindness & \scriptsize integrity & {\color{red}$\times$} & \scriptsize [integrity, hard, trust, truth, fairness] \\
\scriptsize the treaty established lasting & \scriptsize peace & \scriptsize peace & {\color{green!50!black}$\checkmark$} & \scriptsize [peace, security, relations, diplomatic, friendship] \\
\scriptsize the car moved forward & \scriptsize slowly & \scriptsize at & {\color{red}$\times$} & \scriptsize [at, with, ., ,, and] \\
\scriptsize he waited & \scriptsize patiently & \scriptsize for & {\color{red}$\times$} & \scriptsize [for, in, at, and, until] \\
\scriptsize she tied the laces on her & \scriptsize shoes & \scriptsize new & {\color{orange}$\sim$} & \scriptsize [new, shoes, shoe, dress, boots] \\
\scriptsize he wrapped the scarf around his & \scriptsize neck & \scriptsize neck & {\color{green!50!black}$\checkmark$} & \scriptsize [neck, head, waist, shoulders, face] \\
\scriptsize the doctor examined the patient's & \scriptsize heart & \scriptsize chest & {\color{orange}$\sim$} & \scriptsize [chest, heart, throat, \_\_\_\_, stomach] \\
\scriptsize she checked the notifications on her & \scriptsize phone & \scriptsize phone & {\color{green!50!black}$\checkmark$} & \scriptsize [phone, iphone, mobile, smartphone, iPhone] \\
\scriptsize the password must contain at least eight & \scriptsize characters & \scriptsize characters & {\color{green!50!black}$\checkmark$} & \scriptsize [characters, (, digits, letters, character] \\
\scriptsize he plugged the charger into the & \scriptsize socket & \scriptsize wall & {\color{red}$\times$} & \scriptsize [wall, computer, phone, power, outlet] \\
\scriptsize the door began to & \scriptsize open & \scriptsize shake & {\color{orange}$\sim$} & \scriptsize [shake, open, move, c, r] \\
\scriptsize she refused to & \scriptsize quit & \scriptsize accept & {\color{red}$\times$} & \scriptsize [accept, answer, go, give, say] \\
\bottomrule
\end{longtable}

\begin{longtable}{@{}p{3.0cm}p{0.9cm}p{1.1cm}cp{3.5cm}@{}}
\caption{Qwen 1.8B: MLP Ablation ($k=0.2$), all 100 Cloze prompts.} \\
\toprule
Prompt & Exp. & Pred. & & Top-5 \\
\midrule
\endfirsthead
\toprule
Prompt & Exp. & Pred. & & Top-5 \\
\midrule
\endhead
\scriptsize once upon a & \scriptsize time & \scriptsize day & {\color{orange}$\sim$} & \scriptsize [day, summer, time, hill, night] \\
\scriptsize the sun was shining bright in the & \scriptsize sky & \scriptsize sky & {\color{green!50!black}$\checkmark$} & \scriptsize [sky, clouds, , day, blue] \\
\scriptsize she picked up the & \scriptsize ball & \scriptsize ball & {\color{green!50!black}$\checkmark$} & \scriptsize [ball, book, , pen, phone] \\
\scriptsize he put the food in his & \scriptsize mouth & \scriptsize mouth & {\color{green!50!black}$\checkmark$} & \scriptsize [mouth, stomach, , belly, own] \\
\scriptsize the little girl was wearing a pretty & \scriptsize dress & \scriptsize dress & {\color{green!50!black}$\checkmark$} & \scriptsize [dress, , (, red, \_\_\_\_\_\_] \\
\scriptsize the bird flew up into the & \scriptsize sky & \scriptsize trees & {\color{orange}$\sim$} & \scriptsize [trees, sky, air, tree, tall] \\
\scriptsize she gave her friend a big & \scriptsize hug & \scriptsize \_\_\_\_\_\_ & {\color{red}$\times$} & \scriptsize [\_\_\_\_\_\_, surprise, \_\_\_\_\_, birthday, ] \\
\scriptsize the cat jumped on the & \scriptsize table & \scriptsize bed & {\color{red}$\times$} & \scriptsize [bed, ground, top, floor, ] \\
\scriptsize he found a beautiful & \scriptsize flower & \scriptsize girl & {\color{red}$\times$} & \scriptsize [girl, , picture, place, old] \\
\scriptsize they played together in the & \scriptsize park & \scriptsize same & {\color{orange}$\sim$} & \scriptsize [same, park, , past, school] \\
\scriptsize she sang a beautiful & \scriptsize song & \scriptsize song & {\color{green!50!black}$\checkmark$} & \scriptsize [song, tune, , and, poem] \\
\scriptsize he wanted to go & \scriptsize home & \scriptsize to & {\color{red}$\times$} & \scriptsize [to, with, back, and, out] \\
\scriptsize the really big & \scriptsize bear & \scriptsize ones & {\color{red}$\times$} & \scriptsize [ones, thing, deal, stuff, news] \\
\scriptsize a very small & \scriptsize mouse & \scriptsize , & {\color{red}$\times$} & \scriptsize [,, collection, , and, entry] \\
\scriptsize the extremely bright & \scriptsize light & \scriptsize , & {\color{orange}$\sim$} & \scriptsize [,, and, , point, light] \\
\scriptsize he was very happy and started to & \scriptsize laugh & \scriptsize play & {\color{red}$\times$} & \scriptsize [play, , sing, tell, take] \\
\scriptsize the dog saw a cat and started to & \scriptsize bark & \scriptsize chase & {\color{orange}$\sim$} & \scriptsize [chase, follow, run, bark, call] \\
\scriptsize they went outside to & \scriptsize play & \scriptsize play & {\color{green!50!black}$\checkmark$} & \scriptsize [play, see, watch, get, make] \\
\scriptsize he ran as fast as he & \scriptsize could & \scriptsize could & {\color{green!50!black}$\checkmark$} & \scriptsize [could, did, had, was, can] \\
\scriptsize the baby started to & \scriptsize cry & \scriptsize cry & {\color{green!50!black}$\checkmark$} & \scriptsize [cry, grow, start, , make] \\
\scriptsize he carefully & \scriptsize placed & \scriptsize and & {\color{red}$\times$} & \scriptsize [and, ,, as, to, ，] \\
\scriptsize they quickly & \scriptsize ran & \scriptsize became & {\color{red}$\times$} & \scriptsize [became, learned, get, , become] \\
\scriptsize she slowly & \scriptsize walked & \scriptsize began & {\color{red}$\times$} & \scriptsize [began, came, started, , fell] \\
\scriptsize he gently & \scriptsize touched & \scriptsize closed & {\color{orange}$\sim$} & \scriptsize [closed, and, touched, moved, shook] \\
\scriptsize he was so tired that he fell & \scriptsize asleep & \scriptsize into & {\color{orange}$\sim$} & \scriptsize [into, asleep, down, , a] \\
\scriptsize she looked at the stars and felt & \scriptsize happy & \scriptsize the & {\color{red}$\times$} & \scriptsize [the, a, her, that, their] \\
\scriptsize she was really & \scriptsize excited & \scriptsize a & {\color{red}$\times$} & \scriptsize [a, in, , the, good] \\
\scriptsize the water was very & \scriptsize cold & \scriptsize shallow & {\color{red}$\times$} & \scriptsize [shallow, clear, hard, , ,] \\
\scriptsize he was feeling very & \scriptsize sad & \scriptsize  & {\color{red}$\times$} & \scriptsize [, bad, good, \_\_\_\_\_\_, ill] \\
\scriptsize it was an incredibly & \scriptsize beautiful & \scriptsize  & {\color{red}$\times$} & \scriptsize [, hot, busy, day, (] \\
\scriptsize she had a particularly & \scriptsize special & \scriptsize good & {\color{red}$\times$} & \scriptsize [good, hard, strong, bad, (] \\
\scriptsize the surprisingly & \scriptsize gentle & \scriptsize good & {\color{red}$\times$} & \scriptsize [good, small, ,, (, ability] \\
\scriptsize an unusually & \scriptsize quiet & \scriptsize small & {\color{red}$\times$} & \scriptsize [small, warm, long, un, large] \\
\scriptsize the remarkably & \scriptsize clever & \scriptsize , & {\color{red}$\times$} & \scriptsize [,, , (, ability, good] \\
\scriptsize a wonderfully & \scriptsize warm & \scriptsize illustrated & {\color{red}$\times$} & \scriptsize [illustrated, and, full, collection, story] \\
\scriptsize she opened the door and walked & \scriptsize inside & \scriptsize into & {\color{red}$\times$} & \scriptsize [into, in, out, down, onto] \\
\scriptsize the story ended & \scriptsize happily & \scriptsize with & {\color{red}$\times$} & \scriptsize [with, ,, and, in, .] \\
\scriptsize the children were playing & \scriptsize outside & \scriptsize a & {\color{red}$\times$} & \scriptsize [a, with, the, and, ,] \\
\scriptsize the boat sailed across the & \scriptsize ocean & \scriptsize bay & {\color{red}$\times$} & \scriptsize [bay, Bay, , Gulf, river] \\
\scriptsize she hung the picture on the & \scriptsize wall & \scriptsize wall & {\color{green!50!black}$\checkmark$} & \scriptsize [wall, floor, walls, , shelf] \\
\scriptsize the keys were left on the & \scriptsize counter & \scriptsize floor & {\color{red}$\times$} & \scriptsize [floor, ground, keys, table, street] \\
\scriptsize he parked the car in the & \scriptsize garage & \scriptsize middle & {\color{orange}$\sim$} & \scriptsize [middle, corner, driveway, garage, parking] \\
\scriptsize the plane disappeared behind the & \scriptsize clouds & \scriptsize clouds & {\color{green!50!black}$\checkmark$} & \scriptsize [clouds, mountain, mountains, tall, horizon] \\
\scriptsize they arrived early in the & \scriptsize morning & \scriptsize afternoon & {\color{orange}$\sim$} & \scriptsize [afternoon, morning, evening, day, early] \\
\scriptsize the concert lasted until & \scriptsize midnight & \scriptsize  & {\color{orange}$\sim$} & \scriptsize [, the, midnight, about, nine] \\
\scriptsize we will meet again next & \scriptsize week & \scriptsize week & {\color{green!50!black}$\checkmark$} & \scriptsize [week, time, Monday, Friday, weekend] \\
\scriptsize she poured the milk into the & \scriptsize glass & \scriptsize bottle & {\color{red}$\times$} & \scriptsize [bottle, cup, bucket, , pot] \\
\scriptsize he spread butter on the & \scriptsize bread & \scriptsize top & {\color{red}$\times$} & \scriptsize [top, ground, floor, surface, bottom] \\
\scriptsize the chef chopped the & \scriptsize vegetables & \scriptsize meat & {\color{red}$\times$} & \scriptsize [meat, ingredients, whole, pieces, fresh] \\
\scriptsize she stirred the soup with a & \scriptsize spoon & \scriptsize spoon & {\color{green!50!black}$\checkmark$} & \scriptsize [spoon, fork, small, little, hand] \\
\scriptsize the horse galloped across the & \scriptsize field & \scriptsize ground & {\color{orange}$\sim$} & \scriptsize [ground, field, track, fields, trail] \\
\scriptsize the fish swam in the & \scriptsize pond & \scriptsize water & {\color{red}$\times$} & \scriptsize [water, , river, sea, same] \\
\scriptsize the eagle soared above the & \scriptsize mountains & \scriptsize mountain & {\color{orange}$\sim$} & \scriptsize [mountain, mountains, rocky, clouds, tall] \\
\scriptsize after the long hike he was extremely & \scriptsize tired & \scriptsize exhausted & {\color{orange}$\sim$} & \scriptsize [exhausted, tired, relieved, worn, out] \\
\scriptsize the news made everyone very & \scriptsize happy & \scriptsize  & {\color{orange}$\sim$} & \scriptsize [, \_\_\_\_\_\_, happy, , .] \\
\scriptsize the dark alley made her feel & \scriptsize afraid & \scriptsize the & {\color{red}$\times$} & \scriptsize [the, a, both, ., so] \\
\scriptsize the puzzle was incredibly & \scriptsize difficult & \scriptsize difficult & {\color{green!50!black}$\checkmark$} & \scriptsize [difficult, hard, inter, complicated, complex] \\
\scriptsize the audience was completely & \scriptsize silent & \scriptsize in & {\color{red}$\times$} & \scriptsize [in, eng, , \_\_\_\_\_\_, taken] \\
\scriptsize the teacher asked the students to & \scriptsize listen & \scriptsize stand & {\color{red}$\times$} & \scriptsize [stand, , be, turn, stop] \\
\scriptsize she decided to finally & \scriptsize leave & \scriptsize go & {\color{red}$\times$} & \scriptsize [go, get, make, take, start] \\
\scriptsize the alarm clock started to & \scriptsize ring & \scriptsize go & {\color{orange}$\sim$} & \scriptsize [go, ring, be, work, wake] \\
\scriptsize the snow began to & \scriptsize melt & \scriptsize fall & {\color{orange}$\sim$} & \scriptsize [fall, melt, turn, pick, pile] \\
\scriptsize he promised he would never & \scriptsize forget & \scriptsize see & {\color{red}$\times$} & \scriptsize [see, be, come, , say] \\
\scriptsize the children love to & \scriptsize swim & \scriptsize play & {\color{red}$\times$} & \scriptsize [play, go, , tell, be] \\
\scriptsize she wanted to learn how to & \scriptsize dance & \scriptsize play & {\color{orange}$\sim$} & \scriptsize [play, make, be, dance, speak] \\
\scriptsize the mechanic tried to & \scriptsize repair & \scriptsize fix & {\color{orange}$\sim$} & \scriptsize [fix, help, repair, get, make] \\
\scriptsize dark clouds gathered before the & \scriptsize storm & \scriptsize stars & {\color{red}$\times$} & \scriptsize [stars, sun, horizon, star, moon] \\
\scriptsize the flowers bloomed in the & \scriptsize spring & \scriptsize air & {\color{red}$\times$} & \scriptsize [air, sun, ground, water, area] \\
\scriptsize the leaves fall from the trees in & \scriptsize autumn & \scriptsize the & {\color{orange}$\sim$} & \scriptsize [the, a, , late, autumn] \\
\scriptsize the river flowed into the & \scriptsize sea & \scriptsize river & {\color{red}$\times$} & \scriptsize [river, , bay, water, town] \\
\scriptsize the student wrote an & \scriptsize essay & \scriptsize essay & {\color{green!50!black}$\checkmark$} & \scriptsize [essay, letter, book, poem, ] \\
\scriptsize she typed the email on her & \scriptsize keyboard & \scriptsize phone & {\color{red}$\times$} & \scriptsize [phone, computer, own, laptop, ] \\
\scriptsize the professor gave a & \scriptsize lecture & \scriptsize lot & {\color{red}$\times$} & \scriptsize [lot, , very, new, good] \\
\scriptsize he submitted his final & \scriptsize report & \scriptsize exam & {\color{red}$\times$} & \scriptsize [exam, year, essay, draft, defense] \\
\scriptsize the crowd started to & \scriptsize cheer & \scriptsize get & {\color{red}$\times$} & \scriptsize [get, start, turn, go, grow] \\
\scriptsize he bent down to & \scriptsize pick & \scriptsize make & {\color{red}$\times$} & \scriptsize [make, the, listen, take, look] \\
\scriptsize she leaned forward to & \scriptsize whisper & \scriptsize the & {\color{red}$\times$} & \scriptsize [the, face, take, look, me] \\
\scriptsize the wind continued to & \scriptsize blow & \scriptsize be & {\color{orange}$\sim$} & \scriptsize [be, blow, pick, increase, \_\_\_\_\_\_] \\
\scriptsize they gathered around to & \scriptsize celebrate & \scriptsize hear & {\color{red}$\times$} & \scriptsize [hear, listen, help, watch, tell] \\
\scriptsize he struggled to & \scriptsize breathe & \scriptsize keep & {\color{red}$\times$} & \scriptsize [keep, make, find, get, overcome] \\
\scriptsize the sunset was absolutely & \scriptsize stunning & \scriptsize breathtaking & {\color{orange}$\sim$} & \scriptsize [breathtaking, stunning, spectacular, beautiful, amazing] \\
\scriptsize the old house looked rather & \scriptsize creepy & \scriptsize small & {\color{red}$\times$} & \scriptsize [small, old, , (, good] \\
\scriptsize his explanation was perfectly & \scriptsize clear & \scriptsize clear & {\color{green!50!black}$\checkmark$} & \scriptsize [clear, right, in, correct, acceptable] \\
\scriptsize the cake tasted absolutely & \scriptsize delicious & \scriptsize delicious & {\color{green!50!black}$\checkmark$} & \scriptsize [delicious, nothing, amazing, wonderful, no] \\
\scriptsize the mountain trail was extremely & \scriptsize steep & \scriptsize steep & {\color{green!50!black}$\checkmark$} & \scriptsize [steep, difficult, rocky, challenging, hard] \\
\scriptsize the instructions were surprisingly & \scriptsize simple & \scriptsize simple & {\color{green!50!black}$\checkmark$} & \scriptsize [simple, vague, good, small, easy] \\
\scriptsize the scientist published an important & \scriptsize discovery & \scriptsize new & {\color{orange}$\sim$} & \scriptsize [new, discovery, , \_\_\_\_, \_\_\_\_\_\_] \\
\scriptsize the city was known for its rich & \scriptsize history & \scriptsize history & {\color{green!50!black}$\checkmark$} & \scriptsize [history, burial, trading, culture, agricultural] \\
\scriptsize she always valued honesty and & \scriptsize kindness & \scriptsize integrity & {\color{red}$\times$} & \scriptsize [integrity, always, honesty, was, truth] \\
\scriptsize the treaty established lasting & \scriptsize peace & \scriptsize international & {\color{orange}$\sim$} & \scriptsize [international, diplomatic, obligations, ,, peace] \\
\scriptsize the car moved forward & \scriptsize slowly & \scriptsize and & {\color{red}$\times$} & \scriptsize [and, ,, with, to, as] \\
\scriptsize he waited & \scriptsize patiently & \scriptsize for & {\color{red}$\times$} & \scriptsize [for, to, a, in, and] \\
\scriptsize she tied the laces on her & \scriptsize shoes & \scriptsize new & {\color{red}$\times$} & \scriptsize [new, , dress, mother, own] \\
\scriptsize he wrapped the scarf around his & \scriptsize neck & \scriptsize neck & {\color{green!50!black}$\checkmark$} & \scriptsize [neck, nose, face, mouth, left] \\
\scriptsize the doctor examined the patient's & \scriptsize heart & \scriptsize body & {\color{red}$\times$} & \scriptsize [body, blood, medical, tongue, file] \\
\scriptsize she checked the notifications on her & \scriptsize phone & \scriptsize phone & {\color{green!50!black}$\checkmark$} & \scriptsize [phone, cell, iPhone, mobile, iphone] \\
\scriptsize the password must contain at least eight & \scriptsize characters & \scriptsize characters & {\color{green!50!black}$\checkmark$} & \scriptsize [characters, character, digits, consecutive, different] \\
\scriptsize he plugged the charger into the & \scriptsize socket & \scriptsize wall & {\color{red}$\times$} & \scriptsize [wall, , car, outlet, port] \\
\scriptsize the door began to & \scriptsize open & \scriptsize fill & {\color{red}$\times$} & \scriptsize [fill, move, grow, close, make] \\
\scriptsize she refused to & \scriptsize quit & \scriptsize answer & {\color{red}$\times$} & \scriptsize [answer, go, be, take, accept] \\
\bottomrule
\end{longtable}

\begin{longtable}{@{}p{3.0cm}p{0.9cm}p{1.1cm}cp{3.5cm}@{}}
\caption{Qwen 1.8B: Attention Ablation ($k=0.2$), all 100 Cloze prompts.} \\
\toprule
Prompt & Exp. & Pred. & & Top-5 \\
\midrule
\endfirsthead
\toprule
Prompt & Exp. & Pred. & & Top-5 \\
\midrule
\endhead
\scriptsize once upon a & \scriptsize time & \scriptsize time & {\color{green!50!black}$\checkmark$} & \scriptsize [time, child, journey, traveler, great] \\
\scriptsize the sun was shining bright in the & \scriptsize sky & \scriptsize sky & {\color{green!50!black}$\checkmark$} & \scriptsize [sky, park, morning, room, picture] \\
\scriptsize she picked up the & \scriptsize ball & \scriptsize 1 & {\color{red}$\times$} & \scriptsize [1, 0, 2, 3, ] \\
\scriptsize he put the food in his & \scriptsize mouth & \scriptsize mouth & {\color{green!50!black}$\checkmark$} & \scriptsize [mouth, hands, hand, stomach, head] \\
\scriptsize the little girl was wearing a pretty & \scriptsize dress & \scriptsize dress & {\color{green!50!black}$\checkmark$} & \scriptsize [dress, pink, blue, shirt, white] \\
\scriptsize the bird flew up into the & \scriptsize sky & \scriptsize tree & {\color{red}$\times$} & \scriptsize [tree, air, window, house, branches] \\
\scriptsize she gave her friend a big & \scriptsize hug & \scriptsize hug & {\color{green!50!black}$\checkmark$} & \scriptsize [hug, surprise, smile, cake, kiss] \\
\scriptsize the cat jumped on the & \scriptsize table & \scriptsize chair & {\color{orange}$\sim$} & \scriptsize [chair, mouse, table, bed, bag] \\
\scriptsize he found a beautiful & \scriptsize flower & \scriptsize dress & {\color{orange}$\sim$} & \scriptsize [dress, necklace, flower, girl, bird] \\
\scriptsize they played together in the & \scriptsize park & \scriptsize  & {\color{red}$\times$} & \scriptsize [, final, same, championship, national] \\
\scriptsize she sang a beautiful & \scriptsize song & \scriptsize song & {\color{green!50!black}$\checkmark$} & \scriptsize [song, rendition, version, melody, voice] \\
\scriptsize he wanted to go & \scriptsize home & \scriptsize to & {\color{red}$\times$} & \scriptsize [to, and, out, there, with] \\
\scriptsize the really big & \scriptsize bear & \scriptsize deal & {\color{red}$\times$} & \scriptsize [deal, one, question, thing, problem] \\
\scriptsize a very small & \scriptsize mouse & \scriptsize number & {\color{red}$\times$} & \scriptsize [number, amount, fraction, boat, value] \\
\scriptsize the extremely bright & \scriptsize light & \scriptsize star & {\color{orange}$\sim$} & \scriptsize [star, object, light, white, and] \\
\scriptsize he was very happy and started to & \scriptsize laugh & \scriptsize cry & {\color{red}$\times$} & \scriptsize [cry, dance, play, sing, tell] \\
\scriptsize the dog saw a cat and started to & \scriptsize bark & \scriptsize bark & {\color{green!50!black}$\checkmark$} & \scriptsize [bark, chase, run, play, cry] \\
\scriptsize they went outside to & \scriptsize play & \scriptsize play & {\color{green!50!black}$\checkmark$} & \scriptsize [play, see, the, get, look] \\
\scriptsize he ran as fast as he & \scriptsize could & \scriptsize could & {\color{green!50!black}$\checkmark$} & \scriptsize [could, can, possibly, was, had] \\
\scriptsize the baby started to & \scriptsize cry & \scriptsize cry & {\color{green!50!black}$\checkmark$} & \scriptsize [cry, talk, walk, take, make] \\
\scriptsize he carefully & \scriptsize placed & \scriptsize . & {\color{red}$\times$} & \scriptsize [., as, ,, and, to] \\
\scriptsize they quickly & \scriptsize ran & \scriptsize became & {\color{red}$\times$} & \scriptsize [became, and, turned, spread, went] \\
\scriptsize she slowly & \scriptsize walked & \scriptsize and & {\color{orange}$\sim$} & \scriptsize [and, walked, but, moved, \_\_\_\_\_\_] \\
\scriptsize he gently & \scriptsize touched & \scriptsize , & {\color{red}$\times$} & \scriptsize [,, and, opened, ., .] \\
\scriptsize he was so tired that he fell & \scriptsize asleep & \scriptsize asleep & {\color{green!50!black}$\checkmark$} & \scriptsize [asleep, into, a, sleep, sound] \\
\scriptsize she looked at the stars and felt & \scriptsize happy & \scriptsize the & {\color{red}$\times$} & \scriptsize [the, a, like, that, as] \\
\scriptsize she was really & \scriptsize excited & \scriptsize a & {\color{orange}$\sim$} & \scriptsize [a, good, excited, tired, into] \\
\scriptsize the water was very & \scriptsize cold & \scriptsize cold & {\color{green!50!black}$\checkmark$} & \scriptsize [cold, clear, hot, clean, warm] \\
\scriptsize he was feeling very & \scriptsize sad & \scriptsize sick & {\color{orange}$\sim$} & \scriptsize [sick, happy, sad, tired, hungry] \\
\scriptsize it was an incredibly & \scriptsize beautiful & \scriptsize difficult & {\color{red}$\times$} & \scriptsize [difficult, busy, strange, special, exciting] \\
\scriptsize she had a particularly & \scriptsize special & \scriptsize bad & {\color{red}$\times$} & \scriptsize [bad, difficult, good, tough, hard] \\
\scriptsize the surprisingly & \scriptsize gentle & \scriptsize small & {\color{red}$\times$} & \scriptsize [small, short, large, complex, long] \\
\scriptsize an unusually & \scriptsize quiet & \scriptsize large & {\color{red}$\times$} & \scriptsize [large, high, bright, strong, beautiful] \\
\scriptsize the remarkably & \scriptsize clever & \scriptsize good & {\color{red}$\times$} & \scriptsize [good, beautiful, well, low, large] \\
\scriptsize a wonderfully & \scriptsize warm & \scriptsize unique & {\color{red}$\times$} & \scriptsize [unique, illustrated, talented, simple, creative] \\
\scriptsize she opened the door and walked & \scriptsize inside & \scriptsize in & {\color{orange}$\sim$} & \scriptsize [in, into, out, inside, through] \\
\scriptsize the story ended & \scriptsize happily & \scriptsize up & {\color{red}$\times$} & \scriptsize [up, with, in, and, ,] \\
\scriptsize the children were playing & \scriptsize outside & \scriptsize in & {\color{orange}$\sim$} & \scriptsize [in, outside, a, with, on] \\
\scriptsize the boat sailed across the & \scriptsize ocean & \scriptsize river & {\color{red}$\times$} & \scriptsize [river, country, world, bay, water] \\
\scriptsize she hung the picture on the & \scriptsize wall & \scriptsize wall & {\color{green!50!black}$\checkmark$} & \scriptsize [wall, back, mirror, right, door] \\
\scriptsize the keys were left on the & \scriptsize counter & \scriptsize table & {\color{red}$\times$} & \scriptsize [table, door, bathroom, wall, front] \\
\scriptsize he parked the car in the & \scriptsize garage & \scriptsize garage & {\color{green!50!black}$\checkmark$} & \scriptsize [garage, driveway, parking, street, lot] \\
\scriptsize the plane disappeared behind the & \scriptsize clouds & \scriptsize mountain & {\color{orange}$\sim$} & \scriptsize [mountain, hill, mountains, clouds, lake] \\
\scriptsize they arrived early in the & \scriptsize morning & \scriptsize day & {\color{orange}$\sim$} & \scriptsize [day, morning, village, season, town] \\
\scriptsize the concert lasted until & \scriptsize midnight & \scriptsize midnight & {\color{green!50!black}$\checkmark$} & \scriptsize [midnight, the, , it, ten] \\
\scriptsize we will meet again next & \scriptsize week & \scriptsize week & {\color{green!50!black}$\checkmark$} & \scriptsize [week, year, time, month, Monday] \\
\scriptsize she poured the milk into the & \scriptsize glass & \scriptsize blender & {\color{red}$\times$} & \scriptsize [blender, bowl, jug, cake, pan] \\
\scriptsize he spread butter on the & \scriptsize bread & \scriptsize bread & {\color{green!50!black}$\checkmark$} & \scriptsize [bread, pizza, cake, chicken, top] \\
\scriptsize the chef chopped the & \scriptsize vegetables & \scriptsize vegetables & {\color{green!50!black}$\checkmark$} & \scriptsize [vegetables, beef, meat, chicken, fish] \\
\scriptsize she stirred the soup with a & \scriptsize spoon & \scriptsize wooden & {\color{orange}$\sim$} & \scriptsize [wooden, spoon, lad, long, fork] \\
\scriptsize the horse galloped across the & \scriptsize field & \scriptsize field & {\color{green!50!black}$\checkmark$} & \scriptsize [field, sky, pr, grass, ground] \\
\scriptsize the fish swam in the & \scriptsize pond & \scriptsize tank & {\color{red}$\times$} & \scriptsize [tank, water, net, river, fish] \\
\scriptsize the eagle soared above the & \scriptsize mountains & \scriptsize valley & {\color{red}$\times$} & \scriptsize [valley, city, pr, golden, village] \\
\scriptsize after the long hike he was extremely & \scriptsize tired & \scriptsize tired & {\color{green!50!black}$\checkmark$} & \scriptsize [tired, happy, exhausted, fit, fat] \\
\scriptsize the news made everyone very & \scriptsize happy & \scriptsize happy & {\color{green!50!black}$\checkmark$} & \scriptsize [happy, excited, angry, \_\_\_\_\_\_, sad] \\
\scriptsize the dark alley made her feel & \scriptsize afraid & \scriptsize like & {\color{red}$\times$} & \scriptsize [like, safe, safer, as, a] \\
\scriptsize the puzzle was incredibly & \scriptsize difficult & \scriptsize difficult & {\color{green!50!black}$\checkmark$} & \scriptsize [difficult, hard, easy, challenging, satisfying] \\
\scriptsize the audience was completely & \scriptsize silent & \scriptsize capt & {\color{orange}$\sim$} & \scriptsize [capt, absorbed, silent, engaged, and] \\
\scriptsize the teacher asked the students to & \scriptsize listen & \scriptsize write & {\color{red}$\times$} & \scriptsize [write, read, draw, do, make] \\
\scriptsize she decided to finally & \scriptsize leave & \scriptsize get & {\color{red}$\times$} & \scriptsize [get, take, make, go, buy] \\
\scriptsize the alarm clock started to & \scriptsize ring & \scriptsize ring & {\color{green!50!black}$\checkmark$} & \scriptsize [ring, go, beep, be, work] \\
\scriptsize the snow began to & \scriptsize melt & \scriptsize melt & {\color{green!50!black}$\checkmark$} & \scriptsize [melt, fall, sub, disappear, clear] \\
\scriptsize he promised he would never & \scriptsize forget & \scriptsize forget & {\color{green!50!black}$\checkmark$} & \scriptsize [forget, leave, come, let, do] \\
\scriptsize the children love to & \scriptsize swim & \scriptsize play & {\color{red}$\times$} & \scriptsize [play, go, see, have, be] \\
\scriptsize she wanted to learn how to & \scriptsize dance & \scriptsize play & {\color{orange}$\sim$} & \scriptsize [play, make, do, cook, dance] \\
\scriptsize the mechanic tried to & \scriptsize repair & \scriptsize fix & {\color{orange}$\sim$} & \scriptsize [fix, repair, sell, charge, get] \\
\scriptsize dark clouds gathered before the & \scriptsize storm & \scriptsize sun & {\color{orange}$\sim$} & \scriptsize [sun, moon, storm, light, dawn] \\
\scriptsize the flowers bloomed in the & \scriptsize spring & \scriptsize park & {\color{orange}$\sim$} & \scriptsize [park, garden, spring, summer, sun] \\
\scriptsize the leaves fall from the trees in & \scriptsize autumn & \scriptsize the & {\color{orange}$\sim$} & \scriptsize [the, a, autumn, september, late] \\
\scriptsize the river flowed into the & \scriptsize sea & \scriptsize city & {\color{red}$\times$} & \scriptsize [city, lake, dam, reservoir, new] \\
\scriptsize the student wrote an & \scriptsize essay & \scriptsize essay & {\color{green!50!black}$\checkmark$} & \scriptsize [essay, article, assignment, application, exam] \\
\scriptsize she typed the email on her & \scriptsize keyboard & \scriptsize desktop & {\color{red}$\times$} & \scriptsize [desktop, computer, laptop, mobile, phone] \\
\scriptsize the professor gave a & \scriptsize lecture & \scriptsize lecture & {\color{green!50!black}$\checkmark$} & \scriptsize [lecture, lot, wonderful, speech, talk] \\
\scriptsize he submitted his final & \scriptsize report & \scriptsize thesis & {\color{red}$\times$} & \scriptsize [thesis, exam, essay, year, paper] \\
\scriptsize the crowd started to & \scriptsize cheer & \scriptsize dis & {\color{red}$\times$} & \scriptsize [dis, get, die, gather, go] \\
\scriptsize he bent down to & \scriptsize pick & \scriptsize pick & {\color{green!50!black}$\checkmark$} & \scriptsize [pick, look, see, take, make] \\
\scriptsize she leaned forward to & \scriptsize whisper & \scriptsize kiss & {\color{red}$\times$} & \scriptsize [kiss, take, see, give, make] \\
\scriptsize the wind continued to & \scriptsize blow & \scriptsize blow & {\color{green!50!black}$\checkmark$} & \scriptsize [blow, pick, whip, how, be] \\
\scriptsize they gathered around to & \scriptsize celebrate & \scriptsize hear & {\color{red}$\times$} & \scriptsize [hear, see, watch, listen, look] \\
\scriptsize he struggled to & \scriptsize breathe & \scriptsize keep & {\color{orange}$\sim$} & \scriptsize [keep, get, find, make, breathe] \\
\scriptsize the sunset was absolutely & \scriptsize stunning & \scriptsize stunning & {\color{green!50!black}$\checkmark$} & \scriptsize [stunning, beautiful, breathtaking, gorgeous, amazing] \\
\scriptsize the old house looked rather & \scriptsize creepy & \scriptsize for & {\color{red}$\times$} & \scriptsize [for, empty, sad, sh, dull] \\
\scriptsize his explanation was perfectly & \scriptsize clear & \scriptsize clear & {\color{green!50!black}$\checkmark$} & \scriptsize [clear, correct, understandable, reasonable, fine] \\
\scriptsize the cake tasted absolutely & \scriptsize delicious & \scriptsize delicious & {\color{green!50!black}$\checkmark$} & \scriptsize [delicious, wonderful, fantastic, amazing, awful] \\
\scriptsize the mountain trail was extremely & \scriptsize steep & \scriptsize difficult & {\color{orange}$\sim$} & \scriptsize [difficult, crowded, steep, beautiful, hard] \\
\scriptsize the instructions were surprisingly & \scriptsize simple & \scriptsize easy & {\color{orange}$\sim$} & \scriptsize [easy, simple, complex, clear, complicated] \\
\scriptsize the scientist published an important & \scriptsize discovery & \scriptsize discovery & {\color{green!50!black}$\checkmark$} & \scriptsize [discovery, medical, scientific, paper, new] \\
\scriptsize the city was known for its rich & \scriptsize history & \scriptsize history & {\color{green!50!black}$\checkmark$} & \scriptsize [history, and, cultural, people, mineral] \\
\scriptsize she always valued honesty and & \scriptsize kindness & \scriptsize integrity & {\color{red}$\times$} & \scriptsize [integrity, truth, trust, fairness, hard] \\
\scriptsize the treaty established lasting & \scriptsize peace & \scriptsize peace & {\color{green!50!black}$\checkmark$} & \scriptsize [peace, cooperation, relations, diplomatic, friendship] \\
\scriptsize the car moved forward & \scriptsize slowly & \scriptsize at & {\color{red}$\times$} & \scriptsize [at, by, with, ,, and] \\
\scriptsize he waited & \scriptsize patiently & \scriptsize for & {\color{orange}$\sim$} & \scriptsize [for, in, and, ,, patiently] \\
\scriptsize she tied the laces on her & \scriptsize shoes & \scriptsize bike & {\color{red}$\times$} & \scriptsize [bike, own, feet, head, right] \\
\scriptsize he wrapped the scarf around his & \scriptsize neck & \scriptsize head & {\color{orange}$\sim$} & \scriptsize [head, neck, wife, or, face] \\
\scriptsize the doctor examined the patient's & \scriptsize heart & \scriptsize tongue & {\color{orange}$\sim$} & \scriptsize [tongue, eyes, chest, body, heart] \\
\scriptsize she checked the notifications on her & \scriptsize phone & \scriptsize phone & {\color{green!50!black}$\checkmark$} & \scriptsize [phone, iPhone, smartphone, mobile, device] \\
\scriptsize the password must contain at least eight & \scriptsize characters & \scriptsize characters & {\color{green!50!black}$\checkmark$} & \scriptsize [characters, (, letters, character, alphanumeric] \\
\scriptsize he plugged the charger into the & \scriptsize socket & \scriptsize wall & {\color{red}$\times$} & \scriptsize [wall, laptop, computer, charger, phone] \\
\scriptsize the door began to & \scriptsize open & \scriptsize shake & {\color{orange}$\sim$} & \scriptsize [shake, open, c, move, sway] \\
\scriptsize she refused to & \scriptsize quit & \scriptsize say & {\color{red}$\times$} & \scriptsize [say, give, leave, be, let] \\
\bottomrule
\end{longtable}

\subsection{Qwen 0.5B: Full Qualitative Cloze Data}\label{sec:supp_qwen05_cloze}

\begin{longtable}{@{}p{3.0cm}p{0.9cm}p{1.1cm}cp{3.5cm}@{}}
\caption{Qwen 0.5B: Baseline ($k=0$), all 100 Cloze prompts.} \\
\toprule
Prompt & Exp. & Pred. & & Top-5 \\
\midrule
\endfirsthead
\toprule
Prompt & Exp. & Pred. & & Top-5 \\
\midrule
\endhead
\scriptsize once upon a & \scriptsize time & \scriptsize time & {\color{green!50!black}$\checkmark$} & \scriptsize [time, midnight, river, night, mountain] \\
\scriptsize the sun was shining bright in the & \scriptsize sky & \scriptsize sky & {\color{green!50!black}$\checkmark$} & \scriptsize [sky, morning, east, clear, west] \\
\scriptsize she picked up the & \scriptsize ball & \scriptsize book & {\color{red}$\times$} & \scriptsize [book, phone, key, bag, keys] \\
\scriptsize he put the food in his & \scriptsize mouth & \scriptsize mouth & {\color{green!50!black}$\checkmark$} & \scriptsize [mouth, bag, pocket, stomach, hand] \\
\scriptsize the little girl was wearing a pretty & \scriptsize dress & \scriptsize dress & {\color{green!50!black}$\checkmark$} & \scriptsize [dress, pink, red, blue, hat] \\
\scriptsize the bird flew up into the & \scriptsize sky & \scriptsize sky & {\color{green!50!black}$\checkmark$} & \scriptsize [sky, air, tree, trees, blue] \\
\scriptsize she gave her friend a big & \scriptsize hug & \scriptsize hug & {\color{green!50!black}$\checkmark$} & \scriptsize [hug, kiss, present, smile, surprise] \\
\scriptsize the cat jumped on the & \scriptsize table & \scriptsize bed & {\color{orange}$\sim$} & \scriptsize [bed, cat, chair, mat, table] \\
\scriptsize he found a beautiful & \scriptsize flower & \scriptsize place & {\color{orange}$\sim$} & \scriptsize [place, house, garden, flower, bird] \\
\scriptsize they played together in the & \scriptsize park & \scriptsize past & {\color{orange}$\sim$} & \scriptsize [past, same, school, , park] \\
\scriptsize she sang a beautiful & \scriptsize song & \scriptsize song & {\color{green!50!black}$\checkmark$} & \scriptsize [song, voice, ball, piece, and] \\
\scriptsize he wanted to go & \scriptsize home & \scriptsize to & {\color{orange}$\sim$} & \scriptsize [to, home, there, out, back] \\
\scriptsize the really big & \scriptsize bear & \scriptsize one & {\color{red}$\times$} & \scriptsize [one, problem, question, ones, deal] \\
\scriptsize a very small & \scriptsize mouse & \scriptsize amount & {\color{red}$\times$} & \scriptsize [amount, town, number, part, ,] \\
\scriptsize the extremely bright & \scriptsize light & \scriptsize light & {\color{green!50!black}$\checkmark$} & \scriptsize [light, and, lights, star, ,] \\
\scriptsize he was very happy and started to & \scriptsize laugh & \scriptsize sing & {\color{orange}$\sim$} & \scriptsize [sing, dance, play, laugh, talk] \\
\scriptsize the dog saw a cat and started to & \scriptsize bark & \scriptsize run & {\color{red}$\times$} & \scriptsize [run, chase, jump, play, eat] \\
\scriptsize they went outside to & \scriptsize play & \scriptsize play & {\color{green!50!black}$\checkmark$} & \scriptsize [play, have, enjoy, watch, see] \\
\scriptsize he ran as fast as he & \scriptsize could & \scriptsize could & {\color{green!50!black}$\checkmark$} & \scriptsize [could, can, \_\_\_\_\_\_, \_\_\_\_\_, \_\_\_\_] \\
\scriptsize the baby started to & \scriptsize cry & \scriptsize cry & {\color{green!50!black}$\checkmark$} & \scriptsize [cry, sleep, laugh, grow, crawl] \\
\scriptsize he carefully & \scriptsize placed & \scriptsize collected & {\color{red}$\times$} & \scriptsize [collected, read, selected, made, ,] \\
\scriptsize they quickly & \scriptsize ran & \scriptsize and & {\color{red}$\times$} & \scriptsize [and, became, got, get, found] \\
\scriptsize she slowly & \scriptsize walked & \scriptsize walked & {\color{green!50!black}$\checkmark$} & \scriptsize [walked, moved, came, got, made] \\
\scriptsize he gently & \scriptsize touched & \scriptsize touched & {\color{green!50!black}$\checkmark$} & \scriptsize [touched, took, ,, and, picked] \\
\scriptsize he was so tired that he fell & \scriptsize asleep & \scriptsize asleep & {\color{green!50!black}$\checkmark$} & \scriptsize [asleep, into, , down, \_\_\_\_\_\_] \\
\scriptsize she looked at the stars and felt & \scriptsize happy & \scriptsize a & {\color{red}$\times$} & \scriptsize [a, very, the, like, that] \\
\scriptsize she was really & \scriptsize excited & \scriptsize tired & {\color{orange}$\sim$} & \scriptsize [tired, good, a, upset, excited] \\
\scriptsize the water was very & \scriptsize cold & \scriptsize cold & {\color{green!50!black}$\checkmark$} & \scriptsize [cold, dirty, clean, hot, clear] \\
\scriptsize he was feeling very & \scriptsize sad & \scriptsize tired & {\color{orange}$\sim$} & \scriptsize [tired, sad, lonely, sick, ill] \\
\scriptsize it was an incredibly & \scriptsize beautiful & \scriptsize difficult & {\color{orange}$\sim$} & \scriptsize [difficult, busy, long, beautiful, easy] \\
\scriptsize she had a particularly & \scriptsize special & \scriptsize good & {\color{red}$\times$} & \scriptsize [good, bad, strong, nice, long] \\
\scriptsize the surprisingly & \scriptsize gentle & \scriptsize good & {\color{red}$\times$} & \scriptsize [good, little, low, beautiful, long] \\
\scriptsize an unusually & \scriptsize quiet & \scriptsize large & {\color{red}$\times$} & \scriptsize [large, high, good, , bright] \\
\scriptsize the remarkably & \scriptsize clever & \scriptsize good & {\color{red}$\times$} & \scriptsize [good, small, large, low, high] \\
\scriptsize a wonderfully & \scriptsize warm & \scriptsize beautiful & {\color{red}$\times$} & \scriptsize [beautiful, simple, rich, ,, unique] \\
\scriptsize she opened the door and walked & \scriptsize inside & \scriptsize into & {\color{red}$\times$} & \scriptsize [into, in, out, to, towards] \\
\scriptsize the story ended & \scriptsize happily & \scriptsize with & {\color{red}$\times$} & \scriptsize [with, up, on, in, ,] \\
\scriptsize the children were playing & \scriptsize outside & \scriptsize in & {\color{red}$\times$} & \scriptsize [in, on, with, ,, games] \\
\scriptsize the boat sailed across the & \scriptsize ocean & \scriptsize river & {\color{red}$\times$} & \scriptsize [river, lake, sea, water, bay] \\
\scriptsize she hung the picture on the & \scriptsize wall & \scriptsize wall & {\color{green!50!black}$\checkmark$} & \scriptsize [wall, back, window, door, fridge] \\
\scriptsize the keys were left on the & \scriptsize counter & \scriptsize table & {\color{red}$\times$} & \scriptsize [table, desk, shelf, door, top] \\
\scriptsize he parked the car in the & \scriptsize garage & \scriptsize parking & {\color{orange}$\sim$} & \scriptsize [parking, garage, driveway, street, back] \\
\scriptsize the plane disappeared behind the & \scriptsize clouds & \scriptsize clouds & {\color{green!50!black}$\checkmark$} & \scriptsize [clouds, mountains, trees, mountain, hill] \\
\scriptsize they arrived early in the & \scriptsize morning & \scriptsize morning & {\color{green!50!black}$\checkmark$} & \scriptsize [morning, afternoon, evening, day, village] \\
\scriptsize the concert lasted until & \scriptsize midnight & \scriptsize  & {\color{orange}$\sim$} & \scriptsize [, midnight, ten, the, eleven] \\
\scriptsize we will meet again next & \scriptsize week & \scriptsize year & {\color{orange}$\sim$} & \scriptsize [year, week, month, time, weekend] \\
\scriptsize she poured the milk into the & \scriptsize glass & \scriptsize blender & {\color{orange}$\sim$} & \scriptsize [blender, glass, cup, bowl, pan] \\
\scriptsize he spread butter on the & \scriptsize bread & \scriptsize toast & {\color{orange}$\sim$} & \scriptsize [toast, bread, table, cake, plate] \\
\scriptsize the chef chopped the & \scriptsize vegetables & \scriptsize meat & {\color{orange}$\sim$} & \scriptsize [meat, vegetables, chicken, cake, steak] \\
\scriptsize she stirred the soup with a & \scriptsize spoon & \scriptsize spoon & {\color{green!50!black}$\checkmark$} & \scriptsize [spoon, wooden, fork, lad, stick] \\
\scriptsize the horse galloped across the & \scriptsize field & \scriptsize field & {\color{green!50!black}$\checkmark$} & \scriptsize [field, fields, grass, me, open] \\
\scriptsize the fish swam in the & \scriptsize pond & \scriptsize river & {\color{orange}$\sim$} & \scriptsize [river, water, sea, lake, pond] \\
\scriptsize the eagle soared above the & \scriptsize mountains & \scriptsize clouds & {\color{orange}$\sim$} & \scriptsize [clouds, sky, mountains, earth, valley] \\
\scriptsize after the long hike he was extremely & \scriptsize tired & \scriptsize tired & {\color{green!50!black}$\checkmark$} & \scriptsize [tired, thirsty, hungry, exhausted, t] \\
\scriptsize the news made everyone very & \scriptsize happy & \scriptsize \_\_\_\_\_\_ & {\color{orange}$\sim$} & \scriptsize [\_\_\_\_\_\_, happy, excited, \_\_, ] \\
\scriptsize the dark alley made her feel & \scriptsize afraid & \scriptsize like & {\color{red}$\times$} & \scriptsize [like, so, a, very, as] \\
\scriptsize the puzzle was incredibly & \scriptsize difficult & \scriptsize easy & {\color{orange}$\sim$} & \scriptsize [easy, difficult, simple, hard, challenging] \\
\scriptsize the audience was completely & \scriptsize silent & \scriptsize \_\_\_\_\_\_ & {\color{red}$\times$} & \scriptsize [\_\_\_\_\_\_, absorbed, capt, attracted, \_\_\_\_] \\
\scriptsize the teacher asked the students to & \scriptsize listen & \scriptsize write & {\color{red}$\times$} & \scriptsize [write, finish, read, do, \_\_\_\_\_\_] \\
\scriptsize she decided to finally & \scriptsize leave & \scriptsize get & {\color{red}$\times$} & \scriptsize [get, go, take, see, make] \\
\scriptsize the alarm clock started to & \scriptsize ring & \scriptsize ring & {\color{green!50!black}$\checkmark$} & \scriptsize [ring, sound, go, be, make] \\
\scriptsize the snow began to & \scriptsize melt & \scriptsize fall & {\color{orange}$\sim$} & \scriptsize [fall, melt, snow, come, float] \\
\scriptsize he promised he would never & \scriptsize forget & \scriptsize forget & {\color{green!50!black}$\checkmark$} & \scriptsize [forget, leave, be, come, give] \\
\scriptsize the children love to & \scriptsize swim & \scriptsize play & {\color{red}$\times$} & \scriptsize [play, go, see, eat, watch] \\
\scriptsize she wanted to learn how to & \scriptsize dance & \scriptsize cook & {\color{red}$\times$} & \scriptsize [cook, play, drive, make, swim] \\
\scriptsize the mechanic tried to & \scriptsize repair & \scriptsize repair & {\color{green!50!black}$\checkmark$} & \scriptsize [repair, fix, explain, get, diagnose] \\
\scriptsize dark clouds gathered before the & \scriptsize storm & \scriptsize sun & {\color{red}$\times$} & \scriptsize [sun, moon, sky, stars, city] \\
\scriptsize the flowers bloomed in the & \scriptsize spring & \scriptsize garden & {\color{orange}$\sim$} & \scriptsize [garden, spring, park, field, morning] \\
\scriptsize the leaves fall from the trees in & \scriptsize autumn & \scriptsize autumn & {\color{green!50!black}$\checkmark$} & \scriptsize [autumn, the, a, fall, winter] \\
\scriptsize the river flowed into the & \scriptsize sea & \scriptsize lake & {\color{orange}$\sim$} & \scriptsize [lake, sea, river, ocean, pool] \\
\scriptsize the student wrote an & \scriptsize essay & \scriptsize essay & {\color{green!50!black}$\checkmark$} & \scriptsize [essay, article, answer, important, English] \\
\scriptsize she typed the email on her & \scriptsize keyboard & \scriptsize computer & {\color{red}$\times$} & \scriptsize [computer, laptop, phone, cell, mobile] \\
\scriptsize the professor gave a & \scriptsize lecture & \scriptsize lecture & {\color{green!50!black}$\checkmark$} & \scriptsize [lecture, speech, talk, paper, very] \\
\scriptsize he submitted his final & \scriptsize report & \scriptsize report & {\color{green!50!black}$\checkmark$} & \scriptsize [report, answer, draft, paper, papers] \\
\scriptsize the crowd started to & \scriptsize cheer & \scriptsize get & {\color{orange}$\sim$} & \scriptsize [get, cheer, move, gather, shout] \\
\scriptsize he bent down to & \scriptsize pick & \scriptsize pick & {\color{green!50!black}$\checkmark$} & \scriptsize [pick, look, help, the, \_\_\_\_\_\_] \\
\scriptsize she leaned forward to & \scriptsize whisper & \scriptsize look & {\color{red}$\times$} & \scriptsize [look, see, the, kiss, lean] \\
\scriptsize the wind continued to & \scriptsize blow & \scriptsize blow & {\color{green!50!black}$\checkmark$} & \scriptsize [blow, how, change, beat, be] \\
\scriptsize they gathered around to & \scriptsize celebrate & \scriptsize see & {\color{red}$\times$} & \scriptsize [see, watch, listen, hear, talk] \\
\scriptsize he struggled to & \scriptsize breathe & \scriptsize keep & {\color{red}$\times$} & \scriptsize [keep, find, get, make, stay] \\
\scriptsize the sunset was absolutely & \scriptsize stunning & \scriptsize beautiful & {\color{orange}$\sim$} & \scriptsize [beautiful, gorgeous, stunning, breathtaking, amazing] \\
\scriptsize the old house looked rather & \scriptsize creepy & \scriptsize old & {\color{red}$\times$} & \scriptsize [old, small, \_\_\_\_\_\_, , nice] \\
\scriptsize his explanation was perfectly & \scriptsize clear & \scriptsize correct & {\color{orange}$\sim$} & \scriptsize [correct, clear, reasonable, \_\_\_\_\_\_, right] \\
\scriptsize the cake tasted absolutely & \scriptsize delicious & \scriptsize delicious & {\color{green!50!black}$\checkmark$} & \scriptsize [delicious, wonderful, fantastic, terrible, amazing] \\
\scriptsize the mountain trail was extremely & \scriptsize steep & \scriptsize steep & {\color{green!50!black}$\checkmark$} & \scriptsize [steep, difficult, narrow, slippery, dangerous] \\
\scriptsize the instructions were surprisingly & \scriptsize simple & \scriptsize easy & {\color{orange}$\sim$} & \scriptsize [easy, clear, simple, difficult, good] \\
\scriptsize the scientist published an important & \scriptsize discovery & \scriptsize paper & {\color{red}$\times$} & \scriptsize [paper, article, scientific, new, report] \\
\scriptsize the city was known for its rich & \scriptsize history & \scriptsize history & {\color{green!50!black}$\checkmark$} & \scriptsize [history, culture, cultural, and, heritage] \\
\scriptsize she always valued honesty and & \scriptsize kindness & \scriptsize integrity & {\color{red}$\times$} & \scriptsize [integrity, honesty, she, her, was] \\
\scriptsize the treaty established lasting & \scriptsize peace & \scriptsize peace & {\color{green!50!black}$\checkmark$} & \scriptsize [peace, and, economic, relations, diplomatic] \\
\scriptsize the car moved forward & \scriptsize slowly & \scriptsize at & {\color{red}$\times$} & \scriptsize [at, and, ,, with, ] \\
\scriptsize he waited & \scriptsize patiently & \scriptsize for & {\color{red}$\times$} & \scriptsize [for, in, , a, until] \\
\scriptsize she tied the laces on her & \scriptsize shoes & \scriptsize sho & {\color{orange}$\sim$} & \scriptsize [sho, shoes, shoe, horse, hair] \\
\scriptsize he wrapped the scarf around his & \scriptsize neck & \scriptsize neck & {\color{green!50!black}$\checkmark$} & \scriptsize [neck, head, waist, arm, nose] \\
\scriptsize the doctor examined the patient's & \scriptsize heart & \scriptsize body & {\color{red}$\times$} & \scriptsize [body, eyes, leg, feet, foot] \\
\scriptsize she checked the notifications on her & \scriptsize phone & \scriptsize phone & {\color{green!50!black}$\checkmark$} & \scriptsize [phone, computer, mobile, smartphone, iPhone] \\
\scriptsize the password must contain at least eight & \scriptsize characters & \scriptsize characters & {\color{green!50!black}$\checkmark$} & \scriptsize [characters, digits, letters, , of] \\
\scriptsize he plugged the charger into the & \scriptsize socket & \scriptsize wall & {\color{red}$\times$} & \scriptsize [wall, computer, power, outlet, battery] \\
\scriptsize the door began to & \scriptsize open & \scriptsize open & {\color{green!50!black}$\checkmark$} & \scriptsize [open, close, shake, c, sh] \\
\scriptsize she refused to & \scriptsize quit & \scriptsize go & {\color{red}$\times$} & \scriptsize [go, give, take, answer, leave] \\
\bottomrule
\end{longtable}

\begin{longtable}{@{}p{3.0cm}p{0.9cm}p{1.1cm}cp{3.5cm}@{}}
\caption{Qwen 0.5B: MLP Ablation ($k=0.2$), all 100 Cloze prompts.} \\
\toprule
Prompt & Exp. & Pred. & & Top-5 \\
\midrule
\endfirsthead
\toprule
Prompt & Exp. & Pred. & & Top-5 \\
\midrule
\endhead
\scriptsize once upon a & \scriptsize time & \scriptsize once & {\color{red}$\times$} & \scriptsize [once, before, in, when, a] \\
\scriptsize the sun was shining bright in the & \scriptsize sky & \scriptsize sky & {\color{green!50!black}$\checkmark$} & \scriptsize [sky, air, sun, room, morning] \\
\scriptsize she picked up the & \scriptsize ball & \scriptsize things & {\color{red}$\times$} & \scriptsize [things, new, work, book, song] \\
\scriptsize he put the food in his & \scriptsize mouth & \scriptsize 's & {\color{orange}$\sim$} & \scriptsize ['s, room, bag, mouth, ’s] \\
\scriptsize the little girl was wearing a pretty & \scriptsize dress & \scriptsize girl & {\color{red}$\times$} & \scriptsize [girl, little, boy, and, .] \\
\scriptsize the bird flew up into the & \scriptsize sky & \scriptsize sky & {\color{green!50!black}$\checkmark$} & \scriptsize [sky, air, room, clouds, park] \\
\scriptsize she gave her friend a big & \scriptsize hug & \scriptsize and & {\color{red}$\times$} & \scriptsize [and, a, she, what, the] \\
\scriptsize the cat jumped on the & \scriptsize table & \scriptsize dog & {\color{red}$\times$} & \scriptsize [dog, car, train, bed, cat] \\
\scriptsize he found a beautiful & \scriptsize flower & \scriptsize the & {\color{red}$\times$} & \scriptsize [the, he, something, a, book] \\
\scriptsize they played together in the & \scriptsize park & \scriptsize same & {\color{red}$\times$} & \scriptsize [same, room, end, game, classroom] \\
\scriptsize she sang a beautiful & \scriptsize song & \scriptsize she & {\color{orange}$\sim$} & \scriptsize [she, the, song, and, a] \\
\scriptsize he wanted to go & \scriptsize home & \scriptsize to & {\color{red}$\times$} & \scriptsize [to, on, back, for, shopping] \\
\scriptsize the really big & \scriptsize bear & \scriptsize  & {\color{red}$\times$} & \scriptsize [, and, the, , or] \\
\scriptsize a very small & \scriptsize mouse & \scriptsize , & {\color{red}$\times$} & \scriptsize [,, and, space, , things] \\
\scriptsize the extremely bright & \scriptsize light & \scriptsize and & {\color{red}$\times$} & \scriptsize [and, the, -, , ] \\
\scriptsize he was very happy and started to & \scriptsize laugh & \scriptsize work & {\color{red}$\times$} & \scriptsize [work, do, eat, play, make] \\
\scriptsize the dog saw a cat and started to & \scriptsize bark & \scriptsize eat & {\color{red}$\times$} & \scriptsize [eat, play, ., , sleep] \\
\scriptsize they went outside to & \scriptsize play & \scriptsize see & {\color{orange}$\sim$} & \scriptsize [see, the, play, meet, celebrate] \\
\scriptsize he ran as fast as he & \scriptsize could & \scriptsize did & {\color{orange}$\sim$} & \scriptsize [did, in, could, can, as] \\
\scriptsize the baby started to & \scriptsize cry & \scriptsize sleep & {\color{red}$\times$} & \scriptsize [sleep, grow, play, eat, sing] \\
\scriptsize he carefully & \scriptsize placed & \scriptsize he & {\color{red}$\times$} & \scriptsize [he, did, to, she, read] \\
\scriptsize they quickly & \scriptsize ran & \scriptsize get & {\color{red}$\times$} & \scriptsize [get, are, have, see, do] \\
\scriptsize she slowly & \scriptsize walked & \scriptsize came & {\color{red}$\times$} & \scriptsize [came, and, , is, got] \\
\scriptsize he gently & \scriptsize touched & \scriptsize and & {\color{red}$\times$} & \scriptsize [and, ,, touch, to, the] \\
\scriptsize he was so tired that he fell & \scriptsize asleep & \scriptsize asleep & {\color{green!50!black}$\checkmark$} & \scriptsize [asleep, off, ill, into, of] \\
\scriptsize she looked at the stars and felt & \scriptsize happy & \scriptsize the & {\color{red}$\times$} & \scriptsize [the, that, it, a, her] \\
\scriptsize she was really & \scriptsize excited & \scriptsize able & {\color{red}$\times$} & \scriptsize [able, good, happy, very, interested] \\
\scriptsize the water was very & \scriptsize cold & \scriptsize very & {\color{orange}$\sim$} & \scriptsize [very, clean, dirty, ,, cold] \\
\scriptsize he was feeling very & \scriptsize sad & \scriptsize happy & {\color{orange}$\sim$} & \scriptsize [happy, tired, very, well, sad] \\
\scriptsize it was an incredibly & \scriptsize beautiful & \scriptsize , & {\color{red}$\times$} & \scriptsize [,, ., , ..., .] \\
\scriptsize she had a particularly & \scriptsize special & \scriptsize the & {\color{red}$\times$} & \scriptsize [the, and, a, she, interesting] \\
\scriptsize the surprisingly & \scriptsize gentle & \scriptsize , & {\color{red}$\times$} & \scriptsize [,, , (, -, the] \\
\scriptsize an unusually & \scriptsize quiet & \scriptsize  & {\color{red}$\times$} & \scriptsize [, ,, -, and, (] \\
\scriptsize the remarkably & \scriptsize clever & \scriptsize , & {\color{red}$\times$} & \scriptsize [,, , and, ly, the] \\
\scriptsize a wonderfully & \scriptsize warm & \scriptsize , & {\color{red}$\times$} & \scriptsize [,, ，, the, , and] \\
\scriptsize she opened the door and walked & \scriptsize inside & \scriptsize in & {\color{red}$\times$} & \scriptsize [in, into, away, out, through] \\
\scriptsize the story ended & \scriptsize happily & \scriptsize  & {\color{red}$\times$} & \scriptsize [, , ,, 什么意思, with] \\
\scriptsize the children were playing & \scriptsize outside & \scriptsize the & {\color{red}$\times$} & \scriptsize [the, football, games, in, basketball] \\
\scriptsize the boat sailed across the & \scriptsize ocean & \scriptsize river & {\color{red}$\times$} & \scriptsize [river, bridge, road, lake, street] \\
\scriptsize she hung the picture on the & \scriptsize wall & \scriptsize wall & {\color{green!50!black}$\checkmark$} & \scriptsize [wall, bed, table, walls, desk] \\
\scriptsize the keys were left on the & \scriptsize counter & \scriptsize left & {\color{red}$\times$} & \scriptsize [left, door, the, road, in] \\
\scriptsize he parked the car in the & \scriptsize garage & \scriptsize parking & {\color{red}$\times$} & \scriptsize [parking, park, center, centre, same] \\
\scriptsize the plane disappeared behind the & \scriptsize clouds & \scriptsize clouds & {\color{green!50!black}$\checkmark$} & \scriptsize [clouds, sky, sea, , door] \\
\scriptsize they arrived early in the & \scriptsize morning & \scriptsize morning & {\color{green!50!black}$\checkmark$} & \scriptsize [morning, evening, afternoon, early, day] \\
\scriptsize the concert lasted until & \scriptsize midnight & \scriptsize the & {\color{red}$\times$} & \scriptsize [the, , ,, he, until] \\
\scriptsize we will meet again next & \scriptsize week & \scriptsize time & {\color{orange}$\sim$} & \scriptsize [time, day, meeting, week, meet] \\
\scriptsize she poured the milk into the & \scriptsize glass & \scriptsize cup & {\color{orange}$\sim$} & \scriptsize [cup, bowl, glass, basket, bottle] \\
\scriptsize he spread butter on the & \scriptsize bread & \scriptsize bread & {\color{green!50!black}$\checkmark$} & \scriptsize [bread, toast, table, pancakes, cake] \\
\scriptsize the chef chopped the & \scriptsize vegetables & \scriptsize chef & {\color{red}$\times$} & \scriptsize [chef, cake, dog, fish, cook] \\
\scriptsize she stirred the soup with a & \scriptsize spoon & \scriptsize  & {\color{red}$\times$} & \scriptsize [, ., , a, .] \\
\scriptsize the horse galloped across the & \scriptsize field & \scriptsize gap & {\color{orange}$\sim$} & \scriptsize [gap, sky, field, room, road] \\
\scriptsize the fish swam in the & \scriptsize pond & \scriptsize room & {\color{red}$\times$} & \scriptsize [room, park, garden, house, bed] \\
\scriptsize the eagle soared above the & \scriptsize mountains & \scriptsize sky & {\color{red}$\times$} & \scriptsize [sky, clouds, ground, tower, crow] \\
\scriptsize after the long hike he was extremely & \scriptsize tired & \scriptsize tired & {\color{green!50!black}$\checkmark$} & \scriptsize [tired, happy, exhausted, pleased, satisfied] \\
\scriptsize the news made everyone very & \scriptsize happy & \scriptsize happy & {\color{green!50!black}$\checkmark$} & \scriptsize [happy, interesting, careful, glad, much] \\
\scriptsize the dark alley made her feel & \scriptsize afraid & \scriptsize the & {\color{red}$\times$} & \scriptsize [the, his, her, he, it] \\
\scriptsize the puzzle was incredibly & \scriptsize difficult & \scriptsize easy & {\color{orange}$\sim$} & \scriptsize [easy, difficult, challenging, ., hard] \\
\scriptsize the audience was completely & \scriptsize silent & \scriptsize deaf & {\color{orange}$\sim$} & \scriptsize [deaf, silent, satisfied, confused, quiet] \\
\scriptsize the teacher asked the students to & \scriptsize listen & \scriptsize be & {\color{red}$\times$} & \scriptsize [be, the, do, go, sing] \\
\scriptsize she decided to finally & \scriptsize leave & \scriptsize get & {\color{red}$\times$} & \scriptsize [get, go, work, see, start] \\
\scriptsize the alarm clock started to & \scriptsize ring & \scriptsize be & {\color{red}$\times$} & \scriptsize [be, start, work, the, sleep] \\
\scriptsize the snow began to & \scriptsize melt & \scriptsize fall & {\color{red}$\times$} & \scriptsize [fall, snow, be, play, see] \\
\scriptsize he promised he would never & \scriptsize forget & \scriptsize be & {\color{orange}$\sim$} & \scriptsize [be, forget, stop, go, ] \\
\scriptsize the children love to & \scriptsize swim & \scriptsize the & {\color{red}$\times$} & \scriptsize [the, me, eat, play, her] \\
\scriptsize she wanted to learn how to & \scriptsize dance & \scriptsize do & {\color{red}$\times$} & \scriptsize [do, speak, drive, use, play] \\
\scriptsize the mechanic tried to & \scriptsize repair & \scriptsize repair & {\color{green!50!black}$\checkmark$} & \scriptsize [repair, explain, get, help, catch] \\
\scriptsize dark clouds gathered before the & \scriptsize storm & \scriptsize dawn & {\color{red}$\times$} & \scriptsize [dawn, sun, moon, city, earth] \\
\scriptsize the flowers bloomed in the & \scriptsize spring & \scriptsize garden & {\color{orange}$\sim$} & \scriptsize [garden, spring, room, flowers, world] \\
\scriptsize the leaves fall from the trees in & \scriptsize autumn & \scriptsize the & {\color{orange}$\sim$} & \scriptsize [the, spring, their, autumn, it] \\
\scriptsize the river flowed into the & \scriptsize sea & \scriptsize lake & {\color{orange}$\sim$} & \scriptsize [lake, river, city, sea, stream] \\
\scriptsize the student wrote an & \scriptsize essay & \scriptsize teacher & {\color{red}$\times$} & \scriptsize [teacher, , ., , a] \\
\scriptsize she typed the email on her & \scriptsize keyboard & \scriptsize . & {\color{red}$\times$} & \scriptsize [., ., ,, and, computer] \\
\scriptsize the professor gave a & \scriptsize lecture & \scriptsize to & {\color{red}$\times$} & \scriptsize [to, the, a, his, of] \\
\scriptsize he submitted his final & \scriptsize report & \scriptsize  & {\color{red}$\times$} & \scriptsize [, \texttt{<|endoftext|>}, -, his, the] \\
\scriptsize the crowd started to & \scriptsize cheer & \scriptsize sing & {\color{red}$\times$} & \scriptsize [sing, play, listen, eat, laugh] \\
\scriptsize he bent down to & \scriptsize pick & \scriptsize help & {\color{red}$\times$} & \scriptsize [help, his, work, study, the] \\
\scriptsize she leaned forward to & \scriptsize whisper & \scriptsize the & {\color{red}$\times$} & \scriptsize [the, her, , get, go] \\
\scriptsize the wind continued to & \scriptsize blow & \scriptsize blow & {\color{green!50!black}$\checkmark$} & \scriptsize [blow, rain, be, fall, go] \\
\scriptsize they gathered around to & \scriptsize celebrate & \scriptsize celebrate & {\color{green!50!black}$\checkmark$} & \scriptsize [celebrate, listen, see, eat, help] \\
\scriptsize he struggled to & \scriptsize breathe & \scriptsize get & {\color{red}$\times$} & \scriptsize [get, find, make, work, understand] \\
\scriptsize the sunset was absolutely & \scriptsize stunning & \scriptsize fine & {\color{red}$\times$} & \scriptsize [fine, great, the, a, late] \\
\scriptsize the old house looked rather & \scriptsize creepy & \scriptsize well & {\color{red}$\times$} & \scriptsize [well, ., the, very, small] \\
\scriptsize his explanation was perfectly & \scriptsize clear & \scriptsize ． & {\color{red}$\times$} & \scriptsize [．, ., ., and, （] \\
\scriptsize the cake tasted absolutely & \scriptsize delicious & \scriptsize delicious & {\color{green!50!black}$\checkmark$} & \scriptsize [delicious, wonderful, very, well, good] \\
\scriptsize the mountain trail was extremely & \scriptsize steep & \scriptsize difficult & {\color{red}$\times$} & \scriptsize [difficult, easy, close, and, well] \\
\scriptsize the instructions were surprisingly & \scriptsize simple & \scriptsize very & {\color{red}$\times$} & \scriptsize [very, ., , ., easy] \\
\scriptsize the scientist published an important & \scriptsize discovery & \scriptsize something & {\color{red}$\times$} & \scriptsize [something, news, in, , what] \\
\scriptsize the city was known for its rich & \scriptsize history & \scriptsize for & {\color{red}$\times$} & \scriptsize [for, and, city, , people] \\
\scriptsize she always valued honesty and & \scriptsize kindness & \scriptsize her & {\color{red}$\times$} & \scriptsize [her, she, the, honesty, it] \\
\scriptsize the treaty established lasting & \scriptsize peace & \scriptsize  & {\color{red}$\times$} & \scriptsize [, long, lasting, , and] \\
\scriptsize the car moved forward & \scriptsize slowly & \scriptsize  & {\color{red}$\times$} & \scriptsize [, and, ., ., ,] \\
\scriptsize he waited & \scriptsize patiently & \scriptsize for & {\color{red}$\times$} & \scriptsize [for, and, ,, the, ] \\
\scriptsize she tied the laces on her & \scriptsize shoes & \scriptsize and & {\color{red}$\times$} & \scriptsize [and, ., ,, ., ] \\
\scriptsize he wrapped the scarf around his & \scriptsize neck & \scriptsize neck & {\color{green!50!black}$\checkmark$} & \scriptsize [neck, ., body, ,, ] \\
\scriptsize the doctor examined the patient's & \scriptsize heart & \scriptsize the & {\color{red}$\times$} & \scriptsize [the, life, his, health, patient] \\
\scriptsize she checked the notifications on her & \scriptsize phone & \scriptsize phone & {\color{green!50!black}$\checkmark$} & \scriptsize [phone, computer, and, ., ,] \\
\scriptsize the password must contain at least eight & \scriptsize characters & \scriptsize characters & {\color{green!50!black}$\checkmark$} & \scriptsize [characters, digits, words, or, ] \\
\scriptsize he plugged the charger into the & \scriptsize socket & \scriptsize battery & {\color{red}$\times$} & \scriptsize [battery, batteries, car, power, charger] \\
\scriptsize the door began to & \scriptsize open & \scriptsize open & {\color{green!50!black}$\checkmark$} & \scriptsize [open, the, be, walk, go] \\
\scriptsize she refused to & \scriptsize quit & \scriptsize go & {\color{red}$\times$} & \scriptsize [go, eat, accept, help, talk] \\
\bottomrule
\end{longtable}

\begin{longtable}{@{}p{3.0cm}p{0.9cm}p{1.1cm}cp{3.5cm}@{}}
\caption{Qwen 0.5B: Attention Ablation ($k=0.2$), all 100 Cloze prompts.} \\
\toprule
Prompt & Exp. & Pred. & & Top-5 \\
\midrule
\endfirsthead
\toprule
Prompt & Exp. & Pred. & & Top-5 \\
\midrule
\endhead
\scriptsize once upon a & \scriptsize time & \scriptsize time & {\color{green!50!black}$\checkmark$} & \scriptsize [time, man, young, girl, woman] \\
\scriptsize the sun was shining bright in the & \scriptsize sky & \scriptsize  & {\color{red}$\times$} & \scriptsize [, �, -, 要, 我们] \\
\scriptsize she picked up the & \scriptsize ball & \scriptsize following & {\color{red}$\times$} & \scriptsize [following, words, right, photos, food] \\
\scriptsize he put the food in his & \scriptsize mouth & \scriptsize oven & {\color{orange}$\sim$} & \scriptsize [oven, car, room, bag, mouth] \\
\scriptsize the little girl was wearing a pretty & \scriptsize dress & \scriptsize much & {\color{red}$\times$} & \scriptsize [much, , , a, ,] \\
\scriptsize the bird flew up into the & \scriptsize sky & \scriptsize  & {\color{red}$\times$} & \scriptsize [, �, 中, -, ] \\
\scriptsize she gave her friend a big & \scriptsize hug & \scriptsize hug & {\color{green!50!black}$\checkmark$} & \scriptsize [hug, smile, \_\_\_\_\_\_, kiss, surprise] \\
\scriptsize the cat jumped on the & \scriptsize table & \scriptsize  & {\color{red}$\times$} & \scriptsize [, -, *, -, 要] \\
\scriptsize he found a beautiful & \scriptsize flower & \scriptsize bird & {\color{orange}$\sim$} & \scriptsize [bird, flower, girl, butterfly, dog] \\
\scriptsize they played together in the & \scriptsize park & \scriptsize same & {\color{orange}$\sim$} & \scriptsize [same, , classroom, park, garden] \\
\scriptsize she sang a beautiful & \scriptsize song & \scriptsize voice & {\color{orange}$\sim$} & \scriptsize [voice, song, and, melody, ] \\
\scriptsize he wanted to go & \scriptsize home & \scriptsize to & {\color{orange}$\sim$} & \scriptsize [to, home, back, with, on] \\
\scriptsize the really big & \scriptsize bear & \scriptsize most & {\color{red}$\times$} & \scriptsize [most, most, thing, est, ] \\
\scriptsize a very small & \scriptsize mouse & \scriptsize a & {\color{red}$\times$} & \scriptsize [a, ay, est, ., y] \\
\scriptsize the extremely bright & \scriptsize light & \scriptsize most & {\color{red}$\times$} & \scriptsize [most, most, est, the, and] \\
\scriptsize he was very happy and started to & \scriptsize laugh & \scriptsize sing & {\color{red}$\times$} & \scriptsize [sing, play, talk, write, dance] \\
\scriptsize the dog saw a cat and started to & \scriptsize bark & \scriptsize start & {\color{red}$\times$} & \scriptsize [start, do, be, play, eat] \\
\scriptsize they went outside to & \scriptsize play & \scriptsize play & {\color{green!50!black}$\checkmark$} & \scriptsize [play, enjoy, eat, be, see] \\
\scriptsize he ran as fast as he & \scriptsize could & \scriptsize could & {\color{green!50!black}$\checkmark$} & \scriptsize [could, can, jumped, might, thought] \\
\scriptsize the baby started to & \scriptsize cry & \scriptsize  & {\color{red}$\times$} & \scriptsize [, to, , most, 最] \\
\scriptsize he carefully & \scriptsize placed & \scriptsize \_\_\_\_\_ & {\color{red}$\times$} & \scriptsize [\_\_\_\_\_, wrote, \_\_\_\_, \_\_\_\_\_\_, did] \\
\scriptsize they quickly & \scriptsize ran & \scriptsize fall & {\color{red}$\times$} & \scriptsize [fall, , and, add, \_\_\_\_\_\_] \\
\scriptsize she slowly & \scriptsize walked & \scriptsize \_\_\_ & {\color{red}$\times$} & \scriptsize [\_\_\_, \_\_\_\_, \_\_\_\_\_\_, \_\_\_\_, \_\_\_\_\_] \\
\scriptsize he gently & \scriptsize touched & \scriptsize massage & {\color{red}$\times$} & \scriptsize [massage, \_\_\_\_\_\_, remove, wipe, wash] \\
\scriptsize he was so tired that he fell & \scriptsize asleep & \scriptsize asleep & {\color{green!50!black}$\checkmark$} & \scriptsize [asleep, over, sleep, down, ] \\
\scriptsize she looked at the stars and felt & \scriptsize happy & \scriptsize the & {\color{orange}$\sim$} & \scriptsize [the, like, a, sad, happy] \\
\scriptsize she was really & \scriptsize excited & \scriptsize happy & {\color{red}$\times$} & \scriptsize [happy, \_\_\_\_\_\_, good, beautiful, cute] \\
\scriptsize the water was very & \scriptsize cold & \scriptsize  & {\color{red}$\times$} & \scriptsize [, most, most, , much] \\
\scriptsize he was feeling very & \scriptsize sad & \scriptsize sad & {\color{green!50!black}$\checkmark$} & \scriptsize [sad, lonely, tired, bad, happy] \\
\scriptsize it was an incredibly & \scriptsize beautiful & \scriptsize beautiful & {\color{green!50!black}$\checkmark$} & \scriptsize [beautiful, ,, talented, magical, complex] \\
\scriptsize she had a particularly & \scriptsize special & \scriptsize good & {\color{red}$\times$} & \scriptsize [good, bad, nasty, strong, nice] \\
\scriptsize the surprisingly & \scriptsize gentle & \scriptsize most & {\color{red}$\times$} & \scriptsize [most, most, the, , much] \\
\scriptsize an unusually & \scriptsize quiet & \scriptsize * & {\color{red}$\times$} & \scriptsize [*, *, （, ）, ] \\
\scriptsize the remarkably & \scriptsize clever & \scriptsize most & {\color{red}$\times$} & \scriptsize [most, most, est, , the] \\
\scriptsize a wonderfully & \scriptsize warm & \scriptsize * & {\color{red}$\times$} & \scriptsize [*, , **, **, *] \\
\scriptsize she opened the door and walked & \scriptsize inside & \scriptsize in & {\color{red}$\times$} & \scriptsize [in, into, through, \_\_\_\_\_\_, straight] \\
\scriptsize the story ended & \scriptsize happily & \scriptsize  & {\color{red}$\times$} & \scriptsize [, most, , end, �] \\
\scriptsize the children were playing & \scriptsize outside & \scriptsize to & {\color{red}$\times$} & \scriptsize [to, , the, most, are] \\
\scriptsize the boat sailed across the & \scriptsize ocean & \scriptsize 穿过 & {\color{red}$\times$} & \scriptsize [穿过, �, , 飞, ] \\
\scriptsize she hung the picture on the & \scriptsize wall & \scriptsize wall & {\color{green!50!black}$\checkmark$} & \scriptsize [wall, table, board, line, curtain] \\
\scriptsize the keys were left on the & \scriptsize counter & \scriptsize  & {\color{red}$\times$} & \scriptsize [, -, -, , �] \\
\scriptsize he parked the car in the & \scriptsize garage & \scriptsize lot & {\color{red}$\times$} & \scriptsize [lot, driveway, street, car, right] \\
\scriptsize the plane disappeared behind the & \scriptsize clouds & \scriptsize � & {\color{red}$\times$} & \scriptsize [�, , (, , 老师] \\
\scriptsize they arrived early in the & \scriptsize morning & \scriptsize morning & {\color{green!50!black}$\checkmark$} & \scriptsize [morning, world, back, afternoon, evening] \\
\scriptsize the concert lasted until & \scriptsize midnight & \scriptsize most & {\color{red}$\times$} & \scriptsize [most, , 甚至, to, ] \\
\scriptsize we will meet again next & \scriptsize week & \scriptsize to & {\color{red}$\times$} & \scriptsize [to, ., you, ., ,] \\
\scriptsize she poured the milk into the & \scriptsize glass & \scriptsize blender & {\color{red}$\times$} & \scriptsize [blender, fridge, freezer, machine, coffee] \\
\scriptsize he spread butter on the & \scriptsize bread & \scriptsize table & {\color{red}$\times$} & \scriptsize [table, , first, counter, last] \\
\scriptsize the chef chopped the & \scriptsize vegetables & \scriptsize - & {\color{red}$\times$} & \scriptsize [-, , *, -, 1] \\
\scriptsize she stirred the soup with a & \scriptsize spoon & \scriptsize spoon & {\color{green!50!black}$\checkmark$} & \scriptsize [spoon, fork, splash, pinch, string] \\
\scriptsize the horse galloped across the & \scriptsize field & \scriptsize � & {\color{red}$\times$} & \scriptsize [�, , 要, 穿过, -] \\
\scriptsize the fish swam in the & \scriptsize pond & \scriptsize 中 & {\color{red}$\times$} & \scriptsize [中, , �, 要, -] \\
\scriptsize the eagle soared above the & \scriptsize mountains & \scriptsize  & {\color{red}$\times$} & \scriptsize [, �, -, 要, (] \\
\scriptsize after the long hike he was extremely & \scriptsize tired & \scriptsize tired & {\color{green!50!black}$\checkmark$} & \scriptsize [tired, disappointed, nervous, thirsty, proud] \\
\scriptsize the news made everyone very & \scriptsize happy & \scriptsize  & {\color{red}$\times$} & \scriptsize [, first, most, one, ] \\
\scriptsize the dark alley made her feel & \scriptsize afraid & \scriptsize to & {\color{red}$\times$} & \scriptsize [to, , , even, floor] \\
\scriptsize the puzzle was incredibly & \scriptsize difficult & \scriptsize most & {\color{red}$\times$} & \scriptsize [most, , , most, you] \\
\scriptsize the audience was completely & \scriptsize silent & \scriptsize most & {\color{red}$\times$} & \scriptsize [most, the, most, to, ,] \\
\scriptsize the teacher asked the students to & \scriptsize listen & \scriptsize do & {\color{red}$\times$} & \scriptsize [do, be, ask, the, talk] \\
\scriptsize she decided to finally & \scriptsize leave & \scriptsize get & {\color{red}$\times$} & \scriptsize [get, have, make, be, find] \\
\scriptsize the alarm clock started to & \scriptsize ring & \scriptsize  & {\color{red}$\times$} & \scriptsize [, to, most, �, most] \\
\scriptsize the snow began to & \scriptsize melt & \scriptsize most & {\color{red}$\times$} & \scriptsize [most, 雪, snow, , most] \\
\scriptsize he promised he would never & \scriptsize forget & \scriptsize have & {\color{orange}$\sim$} & \scriptsize [have, be, forget, let, know] \\
\scriptsize the children love to & \scriptsize swim & \scriptsize  & {\color{red}$\times$} & \scriptsize [, the, pet, to, most] \\
\scriptsize she wanted to learn how to & \scriptsize dance & \scriptsize be & {\color{red}$\times$} & \scriptsize [be, make, drive, learn, play] \\
\scriptsize the mechanic tried to & \scriptsize repair & \scriptsize the & {\color{red}$\times$} & \scriptsize [the, most, to, most, ] \\
\scriptsize dark clouds gathered before the & \scriptsize storm & \scriptsize storm & {\color{green!50!black}$\checkmark$} & \scriptsize [storm, rain, big, battle, sun] \\
\scriptsize the flowers bloomed in the & \scriptsize spring & \scriptsize - & {\color{red}$\times$} & \scriptsize [-, 中, �, 要, ] \\
\scriptsize the leaves fall from the trees in & \scriptsize autumn & \scriptsize the & {\color{red}$\times$} & \scriptsize [the, which, english, our, their] \\
\scriptsize the river flowed into the & \scriptsize sea & \scriptsize  & {\color{red}$\times$} & \scriptsize [, 与, vest, -, �] \\
\scriptsize the student wrote an & \scriptsize essay & \scriptsize - & {\color{red}$\times$} & \scriptsize [-, , 1, -, 0] \\
\scriptsize she typed the email on her & \scriptsize keyboard & \scriptsize keyboard & {\color{green!50!black}$\checkmark$} & \scriptsize [keyboard, phone, computer, laptop, way] \\
\scriptsize the professor gave a & \scriptsize lecture & \scriptsize  & {\color{red}$\times$} & \scriptsize [, -, , （, �] \\
\scriptsize he submitted his final & \scriptsize report & \scriptsize exam & {\color{red}$\times$} & \scriptsize [exam, answer, examination, exams, test] \\
\scriptsize the crowd started to & \scriptsize cheer & \scriptsize  & {\color{red}$\times$} & \scriptsize [, to, the, most, most] \\
\scriptsize he bent down to & \scriptsize pick & \scriptsize help & {\color{red}$\times$} & \scriptsize [help, pray, love, listen, hear] \\
\scriptsize she leaned forward to & \scriptsize whisper & \scriptsize the & {\color{red}$\times$} & \scriptsize [the, , your, help, me] \\
\scriptsize the wind continued to & \scriptsize blow & \scriptsize most & {\color{red}$\times$} & \scriptsize [most, most, to, , the] \\
\scriptsize they gathered around to & \scriptsize celebrate & \scriptsize see & {\color{red}$\times$} & \scriptsize [see, help, be, learn, watch] \\
\scriptsize he struggled to & \scriptsize breathe & \scriptsize do & {\color{red}$\times$} & \scriptsize [do, get, drive, work, write] \\
\scriptsize the sunset was absolutely & \scriptsize stunning & \scriptsize most & {\color{red}$\times$} & \scriptsize [most, , everything, 要, must] \\
\scriptsize the old house looked rather & \scriptsize creepy & \scriptsize  & {\color{red}$\times$} & \scriptsize [, even, 甚至, �, most] \\
\scriptsize his explanation was perfectly & \scriptsize clear & \scriptsize good & {\color{red}$\times$} & \scriptsize [good, right, reasonable, correct, fine] \\
\scriptsize the cake tasted absolutely & \scriptsize delicious & \scriptsize most & {\color{red}$\times$} & \scriptsize [most, , to, most, the] \\
\scriptsize the mountain trail was extremely & \scriptsize steep & \scriptsize the & {\color{red}$\times$} & \scriptsize [the, mountain, , to, most] \\
\scriptsize the instructions were surprisingly & \scriptsize simple & \scriptsize most & {\color{red}$\times$} & \scriptsize [most, , the, ,, ] \\
\scriptsize the scientist published an important & \scriptsize discovery & \scriptsize is & {\color{red}$\times$} & \scriptsize [is, new, thing, things, natural] \\
\scriptsize the city was known for its rich & \scriptsize history & \scriptsize and & {\color{red}$\times$} & \scriptsize [and, most, , , est] \\
\scriptsize she always valued honesty and & \scriptsize kindness & \scriptsize kindness & {\color{green!50!black}$\checkmark$} & \scriptsize [kindness, integrity, trust, had, respect] \\
\scriptsize the treaty established lasting & \scriptsize peace & \scriptsize  & {\color{red}$\times$} & \scriptsize [, to, last, most, 最后] \\
\scriptsize the car moved forward & \scriptsize slowly & \scriptsize to & {\color{red}$\times$} & \scriptsize [to, , , most, most] \\
\scriptsize he waited & \scriptsize patiently & \scriptsize stoi & {\color{red}$\times$} & \scriptsize [stoi, 弁, /************************, 椅, AdminController] \\
\scriptsize she tied the laces on her & \scriptsize shoes & \scriptsize back & {\color{red}$\times$} & \scriptsize [back, head, dress, body, leg] \\
\scriptsize he wrapped the scarf around his & \scriptsize neck & \scriptsize neck & {\color{green!50!black}$\checkmark$} & \scriptsize [neck, finger, mouth, girlfriend, nose] \\
\scriptsize the doctor examined the patient's & \scriptsize heart & \scriptsize name & {\color{red}$\times$} & \scriptsize [name, health, room, first, handwriting] \\
\scriptsize she checked the notifications on her & \scriptsize phone & \scriptsize phone & {\color{green!50!black}$\checkmark$} & \scriptsize [phone, computer, mobile, iPhone, screen] \\
\scriptsize the password must contain at least eight & \scriptsize characters & \scriptsize most & {\color{red}$\times$} & \scriptsize [most, , �, 最, 注] \\
\scriptsize he plugged the charger into the & \scriptsize socket & \scriptsize wall & {\color{orange}$\sim$} & \scriptsize [wall, socket, outlet, plug, power] \\
\scriptsize the door began to & \scriptsize open & \scriptsize  & {\color{red}$\times$} & \scriptsize [, most, to, most, ] \\
\scriptsize she refused to & \scriptsize quit & \scriptsize be & {\color{red}$\times$} & \scriptsize [be, leave, give, speak, buy] \\
\bottomrule
\end{longtable}

\subsection{AWD-LSTM: Full Qualitative Cloze Data}\label{sec:supp_awd_cloze}

\begin{longtable}{@{}p{3.0cm}p{0.9cm}p{1.1cm}cp{3.5cm}@{}}
\caption{AWD-LSTM: Baseline ($k=0$), all 100 Cloze prompts.} \\
\toprule
Prompt & Exp. & Pred. & & Top-5 \\
\midrule
\endfirsthead
\toprule
Prompt & Exp. & Pred. & & Top-5 \\
\midrule
\endhead
\scriptsize once upon a & \scriptsize time & \scriptsize time & {\color{green!50!black}$\checkmark$} & \scriptsize [time, television, live, single, young] \\
\scriptsize the sun was shining bright in the & \scriptsize sky & \scriptsize sky & {\color{green!50!black}$\checkmark$} & \scriptsize [sky, morning, summer, sun, night] \\
\scriptsize she picked up the & \scriptsize ball & \scriptsize first & {\color{red}$\times$} & \scriptsize [first, idea, news, story, best] \\
\scriptsize he put the food in his & \scriptsize mouth & \scriptsize hands & {\color{red}$\times$} & \scriptsize [hands, kitchen, house, own, car] \\
\scriptsize the little girl was wearing a pretty & \scriptsize dress & \scriptsize black & {\color{orange}$\sim$} & \scriptsize [black, girl, dress, white, long] \\
\scriptsize the bird flew up into the & \scriptsize sky & \scriptsize air & {\color{orange}$\sim$} & \scriptsize [air, sky, ground, forest, water] \\
\scriptsize she gave her friend a big & \scriptsize hug & \scriptsize - & {\color{red}$\times$} & \scriptsize [-, deal, picture, show, drink] \\
\scriptsize the cat jumped on the & \scriptsize table & \scriptsize ground & {\color{red}$\times$} & \scriptsize [ground, floor, roof, ice, screen] \\
\scriptsize he found a beautiful & \scriptsize flower & \scriptsize woman & {\color{red}$\times$} & \scriptsize [woman, ,, and, female, body] \\
\scriptsize they played together in the & \scriptsize park & \scriptsize early & {\color{red}$\times$} & \scriptsize [early, same, 1960s, late, film] \\
\scriptsize she sang a beautiful & \scriptsize song & \scriptsize song & {\color{green!50!black}$\checkmark$} & \scriptsize [song, ballad, ,, version, rendition] \\
\scriptsize he wanted to go & \scriptsize home & \scriptsize to & {\color{red}$\times$} & \scriptsize [to, on, back, into, through] \\
\scriptsize the really big & \scriptsize bear & \scriptsize - & {\color{red}$\times$} & \scriptsize [-, ,, thing, and, video] \\
\scriptsize a very small & \scriptsize mouse & \scriptsize , & {\color{red}$\times$} & \scriptsize [,, and, -, but, (] \\
\scriptsize the extremely bright & \scriptsize light & \scriptsize and & {\color{red}$\times$} & \scriptsize [and, ,, red, black, blue] \\
\scriptsize he was very happy and started to & \scriptsize laugh & \scriptsize feel & {\color{red}$\times$} & \scriptsize [feel, write, develop, make, be] \\
\scriptsize the dog saw a cat and started to & \scriptsize bark & \scriptsize look & {\color{red}$\times$} & \scriptsize [look, fall, show, play, eat] \\
\scriptsize they went outside to & \scriptsize play & \scriptsize meet & {\color{red}$\times$} & \scriptsize [meet, make, perform, work, the] \\
\scriptsize he ran as fast as he & \scriptsize could & \scriptsize could & {\color{green!50!black}$\checkmark$} & \scriptsize [could, had, was, did, would] \\
\scriptsize the baby started to & \scriptsize cry & \scriptsize grow & {\color{red}$\times$} & \scriptsize [grow, die, appear, fall, be] \\
\scriptsize he carefully & \scriptsize placed & \scriptsize kept & {\color{red}$\times$} & \scriptsize [kept, studied, prepared, planned, monitored] \\
\scriptsize they quickly & \scriptsize ran & \scriptsize became & {\color{red}$\times$} & \scriptsize [became, become, won, reached, split] \\
\scriptsize she slowly & \scriptsize walked & \scriptsize became & {\color{red}$\times$} & \scriptsize [became, lost, broke, moved, rose] \\
\scriptsize he gently & \scriptsize touched & \scriptsize sings & {\color{red}$\times$} & \scriptsize [sings, pitched, turned, pushed, breaks] \\
\scriptsize he was so tired that he fell & \scriptsize asleep & \scriptsize out & {\color{orange}$\sim$} & \scriptsize [out, into, in, ill, asleep] \\
\scriptsize she looked at the stars and felt & \scriptsize happy & \scriptsize that & {\color{red}$\times$} & \scriptsize [that, they, the, she, it] \\
\scriptsize she was really & \scriptsize excited & \scriptsize a & {\color{red}$\times$} & \scriptsize [a, well, ,, the, "] \\
\scriptsize the water was very & \scriptsize cold & \scriptsize well & {\color{orange}$\sim$} & \scriptsize [well, similar, cold, good, difficult] \\
\scriptsize he was feeling very & \scriptsize sad & \scriptsize ill & {\color{red}$\times$} & \scriptsize [ill, good, guilty, weak, comfortable] \\
\scriptsize it was an incredibly & \scriptsize beautiful & \scriptsize difficult & {\color{red}$\times$} & \scriptsize [difficult, expensive, popular, complex, controversial] \\
\scriptsize she had a particularly & \scriptsize special & \scriptsize strong & {\color{red}$\times$} & \scriptsize [strong, close, high, difficult, good] \\
\scriptsize the surprisingly & \scriptsize gentle & \scriptsize large & {\color{red}$\times$} & \scriptsize [large, strong, short, rare, successful] \\
\scriptsize an unusually & \scriptsize quiet & \scriptsize large & {\color{red}$\times$} & \scriptsize [large, high, low, long, strong] \\
\scriptsize the remarkably & \scriptsize clever & \scriptsize large & {\color{red}$\times$} & \scriptsize [large, rare, strong, short, high] \\
\scriptsize a wonderfully & \scriptsize warm & \scriptsize developed & {\color{red}$\times$} & \scriptsize [developed, directed, edited, animated, designed] \\
\scriptsize she opened the door and walked & \scriptsize inside & \scriptsize away & {\color{red}$\times$} & \scriptsize [away, the, into, off, out] \\
\scriptsize the story ended & \scriptsize happily & \scriptsize in & {\color{red}$\times$} & \scriptsize [in, with, when, after, on] \\
\scriptsize the children were playing & \scriptsize outside & \scriptsize in & {\color{red}$\times$} & \scriptsize [in, the, with, on, a] \\
\scriptsize the boat sailed across the & \scriptsize ocean & \scriptsize ocean & {\color{green!50!black}$\checkmark$} & \scriptsize [ocean, river, sea, waters, coast] \\
\scriptsize she hung the picture on the & \scriptsize wall & \scriptsize front & {\color{orange}$\sim$} & \scriptsize [front, wall, floor, ceiling, back] \\
\scriptsize the keys were left on the & \scriptsize counter & \scriptsize floor & {\color{red}$\times$} & \scriptsize [floor, ground, door, roof, keys] \\
\scriptsize he parked the car in the & \scriptsize garage & \scriptsize car & {\color{orange}$\sim$} & \scriptsize [car, parking, middle, garage, park] \\
\scriptsize the plane disappeared behind the & \scriptsize clouds & \scriptsize plane & {\color{red}$\times$} & \scriptsize [plane, wreckage, wall, ground, main] \\
\scriptsize they arrived early in the & \scriptsize morning & \scriptsize morning & {\color{green!50!black}$\checkmark$} & \scriptsize [morning, day, year, evening, summer] \\
\scriptsize the concert lasted until & \scriptsize midnight & \scriptsize the & {\color{red}$\times$} & \scriptsize [the, 9, 30, 2011, 2010] \\
\scriptsize we will meet again next & \scriptsize week & \scriptsize year & {\color{orange}$\sim$} & \scriptsize [year, next, week, season, day] \\
\scriptsize she poured the milk into the & \scriptsize glass & \scriptsize water & {\color{red}$\times$} & \scriptsize [water, air, milk, soil, ground] \\
\scriptsize he spread butter on the & \scriptsize bread & \scriptsize streets & {\color{red}$\times$} & \scriptsize [streets, internet, market, ice, ground] \\
\scriptsize the chef chopped the & \scriptsize vegetables & \scriptsize ice & {\color{red}$\times$} & \scriptsize [ice, chef, table, restaurant, cake] \\
\scriptsize she stirred the soup with a & \scriptsize spoon & \scriptsize large & {\color{red}$\times$} & \scriptsize [large, black, few, drink, bottle] \\
\scriptsize the horse galloped across the & \scriptsize field & \scriptsize mountain & {\color{red}$\times$} & \scriptsize [mountain, river, street, streets, desert] \\
\scriptsize the fish swam in the & \scriptsize pond & \scriptsize water & {\color{red}$\times$} & \scriptsize [water, ocean, waters, sea, river] \\
\scriptsize the eagle soared above the & \scriptsize mountains & \scriptsize ground & {\color{red}$\times$} & \scriptsize [ground, earth, eagle, sun, moon] \\
\scriptsize after the long hike he was extremely & \scriptsize tired & \scriptsize popular & {\color{red}$\times$} & \scriptsize [popular, successful, busy, ill, unpopular] \\
\scriptsize the news made everyone very & \scriptsize happy & \scriptsize happy & {\color{green!50!black}$\checkmark$} & \scriptsize [happy, interested, popular, close, comfortable] \\
\scriptsize the dark alley made her feel & \scriptsize afraid & \scriptsize like & {\color{red}$\times$} & \scriptsize [like, more, for, so, very] \\
\scriptsize the puzzle was incredibly & \scriptsize difficult & \scriptsize popular & {\color{red}$\times$} & \scriptsize [popular, complex, well, successful, complicated] \\
\scriptsize the audience was completely & \scriptsize silent & \scriptsize divided & {\color{red}$\times$} & \scriptsize [divided, shocked, upset, surprised, disappointed] \\
\scriptsize the teacher asked the students to & \scriptsize listen & \scriptsize teach & {\color{red}$\times$} & \scriptsize [teach, do, perform, help, take] \\
\scriptsize she decided to finally & \scriptsize leave & \scriptsize make & {\color{red}$\times$} & \scriptsize [make, retire, take, become, go] \\
\scriptsize the alarm clock started to & \scriptsize ring & \scriptsize be & {\color{red}$\times$} & \scriptsize [be, appear, spread, grow, become] \\
\scriptsize the snow began to & \scriptsize melt & \scriptsize fall & {\color{orange}$\sim$} & \scriptsize [fall, appear, burn, grow, melt] \\
\scriptsize he promised he would never & \scriptsize forget & \scriptsize be & {\color{red}$\times$} & \scriptsize [be, return, do, make, have] \\
\scriptsize the children love to & \scriptsize swim & \scriptsize be & {\color{red}$\times$} & \scriptsize [be, play, have, write, love] \\
\scriptsize she wanted to learn how to & \scriptsize dance & \scriptsize do & {\color{red}$\times$} & \scriptsize [do, play, use, make, write] \\
\scriptsize the mechanic tried to & \scriptsize repair & \scriptsize use & {\color{red}$\times$} & \scriptsize [use, make, create, get, solve] \\
\scriptsize dark clouds gathered before the & \scriptsize storm & \scriptsize moon & {\color{red}$\times$} & \scriptsize [moon, dark, sun, night, sky] \\
\scriptsize the flowers bloomed in the & \scriptsize spring & \scriptsize garden & {\color{red}$\times$} & \scriptsize [garden, flowers, sky, trees, tree] \\
\scriptsize the leaves fall from the trees in & \scriptsize autumn & \scriptsize the & {\color{red}$\times$} & \scriptsize [the, a, their, trees, which] \\
\scriptsize the river flowed into the & \scriptsize sea & \scriptsize river & {\color{orange}$\sim$} & \scriptsize [river, city, sea, valley, lake] \\
\scriptsize the student wrote an & \scriptsize essay & \scriptsize article & {\color{orange}$\sim$} & \scriptsize [article, essay, autobiography, editorial, account] \\
\scriptsize she typed the email on her & \scriptsize keyboard & \scriptsize phone & {\color{red}$\times$} & \scriptsize [phone, website, blog, own, desk] \\
\scriptsize the professor gave a & \scriptsize lecture & \scriptsize speech & {\color{orange}$\sim$} & \scriptsize [speech, lecture, series, positive, list] \\
\scriptsize he submitted his final & \scriptsize report & \scriptsize draft & {\color{red}$\times$} & \scriptsize [draft, book, decision, work, proposal] \\
\scriptsize the crowd started to & \scriptsize cheer & \scriptsize cry & {\color{red}$\times$} & \scriptsize [cry, riot, fall, fight, play] \\
\scriptsize he bent down to & \scriptsize pick & \scriptsize the & {\color{red}$\times$} & \scriptsize [the, make, a, end, start] \\
\scriptsize she leaned forward to & \scriptsize whisper & \scriptsize a & {\color{red}$\times$} & \scriptsize [a, the, go, an, her] \\
\scriptsize the wind continued to & \scriptsize blow & \scriptsize strike & {\color{red}$\times$} & \scriptsize [strike, rise, influence, grow, move] \\
\scriptsize they gathered around to & \scriptsize celebrate & \scriptsize the & {\color{red}$\times$} & \scriptsize [the, a, make, show, one] \\
\scriptsize he struggled to & \scriptsize breathe & \scriptsize make & {\color{red}$\times$} & \scriptsize [make, maintain, find, achieve, get] \\
\scriptsize the sunset was absolutely & \scriptsize stunning & \scriptsize impossible & {\color{red}$\times$} & \scriptsize [impossible, perfect, destroyed, clear, forgotten] \\
\scriptsize the old house looked rather & \scriptsize creepy & \scriptsize like & {\color{red}$\times$} & \scriptsize [like, similar, different, than, as] \\
\scriptsize his explanation was perfectly & \scriptsize clear & \scriptsize true & {\color{orange}$\sim$} & \scriptsize [true, correct, accurate, clear, accepted] \\
\scriptsize the cake tasted absolutely & \scriptsize delicious & \scriptsize good & {\color{red}$\times$} & \scriptsize [good, sweet, perfect, like, fresh] \\
\scriptsize the mountain trail was extremely & \scriptsize steep & \scriptsize popular & {\color{red}$\times$} & \scriptsize [popular, difficult, well, important, dangerous] \\
\scriptsize the instructions were surprisingly & \scriptsize simple & \scriptsize set & {\color{orange}$\sim$} & \scriptsize [set, good, simple, given, well] \\
\scriptsize the scientist published an important & \scriptsize discovery & \scriptsize paper & {\color{red}$\times$} & \scriptsize [paper, article, work, book, essay] \\
\scriptsize the city was known for its rich & \scriptsize history & \scriptsize and & {\color{red}$\times$} & \scriptsize [and, ,, oil, natural, use] \\
\scriptsize she always valued honesty and & \scriptsize kindness & \scriptsize respect & {\color{red}$\times$} & \scriptsize [respect, integrity, honesty, money, freedom] \\
\scriptsize the treaty established lasting & \scriptsize peace & \scriptsize status & {\color{orange}$\sim$} & \scriptsize [status, peace, effect, for, boundaries] \\
\scriptsize the car moved forward & \scriptsize slowly & \scriptsize in & {\color{red}$\times$} & \scriptsize [in, to, from, on, during] \\
\scriptsize he waited & \scriptsize patiently & \scriptsize for & {\color{red}$\times$} & \scriptsize [for, until, a, at, in] \\
\scriptsize she tied the laces on her & \scriptsize shoes & \scriptsize own & {\color{red}$\times$} & \scriptsize [own, head, wedding, home, second] \\
\scriptsize he wrapped the scarf around his & \scriptsize neck & \scriptsize head & {\color{orange}$\sim$} & \scriptsize [head, neck, shoulder, body, chest] \\
\scriptsize the doctor examined the patient's & \scriptsize heart & \scriptsize body & {\color{orange}$\sim$} & \scriptsize [body, brain, condition, blood, heart] \\
\scriptsize she checked the notifications on her & \scriptsize phone & \scriptsize website & {\color{orange}$\sim$} & \scriptsize [website, phone, own, official, blog] \\
\scriptsize the password must contain at least eight & \scriptsize characters & \scriptsize cards & {\color{red}$\times$} & \scriptsize [cards, possible, different, other, pages] \\
\scriptsize he plugged the charger into the & \scriptsize socket & \scriptsize new & {\color{red}$\times$} & \scriptsize [new, current, first, second, floor] \\
\scriptsize the door began to & \scriptsize open & \scriptsize fall & {\color{orange}$\sim$} & \scriptsize [fall, be, open, appear, collapse] \\
\scriptsize she refused to & \scriptsize quit & \scriptsize do & {\color{red}$\times$} & \scriptsize [do, give, perform, leave, make] \\
\bottomrule
\end{longtable}

\begin{longtable}{@{}p{3.0cm}p{0.9cm}p{1.1cm}cp{3.5cm}@{}}
\caption{AWD-LSTM: Whole-Unit Removal (\texttt{weight\_ih} + \texttt{weight\_hh}, $k=0.2$, main-paper target), all 100 Cloze prompts.} \\
\toprule
Prompt & Exp. & Pred. & & Top-5 \\
\midrule
\endfirsthead
\toprule
Prompt & Exp. & Pred. & & Top-5 \\
\midrule
\endhead
\scriptsize once upon a & \scriptsize time & \scriptsize a & {\color{red}$\times$} & \scriptsize [a, home, family, in, -] \\
\scriptsize the sun was shining bright in the & \scriptsize sky & \scriptsize summer & {\color{red}$\times$} & \scriptsize [summer, year, day, late, time] \\
\scriptsize she picked up the & \scriptsize ball & \scriptsize number & {\color{red}$\times$} & \scriptsize [number, world, most, best, first] \\
\scriptsize he put the food in his & \scriptsize mouth & \scriptsize family & {\color{red}$\times$} & \scriptsize [family, ,, book, death, own] \\
\scriptsize the little girl was wearing a pretty & \scriptsize dress & \scriptsize , & {\color{red}$\times$} & \scriptsize [,, -, and, black, woman] \\
\scriptsize the bird flew up into the & \scriptsize sky & \scriptsize air & {\color{red}$\times$} & \scriptsize [air, sea, ground, ocean, field] \\
\scriptsize she gave her friend a big & \scriptsize hug & \scriptsize , & {\color{red}$\times$} & \scriptsize [,, -, picture, and, box] \\
\scriptsize the cat jumped on the & \scriptsize table & \scriptsize ice & {\color{red}$\times$} & \scriptsize [ice, world, field, air, water] \\
\scriptsize he found a beautiful & \scriptsize flower & \scriptsize , & {\color{red}$\times$} & \scriptsize [,, and, body, new, "] \\
\scriptsize they played together in the & \scriptsize park & \scriptsize same & {\color{red}$\times$} & \scriptsize [same, early, 1970s, summer, 1960s] \\
\scriptsize she sang a beautiful & \scriptsize song & \scriptsize , & {\color{orange}$\sim$} & \scriptsize [,, and, song, ballad, woman] \\
\scriptsize he wanted to go & \scriptsize home & \scriptsize on & {\color{red}$\times$} & \scriptsize [on, to, back, in, ,] \\
\scriptsize the really big & \scriptsize bear & \scriptsize , & {\color{red}$\times$} & \scriptsize [,, -, and, things, new] \\
\scriptsize a very small & \scriptsize mouse & \scriptsize , & {\color{red}$\times$} & \scriptsize [,, and, -, (, "] \\
\scriptsize the extremely bright & \scriptsize light & \scriptsize and & {\color{red}$\times$} & \scriptsize [and, ,, -, =, new] \\
\scriptsize he was very happy and started to & \scriptsize laugh & \scriptsize make & {\color{red}$\times$} & \scriptsize [make, have, ", become, get] \\
\scriptsize the dog saw a cat and started to & \scriptsize bark & \scriptsize make & {\color{red}$\times$} & \scriptsize [make, fall, be, the, look] \\
\scriptsize they went outside to & \scriptsize play & \scriptsize the & {\color{red}$\times$} & \scriptsize [the, a, their, make, together] \\
\scriptsize he ran as fast as he & \scriptsize could & \scriptsize was & {\color{red}$\times$} & \scriptsize [was, is, had, did, and] \\
\scriptsize the baby started to & \scriptsize cry & \scriptsize fall & {\color{red}$\times$} & \scriptsize [fall, become, be, make, have] \\
\scriptsize he carefully & \scriptsize placed & \scriptsize kept & {\color{red}$\times$} & \scriptsize [kept, -, put, considered, selected] \\
\scriptsize they quickly & \scriptsize ran & \scriptsize become & {\color{red}$\times$} & \scriptsize [become, =, spread, became, met] \\
\scriptsize she slowly & \scriptsize walked & \scriptsize re & {\color{red}$\times$} & \scriptsize [re, =, and, began, became] \\
\scriptsize he gently & \scriptsize touched & \scriptsize and & {\color{red}$\times$} & \scriptsize [and, put, =, broke, beat] \\
\scriptsize he was so tired that he fell & \scriptsize asleep & \scriptsize to & {\color{red}$\times$} & \scriptsize [to, in, out, ill, and] \\
\scriptsize she looked at the stars and felt & \scriptsize happy & \scriptsize she & {\color{red}$\times$} & \scriptsize [she, that, the, ", it] \\
\scriptsize she was really & \scriptsize excited & \scriptsize a & {\color{red}$\times$} & \scriptsize [a, well, ,, close, in] \\
\scriptsize the water was very & \scriptsize cold & \scriptsize well & {\color{red}$\times$} & \scriptsize [well, important, popular, common, close] \\
\scriptsize he was feeling very & \scriptsize sad & \scriptsize ill & {\color{red}$\times$} & \scriptsize [ill, upset, close, ,, critical] \\
\scriptsize it was an incredibly & \scriptsize beautiful & \scriptsize close & {\color{red}$\times$} & \scriptsize [close, high, ,, difficult, important] \\
\scriptsize she had a particularly & \scriptsize special & \scriptsize close & {\color{red}$\times$} & \scriptsize [close, well, strong, great, popular] \\
\scriptsize the surprisingly & \scriptsize gentle & \scriptsize first & {\color{red}$\times$} & \scriptsize [first, best, high, ,, great] \\
\scriptsize an unusually & \scriptsize quiet & \scriptsize high & {\color{red}$\times$} & \scriptsize [high, large, -, first, ,] \\
\scriptsize the remarkably & \scriptsize clever & \scriptsize first & {\color{red}$\times$} & \scriptsize [first, -, best, long, new] \\
\scriptsize a wonderfully & \scriptsize warm & \scriptsize , & {\color{red}$\times$} & \scriptsize [,, i, ", personal, natural] \\
\scriptsize she opened the door and walked & \scriptsize inside & \scriptsize the & {\color{red}$\times$} & \scriptsize [the, in, on, at, ,] \\
\scriptsize the story ended & \scriptsize happily & \scriptsize in & {\color{red}$\times$} & \scriptsize [in, with, on, the, at] \\
\scriptsize the children were playing & \scriptsize outside & \scriptsize in & {\color{red}$\times$} & \scriptsize [in, on, a, at, the] \\
\scriptsize the boat sailed across the & \scriptsize ocean & \scriptsize western & {\color{red}$\times$} & \scriptsize [western, island, sea, country, coast] \\
\scriptsize she hung the picture on the & \scriptsize wall & \scriptsize back & {\color{red}$\times$} & \scriptsize [back, body, top, floor, family] \\
\scriptsize the keys were left on the & \scriptsize counter & \scriptsize body & {\color{red}$\times$} & \scriptsize [body, to, way, market, family] \\
\scriptsize he parked the car in the & \scriptsize garage & \scriptsize same & {\color{red}$\times$} & \scriptsize [same, late, house, summer, early] \\
\scriptsize the plane disappeared behind the & \scriptsize clouds & \scriptsize body & {\color{red}$\times$} & \scriptsize [body, plane, two, and, death] \\
\scriptsize they arrived early in the & \scriptsize morning & \scriptsize year & {\color{orange}$\sim$} & \scriptsize [year, 20th, day, 19th, morning] \\
\scriptsize the concert lasted until & \scriptsize midnight & \scriptsize the & {\color{red}$\times$} & \scriptsize [the, its, 2006, 1997, 1995] \\
\scriptsize we will meet again next & \scriptsize week & \scriptsize , & {\color{orange}$\sim$} & \scriptsize [,, year, week, in, next] \\
\scriptsize she poured the milk into the & \scriptsize glass & \scriptsize water & {\color{red}$\times$} & \scriptsize [water, body, air, production, sea] \\
\scriptsize he spread butter on the & \scriptsize bread & \scriptsize area & {\color{red}$\times$} & \scriptsize [area, market, field, way, country] \\
\scriptsize the chef chopped the & \scriptsize vegetables & \scriptsize way & {\color{red}$\times$} & \scriptsize [way, ice, back, old, and] \\
\scriptsize she stirred the soup with a & \scriptsize spoon & \scriptsize " & {\color{red}$\times$} & \scriptsize [", few, black, new, ,] \\
\scriptsize the horse galloped across the & \scriptsize field & \scriptsize country & {\color{red}$\times$} & \scriptsize [country, world, land, western, line] \\
\scriptsize the fish swam in the & \scriptsize pond & \scriptsize water & {\color{red}$\times$} & \scriptsize [water, ice, sea, area, river] \\
\scriptsize the eagle soared above the & \scriptsize mountains & \scriptsize high & {\color{red}$\times$} & \scriptsize [high, ground, level, head, sea] \\
\scriptsize after the long hike he was extremely & \scriptsize tired & \scriptsize ill & {\color{red}$\times$} & \scriptsize [ill, involved, critical, well, close] \\
\scriptsize the news made everyone very & \scriptsize happy & \scriptsize interested & {\color{red}$\times$} & \scriptsize [interested, ,, close, upset, at] \\
\scriptsize the dark alley made her feel & \scriptsize afraid & \scriptsize , & {\color{red}$\times$} & \scriptsize [,, and, ", in, as] \\
\scriptsize the puzzle was incredibly & \scriptsize difficult & \scriptsize , & {\color{red}$\times$} & \scriptsize [,, popular, well, a, part] \\
\scriptsize the audience was completely & \scriptsize silent & \scriptsize upset & {\color{red}$\times$} & \scriptsize [upset, and, disappointed, divided, affected] \\
\scriptsize the teacher asked the students to & \scriptsize listen & \scriptsize make & {\color{red}$\times$} & \scriptsize [make, use, ", the, be] \\
\scriptsize she decided to finally & \scriptsize leave & \scriptsize make & {\color{red}$\times$} & \scriptsize [make, become, ,, marry, take] \\
\scriptsize the alarm clock started to & \scriptsize ring & \scriptsize the & {\color{red}$\times$} & \scriptsize [the, have, fall, change, use] \\
\scriptsize the snow began to & \scriptsize melt & \scriptsize fall & {\color{red}$\times$} & \scriptsize [fall, have, become, spread, rise] \\
\scriptsize he promised he would never & \scriptsize forget & \scriptsize be & {\color{red}$\times$} & \scriptsize [be, have, make, do, go] \\
\scriptsize the children love to & \scriptsize swim & \scriptsize a & {\color{red}$\times$} & \scriptsize [a, the, an, be, love] \\
\scriptsize she wanted to learn how to & \scriptsize dance & \scriptsize use & {\color{red}$\times$} & \scriptsize [use, play, ", make, live] \\
\scriptsize the mechanic tried to & \scriptsize repair & \scriptsize make & {\color{red}$\times$} & \scriptsize [make, create, use, beat, =] \\
\scriptsize dark clouds gathered before the & \scriptsize storm & \scriptsize event & {\color{red}$\times$} & \scriptsize [event, day, end, first, fall] \\
\scriptsize the flowers bloomed in the & \scriptsize spring & \scriptsize family & {\color{red}$\times$} & \scriptsize [family, house, city, same, country] \\
\scriptsize the leaves fall from the trees in & \scriptsize autumn & \scriptsize the & {\color{red}$\times$} & \scriptsize [the, a, in, which, and] \\
\scriptsize the river flowed into the & \scriptsize sea & \scriptsize river & {\color{orange}$\sim$} & \scriptsize [river, city, area, sea, water] \\
\scriptsize the student wrote an & \scriptsize essay & \scriptsize " & {\color{orange}$\sim$} & \scriptsize [", article, essay, autobiography, original] \\
\scriptsize she typed the email on her & \scriptsize keyboard & \scriptsize website & {\color{red}$\times$} & \scriptsize [website, family, ,, blog, personal] \\
\scriptsize the professor gave a & \scriptsize lecture & \scriptsize " & {\color{red}$\times$} & \scriptsize [", more, similar, history, positive] \\
\scriptsize he submitted his final & \scriptsize report & \scriptsize two & {\color{red}$\times$} & \scriptsize [two, book, to, draft, autobiography] \\
\scriptsize the crowd started to & \scriptsize cheer & \scriptsize rise & {\color{red}$\times$} & \scriptsize [rise, fall, build, make, move] \\
\scriptsize he bent down to & \scriptsize pick & \scriptsize the & {\color{red}$\times$} & \scriptsize [the, a, death, his, down] \\
\scriptsize she leaned forward to & \scriptsize whisper & \scriptsize a & {\color{red}$\times$} & \scriptsize [a, the, her, his, an] \\
\scriptsize the wind continued to & \scriptsize blow & \scriptsize the & {\color{red}$\times$} & \scriptsize [the, rise, fall, have, grow] \\
\scriptsize they gathered around to & \scriptsize celebrate & \scriptsize the & {\color{red}$\times$} & \scriptsize [the, a, an, in, on] \\
\scriptsize he struggled to & \scriptsize breathe & \scriptsize make & {\color{red}$\times$} & \scriptsize [make, achieve, get, =, do] \\
\scriptsize the sunset was absolutely & \scriptsize stunning & \scriptsize , & {\color{red}$\times$} & \scriptsize [,, the, important, a, as] \\
\scriptsize the old house looked rather & \scriptsize creepy & \scriptsize like & {\color{red}$\times$} & \scriptsize [like, than, =, ", on] \\
\scriptsize his explanation was perfectly & \scriptsize clear & \scriptsize clear & {\color{green!50!black}$\checkmark$} & \scriptsize [clear, correct, detailed, true, consistent] \\
\scriptsize the cake tasted absolutely & \scriptsize delicious & \scriptsize , & {\color{red}$\times$} & \scriptsize [,, , like, great, grown] \\
\scriptsize the mountain trail was extremely & \scriptsize steep & \scriptsize popular & {\color{red}$\times$} & \scriptsize [popular, well, successful, close, important] \\
\scriptsize the instructions were surprisingly & \scriptsize simple & \scriptsize , & {\color{red}$\times$} & \scriptsize [,, set, used, for, put] \\
\scriptsize the scientist published an important & \scriptsize discovery & \scriptsize paper & {\color{red}$\times$} & \scriptsize [paper, book, information, source, biography] \\
\scriptsize the city was known for its rich & \scriptsize history & \scriptsize and & {\color{orange}$\sim$} & \scriptsize [and, ,, history, use, natural] \\
\scriptsize she always valued honesty and & \scriptsize kindness & \scriptsize the & {\color{red}$\times$} & \scriptsize [the, ", ,, respect, self] \\
\scriptsize the treaty established lasting & \scriptsize peace & \scriptsize years & {\color{red}$\times$} & \scriptsize [years, over, the, decades, until] \\
\scriptsize the car moved forward & \scriptsize slowly & \scriptsize in & {\color{red}$\times$} & \scriptsize [in, and, ,, on, from] \\
\scriptsize he waited & \scriptsize patiently & \scriptsize to & {\color{red}$\times$} & \scriptsize [to, for, in, on, until] \\
\scriptsize she tied the laces on her & \scriptsize shoes & \scriptsize , & {\color{red}$\times$} & \scriptsize [,, family, home, body, and] \\
\scriptsize he wrapped the scarf around his & \scriptsize neck & \scriptsize head & {\color{orange}$\sim$} & \scriptsize [head, body, death, ,, neck] \\
\scriptsize the doctor examined the patient's & \scriptsize heart & \scriptsize body & {\color{red}$\times$} & \scriptsize [body, life, treatment, death, health] \\
\scriptsize she checked the notifications on her & \scriptsize phone & \scriptsize website & {\color{red}$\times$} & \scriptsize [website, home, family, ,, personal] \\
\scriptsize the password must contain at least eight & \scriptsize characters & \scriptsize different & {\color{red}$\times$} & \scriptsize [different, other, ,, ", people] \\
\scriptsize he plugged the charger into the & \scriptsize socket & \scriptsize new & {\color{red}$\times$} & \scriptsize [new, same, ", modern, body] \\
\scriptsize the door began to & \scriptsize open & \scriptsize have & {\color{orange}$\sim$} & \scriptsize [have, become, fall, fire, open] \\
\scriptsize she refused to & \scriptsize quit & \scriptsize do & {\color{red}$\times$} & \scriptsize [do, be, have, hold, make] \\
\bottomrule
\end{longtable}

\begin{longtable}{@{}p{3.0cm}p{0.9cm}p{1.1cm}cp{3.5cm}@{}}
\caption{AWD-LSTM: Recurrent-Connection Pruning (\texttt{weight\_hh}, $k=0.2$, supplementary control), all 100 Cloze prompts.} \\
\toprule
Prompt & Exp. & Pred. & & Top-5 \\
\midrule
\endfirsthead
\toprule
Prompt & Exp. & Pred. & & Top-5 \\
\midrule
\endhead
\scriptsize once upon a & \scriptsize time & \scriptsize time & {\color{green!50!black}$\checkmark$} & \scriptsize [time, ship, a, single, surface] \\
\scriptsize the sun was shining bright in the & \scriptsize sky & \scriptsize sky & {\color{green!50!black}$\checkmark$} & \scriptsize [sky, sun, light, skies, dark] \\
\scriptsize she picked up the & \scriptsize ball & \scriptsize first & {\color{red}$\times$} & \scriptsize [first, ", last, next, most] \\
\scriptsize he put the food in his & \scriptsize mouth & \scriptsize house & {\color{red}$\times$} & \scriptsize [house, car, mind, book, own] \\
\scriptsize the little girl was wearing a pretty & \scriptsize dress & \scriptsize little & {\color{red}$\times$} & \scriptsize [little, black, girl, long, white] \\
\scriptsize the bird flew up into the & \scriptsize sky & \scriptsize air & {\color{orange}$\sim$} & \scriptsize [air, sky, desert, ground, sea] \\
\scriptsize she gave her friend a big & \scriptsize hug & \scriptsize - & {\color{red}$\times$} & \scriptsize [-, deal, show, name, piece] \\
\scriptsize the cat jumped on the & \scriptsize table & \scriptsize ground & {\color{red}$\times$} & \scriptsize [ground, screen, ice, moon, roof] \\
\scriptsize he found a beautiful & \scriptsize flower & \scriptsize woman & {\color{red}$\times$} & \scriptsize [woman, and, body, ,, female] \\
\scriptsize they played together in the & \scriptsize park & \scriptsize same & {\color{red}$\times$} & \scriptsize [same, first, early, late, 1960s] \\
\scriptsize she sang a beautiful & \scriptsize song & \scriptsize song & {\color{green!50!black}$\checkmark$} & \scriptsize [song, version, woman, ballad, voice] \\
\scriptsize he wanted to go & \scriptsize home & \scriptsize to & {\color{red}$\times$} & \scriptsize [to, back, on, into, in] \\
\scriptsize the really big & \scriptsize bear & \scriptsize - & {\color{red}$\times$} & \scriptsize [-, ,, thing, things, big] \\
\scriptsize a very small & \scriptsize mouse & \scriptsize small & {\color{red}$\times$} & \scriptsize [small, ,, number, large, short] \\
\scriptsize the extremely bright & \scriptsize light & \scriptsize red & {\color{orange}$\sim$} & \scriptsize [red, light, and, black, ,] \\
\scriptsize he was very happy and started to & \scriptsize laugh & \scriptsize make & {\color{red}$\times$} & \scriptsize [make, feel, write, think, do] \\
\scriptsize the dog saw a cat and started to & \scriptsize bark & \scriptsize show & {\color{red}$\times$} & \scriptsize [show, play, look, be, appear] \\
\scriptsize they went outside to & \scriptsize play & \scriptsize the & {\color{red}$\times$} & \scriptsize [the, work, a, make, their] \\
\scriptsize he ran as fast as he & \scriptsize could & \scriptsize could & {\color{green!50!black}$\checkmark$} & \scriptsize [could, was, had, did, ran] \\
\scriptsize the baby started to & \scriptsize cry & \scriptsize appear & {\color{red}$\times$} & \scriptsize [appear, grow, be, play, become] \\
\scriptsize he carefully & \scriptsize placed & \scriptsize studied & {\color{red}$\times$} & \scriptsize [studied, kept, tried, planned, researched] \\
\scriptsize they quickly & \scriptsize ran & \scriptsize became & {\color{red}$\times$} & \scriptsize [became, become, lost, went, began] \\
\scriptsize she slowly & \scriptsize walked & \scriptsize became & {\color{red}$\times$} & \scriptsize [became, lost, fell, died, began] \\
\scriptsize he gently & \scriptsize touched & \scriptsize sings & {\color{red}$\times$} & \scriptsize [sings, uses, turns, throws, pitched] \\
\scriptsize he was so tired that he fell & \scriptsize asleep & \scriptsize in & {\color{red}$\times$} & \scriptsize [in, out, to, into, ill] \\
\scriptsize she looked at the stars and felt & \scriptsize happy & \scriptsize that & {\color{red}$\times$} & \scriptsize [that, the, they, she, like] \\
\scriptsize she was really & \scriptsize excited & \scriptsize a & {\color{red}$\times$} & \scriptsize [a, the, an, born, made] \\
\scriptsize the water was very & \scriptsize cold & \scriptsize similar & {\color{red}$\times$} & \scriptsize [similar, good, difficult, much, high] \\
\scriptsize he was feeling very & \scriptsize sad & \scriptsize much & {\color{red}$\times$} & \scriptsize [much, good, ill, similar, guilty] \\
\scriptsize it was an incredibly & \scriptsize beautiful & \scriptsize complex & {\color{red}$\times$} & \scriptsize [complex, difficult, small, large, powerful] \\
\scriptsize she had a particularly & \scriptsize special & \scriptsize strong & {\color{red}$\times$} & \scriptsize [strong, controversial, difficult, great, good] \\
\scriptsize the surprisingly & \scriptsize gentle & \scriptsize successful & {\color{red}$\times$} & \scriptsize [successful, popular, large, strong, powerful] \\
\scriptsize an unusually & \scriptsize quiet & \scriptsize large & {\color{red}$\times$} & \scriptsize [large, high, low, long, small] \\
\scriptsize the remarkably & \scriptsize clever & \scriptsize large & {\color{red}$\times$} & \scriptsize [large, successful, powerful, popular, new] \\
\scriptsize a wonderfully & \scriptsize warm & \scriptsize popular & {\color{red}$\times$} & \scriptsize [popular, successful, animated, martial, large] \\
\scriptsize she opened the door and walked & \scriptsize inside & \scriptsize the & {\color{red}$\times$} & \scriptsize [the, out, through, in, away] \\
\scriptsize the story ended & \scriptsize happily & \scriptsize with & {\color{red}$\times$} & \scriptsize [with, in, up, when, at] \\
\scriptsize the children were playing & \scriptsize outside & \scriptsize in & {\color{red}$\times$} & \scriptsize [in, the, on, a, at] \\
\scriptsize the boat sailed across the & \scriptsize ocean & \scriptsize sea & {\color{orange}$\sim$} & \scriptsize [sea, ocean, waters, river, country] \\
\scriptsize she hung the picture on the & \scriptsize wall & \scriptsize screen & {\color{orange}$\sim$} & \scriptsize [screen, front, wall, set, floor] \\
\scriptsize the keys were left on the & \scriptsize counter & \scriptsize floor & {\color{red}$\times$} & \scriptsize [floor, ground, wall, back, board] \\
\scriptsize he parked the car in the & \scriptsize garage & \scriptsize car & {\color{orange}$\sim$} & \scriptsize [car, same, garage, house, park] \\
\scriptsize the plane disappeared behind the & \scriptsize clouds & \scriptsize scenes & {\color{red}$\times$} & \scriptsize [scenes, plane, scene, wreckage, body] \\
\scriptsize they arrived early in the & \scriptsize morning & \scriptsize first & {\color{orange}$\sim$} & \scriptsize [first, morning, early, 20th, second] \\
\scriptsize the concert lasted until & \scriptsize midnight & \scriptsize the & {\color{red}$\times$} & \scriptsize [the, 30, 9, 14, 21] \\
\scriptsize we will meet again next & \scriptsize week & \scriptsize year & {\color{red}$\times$} & \scriptsize [year, ,, next, in, to] \\
\scriptsize she poured the milk into the & \scriptsize glass & \scriptsize water & {\color{red}$\times$} & \scriptsize [water, air, milk, blood, mud] \\
\scriptsize he spread butter on the & \scriptsize bread & \scriptsize ice & {\color{red}$\times$} & \scriptsize [ice, sea, water, ground, air] \\
\scriptsize the chef chopped the & \scriptsize vegetables & \scriptsize ice & {\color{red}$\times$} & \scriptsize [ice, table, pie, meat, dish] \\
\scriptsize she stirred the soup with a & \scriptsize spoon & \scriptsize black & {\color{red}$\times$} & \scriptsize [black, series, ", large, red] \\
\scriptsize the horse galloped across the & \scriptsize field & \scriptsize street & {\color{red}$\times$} & \scriptsize [street, mountain, road, streets, river] \\
\scriptsize the fish swam in the & \scriptsize pond & \scriptsize water & {\color{red}$\times$} & \scriptsize [water, ocean, waters, sea, river] \\
\scriptsize the eagle soared above the & \scriptsize mountains & \scriptsize ground & {\color{red}$\times$} & \scriptsize [ground, water, earth, sea, sky] \\
\scriptsize after the long hike he was extremely & \scriptsize tired & \scriptsize popular & {\color{red}$\times$} & \scriptsize [popular, unpopular, successful, busy, difficult] \\
\scriptsize the news made everyone very & \scriptsize happy & \scriptsize happy & {\color{green!50!black}$\checkmark$} & \scriptsize [happy, angry, uncomfortable, excited, interested] \\
\scriptsize the dark alley made her feel & \scriptsize afraid & \scriptsize like & {\color{red}$\times$} & \scriptsize [like, more, as, very, to] \\
\scriptsize the puzzle was incredibly & \scriptsize difficult & \scriptsize complicated & {\color{orange}$\sim$} & \scriptsize [complicated, popular, difficult, controversial, complex] \\
\scriptsize the audience was completely & \scriptsize silent & \scriptsize divided & {\color{red}$\times$} & \scriptsize [divided, ignored, unknown, shocked, surprised] \\
\scriptsize the teacher asked the students to & \scriptsize listen & \scriptsize read & {\color{red}$\times$} & \scriptsize [read, teach, write, speak, do] \\
\scriptsize she decided to finally & \scriptsize leave & \scriptsize make & {\color{red}$\times$} & \scriptsize [make, take, become, go, return] \\
\scriptsize the alarm clock started to & \scriptsize ring & \scriptsize appear & {\color{red}$\times$} & \scriptsize [appear, be, become, spread, run] \\
\scriptsize the snow began to & \scriptsize melt & \scriptsize appear & {\color{red}$\times$} & \scriptsize [appear, spread, fall, become, be] \\
\scriptsize he promised he would never & \scriptsize forget & \scriptsize do & {\color{red}$\times$} & \scriptsize [do, have, be, make, return] \\
\scriptsize the children love to & \scriptsize swim & \scriptsize be & {\color{red}$\times$} & \scriptsize [be, have, play, live, love] \\
\scriptsize she wanted to learn how to & \scriptsize dance & \scriptsize do & {\color{red}$\times$} & \scriptsize [do, make, play, write, work] \\
\scriptsize the mechanic tried to & \scriptsize repair & \scriptsize use & {\color{red}$\times$} & \scriptsize [use, make, create, do, kill] \\
\scriptsize dark clouds gathered before the & \scriptsize storm & \scriptsize night & {\color{orange}$\sim$} & \scriptsize [night, dark, dawn, clouds, storm] \\
\scriptsize the flowers bloomed in the & \scriptsize spring & \scriptsize garden & {\color{red}$\times$} & \scriptsize [garden, house, city, park, gardens] \\
\scriptsize the leaves fall from the trees in & \scriptsize autumn & \scriptsize the & {\color{red}$\times$} & \scriptsize [the, a, front, their, an] \\
\scriptsize the river flowed into the & \scriptsize sea & \scriptsize river & {\color{orange}$\sim$} & \scriptsize [river, sea, water, city, valley] \\
\scriptsize the student wrote an & \scriptsize essay & \scriptsize article & {\color{orange}$\sim$} & \scriptsize [article, essay, autobiography, editorial, account] \\
\scriptsize she typed the email on her & \scriptsize keyboard & \scriptsize phone & {\color{red}$\times$} & \scriptsize [phone, blog, own, letter, website] \\
\scriptsize the professor gave a & \scriptsize lecture & \scriptsize speech & {\color{orange}$\sim$} & \scriptsize [speech, lecture, ", positive, similar] \\
\scriptsize he submitted his final & \scriptsize report & \scriptsize work & {\color{red}$\times$} & \scriptsize [work, book, decision, attempt, draft] \\
\scriptsize the crowd started to & \scriptsize cheer & \scriptsize play & {\color{red}$\times$} & \scriptsize [play, sing, show, feel, cry] \\
\scriptsize he bent down to & \scriptsize pick & \scriptsize the & {\color{red}$\times$} & \scriptsize [the, a, his, make, one] \\
\scriptsize she leaned forward to & \scriptsize whisper & \scriptsize a & {\color{red}$\times$} & \scriptsize [a, the, his, make, go] \\
\scriptsize the wind continued to & \scriptsize blow & \scriptsize be & {\color{red}$\times$} & \scriptsize [be, rise, influence, increase, strike] \\
\scriptsize they gathered around to & \scriptsize celebrate & \scriptsize the & {\color{orange}$\sim$} & \scriptsize [the, a, celebrate, work, be] \\
\scriptsize he struggled to & \scriptsize breathe & \scriptsize make & {\color{red}$\times$} & \scriptsize [make, find, maintain, do, achieve] \\
\scriptsize the sunset was absolutely & \scriptsize stunning & \scriptsize no & {\color{red}$\times$} & \scriptsize [no, impossible, never, a, not] \\
\scriptsize the old house looked rather & \scriptsize creepy & \scriptsize like & {\color{red}$\times$} & \scriptsize [like, on, similar, than, at] \\
\scriptsize his explanation was perfectly & \scriptsize clear & \scriptsize well & {\color{orange}$\sim$} & \scriptsize [well, true, correct, accepted, clear] \\
\scriptsize the cake tasted absolutely & \scriptsize delicious & \scriptsize so & {\color{red}$\times$} & \scriptsize [so, great, nothing, no, right] \\
\scriptsize the mountain trail was extremely & \scriptsize steep & \scriptsize popular & {\color{red}$\times$} & \scriptsize [popular, difficult, well, controversial, dangerous] \\
\scriptsize the instructions were surprisingly & \scriptsize simple & \scriptsize set & {\color{red}$\times$} & \scriptsize [set, made, limited, not, good] \\
\scriptsize the scientist published an important & \scriptsize discovery & \scriptsize work & {\color{red}$\times$} & \scriptsize [work, book, article, research, essay] \\
\scriptsize the city was known for its rich & \scriptsize history & \scriptsize and & {\color{orange}$\sim$} & \scriptsize [and, ,, history, rich, reputation] \\
\scriptsize she always valued honesty and & \scriptsize kindness & \scriptsize respect & {\color{red}$\times$} & \scriptsize [respect, integrity, honesty, self, personal] \\
\scriptsize the treaty established lasting & \scriptsize peace & \scriptsize for & {\color{red}$\times$} & \scriptsize [for, the, long, a, two] \\
\scriptsize the car moved forward & \scriptsize slowly & \scriptsize to & {\color{red}$\times$} & \scriptsize [to, from, in, ,, on] \\
\scriptsize he waited & \scriptsize patiently & \scriptsize for & {\color{red}$\times$} & \scriptsize [for, in, until, at, a] \\
\scriptsize she tied the laces on her & \scriptsize shoes & \scriptsize own & {\color{red}$\times$} & \scriptsize [own, head, body, husband, death] \\
\scriptsize he wrapped the scarf around his & \scriptsize neck & \scriptsize head & {\color{orange}$\sim$} & \scriptsize [head, body, neck, eyes, hair] \\
\scriptsize the doctor examined the patient's & \scriptsize heart & \scriptsize body & {\color{red}$\times$} & \scriptsize [body, brain, condition, blood, right] \\
\scriptsize she checked the notifications on her & \scriptsize phone & \scriptsize phone & {\color{green!50!black}$\checkmark$} & \scriptsize [phone, own, website, radio, blog] \\
\scriptsize the password must contain at least eight & \scriptsize characters & \scriptsize or & {\color{red}$\times$} & \scriptsize [or, hundred, times, different, other] \\
\scriptsize he plugged the charger into the & \scriptsize socket & \scriptsize " & {\color{red}$\times$} & \scriptsize [", current, modern, new, first] \\
\scriptsize the door began to & \scriptsize open & \scriptsize be & {\color{orange}$\sim$} & \scriptsize [be, become, open, appear, collapse] \\
\scriptsize she refused to & \scriptsize quit & \scriptsize do & {\color{red}$\times$} & \scriptsize [do, leave, give, make, take] \\
\bottomrule
\end{longtable}

\begin{longtable}{@{}p{3.0cm}p{0.9cm}p{1.1cm}cp{3.5cm}@{}}
\caption{AWD-LSTM: Input-Connection Pruning (\texttt{weight\_ih}, $k=0.2$, supplementary control), all 100 Cloze prompts.} \\
\toprule
Prompt & Exp. & Pred. & & Top-5 \\
\midrule
\endfirsthead
\toprule
Prompt & Exp. & Pred. & & Top-5 \\
\midrule
\endhead
\scriptsize once upon a & \scriptsize time & \scriptsize life & {\color{red}$\times$} & \scriptsize [life, live, film, day, history] \\
\scriptsize the sun was shining bright in the & \scriptsize sky & \scriptsize sky & {\color{green!50!black}$\checkmark$} & \scriptsize [sky, black, day, life, days] \\
\scriptsize she picked up the & \scriptsize ball & \scriptsize idea & {\color{red}$\times$} & \scriptsize [idea, film, girls, women, action] \\
\scriptsize he put the food in his & \scriptsize mouth & \scriptsize family & {\color{red}$\times$} & \scriptsize [family, life, way, case, place] \\
\scriptsize the little girl was wearing a pretty & \scriptsize dress & \scriptsize girl & {\color{red}$\times$} & \scriptsize [girl, black, -, woman, ,] \\
\scriptsize the bird flew up into the & \scriptsize sky & \scriptsize air & {\color{red}$\times$} & \scriptsize [air, space, flight, city, black] \\
\scriptsize she gave her friend a big & \scriptsize hug & \scriptsize - & {\color{red}$\times$} & \scriptsize [-, life, to, name, affair] \\
\scriptsize the cat jumped on the & \scriptsize table & \scriptsize moon & {\color{red}$\times$} & \scriptsize [moon, life, ice, glass, black] \\
\scriptsize he found a beautiful & \scriptsize flower & \scriptsize life & {\color{red}$\times$} & \scriptsize [life, -, woman, girl, black] \\
\scriptsize they played together in the & \scriptsize park & \scriptsize summer & {\color{red}$\times$} & \scriptsize [summer, early, same, late, 2012] \\
\scriptsize she sang a beautiful & \scriptsize song & \scriptsize version & {\color{orange}$\sim$} & \scriptsize [version, -, ,, of, song] \\
\scriptsize he wanted to go & \scriptsize home & \scriptsize to & {\color{red}$\times$} & \scriptsize [to, in, on, and, back] \\
\scriptsize the really big & \scriptsize bear & \scriptsize - & {\color{red}$\times$} & \scriptsize [-, city, and, of, ,] \\
\scriptsize a very small & \scriptsize mouse & \scriptsize - & {\color{red}$\times$} & \scriptsize [-, action, nature, human, city] \\
\scriptsize the extremely bright & \scriptsize light & \scriptsize black & {\color{red}$\times$} & \scriptsize [black, of, red, -, blue] \\
\scriptsize he was very happy and started to & \scriptsize laugh & \scriptsize work & {\color{red}$\times$} & \scriptsize [work, life, be, his, love] \\
\scriptsize the dog saw a cat and started to & \scriptsize bark & \scriptsize die & {\color{red}$\times$} & \scriptsize [die, look, live, work, eat] \\
\scriptsize they went outside to & \scriptsize play & \scriptsize live & {\color{orange}$\sim$} & \scriptsize [live, work, play, make, be] \\
\scriptsize he ran as fast as he & \scriptsize could & \scriptsize was & {\color{orange}$\sim$} & \scriptsize [was, could, had, is, can] \\
\scriptsize the baby started to & \scriptsize cry & \scriptsize live & {\color{red}$\times$} & \scriptsize [live, die, be, have, fall] \\
\scriptsize he carefully & \scriptsize placed & \scriptsize - & {\color{red}$\times$} & \scriptsize [-, studied, photography, found, "] \\
\scriptsize they quickly & \scriptsize ran & \scriptsize become & {\color{red}$\times$} & \scriptsize [become, fall, share, -, end] \\
\scriptsize she slowly & \scriptsize walked & \scriptsize found & {\color{red}$\times$} & \scriptsize [found, -, falls, development, lost] \\
\scriptsize he gently & \scriptsize touched & \scriptsize - & {\color{red}$\times$} & \scriptsize [-, charge, heart, fall, and] \\
\scriptsize he was so tired that he fell & \scriptsize asleep & \scriptsize in & {\color{red}$\times$} & \scriptsize [in, into, out, from, to] \\
\scriptsize she looked at the stars and felt & \scriptsize happy & \scriptsize the & {\color{red}$\times$} & \scriptsize [the, her, ", a, that] \\
\scriptsize she was really & \scriptsize excited & \scriptsize a & {\color{red}$\times$} & \scriptsize [a, ", -, an, ,] \\
\scriptsize the water was very & \scriptsize cold & \scriptsize a & {\color{red}$\times$} & \scriptsize [a, popular, important, expensive, similar] \\
\scriptsize he was feeling very & \scriptsize sad & \scriptsize good & {\color{red}$\times$} & \scriptsize [good, great, comfortable, ill, much] \\
\scriptsize it was an incredibly & \scriptsize beautiful & \scriptsize great & {\color{red}$\times$} & \scriptsize [great, complex, good, fun, expensive] \\
\scriptsize she had a particularly & \scriptsize special & \scriptsize life & {\color{red}$\times$} & \scriptsize [life, summer, great, -, successful] \\
\scriptsize the surprisingly & \scriptsize gentle & \scriptsize - & {\color{red}$\times$} & \scriptsize [-, final, success, series, ,] \\
\scriptsize an unusually & \scriptsize quiet & \scriptsize - & {\color{red}$\times$} & \scriptsize [-, performance, early, flight, appearance] \\
\scriptsize the remarkably & \scriptsize clever & \scriptsize - & {\color{red}$\times$} & \scriptsize [-, success, complex, experience, impact] \\
\scriptsize a wonderfully & \scriptsize warm & \scriptsize adaptation & {\color{red}$\times$} & \scriptsize [adaptation, action, -, ,, science] \\
\scriptsize she opened the door and walked & \scriptsize inside & \scriptsize the & {\color{red}$\times$} & \scriptsize [the, her, in, ,, a] \\
\scriptsize the story ended & \scriptsize happily & \scriptsize in & {\color{red}$\times$} & \scriptsize [in, the, with, on, as] \\
\scriptsize the children were playing & \scriptsize outside & \scriptsize in & {\color{red}$\times$} & \scriptsize [in, the, a, for, by] \\
\scriptsize the boat sailed across the & \scriptsize ocean & \scriptsize island & {\color{orange}$\sim$} & \scriptsize [island, river, bay, ocean, city] \\
\scriptsize she hung the picture on the & \scriptsize wall & \scriptsize wall & {\color{green!50!black}$\checkmark$} & \scriptsize [wall, floor, ground, moon, red] \\
\scriptsize the keys were left on the & \scriptsize counter & \scriptsize ground & {\color{red}$\times$} & \scriptsize [ground, to, road, line, floor] \\
\scriptsize he parked the car in the & \scriptsize garage & \scriptsize park & {\color{red}$\times$} & \scriptsize [park, city, early, summer, office] \\
\scriptsize the plane disappeared behind the & \scriptsize clouds & \scriptsize plane & {\color{red}$\times$} & \scriptsize [plane, black, flight, scenes, bridge] \\
\scriptsize they arrived early in the & \scriptsize morning & \scriptsize summer & {\color{red}$\times$} & \scriptsize [summer, history, life, day, 20th] \\
\scriptsize the concert lasted until & \scriptsize midnight & \scriptsize the & {\color{red}$\times$} & \scriptsize [the, 2005, 2000, 2011, 2006] \\
\scriptsize we will meet again next & \scriptsize week & \scriptsize year & {\color{orange}$\sim$} & \scriptsize [year, season, ,, week, time] \\
\scriptsize she poured the milk into the & \scriptsize glass & \scriptsize water & {\color{orange}$\sim$} & \scriptsize [water, life, ice, glass, fire] \\
\scriptsize he spread butter on the & \scriptsize bread & \scriptsize street & {\color{red}$\times$} & \scriptsize [street, streets, road, ice, ground] \\
\scriptsize the chef chopped the & \scriptsize vegetables & \scriptsize glass & {\color{red}$\times$} & \scriptsize [glass, ice, leg, up, chain] \\
\scriptsize she stirred the soup with a & \scriptsize spoon & \scriptsize black & {\color{red}$\times$} & \scriptsize [black, ", series, life, wife] \\
\scriptsize the horse galloped across the & \scriptsize field & \scriptsize street & {\color{red}$\times$} & \scriptsize [street, city, road, streets, town] \\
\scriptsize the fish swam in the & \scriptsize pond & \scriptsize water & {\color{red}$\times$} & \scriptsize [water, river, summer, ice, ocean] \\
\scriptsize the eagle soared above the & \scriptsize mountains & \scriptsize ground & {\color{red}$\times$} & \scriptsize [ground, earth, moon, flight, rock] \\
\scriptsize after the long hike he was extremely & \scriptsize tired & \scriptsize popular & {\color{red}$\times$} & \scriptsize [popular, successful, a, ill, angry] \\
\scriptsize the news made everyone very & \scriptsize happy & \scriptsize popular & {\color{red}$\times$} & \scriptsize [popular, famous, public, -, much] \\
\scriptsize the dark alley made her feel & \scriptsize afraid & \scriptsize of & {\color{red}$\times$} & \scriptsize [of, a, ", -, the] \\
\scriptsize the puzzle was incredibly & \scriptsize difficult & \scriptsize complex & {\color{red}$\times$} & \scriptsize [complex, a, popular, fun, "] \\
\scriptsize the audience was completely & \scriptsize silent & \scriptsize black & {\color{red}$\times$} & \scriptsize [black, a, upset, cast, "] \\
\scriptsize the teacher asked the students to & \scriptsize listen & \scriptsize be & {\color{red}$\times$} & \scriptsize [be, become, work, take, go] \\
\scriptsize she decided to finally & \scriptsize leave & \scriptsize marry & {\color{red}$\times$} & \scriptsize [marry, become, do, take, be] \\
\scriptsize the alarm clock started to & \scriptsize ring & \scriptsize be & {\color{red}$\times$} & \scriptsize [be, take, fall, work, live] \\
\scriptsize the snow began to & \scriptsize melt & \scriptsize live & {\color{red}$\times$} & \scriptsize [live, fall, be, exist, develop] \\
\scriptsize he promised he would never & \scriptsize forget & \scriptsize be & {\color{red}$\times$} & \scriptsize [be, have, get, do, come] \\
\scriptsize the children love to & \scriptsize swim & \scriptsize be & {\color{red}$\times$} & \scriptsize [be, play, death, make, art] \\
\scriptsize she wanted to learn how to & \scriptsize dance & \scriptsize work & {\color{red}$\times$} & \scriptsize [work, write, make, play, do] \\
\scriptsize the mechanic tried to & \scriptsize repair & \scriptsize make & {\color{red}$\times$} & \scriptsize [make, create, build, move, get] \\
\scriptsize dark clouds gathered before the & \scriptsize storm & \scriptsize human & {\color{red}$\times$} & \scriptsize [human, sky, moon, planet, x] \\
\scriptsize the flowers bloomed in the & \scriptsize spring & \scriptsize red & {\color{red}$\times$} & \scriptsize [red, black, garden, green, summer] \\
\scriptsize the leaves fall from the trees in & \scriptsize autumn & \scriptsize the & {\color{red}$\times$} & \scriptsize [the, a, an, their, which] \\
\scriptsize the river flowed into the & \scriptsize sea & \scriptsize river & {\color{red}$\times$} & \scriptsize [river, city, way, south, western] \\
\scriptsize the student wrote an & \scriptsize essay & \scriptsize in & {\color{red}$\times$} & \scriptsize [in, episode, article, on, at] \\
\scriptsize she typed the email on her & \scriptsize keyboard & \scriptsize book & {\color{red}$\times$} & \scriptsize [book, family, blog, paper, mother] \\
\scriptsize the professor gave a & \scriptsize lecture & \scriptsize history & {\color{red}$\times$} & \scriptsize [history, speech, life, series, live] \\
\scriptsize he submitted his final & \scriptsize report & \scriptsize case & {\color{red}$\times$} & \scriptsize [case, book, campaign, series, bid] \\
\scriptsize the crowd started to & \scriptsize cheer & \scriptsize live & {\color{red}$\times$} & \scriptsize [live, play, move, lose, fall] \\
\scriptsize he bent down to & \scriptsize pick & \scriptsize the & {\color{red}$\times$} & \scriptsize [the, a, win, his, run] \\
\scriptsize she leaned forward to & \scriptsize whisper & \scriptsize her & {\color{red}$\times$} & \scriptsize [her, the, a, his, swimming] \\
\scriptsize the wind continued to & \scriptsize blow & \scriptsize the & {\color{red}$\times$} & \scriptsize [the, be, produce, grow, move] \\
\scriptsize they gathered around to & \scriptsize celebrate & \scriptsize live & {\color{red}$\times$} & \scriptsize [live, the, start, a, view] \\
\scriptsize he struggled to & \scriptsize breathe & \scriptsize find & {\color{red}$\times$} & \scriptsize [find, get, make, do, live] \\
\scriptsize the sunset was absolutely & \scriptsize stunning & \scriptsize - & {\color{red}$\times$} & \scriptsize [-, a, the, black, fun] \\
\scriptsize the old house looked rather & \scriptsize creepy & \scriptsize a & {\color{red}$\times$} & \scriptsize [a, like, ", -, as] \\
\scriptsize his explanation was perfectly & \scriptsize clear & \scriptsize - & {\color{red}$\times$} & \scriptsize [-, good, accurate, true, perfect] \\
\scriptsize the cake tasted absolutely & \scriptsize delicious & \scriptsize fun & {\color{red}$\times$} & \scriptsize [fun, good, fair, black, great] \\
\scriptsize the mountain trail was extremely & \scriptsize steep & \scriptsize popular & {\color{red}$\times$} & \scriptsize [popular, a, active, famous, important] \\
\scriptsize the instructions were surprisingly & \scriptsize simple & \scriptsize set & {\color{red}$\times$} & \scriptsize [set, to, made, written, put] \\
\scriptsize the scientist published an important & \scriptsize discovery & \scriptsize work & {\color{red}$\times$} & \scriptsize [work, history, paper, book, project] \\
\scriptsize the city was known for its rich & \scriptsize history & \scriptsize history & {\color{green!50!black}$\checkmark$} & \scriptsize [history, and, culture, ,, natural] \\
\scriptsize she always valued honesty and & \scriptsize kindness & \scriptsize the & {\color{red}$\times$} & \scriptsize [the, love, other, her, a] \\
\scriptsize the treaty established lasting & \scriptsize peace & \scriptsize the & {\color{red}$\times$} & \scriptsize [the, status, place, independence, a] \\
\scriptsize the car moved forward & \scriptsize slowly & \scriptsize to & {\color{red}$\times$} & \scriptsize [to, from, in, ,, by] \\
\scriptsize he waited & \scriptsize patiently & \scriptsize for & {\color{red}$\times$} & \scriptsize [for, in, at, a, on] \\
\scriptsize she tied the laces on her & \scriptsize shoes & \scriptsize life & {\color{red}$\times$} & \scriptsize [life, death, own, family, the] \\
\scriptsize he wrapped the scarf around his & \scriptsize neck & \scriptsize life & {\color{red}$\times$} & \scriptsize [life, body, left, hair, new] \\
\scriptsize the doctor examined the patient's & \scriptsize heart & \scriptsize body & {\color{red}$\times$} & \scriptsize [body, life, family, doctors, eye] \\
\scriptsize she checked the notifications on her & \scriptsize phone & \scriptsize family & {\color{red}$\times$} & \scriptsize [family, life, mother, wedding, career] \\
\scriptsize the password must contain at least eight & \scriptsize characters & \scriptsize million & {\color{red}$\times$} & \scriptsize [million, of, life, species, points] \\
\scriptsize he plugged the charger into the & \scriptsize socket & \scriptsize office & {\color{red}$\times$} & \scriptsize [office, room, glass, life, building] \\
\scriptsize the door began to & \scriptsize open & \scriptsize be & {\color{red}$\times$} & \scriptsize [be, fall, fire, look, have] \\
\scriptsize she refused to & \scriptsize quit & \scriptsize be & {\color{red}$\times$} & \scriptsize [be, give, do, marry, attend] \\
\bottomrule
\end{longtable}

\subsection{Small LSTM: Full Qualitative Cloze Data}\label{sec:supp_small_lstm_cloze}

\begin{longtable}{@{}p{3.0cm}p{0.9cm}p{1.1cm}cp{3.5cm}@{}}
\caption{Small LSTM: Baseline ($k=0$), all 100 Cloze prompts.} \\
\toprule
Prompt & Exp. & Pred. & & Top-5 \\
\midrule
\endfirsthead
\toprule
Prompt & Exp. & Pred. & & Top-5 \\
\midrule
\endhead
\scriptsize once upon a & \scriptsize time & \scriptsize year & {\color{orange}$\sim$} & \scriptsize [year, period, time, new, few] \\
\scriptsize the sun was shining bright in the & \scriptsize sky & \scriptsize middle & {\color{red}$\times$} & \scriptsize [middle, same, face, first, second] \\
\scriptsize she picked up the & \scriptsize ball & \scriptsize first & {\color{red}$\times$} & \scriptsize [first, ", the, second, new] \\
\scriptsize he put the food in his & \scriptsize mouth & \scriptsize own & {\color{red}$\times$} & \scriptsize [own, home, car, first, life] \\
\scriptsize the little girl was wearing a pretty & \scriptsize dress & \scriptsize long & {\color{red}$\times$} & \scriptsize [long, little, large, high, a] \\
\scriptsize the bird flew up into the & \scriptsize sky & \scriptsize city & {\color{red}$\times$} & \scriptsize [city, united, country, area, first] \\
\scriptsize she gave her friend a big & \scriptsize hug & \scriptsize day & {\color{red}$\times$} & \scriptsize [day, -, job, chance, role] \\
\scriptsize the cat jumped on the & \scriptsize table & \scriptsize ground & {\color{red}$\times$} & \scriptsize [ground, side, back, island, field] \\
\scriptsize he found a beautiful & \scriptsize flower & \scriptsize man & {\color{red}$\times$} & \scriptsize [man, -, ,, ", and] \\
\scriptsize they played together in the & \scriptsize park & \scriptsize united & {\color{red}$\times$} & \scriptsize [united, first, early, second, same] \\
\scriptsize she sang a beautiful & \scriptsize song & \scriptsize man & {\color{red}$\times$} & \scriptsize [man, -, ,, ", day] \\
\scriptsize he wanted to go & \scriptsize home & \scriptsize to & {\color{red}$\times$} & \scriptsize [to, into, on, in, out] \\
\scriptsize the really big & \scriptsize bear & \scriptsize - & {\color{red}$\times$} & \scriptsize [-, ,, of, and, "] \\
\scriptsize a very small & \scriptsize mouse & \scriptsize number & {\color{red}$\times$} & \scriptsize [number, -, ,, amount, and] \\
\scriptsize the extremely bright & \scriptsize light & \scriptsize - & {\color{red}$\times$} & \scriptsize [-, ,, and, of, "] \\
\scriptsize he was very happy and started to & \scriptsize laugh & \scriptsize be & {\color{red}$\times$} & \scriptsize [be, make, have, take, work] \\
\scriptsize the dog saw a cat and started to & \scriptsize bark & \scriptsize be & {\color{red}$\times$} & \scriptsize [be, have, make, take, play] \\
\scriptsize they went outside to & \scriptsize play & \scriptsize the & {\color{red}$\times$} & \scriptsize [the, a, be, their, take] \\
\scriptsize he ran as fast as he & \scriptsize could & \scriptsize was & {\color{orange}$\sim$} & \scriptsize [was, had, would, could, did] \\
\scriptsize the baby started to & \scriptsize cry & \scriptsize be & {\color{red}$\times$} & \scriptsize [be, have, take, make, use] \\
\scriptsize he carefully & \scriptsize placed & \scriptsize had & {\color{red}$\times$} & \scriptsize [had, made, took, became, defended] \\
\scriptsize they quickly & \scriptsize ran & \scriptsize had & {\color{red}$\times$} & \scriptsize [had, became, were, began, came] \\
\scriptsize she slowly & \scriptsize walked & \scriptsize became & {\color{red}$\times$} & \scriptsize [became, had, took, was, began] \\
\scriptsize he gently & \scriptsize touched & \scriptsize had & {\color{red}$\times$} & \scriptsize [had, took, became, defended, made] \\
\scriptsize he was so tired that he fell & \scriptsize asleep & \scriptsize to & {\color{red}$\times$} & \scriptsize [to, out, in, back, into] \\
\scriptsize she looked at the stars and felt & \scriptsize happy & \scriptsize that & {\color{red}$\times$} & \scriptsize [that, the, it, he, "] \\
\scriptsize she was really & \scriptsize excited & \scriptsize a & {\color{red}$\times$} & \scriptsize [a, the, able, interested, involved] \\
\scriptsize the water was very & \scriptsize cold & \scriptsize much & {\color{red}$\times$} & \scriptsize [much, small, large, expensive, successful] \\
\scriptsize he was feeling very & \scriptsize sad & \scriptsize much & {\color{red}$\times$} & \scriptsize [much, weak, close, good, well] \\
\scriptsize it was an incredibly & \scriptsize beautiful & \scriptsize successful & {\color{red}$\times$} & \scriptsize [successful, important, expensive, popular, well] \\
\scriptsize she had a particularly & \scriptsize special & \scriptsize successful & {\color{red}$\times$} & \scriptsize [successful, strong, good, positive, powerful] \\
\scriptsize the surprisingly & \scriptsize gentle & \scriptsize successful & {\color{red}$\times$} & \scriptsize [successful, common, high, difficult, large] \\
\scriptsize an unusually & \scriptsize quiet & \scriptsize large & {\color{red}$\times$} & \scriptsize [large, low, high, small, successful] \\
\scriptsize the remarkably & \scriptsize clever & \scriptsize high & {\color{red}$\times$} & \scriptsize [high, large, successful, small, common] \\
\scriptsize a wonderfully & \scriptsize warm & \scriptsize successful & {\color{red}$\times$} & \scriptsize [successful, -, high, important, different] \\
\scriptsize she opened the door and walked & \scriptsize inside & \scriptsize out & {\color{red}$\times$} & \scriptsize [out, into, to, down, the] \\
\scriptsize the story ended & \scriptsize happily & \scriptsize in & {\color{red}$\times$} & \scriptsize [in, by, with, ., on] \\
\scriptsize the children were playing & \scriptsize outside & \scriptsize the & {\color{red}$\times$} & \scriptsize [the, a, in, ., on] \\
\scriptsize the boat sailed across the & \scriptsize ocean & \scriptsize country & {\color{red}$\times$} & \scriptsize [country, river, united, city, world] \\
\scriptsize she hung the picture on the & \scriptsize wall & \scriptsize ground & {\color{red}$\times$} & \scriptsize [ground, side, first, other, same] \\
\scriptsize the keys were left on the & \scriptsize counter & \scriptsize ground & {\color{red}$\times$} & \scriptsize [ground, side, first, island, field] \\
\scriptsize he parked the car in the & \scriptsize garage & \scriptsize house & {\color{red}$\times$} & \scriptsize [house, first, area, middle, city] \\
\scriptsize the plane disappeared behind the & \scriptsize clouds & \scriptsize island & {\color{red}$\times$} & \scriptsize [island, river, city, the, road] \\
\scriptsize they arrived early in the & \scriptsize morning & \scriptsize year & {\color{red}$\times$} & \scriptsize [year, day, night, war, month] \\
\scriptsize the concert lasted until & \scriptsize midnight & \scriptsize the & {\color{red}$\times$} & \scriptsize [the, november, october, april, june] \\
\scriptsize we will meet again next & \scriptsize week & \scriptsize to & {\color{red}$\times$} & \scriptsize [to, ., year, ,, and] \\
\scriptsize she poured the milk into the & \scriptsize glass & \scriptsize house & {\color{red}$\times$} & \scriptsize [house, head, water, field, back] \\
\scriptsize he spread butter on the & \scriptsize bread & \scriptsize ground & {\color{red}$\times$} & \scriptsize [ground, side, back, island, field] \\
\scriptsize the chef chopped the & \scriptsize vegetables & \scriptsize ball & {\color{red}$\times$} & \scriptsize [ball, house, entire, money, field] \\
\scriptsize she stirred the soup with a & \scriptsize spoon & \scriptsize " & {\color{red}$\times$} & \scriptsize [", new, large, long, little] \\
\scriptsize the horse galloped across the & \scriptsize field & \scriptsize city & {\color{red}$\times$} & \scriptsize [city, world, river, country, area] \\
\scriptsize the fish swam in the & \scriptsize pond & \scriptsize area & {\color{red}$\times$} & \scriptsize [area, united, city, middle, house] \\
\scriptsize the eagle soared above the & \scriptsize mountains & \scriptsize end & {\color{red}$\times$} & \scriptsize [end, ground, top, field, same] \\
\scriptsize after the long hike he was extremely & \scriptsize tired & \scriptsize successful & {\color{red}$\times$} & \scriptsize [successful, busy, close, well, ill] \\
\scriptsize the news made everyone very & \scriptsize happy & \scriptsize much & {\color{red}$\times$} & \scriptsize [much, close, difficult, good, a] \\
\scriptsize the dark alley made her feel & \scriptsize afraid & \scriptsize in & {\color{red}$\times$} & \scriptsize [in, a, on, the, as] \\
\scriptsize the puzzle was incredibly & \scriptsize difficult & \scriptsize successful & {\color{orange}$\sim$} & \scriptsize [successful, well, important, difficult, popular] \\
\scriptsize the audience was completely & \scriptsize silent & \scriptsize damaged & {\color{red}$\times$} & \scriptsize [damaged, the, a, involved, in] \\
\scriptsize the teacher asked the students to & \scriptsize listen & \scriptsize be & {\color{red}$\times$} & \scriptsize [be, take, make, go, have] \\
\scriptsize she decided to finally & \scriptsize leave & \scriptsize be & {\color{red}$\times$} & \scriptsize [be, make, take, have, get] \\
\scriptsize the alarm clock started to & \scriptsize ring & \scriptsize be & {\color{red}$\times$} & \scriptsize [be, have, take, make, the] \\
\scriptsize the snow began to & \scriptsize melt & \scriptsize be & {\color{red}$\times$} & \scriptsize [be, have, take, make, use] \\
\scriptsize he promised he would never & \scriptsize forget & \scriptsize be & {\color{red}$\times$} & \scriptsize [be, have, want, do, get] \\
\scriptsize the children love to & \scriptsize swim & \scriptsize be & {\color{red}$\times$} & \scriptsize [be, the, have, make, take] \\
\scriptsize she wanted to learn how to & \scriptsize dance & \scriptsize be & {\color{red}$\times$} & \scriptsize [be, make, do, take, get] \\
\scriptsize the mechanic tried to & \scriptsize repair & \scriptsize make & {\color{red}$\times$} & \scriptsize [make, be, take, create, get] \\
\scriptsize dark clouds gathered before the & \scriptsize storm & \scriptsize first & {\color{red}$\times$} & \scriptsize [first, death, battle, end, second] \\
\scriptsize the flowers bloomed in the & \scriptsize spring & \scriptsize first & {\color{red}$\times$} & \scriptsize [first, same, middle, united, world] \\
\scriptsize the leaves fall from the trees in & \scriptsize autumn & \scriptsize the & {\color{red}$\times$} & \scriptsize [the, a, an, which, his] \\
\scriptsize the river flowed into the & \scriptsize sea & \scriptsize city & {\color{red}$\times$} & \scriptsize [city, area, river, village, town] \\
\scriptsize the student wrote an & \scriptsize essay & \scriptsize " & {\color{red}$\times$} & \scriptsize [", important, article, additional, annual] \\
\scriptsize she typed the email on her & \scriptsize keyboard & \scriptsize own & {\color{red}$\times$} & \scriptsize [own, first, behalf, return, second] \\
\scriptsize the professor gave a & \scriptsize lecture & \scriptsize new & {\color{red}$\times$} & \scriptsize [new, ", letter, total, telegram] \\
\scriptsize he submitted his final & \scriptsize report & \scriptsize work & {\color{red}$\times$} & \scriptsize [work, appearance, book, role, ,] \\
\scriptsize the crowd started to & \scriptsize cheer & \scriptsize be & {\color{red}$\times$} & \scriptsize [be, have, take, make, use] \\
\scriptsize he bent down to & \scriptsize pick & \scriptsize the & {\color{red}$\times$} & \scriptsize [the, a, be, his, an] \\
\scriptsize she leaned forward to & \scriptsize whisper & \scriptsize the & {\color{red}$\times$} & \scriptsize [the, a, his, an, her] \\
\scriptsize the wind continued to & \scriptsize blow & \scriptsize be & {\color{red}$\times$} & \scriptsize [be, have, make, take, reach] \\
\scriptsize they gathered around to & \scriptsize celebrate & \scriptsize the & {\color{red}$\times$} & \scriptsize [the, be, a, take, their] \\
\scriptsize he struggled to & \scriptsize breathe & \scriptsize make & {\color{red}$\times$} & \scriptsize [make, be, take, get, find] \\
\scriptsize the sunset was absolutely & \scriptsize stunning & \scriptsize a & {\color{red}$\times$} & \scriptsize [a, the, an, successful, in] \\
\scriptsize the old house looked rather & \scriptsize creepy & \scriptsize than & {\color{red}$\times$} & \scriptsize [than, to, ., in, -] \\
\scriptsize his explanation was perfectly & \scriptsize clear & \scriptsize a & {\color{red}$\times$} & \scriptsize [a, successful, the, positive, .] \\
\scriptsize the cake tasted absolutely & \scriptsize delicious & \scriptsize a & {\color{red}$\times$} & \scriptsize [a, the, more, to, an] \\
\scriptsize the mountain trail was extremely & \scriptsize steep & \scriptsize successful & {\color{red}$\times$} & \scriptsize [successful, well, small, large, close] \\
\scriptsize the instructions were surprisingly & \scriptsize simple & \scriptsize successful & {\color{red}$\times$} & \scriptsize [successful, well, a, difficult, important] \\
\scriptsize the scientist published an important & \scriptsize discovery & \scriptsize role & {\color{red}$\times$} & \scriptsize [role, part, review, statement, theme] \\
\scriptsize the city was known for its rich & \scriptsize history & \scriptsize and & {\color{red}$\times$} & \scriptsize [and, ,, -, ., "] \\
\scriptsize she always valued honesty and & \scriptsize kindness & \scriptsize the & {\color{red}$\times$} & \scriptsize [the, a, was, is, his] \\
\scriptsize the treaty established lasting & \scriptsize peace & \scriptsize to & {\color{red}$\times$} & \scriptsize [to, the, in, from, by] \\
\scriptsize the car moved forward & \scriptsize slowly & \scriptsize to & {\color{red}$\times$} & \scriptsize [to, ., ,, and, into] \\
\scriptsize he waited & \scriptsize patiently & \scriptsize for & {\color{red}$\times$} & \scriptsize [for, to, in, a, .] \\
\scriptsize she tied the laces on her & \scriptsize shoes & \scriptsize own & {\color{red}$\times$} & \scriptsize [own, first, side, back, second] \\
\scriptsize he wrapped the scarf around his & \scriptsize neck & \scriptsize head & {\color{orange}$\sim$} & \scriptsize [head, own, body, face, neck] \\
\scriptsize the doctor examined the patient's & \scriptsize heart & \scriptsize life & {\color{red}$\times$} & \scriptsize [life, ", death, body, first] \\
\scriptsize she checked the notifications on her & \scriptsize phone & \scriptsize own & {\color{red}$\times$} & \scriptsize [own, first, side, way, home] \\
\scriptsize the password must contain at least eight & \scriptsize characters & \scriptsize of & {\color{red}$\times$} & \scriptsize [of, ,, ., years, days] \\
\scriptsize he plugged the charger into the & \scriptsize socket & \scriptsize first & {\color{red}$\times$} & \scriptsize [first, house, ", air, second] \\
\scriptsize the door began to & \scriptsize open & \scriptsize be & {\color{red}$\times$} & \scriptsize [be, have, take, make, use] \\
\scriptsize she refused to & \scriptsize quit & \scriptsize make & {\color{red}$\times$} & \scriptsize [make, be, take, do, return] \\
\bottomrule
\end{longtable}

\begin{longtable}{@{}p{3.0cm}p{0.9cm}p{1.1cm}cp{3.5cm}@{}}
\caption{Small LSTM: Whole-Unit Removal (\texttt{weight\_ih} + \texttt{weight\_hh}, $k=0.2$, main-paper target), all 100 Cloze prompts.} \\
\toprule
Prompt & Exp. & Pred. & & Top-5 \\
\midrule
\endfirsthead
\toprule
Prompt & Exp. & Pred. & & Top-5 \\
\midrule
\endhead
\scriptsize once upon a & \scriptsize time & \scriptsize " & {\color{red}$\times$} & \scriptsize [", year, day, long, few] \\
\scriptsize the sun was shining bright in the & \scriptsize sky & \scriptsize first & {\color{red}$\times$} & \scriptsize [first, world, same, united, middle] \\
\scriptsize she picked up the & \scriptsize ball & \scriptsize first & {\color{red}$\times$} & \scriptsize [first, ", the, second, same] \\
\scriptsize he put the food in his & \scriptsize mouth & \scriptsize own & {\color{red}$\times$} & \scriptsize [own, first, life, home, time] \\
\scriptsize the little girl was wearing a pretty & \scriptsize dress & \scriptsize - & {\color{red}$\times$} & \scriptsize [-, ,, a, ., the] \\
\scriptsize the bird flew up into the & \scriptsize sky & \scriptsize first & {\color{red}$\times$} & \scriptsize [first, world, second, the, "] \\
\scriptsize she gave her friend a big & \scriptsize hug & \scriptsize , & {\color{red}$\times$} & \scriptsize [,, ., and, -, in] \\
\scriptsize the cat jumped on the & \scriptsize table & \scriptsize first & {\color{red}$\times$} & \scriptsize [first, ground, second, other, field] \\
\scriptsize he found a beautiful & \scriptsize flower & \scriptsize , & {\color{red}$\times$} & \scriptsize [,, -, and, ., "] \\
\scriptsize they played together in the & \scriptsize park & \scriptsize first & {\color{red}$\times$} & \scriptsize [first, second, series, early, same] \\
\scriptsize she sang a beautiful & \scriptsize song & \scriptsize , & {\color{red}$\times$} & \scriptsize [,, -, and, ., in] \\
\scriptsize he wanted to go & \scriptsize home & \scriptsize to & {\color{red}$\times$} & \scriptsize [to, in, on, into, ,] \\
\scriptsize the really big & \scriptsize bear & \scriptsize , & {\color{red}$\times$} & \scriptsize [,, -, of, and, .] \\
\scriptsize a very small & \scriptsize mouse & \scriptsize , & {\color{red}$\times$} & \scriptsize [,, -, ., and, in] \\
\scriptsize the extremely bright & \scriptsize light & \scriptsize , & {\color{red}$\times$} & \scriptsize [,, ., -, and, of] \\
\scriptsize he was very happy and started to & \scriptsize laugh & \scriptsize be & {\color{red}$\times$} & \scriptsize [be, have, make, the, use] \\
\scriptsize the dog saw a cat and started to & \scriptsize bark & \scriptsize be & {\color{red}$\times$} & \scriptsize [be, have, the, use, make] \\
\scriptsize they went outside to & \scriptsize play & \scriptsize the & {\color{red}$\times$} & \scriptsize [the, a, be, their, an] \\
\scriptsize he ran as fast as he & \scriptsize could & \scriptsize was & {\color{orange}$\sim$} & \scriptsize [was, had, would, ,, could] \\
\scriptsize the baby started to & \scriptsize cry & \scriptsize be & {\color{red}$\times$} & \scriptsize [be, have, the, use, work] \\
\scriptsize he carefully & \scriptsize placed & \scriptsize had & {\color{red}$\times$} & \scriptsize [had, made, ,, used, found] \\
\scriptsize they quickly & \scriptsize ran & \scriptsize had & {\color{red}$\times$} & \scriptsize [had, ,, were, been, the] \\
\scriptsize she slowly & \scriptsize walked & \scriptsize had & {\color{red}$\times$} & \scriptsize [had, was, ,, became, in] \\
\scriptsize he gently & \scriptsize touched & \scriptsize had & {\color{red}$\times$} & \scriptsize [had, to, ,, made, was] \\
\scriptsize he was so tired that he fell & \scriptsize asleep & \scriptsize to & {\color{red}$\times$} & \scriptsize [to, ., in, ,, and] \\
\scriptsize she looked at the stars and felt & \scriptsize happy & \scriptsize that & {\color{red}$\times$} & \scriptsize [that, the, ,, it, .] \\
\scriptsize she was really & \scriptsize excited & \scriptsize the & {\color{red}$\times$} & \scriptsize [the, a, in, to, given] \\
\scriptsize the water was very & \scriptsize cold & \scriptsize the & {\color{red}$\times$} & \scriptsize [the, a, well, successful, ,] \\
\scriptsize he was feeling very & \scriptsize sad & \scriptsize the & {\color{red}$\times$} & \scriptsize [the, to, a, in, well] \\
\scriptsize it was an incredibly & \scriptsize beautiful & \scriptsize successful & {\color{red}$\times$} & \scriptsize [successful, -, important, the, a] \\
\scriptsize she had a particularly & \scriptsize special & \scriptsize - & {\color{red}$\times$} & \scriptsize [-, ,, of, ., in] \\
\scriptsize the surprisingly & \scriptsize gentle & \scriptsize - & {\color{red}$\times$} & \scriptsize [-, the, of, and, ,] \\
\scriptsize an unusually & \scriptsize quiet & \scriptsize - & {\color{red}$\times$} & \scriptsize [-, successful, the, a, long] \\
\scriptsize the remarkably & \scriptsize clever & \scriptsize the & {\color{red}$\times$} & \scriptsize [the, -, and, a, of] \\
\scriptsize a wonderfully & \scriptsize warm & \scriptsize , & {\color{red}$\times$} & \scriptsize [,, ., -, and, in] \\
\scriptsize she opened the door and walked & \scriptsize inside & \scriptsize to & {\color{red}$\times$} & \scriptsize [to, the, ., a, in] \\
\scriptsize the story ended & \scriptsize happily & \scriptsize in & {\color{red}$\times$} & \scriptsize [in, ., ,, to, the] \\
\scriptsize the children were playing & \scriptsize outside & \scriptsize the & {\color{red}$\times$} & \scriptsize [the, a, in, ,, .] \\
\scriptsize the boat sailed across the & \scriptsize ocean & \scriptsize city & {\color{red}$\times$} & \scriptsize [city, country, world, new, area] \\
\scriptsize she hung the picture on the & \scriptsize wall & \scriptsize first & {\color{red}$\times$} & \scriptsize [first, ground, ", other, show] \\
\scriptsize the keys were left on the & \scriptsize counter & \scriptsize first & {\color{red}$\times$} & \scriptsize [first, ground, field, other, second] \\
\scriptsize he parked the car in the & \scriptsize garage & \scriptsize first & {\color{red}$\times$} & \scriptsize [first, world, area, air, same] \\
\scriptsize the plane disappeared behind the & \scriptsize clouds & \scriptsize first & {\color{red}$\times$} & \scriptsize [first, ", the, new, original] \\
\scriptsize they arrived early in the & \scriptsize morning & \scriptsize first & {\color{red}$\times$} & \scriptsize [first, early, year, second, same] \\
\scriptsize the concert lasted until & \scriptsize midnight & \scriptsize the & {\color{red}$\times$} & \scriptsize [the, a, its, that, ,] \\
\scriptsize we will meet again next & \scriptsize week & \scriptsize , & {\color{red}$\times$} & \scriptsize [,, ., to, and, the] \\
\scriptsize she poured the milk into the & \scriptsize glass & \scriptsize first & {\color{red}$\times$} & \scriptsize [first, the, ", air, film] \\
\scriptsize he spread butter on the & \scriptsize bread & \scriptsize ground & {\color{red}$\times$} & \scriptsize [ground, first, field, other, island] \\
\scriptsize the chef chopped the & \scriptsize vegetables & \scriptsize first & {\color{red}$\times$} & \scriptsize [first, the, ", same, game] \\
\scriptsize she stirred the soup with a & \scriptsize spoon & \scriptsize " & {\color{red}$\times$} & \scriptsize [", new, single, long, large] \\
\scriptsize the horse galloped across the & \scriptsize field & \scriptsize field & {\color{green!50!black}$\checkmark$} & \scriptsize [field, city, world, house, area] \\
\scriptsize the fish swam in the & \scriptsize pond & \scriptsize first & {\color{red}$\times$} & \scriptsize [first, world, united, area, same] \\
\scriptsize the eagle soared above the & \scriptsize mountains & \scriptsize end & {\color{red}$\times$} & \scriptsize [end, first, same, the, time] \\
\scriptsize after the long hike he was extremely & \scriptsize tired & \scriptsize the & {\color{red}$\times$} & \scriptsize [the, successful, a, well, in] \\
\scriptsize the news made everyone very & \scriptsize happy & \scriptsize the & {\color{red}$\times$} & \scriptsize [the, a, ., to, ,] \\
\scriptsize the dark alley made her feel & \scriptsize afraid & \scriptsize . & {\color{red}$\times$} & \scriptsize [., ,, and, in, the] \\
\scriptsize the puzzle was incredibly & \scriptsize difficult & \scriptsize the & {\color{red}$\times$} & \scriptsize [the, a, well, successful, of] \\
\scriptsize the audience was completely & \scriptsize silent & \scriptsize the & {\color{red}$\times$} & \scriptsize [the, a, in, to, used] \\
\scriptsize the teacher asked the students to & \scriptsize listen & \scriptsize be & {\color{red}$\times$} & \scriptsize [be, the, use, have, make] \\
\scriptsize she decided to finally & \scriptsize leave & \scriptsize be & {\color{red}$\times$} & \scriptsize [be, take, use, make, have] \\
\scriptsize the alarm clock started to & \scriptsize ring & \scriptsize be & {\color{red}$\times$} & \scriptsize [be, have, the, use, place] \\
\scriptsize the snow began to & \scriptsize melt & \scriptsize be & {\color{red}$\times$} & \scriptsize [be, have, the, use, make] \\
\scriptsize he promised he would never & \scriptsize forget & \scriptsize be & {\color{red}$\times$} & \scriptsize [be, have, work, want, take] \\
\scriptsize the children love to & \scriptsize swim & \scriptsize be & {\color{red}$\times$} & \scriptsize [be, the, have, a, take] \\
\scriptsize she wanted to learn how to & \scriptsize dance & \scriptsize be & {\color{red}$\times$} & \scriptsize [be, make, the, take, get] \\
\scriptsize the mechanic tried to & \scriptsize repair & \scriptsize be & {\color{red}$\times$} & \scriptsize [be, make, take, use, the] \\
\scriptsize dark clouds gathered before the & \scriptsize storm & \scriptsize first & {\color{red}$\times$} & \scriptsize [first, ", end, second, the] \\
\scriptsize the flowers bloomed in the & \scriptsize spring & \scriptsize first & {\color{red}$\times$} & \scriptsize [first, united, world, new, the] \\
\scriptsize the leaves fall from the trees in & \scriptsize autumn & \scriptsize the & {\color{red}$\times$} & \scriptsize [the, a, their, its, an] \\
\scriptsize the river flowed into the & \scriptsize sea & \scriptsize area & {\color{red}$\times$} & \scriptsize [area, city, world, the, field] \\
\scriptsize the student wrote an & \scriptsize essay & \scriptsize important & {\color{red}$\times$} & \scriptsize [important, ", additional, attempt, annual] \\
\scriptsize she typed the email on her & \scriptsize keyboard & \scriptsize . & {\color{red}$\times$} & \scriptsize [., ,, the, to, and] \\
\scriptsize the professor gave a & \scriptsize lecture & \scriptsize " & {\color{red}$\times$} & \scriptsize [", new, total, single, ,] \\
\scriptsize he submitted his final & \scriptsize report & \scriptsize , & {\color{red}$\times$} & \scriptsize [,, ., and, -, the] \\
\scriptsize the crowd started to & \scriptsize cheer & \scriptsize be & {\color{red}$\times$} & \scriptsize [be, have, the, use, make] \\
\scriptsize he bent down to & \scriptsize pick & \scriptsize the & {\color{red}$\times$} & \scriptsize [the, a, be, an, his] \\
\scriptsize she leaned forward to & \scriptsize whisper & \scriptsize the & {\color{red}$\times$} & \scriptsize [the, a, his, an, ,] \\
\scriptsize the wind continued to & \scriptsize blow & \scriptsize be & {\color{red}$\times$} & \scriptsize [be, the, have, make, end] \\
\scriptsize they gathered around to & \scriptsize celebrate & \scriptsize the & {\color{red}$\times$} & \scriptsize [the, be, a, have, take] \\
\scriptsize he struggled to & \scriptsize breathe & \scriptsize be & {\color{red}$\times$} & \scriptsize [be, make, the, take, play] \\
\scriptsize the sunset was absolutely & \scriptsize stunning & \scriptsize the & {\color{red}$\times$} & \scriptsize [the, a, in, ,, used] \\
\scriptsize the old house looked rather & \scriptsize creepy & \scriptsize . & {\color{red}$\times$} & \scriptsize [., ,, to, in, and] \\
\scriptsize his explanation was perfectly & \scriptsize clear & \scriptsize the & {\color{red}$\times$} & \scriptsize [the, ., a, ,, in] \\
\scriptsize the cake tasted absolutely & \scriptsize delicious & \scriptsize the & {\color{red}$\times$} & \scriptsize [the, a, ., in, to] \\
\scriptsize the mountain trail was extremely & \scriptsize steep & \scriptsize the & {\color{red}$\times$} & \scriptsize [the, a, well, successful, of] \\
\scriptsize the instructions were surprisingly & \scriptsize simple & \scriptsize the & {\color{red}$\times$} & \scriptsize [the, a, in, to, used] \\
\scriptsize the scientist published an important & \scriptsize discovery & \scriptsize , & {\color{red}$\times$} & \scriptsize [,, ", and, the, -] \\
\scriptsize the city was known for its rich & \scriptsize history & \scriptsize , & {\color{red}$\times$} & \scriptsize [,, and, ., -, in] \\
\scriptsize she always valued honesty and & \scriptsize kindness & \scriptsize the & {\color{red}$\times$} & \scriptsize [the, a, his, was, in] \\
\scriptsize the treaty established lasting & \scriptsize peace & \scriptsize the & {\color{red}$\times$} & \scriptsize [the, to, a, ,, in] \\
\scriptsize the car moved forward & \scriptsize slowly & \scriptsize to & {\color{red}$\times$} & \scriptsize [to, ,, ., and, in] \\
\scriptsize he waited & \scriptsize patiently & \scriptsize to & {\color{red}$\times$} & \scriptsize [to, the, a, ., ,] \\
\scriptsize she tied the laces on her & \scriptsize shoes & \scriptsize . & {\color{red}$\times$} & \scriptsize [., ,, to, own, and] \\
\scriptsize he wrapped the scarf around his & \scriptsize neck & \scriptsize own & {\color{red}$\times$} & \scriptsize [own, first, life, home, the] \\
\scriptsize the doctor examined the patient's & \scriptsize heart & \scriptsize " & {\color{red}$\times$} & \scriptsize [", first, the, and, .] \\
\scriptsize she checked the notifications on her & \scriptsize phone & \scriptsize . & {\color{red}$\times$} & \scriptsize [., own, ,, to, the] \\
\scriptsize the password must contain at least eight & \scriptsize characters & \scriptsize , & {\color{red}$\times$} & \scriptsize [,, ., of, in, and] \\
\scriptsize he plugged the charger into the & \scriptsize socket & \scriptsize first & {\color{red}$\times$} & \scriptsize [first, ", the, world, second] \\
\scriptsize the door began to & \scriptsize open & \scriptsize be & {\color{red}$\times$} & \scriptsize [be, have, the, use, make] \\
\scriptsize she refused to & \scriptsize quit & \scriptsize be & {\color{red}$\times$} & \scriptsize [be, work, take, make, play] \\
\bottomrule
\end{longtable}
\begin{longtable}{@{}p{3.0cm}p{0.9cm}p{1.1cm}cp{3.5cm}@{}}
\caption{Small LSTM: Recurrent-Connection Pruning (\texttt{weight\_hh}, $k=0.2$, supplementary control), all 100 Cloze prompts.} \\
\toprule
Prompt & Exp. & Pred. & & Top-5 \\
\midrule
\endfirsthead
\toprule
Prompt & Exp. & Pred. & & Top-5 \\
\midrule
\endhead
\scriptsize once upon a & \scriptsize time & \scriptsize new & {\color{red}$\times$} & \scriptsize [new, few, ", brief, single] \\
\scriptsize the sun was shining bright in the & \scriptsize sky & \scriptsize same & {\color{red}$\times$} & \scriptsize [same, area, first, middle, house] \\
\scriptsize she picked up the & \scriptsize ball & \scriptsize end & {\color{red}$\times$} & \scriptsize [end, the, ", first, new] \\
\scriptsize he put the food in his & \scriptsize mouth & \scriptsize own & {\color{red}$\times$} & \scriptsize [own, home, first, life, family] \\
\scriptsize the little girl was wearing a pretty & \scriptsize dress & \scriptsize long & {\color{red}$\times$} & \scriptsize [long, large, -, much, a] \\
\scriptsize the bird flew up into the & \scriptsize sky & \scriptsize first & {\color{red}$\times$} & \scriptsize [first, new, the, ", end] \\
\scriptsize she gave her friend a big & \scriptsize hug & \scriptsize point & {\color{red}$\times$} & \scriptsize [point, man, -, two, ,] \\
\scriptsize the cat jumped on the & \scriptsize table & \scriptsize the & {\color{red}$\times$} & \scriptsize [the, ground, first, stage, side] \\
\scriptsize he found a beautiful & \scriptsize flower & \scriptsize - & {\color{red}$\times$} & \scriptsize [-, ,, man, and, woman] \\
\scriptsize they played together in the & \scriptsize park & \scriptsize first & {\color{red}$\times$} & \scriptsize [first, second, same, new, early] \\
\scriptsize she sang a beautiful & \scriptsize song & \scriptsize - & {\color{red}$\times$} & \scriptsize [-, ,, man, and, "] \\
\scriptsize he wanted to go & \scriptsize home & \scriptsize to & {\color{red}$\times$} & \scriptsize [to, on, into, back, out] \\
\scriptsize the really big & \scriptsize bear & \scriptsize - & {\color{red}$\times$} & \scriptsize [-, ,, and, the, part] \\
\scriptsize a very small & \scriptsize mouse & \scriptsize - & {\color{red}$\times$} & \scriptsize [-, number, ,, part, amount] \\
\scriptsize the extremely bright & \scriptsize light & \scriptsize - & {\color{red}$\times$} & \scriptsize [-, ,, and, of, hand] \\
\scriptsize he was very happy and started to & \scriptsize laugh & \scriptsize be & {\color{red}$\times$} & \scriptsize [be, have, make, take, get] \\
\scriptsize the dog saw a cat and started to & \scriptsize bark & \scriptsize be & {\color{red}$\times$} & \scriptsize [be, have, make, take, the] \\
\scriptsize they went outside to & \scriptsize play & \scriptsize the & {\color{red}$\times$} & \scriptsize [the, a, their, be, an] \\
\scriptsize he ran as fast as he & \scriptsize could & \scriptsize was & {\color{red}$\times$} & \scriptsize [was, had, used, ,, would] \\
\scriptsize the baby started to & \scriptsize cry & \scriptsize be & {\color{red}$\times$} & \scriptsize [be, have, the, make, take] \\
\scriptsize he carefully & \scriptsize placed & \scriptsize had & {\color{red}$\times$} & \scriptsize [had, was, made, took, killed] \\
\scriptsize they quickly & \scriptsize ran & \scriptsize had & {\color{red}$\times$} & \scriptsize [had, were, took, became, began] \\
\scriptsize she slowly & \scriptsize walked & \scriptsize became & {\color{red}$\times$} & \scriptsize [became, had, took, was, moved] \\
\scriptsize he gently & \scriptsize touched & \scriptsize had & {\color{red}$\times$} & \scriptsize [had, took, was, became, made] \\
\scriptsize he was so tired that he fell & \scriptsize asleep & \scriptsize out & {\color{red}$\times$} & \scriptsize [out, to, down, in, back] \\
\scriptsize she looked at the stars and felt & \scriptsize happy & \scriptsize that & {\color{red}$\times$} & \scriptsize [that, it, the, they, to] \\
\scriptsize she was really & \scriptsize excited & \scriptsize a & {\color{red}$\times$} & \scriptsize [a, the, able, seen, in] \\
\scriptsize the water was very & \scriptsize cold & \scriptsize much & {\color{red}$\times$} & \scriptsize [much, small, expensive, large, a] \\
\scriptsize he was feeling very & \scriptsize sad & \scriptsize much & {\color{red}$\times$} & \scriptsize [much, weak, little, small, large] \\
\scriptsize it was an incredibly & \scriptsize beautiful & \scriptsize successful & {\color{red}$\times$} & \scriptsize [successful, important, strong, expensive, long] \\
\scriptsize she had a particularly & \scriptsize special & \scriptsize successful & {\color{red}$\times$} & \scriptsize [successful, strong, good, large, great] \\
\scriptsize the surprisingly & \scriptsize gentle & \scriptsize the & {\color{red}$\times$} & \scriptsize [the, large, high, a, low] \\
\scriptsize an unusually & \scriptsize quiet & \scriptsize long & {\color{red}$\times$} & \scriptsize [long, much, large, high, low] \\
\scriptsize the remarkably & \scriptsize clever & \scriptsize large & {\color{red}$\times$} & \scriptsize [large, much, few, long, high] \\
\scriptsize a wonderfully & \scriptsize warm & \scriptsize successful & {\color{red}$\times$} & \scriptsize [successful, -, different, high, short] \\
\scriptsize she opened the door and walked & \scriptsize inside & \scriptsize out & {\color{red}$\times$} & \scriptsize [out, to, down, into, on] \\
\scriptsize the story ended & \scriptsize happily & \scriptsize in & {\color{red}$\times$} & \scriptsize [in, with, by, ., on] \\
\scriptsize the children were playing & \scriptsize outside & \scriptsize the & {\color{red}$\times$} & \scriptsize [the, a, in, and, on] \\
\scriptsize the boat sailed across the & \scriptsize ocean & \scriptsize country & {\color{red}$\times$} & \scriptsize [country, city, world, united, area] \\
\scriptsize she hung the picture on the & \scriptsize wall & \scriptsize the & {\color{red}$\times$} & \scriptsize [the, first, new, same, stage] \\
\scriptsize the keys were left on the & \scriptsize counter & \scriptsize first & {\color{red}$\times$} & \scriptsize [first, end, the, same, second] \\
\scriptsize he parked the car in the & \scriptsize garage & \scriptsize area & {\color{red}$\times$} & \scriptsize [area, first, world, new, house] \\
\scriptsize the plane disappeared behind the & \scriptsize clouds & \scriptsize the & {\color{red}$\times$} & \scriptsize [the, new, two, city, other] \\
\scriptsize they arrived early in the & \scriptsize morning & \scriptsize first & {\color{red}$\times$} & \scriptsize [first, same, early, second, summer] \\
\scriptsize the concert lasted until & \scriptsize midnight & \scriptsize the & {\color{red}$\times$} & \scriptsize [the, they, it, his, its] \\
\scriptsize we will meet again next & \scriptsize week & \scriptsize . & {\color{red}$\times$} & \scriptsize [., to, ,, and, until] \\
\scriptsize she poured the milk into the & \scriptsize glass & \scriptsize house & {\color{red}$\times$} & \scriptsize [house, the, water, air, car] \\
\scriptsize he spread butter on the & \scriptsize bread & \scriptsize ground & {\color{red}$\times$} & \scriptsize [ground, other, the, side, new] \\
\scriptsize the chef chopped the & \scriptsize vegetables & \scriptsize house & {\color{red}$\times$} & \scriptsize [house, the, ,, car, ball] \\
\scriptsize she stirred the soup with a & \scriptsize spoon & \scriptsize " & {\color{red}$\times$} & \scriptsize [", new, large, number, single] \\
\scriptsize the horse galloped across the & \scriptsize field & \scriptsize city & {\color{red}$\times$} & \scriptsize [city, area, world, country, river] \\
\scriptsize the fish swam in the & \scriptsize pond & \scriptsize area & {\color{red}$\times$} & \scriptsize [area, city, united, new, world] \\
\scriptsize the eagle soared above the & \scriptsize mountains & \scriptsize the & {\color{red}$\times$} & \scriptsize [the, end, ground, line, field] \\
\scriptsize after the long hike he was extremely & \scriptsize tired & \scriptsize close & {\color{red}$\times$} & \scriptsize [close, successful, busy, small, much] \\
\scriptsize the news made everyone very & \scriptsize happy & \scriptsize much & {\color{red}$\times$} & \scriptsize [much, close, long, far, different] \\
\scriptsize the dark alley made her feel & \scriptsize afraid & \scriptsize to & {\color{red}$\times$} & \scriptsize [to, ", a, in, ,] \\
\scriptsize the puzzle was incredibly & \scriptsize difficult & \scriptsize successful & {\color{orange}$\sim$} & \scriptsize [successful, well, much, popular, difficult] \\
\scriptsize the audience was completely & \scriptsize silent & \scriptsize destroyed & {\color{red}$\times$} & \scriptsize [destroyed, damaged, used, able, the] \\
\scriptsize the teacher asked the students to & \scriptsize listen & \scriptsize be & {\color{red}$\times$} & \scriptsize [be, have, take, make, go] \\
\scriptsize she decided to finally & \scriptsize leave & \scriptsize be & {\color{red}$\times$} & \scriptsize [be, the, make, take, become] \\
\scriptsize the alarm clock started to & \scriptsize ring & \scriptsize be & {\color{red}$\times$} & \scriptsize [be, have, the, make, take] \\
\scriptsize the snow began to & \scriptsize melt & \scriptsize be & {\color{red}$\times$} & \scriptsize [be, have, make, the, take] \\
\scriptsize he promised he would never & \scriptsize forget & \scriptsize be & {\color{red}$\times$} & \scriptsize [be, have, do, go, want] \\
\scriptsize the children love to & \scriptsize swim & \scriptsize be & {\color{red}$\times$} & \scriptsize [be, the, have, make, a] \\
\scriptsize she wanted to learn how to & \scriptsize dance & \scriptsize be & {\color{red}$\times$} & \scriptsize [be, make, get, take, go] \\
\scriptsize the mechanic tried to & \scriptsize repair & \scriptsize be & {\color{red}$\times$} & \scriptsize [be, make, take, have, create] \\
\scriptsize dark clouds gathered before the & \scriptsize storm & \scriptsize first & {\color{red}$\times$} & \scriptsize [first, the, two, ,, second] \\
\scriptsize the flowers bloomed in the & \scriptsize spring & \scriptsize new & {\color{red}$\times$} & \scriptsize [new, same, first, area, early] \\
\scriptsize the leaves fall from the trees in & \scriptsize autumn & \scriptsize the & {\color{red}$\times$} & \scriptsize [the, a, an, which, their] \\
\scriptsize the river flowed into the & \scriptsize sea & \scriptsize area & {\color{red}$\times$} & \scriptsize [area, city, world, village, town] \\
\scriptsize the student wrote an & \scriptsize essay & \scriptsize " & {\color{red}$\times$} & \scriptsize [", important, additional, open, article] \\
\scriptsize she typed the email on her & \scriptsize keyboard & \scriptsize own & {\color{red}$\times$} & \scriptsize [own, death, first, behalf, the] \\
\scriptsize the professor gave a & \scriptsize lecture & \scriptsize new & {\color{red}$\times$} & \scriptsize [new, ", number, total, more] \\
\scriptsize he submitted his final & \scriptsize report & \scriptsize time & {\color{red}$\times$} & \scriptsize [time, two, day, work, first] \\
\scriptsize the crowd started to & \scriptsize cheer & \scriptsize be & {\color{red}$\times$} & \scriptsize [be, have, the, make, take] \\
\scriptsize he bent down to & \scriptsize pick & \scriptsize the & {\color{red}$\times$} & \scriptsize [the, a, his, be, an] \\
\scriptsize she leaned forward to & \scriptsize whisper & \scriptsize the & {\color{red}$\times$} & \scriptsize [the, a, his, an, be] \\
\scriptsize the wind continued to & \scriptsize blow & \scriptsize be & {\color{red}$\times$} & \scriptsize [be, have, make, the, take] \\
\scriptsize they gathered around to & \scriptsize celebrate & \scriptsize the & {\color{red}$\times$} & \scriptsize [the, a, be, their, an] \\
\scriptsize he struggled to & \scriptsize breathe & \scriptsize be & {\color{red}$\times$} & \scriptsize [be, make, the, take, have] \\
\scriptsize the sunset was absolutely & \scriptsize stunning & \scriptsize a & {\color{red}$\times$} & \scriptsize [a, the, successful, more, "] \\
\scriptsize the old house looked rather & \scriptsize creepy & \scriptsize than & {\color{red}$\times$} & \scriptsize [than, much, far, to, a] \\
\scriptsize his explanation was perfectly & \scriptsize clear & \scriptsize . & {\color{red}$\times$} & \scriptsize [., the, to, in, that] \\
\scriptsize the cake tasted absolutely & \scriptsize delicious & \scriptsize the & {\color{red}$\times$} & \scriptsize [the, a, more, to, "] \\
\scriptsize the mountain trail was extremely & \scriptsize steep & \scriptsize much & {\color{red}$\times$} & \scriptsize [much, small, well, close, successful] \\
\scriptsize the instructions were surprisingly & \scriptsize simple & \scriptsize well & {\color{red}$\times$} & \scriptsize [well, ., much, successful, long] \\
\scriptsize the scientist published an important & \scriptsize discovery & \scriptsize book & {\color{red}$\times$} & \scriptsize [book, statement, article, time, account] \\
\scriptsize the city was known for its rich & \scriptsize history & \scriptsize and & {\color{red}$\times$} & \scriptsize [and, ,, -, use, "] \\
\scriptsize she always valued honesty and & \scriptsize kindness & \scriptsize the & {\color{red}$\times$} & \scriptsize [the, a, his, it, not] \\
\scriptsize the treaty established lasting & \scriptsize peace & \scriptsize a & {\color{red}$\times$} & \scriptsize [a, the, two, 10, to] \\
\scriptsize the car moved forward & \scriptsize slowly & \scriptsize to & {\color{red}$\times$} & \scriptsize [to, and, ,, with, in] \\
\scriptsize he waited & \scriptsize patiently & \scriptsize to & {\color{red}$\times$} & \scriptsize [to, for, and, in, from] \\
\scriptsize she tied the laces on her & \scriptsize shoes & \scriptsize own & {\color{red}$\times$} & \scriptsize [own, first, death, second, father] \\
\scriptsize he wrapped the scarf around his & \scriptsize neck & \scriptsize own & {\color{red}$\times$} & \scriptsize [own, head, home, father, first] \\
\scriptsize the doctor examined the patient's & \scriptsize heart & \scriptsize first & {\color{red}$\times$} & \scriptsize [first, own, home, ", the] \\
\scriptsize she checked the notifications on her & \scriptsize phone & \scriptsize own & {\color{red}$\times$} & \scriptsize [own, first, death, behalf, home] \\
\scriptsize the password must contain at least eight & \scriptsize characters & \scriptsize years & {\color{red}$\times$} & \scriptsize [years, days, -, times, months] \\
\scriptsize he plugged the charger into the & \scriptsize socket & \scriptsize first & {\color{red}$\times$} & \scriptsize [first, the, new, ", two] \\
\scriptsize the door began to & \scriptsize open & \scriptsize be & {\color{red}$\times$} & \scriptsize [be, have, the, make, take] \\
\scriptsize she refused to & \scriptsize quit & \scriptsize be & {\color{red}$\times$} & \scriptsize [be, make, take, give, have] \\
\bottomrule
\end{longtable}

\begin{longtable}{@{}p{3.0cm}p{0.9cm}p{1.1cm}cp{3.5cm}@{}}
\caption{Small LSTM: Input-Connection Pruning (\texttt{weight\_ih}, $k=0.2$, supplementary control), all 100 Cloze prompts.} \\
\toprule
Prompt & Exp. & Pred. & & Top-5 \\
\midrule
\endfirsthead
\toprule
Prompt & Exp. & Pred. & & Top-5 \\
\midrule
\endhead
\scriptsize once upon a & \scriptsize time & \scriptsize part & {\color{orange}$\sim$} & \scriptsize [part, time, ", the, year] \\
\scriptsize the sun was shining bright in the & \scriptsize sky & \scriptsize the & {\color{red}$\times$} & \scriptsize [the, and, year, world, season] \\
\scriptsize she picked up the & \scriptsize ball & \scriptsize the & {\color{red}$\times$} & \scriptsize [the, a, to, ,, .] \\
\scriptsize he put the food in his & \scriptsize mouth & \scriptsize , & {\color{red}$\times$} & \scriptsize [,, ., and, to, -] \\
\scriptsize the little girl was wearing a pretty & \scriptsize dress & \scriptsize - & {\color{red}$\times$} & \scriptsize [-, ,, ., and, in] \\
\scriptsize the bird flew up into the & \scriptsize sky & \scriptsize the & {\color{red}$\times$} & \scriptsize [the, world, -, ,, and] \\
\scriptsize she gave her friend a big & \scriptsize hug & \scriptsize . & {\color{red}$\times$} & \scriptsize [., ,, and, -, of] \\
\scriptsize the cat jumped on the & \scriptsize table & \scriptsize the & {\color{red}$\times$} & \scriptsize [the, ", ,, -, and] \\
\scriptsize he found a beautiful & \scriptsize flower & \scriptsize , & {\color{red}$\times$} & \scriptsize [,, ., and, -, the] \\
\scriptsize they played together in the & \scriptsize park & \scriptsize the & {\color{red}$\times$} & \scriptsize [the, a, ,, time, an] \\
\scriptsize she sang a beautiful & \scriptsize song & \scriptsize , & {\color{red}$\times$} & \scriptsize [,, ., and, -, in] \\
\scriptsize he wanted to go & \scriptsize home & \scriptsize to & {\color{red}$\times$} & \scriptsize [to, the, a, ., in] \\
\scriptsize the really big & \scriptsize bear & \scriptsize - & {\color{red}$\times$} & \scriptsize [-, ,, ., and, the] \\
\scriptsize a very small & \scriptsize mouse & \scriptsize - & {\color{red}$\times$} & \scriptsize [-, ,, of, ., day] \\
\scriptsize the extremely bright & \scriptsize light & \scriptsize - & {\color{red}$\times$} & \scriptsize [-, ,, and, the, .] \\
\scriptsize he was very happy and started to & \scriptsize laugh & \scriptsize the & {\color{red}$\times$} & \scriptsize [the, a, was, it, "] \\
\scriptsize the dog saw a cat and started to & \scriptsize bark & \scriptsize the & {\color{red}$\times$} & \scriptsize [the, a, an, first, in] \\
\scriptsize they went outside to & \scriptsize play & \scriptsize the & {\color{red}$\times$} & \scriptsize [the, a, to, ,, and] \\
\scriptsize he ran as fast as he & \scriptsize could & \scriptsize 's & {\color{red}$\times$} & \scriptsize ['s, ., and, is, ,] \\
\scriptsize the baby started to & \scriptsize cry & \scriptsize the & {\color{red}$\times$} & \scriptsize [the, a, to, ., in] \\
\scriptsize he carefully & \scriptsize placed & \scriptsize . & {\color{red}$\times$} & \scriptsize [., ,, the, and, was] \\
\scriptsize they quickly & \scriptsize ran & \scriptsize , & {\color{red}$\times$} & \scriptsize [,, ., had, and, the] \\
\scriptsize she slowly & \scriptsize walked & \scriptsize . & {\color{red}$\times$} & \scriptsize [., ,, the, had, was] \\
\scriptsize he gently & \scriptsize touched & \scriptsize , & {\color{red}$\times$} & \scriptsize [,, ., the, and, was] \\
\scriptsize he was so tired that he fell & \scriptsize asleep & \scriptsize . & {\color{red}$\times$} & \scriptsize [., to, as, ", and] \\
\scriptsize she looked at the stars and felt & \scriptsize happy & \scriptsize . & {\color{red}$\times$} & \scriptsize [., ,, that, ", to] \\
\scriptsize she was really & \scriptsize excited & \scriptsize to & {\color{red}$\times$} & \scriptsize [to, the, a, ., in] \\
\scriptsize the water was very & \scriptsize cold & \scriptsize . & {\color{red}$\times$} & \scriptsize [., to, ,, and, the] \\
\scriptsize he was feeling very & \scriptsize sad & \scriptsize to & {\color{red}$\times$} & \scriptsize [to, the, ., ,, that] \\
\scriptsize it was an incredibly & \scriptsize beautiful & \scriptsize . & {\color{red}$\times$} & \scriptsize [., and, to, -, ,] \\
\scriptsize she had a particularly & \scriptsize special & \scriptsize . & {\color{red}$\times$} & \scriptsize [., to, -, and, ,] \\
\scriptsize the surprisingly & \scriptsize gentle & \scriptsize - & {\color{red}$\times$} & \scriptsize [-, ,, the, of, and] \\
\scriptsize an unusually & \scriptsize quiet & \scriptsize the & {\color{red}$\times$} & \scriptsize [the, -, of, ", ,] \\
\scriptsize the remarkably & \scriptsize clever & \scriptsize - & {\color{red}$\times$} & \scriptsize [-, ,, the, of, and] \\
\scriptsize a wonderfully & \scriptsize warm & \scriptsize - & {\color{red}$\times$} & \scriptsize [-, of, ,, and, in] \\
\scriptsize she opened the door and walked & \scriptsize inside & \scriptsize . & {\color{red}$\times$} & \scriptsize [., ,, to, in, and] \\
\scriptsize the story ended & \scriptsize happily & \scriptsize . & {\color{red}$\times$} & \scriptsize [., in, ,, the, and] \\
\scriptsize the children were playing & \scriptsize outside & \scriptsize . & {\color{red}$\times$} & \scriptsize [., ,, in, and, the] \\
\scriptsize the boat sailed across the & \scriptsize ocean & \scriptsize the & {\color{red}$\times$} & \scriptsize [the, ,, -, and, a] \\
\scriptsize she hung the picture on the & \scriptsize wall & \scriptsize " & {\color{red}$\times$} & \scriptsize [", new, the, -, early] \\
\scriptsize the keys were left on the & \scriptsize counter & \scriptsize the & {\color{red}$\times$} & \scriptsize [the, ,, ", and, -] \\
\scriptsize he parked the car in the & \scriptsize garage & \scriptsize the & {\color{red}$\times$} & \scriptsize [the, ,, and, south, west] \\
\scriptsize the plane disappeared behind the & \scriptsize clouds & \scriptsize the & {\color{red}$\times$} & \scriptsize [the, ,, a, and, -] \\
\scriptsize they arrived early in the & \scriptsize morning & \scriptsize the & {\color{red}$\times$} & \scriptsize [the, a, ,, early, time] \\
\scriptsize the concert lasted until & \scriptsize midnight & \scriptsize . & {\color{red}$\times$} & \scriptsize [., ,, the, to, that] \\
\scriptsize we will meet again next & \scriptsize week & \scriptsize the & {\color{red}$\times$} & \scriptsize [the, to, and, ,, a] \\
\scriptsize she poured the milk into the & \scriptsize glass & \scriptsize the & {\color{red}$\times$} & \scriptsize [the, -, ,, and, a] \\
\scriptsize he spread butter on the & \scriptsize bread & \scriptsize " & {\color{red}$\times$} & \scriptsize [", the, a, -, early] \\
\scriptsize the chef chopped the & \scriptsize vegetables & \scriptsize the & {\color{red}$\times$} & \scriptsize [the, and, ,, a, .] \\
\scriptsize she stirred the soup with a & \scriptsize spoon & \scriptsize - & {\color{red}$\times$} & \scriptsize [-, and, ,, ", new] \\
\scriptsize the horse galloped across the & \scriptsize field & \scriptsize the & {\color{red}$\times$} & \scriptsize [the, ,, and, -, world] \\
\scriptsize the fish swam in the & \scriptsize pond & \scriptsize the & {\color{red}$\times$} & \scriptsize [the, world, and, ,, new] \\
\scriptsize the eagle soared above the & \scriptsize mountains & \scriptsize the & {\color{red}$\times$} & \scriptsize [the, ,, and, ", a] \\
\scriptsize after the long hike he was extremely & \scriptsize tired & \scriptsize the & {\color{red}$\times$} & \scriptsize [the, to, ., and, ,] \\
\scriptsize the news made everyone very & \scriptsize happy & \scriptsize the & {\color{red}$\times$} & \scriptsize [the, a, to, that, .] \\
\scriptsize the dark alley made her feel & \scriptsize afraid & \scriptsize . & {\color{red}$\times$} & \scriptsize [., ,, the, and, of] \\
\scriptsize the puzzle was incredibly & \scriptsize difficult & \scriptsize . & {\color{red}$\times$} & \scriptsize [., and, to, ,, the] \\
\scriptsize the audience was completely & \scriptsize silent & \scriptsize . & {\color{red}$\times$} & \scriptsize [., and, to, ,, the] \\
\scriptsize the teacher asked the students to & \scriptsize listen & \scriptsize the & {\color{red}$\times$} & \scriptsize [the, be, to, ", ,] \\
\scriptsize she decided to finally & \scriptsize leave & \scriptsize the & {\color{red}$\times$} & \scriptsize [the, to, a, ., ,] \\
\scriptsize the alarm clock started to & \scriptsize ring & \scriptsize the & {\color{red}$\times$} & \scriptsize [the, ,, ., a, to] \\
\scriptsize the snow began to & \scriptsize melt & \scriptsize the & {\color{red}$\times$} & \scriptsize [the, ., ,, a, to] \\
\scriptsize he promised he would never & \scriptsize forget & \scriptsize to & {\color{red}$\times$} & \scriptsize [to, the, be, been, .] \\
\scriptsize the children love to & \scriptsize swim & \scriptsize the & {\color{red}$\times$} & \scriptsize [the, ., ,, a, "] \\
\scriptsize she wanted to learn how to & \scriptsize dance & \scriptsize the & {\color{red}$\times$} & \scriptsize [the, a, ,, be, .] \\
\scriptsize the mechanic tried to & \scriptsize repair & \scriptsize the & {\color{red}$\times$} & \scriptsize [the, a, to, be, "] \\
\scriptsize dark clouds gathered before the & \scriptsize storm & \scriptsize the & {\color{red}$\times$} & \scriptsize [the, a, ,, ", time] \\
\scriptsize the flowers bloomed in the & \scriptsize spring & \scriptsize the & {\color{red}$\times$} & \scriptsize [the, ,, world, and, "] \\
\scriptsize the leaves fall from the trees in & \scriptsize autumn & \scriptsize the & {\color{red}$\times$} & \scriptsize [the, ,, a, -, .] \\
\scriptsize the river flowed into the & \scriptsize sea & \scriptsize the & {\color{red}$\times$} & \scriptsize [the, and, ,, world, -] \\
\scriptsize the student wrote an & \scriptsize essay & \scriptsize the & {\color{red}$\times$} & \scriptsize [the, a, ,, ., that] \\
\scriptsize she typed the email on her & \scriptsize keyboard & \scriptsize . & {\color{red}$\times$} & \scriptsize [., ,, to, the, as] \\
\scriptsize the professor gave a & \scriptsize lecture & \scriptsize " & {\color{red}$\times$} & \scriptsize [", the, name, -, a] \\
\scriptsize he submitted his final & \scriptsize report & \scriptsize to & {\color{red}$\times$} & \scriptsize [to, ,, the, ., a] \\
\scriptsize the crowd started to & \scriptsize cheer & \scriptsize the & {\color{red}$\times$} & \scriptsize [the, a, to, ., and] \\
\scriptsize he bent down to & \scriptsize pick & \scriptsize the & {\color{red}$\times$} & \scriptsize [the, a, to, ", .] \\
\scriptsize she leaned forward to & \scriptsize whisper & \scriptsize a & {\color{red}$\times$} & \scriptsize [a, the, to, in, one] \\
\scriptsize the wind continued to & \scriptsize blow & \scriptsize the & {\color{red}$\times$} & \scriptsize [the, a, ., to, and] \\
\scriptsize they gathered around to & \scriptsize celebrate & \scriptsize the & {\color{red}$\times$} & \scriptsize [the, a, to, ,, .] \\
\scriptsize he struggled to & \scriptsize breathe & \scriptsize the & {\color{red}$\times$} & \scriptsize [the, to, a, ", be] \\
\scriptsize the sunset was absolutely & \scriptsize stunning & \scriptsize . & {\color{red}$\times$} & \scriptsize [., ,, and, to, -] \\
\scriptsize the old house looked rather & \scriptsize creepy & \scriptsize . & {\color{red}$\times$} & \scriptsize [., -, ,, to, the] \\
\scriptsize his explanation was perfectly & \scriptsize clear & \scriptsize . & {\color{red}$\times$} & \scriptsize [., ,, and, to, as] \\
\scriptsize the cake tasted absolutely & \scriptsize delicious & \scriptsize the & {\color{red}$\times$} & \scriptsize [the, ., to, ,, and] \\
\scriptsize the mountain trail was extremely & \scriptsize steep & \scriptsize . & {\color{red}$\times$} & \scriptsize [., and, ,, to, the] \\
\scriptsize the instructions were surprisingly & \scriptsize simple & \scriptsize the & {\color{red}$\times$} & \scriptsize [the, ., and, to, ,] \\
\scriptsize the scientist published an important & \scriptsize discovery & \scriptsize , & {\color{red}$\times$} & \scriptsize [,, the, -, and, in] \\
\scriptsize the city was known for its rich & \scriptsize history & \scriptsize . & {\color{red}$\times$} & \scriptsize [., ,, -, and, the] \\
\scriptsize she always valued honesty and & \scriptsize kindness & \scriptsize the & {\color{red}$\times$} & \scriptsize [the, to, a, that, "] \\
\scriptsize the treaty established lasting & \scriptsize peace & \scriptsize the & {\color{red}$\times$} & \scriptsize [the, to, ,, ., and] \\
\scriptsize the car moved forward & \scriptsize slowly & \scriptsize to & {\color{red}$\times$} & \scriptsize [to, and, ., the, ,] \\
\scriptsize he waited & \scriptsize patiently & \scriptsize to & {\color{red}$\times$} & \scriptsize [to, for, the, in, .] \\
\scriptsize she tied the laces on her & \scriptsize shoes & \scriptsize . & {\color{red}$\times$} & \scriptsize [., and, to, ,, as] \\
\scriptsize he wrapped the scarf around his & \scriptsize neck & \scriptsize . & {\color{red}$\times$} & \scriptsize [., ,, and, to, the] \\
\scriptsize the doctor examined the patient's & \scriptsize heart & \scriptsize the & {\color{red}$\times$} & \scriptsize [the, -, of, in, ,] \\
\scriptsize she checked the notifications on her & \scriptsize phone & \scriptsize . & {\color{red}$\times$} & \scriptsize [., ,, and, the, to] \\
\scriptsize the password must contain at least eight & \scriptsize characters & \scriptsize , & {\color{red}$\times$} & \scriptsize [,, the, ., to, and] \\
\scriptsize he plugged the charger into the & \scriptsize socket & \scriptsize the & {\color{red}$\times$} & \scriptsize [the, ", old, new, a] \\
\scriptsize the door began to & \scriptsize open & \scriptsize the & {\color{red}$\times$} & \scriptsize [the, a, ., to, ,] \\
\scriptsize she refused to & \scriptsize quit & \scriptsize to & {\color{red}$\times$} & \scriptsize [to, the, be, a, "] \\
\bottomrule
\end{longtable}

\end{document}